\RequirePackage{fix-cm}
\documentclass[oneside,phd,4x6]{snuthesis}

\usepackage[titles]{tocloft} % when you don't use subfigure package
\makeatletter
\if@snu@ko
\renewcommand\cftchappresnum{제~}
\renewcommand\cftchapaftersnum{~장}
\renewcommand\cftfigpresnum{그림~}
\renewcommand\cfttabpresnum{표~}
\else
\renewcommand\cftchappresnum{}
\renewcommand\cftfigpresnum{}
\renewcommand\cfttabpresnum{}
\fi
\makeatother
\newlength{\mytmplen}
\makeatletter
\g@addto@macro\appendix{%
	\addtocontents{toc}{%
		\settowidth{\mytmplen}{\bfseries\protect\cftchappresnum\protect\cftchapaftersnum}%
		\addtolength{\cftchapnumwidth}{-\mytmplen}%
		\protect\renewcommand{\protect\cftchappresnum}{\appendixname~}%
		\protect\renewcommand{\protect\cftchapaftersnum}{}%
		\settowidth{\mytmplen}{\bfseries\protect\cftchappresnum\protect\cftchapaftersnum}%
		\addtolength{\cftchapnumwidth}{\mytmplen}%
	}%
}
\makeatother

\newlength\longest

\ifpdf
\input glyphtounicode\pdfgentounicode=1 %type 1 font
\usepackage[pdftex]{graphicx}
\else
\usepackage[dvipdfmx]{graphicx}
\fi

\renewcommand{\cftpartpresnum}{Part~}
\let\cftoldpartfont\cftpartfont
\renewcommand{\cftpartfont}{\cftoldpartfont\cftpartpresnum} %part in front of part number
\cftpagenumbersoff{part} % page number off for parts

\usepackage[defaultlines=2,all]{nowidow} % do you hate poor widows?
\usepackage{pifont}% http://ctan.org/pkg/pifont
\usepackage{xpatch}
\usepackage{makecell}
\usepackage{booktabs} % to use \toprule \midrule \bottomrule commands in tabular
\usepackage[backend=bibtex,style=numeric,abbreviate=false,backref=true,maxbibnames=12,sortcites=true]{biblatex}
\xpatchbibmacro{date+extradate}{%
	\printtext[parens]%
}{%
	\setunit*{\addcomma\space}%
	\printtext%
}{}{} 
\renewbibmacro{in:}{}
\DeclareFieldFormat[article,incollection,inproceedings]{pages}{#1}
\DeclareFieldFormat[book,inbook]{pages}{#1\addspace pp\adddot} % pp. for books
\DeclareDelimFormat[bib,biblist]{nametitledelim}{\addcolon\space} % colon after year
\DeclareFieldFormat
[article,incollection,inproceedings]
{title}{#1}
\renewbibmacro*{journal+issuetitle}{%
	\usebibmacro{journal}%
	\setunit*{\addcomma\space}%
	\iffieldundef{series}
	{}
	{\newunit
		\printfield{series}%
		\setunit{\addspace}}%
	\usebibmacro{volume+number+eid}%
	\setunit{\addspace}%
	\usebibmacro{issue+date}%
	\setunit{\addcolon\space}%
	\usebibmacro{issue}%
	\newunit}
\DeclareFieldFormat{journaltitle}{\mkbibemph{#1\isdot}}
\DeclareFieldFormat[article]{volume}{\bfseries #1}
\renewbibmacro{volume+number+eid}{% removing number entry
	\printfield{volume}%
	\setunit{\addcomma\space}%
	\printfield{eid}}
\DeclareFieldFormat{titlecase}{\MakeTitleCase{#1}}
\newrobustcmd{\MakeTitleCase}[1]{%
	\ifthenelse{\ifcurrentfield{booktitle}\OR\ifcurrentfield{booksubtitle}%
		\OR\ifcurrentfield{maintitle}\OR\ifcurrentfield{mainsubtitle}%
		\OR\ifcurrentfield{journaltitle}\OR\ifcurrentfield{journalsubtitle}%
		\OR\ifcurrentfield{issuetitle}\OR\ifcurrentfield{issuesubtitle}%
		\OR\ifentrytype{book}\OR\ifentrytype{mvbook}\OR\ifentrytype{bookinbook}%
		\OR\ifentrytype{booklet}\OR\ifentrytype{suppbook}%
		\OR\ifentrytype{collection}\OR\ifentrytype{mvcollection}%
		\OR\ifentrytype{suppcollection}\OR\ifentrytype{manual}%
		\OR\ifentrytype{periodical}\OR\ifentrytype{suppperiodical}%
		\OR\ifentrytype{proceedings}\OR\ifentrytype{mvproceedings}%
		\OR\ifentrytype{reference}\OR\ifentrytype{mvreference}%
		\OR\ifentrytype{report}\OR\ifentrytype{thesis}}
	{#1}
	{\MakeSentenceCase{#1}}} % make only first letter of article title capital only
\AtBeginBibliography{%
}
\usepackage{amsmath} \allowdisplaybreaks[4] % this will allow page breaks for align. Likelihood options: 0-4
\usepackage{amssymb}
\usepackage{bbm}
\usepackage{amsthm}
\newtheoremstyle{newdef}% name of the style to be used
{}% measure of space to leave above the theorem. E.g.: 3pt
{}% measure of space to leave below the theorem. E.g.: 3pt
{}% name of the font to use in the body of the theorem
{}% measure of space to indent
{\bfseries}% name of head font
{}% punctuation between head and body
{1em}% space after theorem head; " " = normal interword space
{}
\theoremstyle{newdef}

\makeatletter
\def\thm@space@setup{%
	\thm@preskip=0.5\baselineskip % modifying buffer for theorems and definitions
	\thm@postskip=\thm@preskip % or whatever, if you don't want them to be equal
}
\makeatother
\newtheoremstyle{slthm}% name of the style to be used
{}% measure of space to leave above the theorem. E.g.: 3pt
{}% measure of space to leave below the theorem. E.g.: 3pt
{\slshape}% name of the font to use in the body of the theorem
{}% measure of space to indent
{\bfseries}% name of head font
{}% punctuation between head and body
{1em}% space after theorem head; " " = normal interword space
{}
\theoremstyle{slthm}

\usepackage{commath}

\usepackage{enumitem}
\setlist{nolistsep} % remove spacing between items in a list
\usepackage{upgreek}
\usepackage{mathrsfs}
\usepackage{siunitx}
\usepackage[dvipsnames]{xcolor}
\usepackage{tikz}
\usetikzlibrary{positioning}
\usetikzlibrary{trees}
\tikzset{>=latex}
\usepackage{tabularx}
\usepackage{multirow}
\usepackage[flushleft]{threeparttable}
\usepackage[nodisplayskipstretch]{setspace} 
\usepackage[labelfont=bf]{caption} % change number for spacing modification in: font={stretch=1.1}
\usepackage{tocloft} % table of contents
\usepackage{xpatch}

\makeatletter
\xpatchcmd{\@chapter}{%
	\addtocontents{lof}{\protect\addvspace{10\p@}}%
	\addtocontents{lot}{\protect\addvspace{10\p@}}%
}{}{}{}
\makeatother
\usepackage{etoolbox}
\AtBeginEnvironment{quote}{\singlespace\vspace{-\topsep}}
\AtEndEnvironment{quote}{\vspace{-\topsep}\endsinglespace}
\usepackage[activate={true,nocompatibility},final,tracking=true,kerning=true,spacing=true,factor=1100,stretch=10,shrink=10]{microtype}
\SetTracking{encoding={*}, shape=sc}{40} % reduce spacing in scshape
\usepackage{mathtools} % for dashed and dotted lines %\sampleline{},\sampleline{dashed},\sampleline{dotted},\sampleline{dash pattern=on .7em off .2em on .05em off .2em}
\DeclareRobustCommand\sampleline[1]{%
	\tikz\draw[#1,line width=0.35mm] (0,0) (0,\the\dimexpr\fontdimen22\textfont2\relax)
	-- (1.5em,\the\dimexpr\fontdimen22\textfont2\relax);%
}
\usepackage[figuresright]{rotating}
\usepackage{imakeidx}
\makeindex[columns=2, title=Index, 
options= -s mystyle.ist]
\makeatletter
\def\@idxitem{\par\hangindent 0pt}
\makeatother
\indexsetup{othercode=\small}

\usepackage{lettrine}
\usepackage{GoudyIn}

\usepackage{pgfornament}

\usepackage{needspace} % for unusual circumstances requiring more space in page (ex. \needspace{4\baselineskip})

\usepackage{titlesec}
\titleformat{\section}{\normalfont\fontsize{14}{14}\bfseries}{\thesection}{1em}{\hspace{0em}}{}
\titleformat{\chapter}[display]{\normalfont\fontsize{16}{16}\bfseries}{\chaptertitlename\ \thechapter}{0em}{\huge}
\titlespacing*{\chapter}{0pt}{30pt}{40pt}

\usepackage{afterpage} % needed for full page figures
\makeatletter
\@fpsep\textheight
\afterpage{\global\setlength\@fpsep{8\p@ \@plus 2fil}}
\makeatother


\usepackage{pgfplots,pgfplotstable}
\usepgflibrary{patterns}
\usetikzlibrary{patterns}
\usepackage{stfloats}
\usepackage{placeins}
\usepackage{calc}
\usepackage{array, booktabs, makecell}
\usepackage{amsfonts}
\usepackage{algorithmic}
\usepackage[caption=false]{subfig}
\usepackage{graphicx}
\usepackage{textcomp}

\usepackage{algorithm}
\usepackage{xcolor}

\pgfplotsset{compat=newest,width=8cm,height=4.8cm,
	every axis/.append style={
		tick label style={font=\fontsize{6}{6.5}\selectfont},
		label style={font=\fontsize{6}{6.5}\selectfont}
	},
	legend image code/.code={
		\draw[mark repeat=2,mark phase=2]
		plot coordinates {
			(0cm,0cm)
			(0.15cm,0cm)        
			(0.3cm,0cm)
		};
	},
	major tick length=0.03cm,
	xtick align=outside,ytick align=outside,
	axis x line*=bottom,axis y line*=left,axis line style=ultra thin
}

\DeclareMathOperator*{\argmax}{arg\,max}
\DeclareMathOperator*{\argmin}{arg\,min}

\definecolor{DarkOrange}{RGB}{209,78,1}
\definecolor{DarkGreen}{RGB}{3,117,3}
\definecolor{DarkBlue}{RGB}{20,68,106}

\definecolor{Orange}{RGB}{255,127,14}
\definecolor{Green}{RGB}{65,197,114}
\definecolor{Blue}{RGB}{38,162,237}

\title{ Enhancing Explainability of Graph Neural Networks Through Conceptual and Structural Analyses and Their Extensions} 
\title*{그래프 신경망의 설명 가능성 향상을 위한 \\  개념적 및 구조적 분석과 확장 연구}  
\allcapstitle{ENHANCING EXPLAINABILITY OF GRAPH}{NEURAL NETWORKS THROUGH CONCEPTUAL AND STRUCTURAL}{ANALYSES AND THEIR EXTENSIONS} 

\academicko{공학}
\schoolen{Seoul National University}
\departmenten{Electrical and Computer Engineering}
\departmentko{전기정보공학부}
\bachelors{Hanoi University of Science and Technology}
\bsyear{2014}
\masters{Seoul National University}
\msyear{2019}

\author{Bui Tien Cuong} % English Name
\author*{부이~쿠옹~티엔} % Korean Name
\authorallcaps{BUI TIEN CUONG} % All caps English name
\studentnumber{2019-35731}

\advisor{Wen-Syan Li}
\advisor*{Wen-Syan Li}

\graddate{2024년~~8월} % Graduation Year and Month in Korean
\endate{August 2024} % Graduation Month in English
\defensedate{2024년~~06월~~04일} % Defense Date in Korean
\gradyear{2024} % Year of Graduation

\submissiondate{2024~년~6~월}

\approvaldate{2024~년~~~~~월}

\committeemembers%
{정~교~민~~~(인)}%
{웬~샨~리~~~(인)}%
{박~현~우~~~(인)}%
{심~준~호~~~(인)}%
{이~상~학~~~(인)}% 이상학
\encommitteemembers%
{Professor Kyomin Jung}%
{Professor Wen-Syan Li}%
{Professor Hyunwoo Park}%
{Professor Junho Shim}%
{Professor Sanghack Lee}%

\usepackage[]{hyperref}
\hypersetup{
	pdftitle={},
	pdfauthor={Tien-Cuong Bui},
	pdfsubject={Ph.D. Dissertation},
	pdfkeywords={Graph Neural Networks, eXplainable AI, Human-in-the-loop ML, Case-based Reasoning, Knowledge Distillation, Large Language Models},
	bookmarksnumbered=true,     
	bookmarksopen=true,         
	bookmarksopenlevel=1,       
	colorlinks=false, % do not use links for references and footnotes
	hidelinks, % remove ugly boxed windows indicating links           
	pdfstartview=Fit, % choose "Fit" to start with the fit window            
	pdfpagemode=UseOutlines,    % outlines on a pdf document
	pdfpagelayout=OneColumn % two-page viewer with odd number pages to the right/ other option: "OneColumn" for scrolling
}
\usepackage{cleveref}
\crefname{figure}{Figure}{Figures}
\crefname{table}{Table}{Tables}
\crefname{algorithm}{Algorithm}{Algorithms}
\crefname{equation}{Equation}{Equations}
\crefname{section}{Section}{Sections}
\crefname{chapter}{Chapter}{Chapters}

\usepackage{pdfpages} %z activate for inserting injun.pdf

\begin{document}
	\pagenumbering{Roman}

	%%%%%%%%%%%%%%%%%%%%%%%%%%%%%%%%%%%%%%%%%%%%%%%%%%%%%%%%%%%%%%%
	%% FRONT MATTER
	%%%%%%%%%%%%%%%%%%%%%%%%%%%%%%%%%%%%%%%%%%%%%%%%%%%%%%%%%%%%%%%
	
	%% Cover to be used for thesis defense (dddd)
	%\presentation{최종} % options: 최종, 중간, 제안
    %\makefrontcover % final defence
	
	% Hardcover (English version) (ssss)
	\hardcoveren
	\cleardoublepage
	
	% Hardcover (Default - Korean based) (ssss)f
	% \hardcover
	% \cleardoublepage
	
	% \includepdf{Injun/phd_injun_66.pdf}
	
	%% Turn on inner cover (English title and committee member names) options: jadae vs. non-jadae (ssss)
	\makefrontcoveren % for those who did not go to snu for undergrad and master's
	% \makefrontcoverenjadae % jadae option for those who went to snu for underground

	%% REQUIRED
	\makeapproval % for some reason, page margins get messed up without this command

	%% Colophon (ssss)
	\publications
	{Tien-Cuong Bui, Van-Duc Le, Wen-Syan Li, 2023. \emph{IEEE Access} DOI:10.1109/ACCESS.2023.3270385; } 
	{Tien-Cuong Bui and Wen-Syan Li, 2023. \emph{IEEE ICDM 2023} DOI:10.1109/ICDM58522.2023.00106}
	{Tien-Cuong Bui and Wen-Syan Li, 2024. \emph{PAKDD 2024} DOI:10.1007/978-981-97-2242-6\_9}
	{Tien-Cuong Bui and Wen-Syan Li, 2024. \emph{PAKDD 2024}  DOI:10.1007/978-981-97-2650-9\_2}{}{}
	\makecolophon
	\clearpage
	
	%%%%%%%%%%%%%%%%%%%%%%%%%%%%%%%%%%%%%%%%%%%%%%%%%%%%%%%%%%%%%%%%%%%%
	%% ABSTRACT
	%%%%%%%%%%%%%%%%%%%%%%%%%%%%%%%%%%%%%%%%%%%%%%%%%%%%%%%%%%%%%%%%%%%%
	\pagenumbering{Roman} % start page numbering (first in Roman numeral)
	
	
	\begin{abstract}
	Graph Neural Networks (GNNs) have become a powerful tool for modeling and analyzing data with graph structures. The wide adoption in numerous applications underscores the value of these models. However, the complexity of these methods often impedes understanding their decision-making processes.  
	Current Explainable AI (XAI) methods struggle to untangle the intricate relationships and interactions within graphs. Several methods have tried to bridge this gap via a post-hoc approach or self-interpretable design. Most of them focus on graph structure analysis to determine essential patterns that correlate with prediction outcomes. While post-hoc explanation methods are adaptable, they require extra computational resources and may be less reliable due to limited access to the model's internal workings. Conversely, Interpretable models can provide immediate explanations, but their generalizability to different scenarios remains a major concern. 
	
	To address these shortcomings, this thesis seeks to develop a novel XAI framework tailored for graph-based machine learning. 
	The proposed framework aims to offer adaptable, computationally efficient explanations for GNNs, moving beyond individual feature analysis to capture how graph structure influences predictions. It presents a general approach to enhance the interpretability of existing GNN architectures by training multiple specialty learners, each capturing specific types of interactions within graphs, such as features or message-passing processes. Later, multiple explainers are constructed to offer various explanation modalities based on trained specialty learners.
	The effectiveness of example-based explanations and the natural interpretability of the KNN algorithm motivate the creation of novel interpretable GNNs. The framework extracts frequently occurring ``concepts'' (substructures) from training graphs, used as a basis for inferring predictions and generating explanations. The goal is a multifaceted explanation system offering compact, user-centric insights. Additionally, the framework proposes an approximation method for structure similarity between two graphs via Earth Mover Distance optimal transport, which enhances both predictive performance and the user comprehension of reference selection. 
	Diverse explanation modalities provide users with meaningful insights into the internal logic of models, which can be leveraged for model debugging, debiasing, and improvement.
	Building upon this intuition, the framework aims to incorporate domain knowledge to guide GNNs toward more human-understandable representations, fostering trust and ethical use of this technology. Specifically, it allows domain experts to actively verify and control representation learning and reference selection processes by providing multi-level knowledge-guided constraints.
	The thesis presents extensive experimental results and findings that underscore the efficiency and effectiveness of the proposed framework. Finally, it concludes with a thorough discussion of possible avenues for future work, practical applications, and potential extensions into the latest advanced fields like large language models.
\end{abstract}
	
	%% Turn on for acknowledgment in the front matter (English)
	%\cleardoublepage
	%\acknowledgment
	%Thank you.
	\clearpage
	
	%%%%%%%%%%%%%%%%%%%%%%%%%%%%%%%%%%%%%%%%%%%%%%%%%%%%%%%%%%%%%%%%%%%%%%%%%%%
	%% TABLE OF CONTENTS, LIST OF FIGURES, ETC (NO NEED TO MODIFY; ALREADY FORMATTED)
	%%%%%%%%%%%%%%%%%%%%%%%%%%%%%%%%%%%%%%%%%%%%%%%%%%%%%%%%%%%%%%%%%%%%%%%%%%%
	\begingroup
	\setstretch{1.1} % make multiline chapter titles single spaced, also turn on tocloft options
	\cftsetpnumwidth{2.0em} \cftsetrmarg{3.55em} % adjust spacing between page number and title
	\phantomsection\clearpage\addcontentsline{toc}{chapter}{Contents}%
	%	\tableofcontents
	{\def\makebox[#1][#2]#3{#3}\tableofcontents}
	\endgroup
	% some tolerance for nicer pagination
	\addtolength{\topskip}{0pt plus 5pt minus 5pt}
	\addtolength{\parskip}{0pt plus 5pt minus 5pt}
	% make captions single spacing in the list of figures and list of tables
	% be sure to also modify tocloft options for item separation
	\begin{sloppypar} % sloppy command here to avoid overfull hbox
		\begingroup
		\emergencystretch 1em % reduce the spacing between words under sloppy
		\countdef\interlinepenalty255 %allow page breaks for LoF figure captions
		\setstretch{1.1} 
		\cftsetpnumwidth{2.0em} \cftsetrmarg{3.55em} % adjust spacing between page number and caption
		\clearpage\phantomsection\clearpage\addcontentsline{toc}{chapter}{List of Figures}%
		%	\listoffigures
		{\def\makebox[#1][#2]#3{#3}\listoffigures}
		\clearpage\phantomsection\clearpage\addcontentsline{toc}{chapter}{List of Tables}%
		%	\listoftables
		{\def\makebox[#1][#2]#3{#3}\listoftables}
		\endgroup
	\end{sloppypar}
	
	\flushbottom % For consistent bottom page margins, use with caution! to undo, use \raggedbottom
	\addtolength{\baselineskip}{0pt plus 1pt minus 1pt} % with flush bottom, give linespace some room
	
	\cleardoublepage % fresh start for chapter 1
	\pagenumbering{arabic}
	\addtolength{\topskip}{0pt plus-3pt minus-3pt} %undo too much wiggle room
	\addtolength{\parskip}{0pt plus-4pt minus-4pt}
	%\setcounter{chapter}{-1} % Chapter 0 for "Overview" (per tradition).
	
	%%%%%%%%%%%%%%%%%%%%%%%%%%%%%%%%%%%%%%%%%%%%%%%%%%%%%%%%%%%%%%%%%%%%%%%%%%%%%%%%%
	%% MAIN CONTENTS (START WRITING YOUR DISSERTATION!)
	%%%%%%%%%%%%%%%%%%%%%%%%%%%%%%%%%%%%%%%%%%%%%%%%%%%%%%%%%%%%%%%%%%%%%%%%%%%%%%%%%
	\chapter{Introduction}
% summarize of GNN
Graph Neural Networks (GNNs) \cite{zhou2020graph,zhang2020deep, xia2021graph} are effective for extracting insights from graph data, proving valuable in diverse domains, like social networks \cite{fan2019graph, sankar2021graph}, bio-informatics \cite{zhang2021graph}, and recommender systems \cite{wu2020graph}. Their unique ability to model complex relationships and dependencies in graphs allows for more accurate predictions and deeper insights than traditional methods. Advancements in GNN architectures have further enhanced their scalability and performance, making them practical for analyzing large-scale, real-world graph data. Ongoing research is continually expanding GNN capabilities, delving into areas like dynamic graph processing, higher-order graph representations, and integration with other deep learning or interpretable methods. This progress promises even broader applicability and impact. The widespread adoption of GNNs, particularly in critical decision-making scenarios, has highlighted the need for the interpretability of these models.  As a result, recent research efforts have prioritized enhancing the transparency and explainability of GNNs. 

% summarize of XAI
Explainable AI (XAI) methods \cite{dwivedi2023explainable, hassija2024interpreting}  provide valuable insights into the decision-making processes of AI models through various approaches categorized by timing (ante-hoc vs. post-hoc), scope (global vs. local), and the focus of the explanation (model-level vs. instance-level). Ante-hoc methods, which include transparent models like linear regression and decision trees, are designed to be inherently interpretable and are suited for scenarios that demand high levels of interpretability and trust, although they may sacrifice complexity and accuracy. Post-hoc methods, in contrast, treat models as opaque systems and focus on interpreting input-output relations; these methods are prevalent for their adaptability across different models and include techniques like SHAP \cite{lundberg2017unified} and LIME \cite{ribeiro2016should}, which, while computationally demanding, help clarify complex models. On a broader scale, global explanations aim to outline the overall logic of AI models, enhancing a broad understanding of their behavior, whereas local explanations delve into the root causes of individual predictions, offering detailed insights into particular decisions through techniques like surrogate, perturbation, gradient-based, and counterfactual analysis. The selection between these explanatory approaches varies with the goals of transparency and the application's requirements, providing either a holistic view or a detailed analysis of how specific input features affect predictions. Additionally, example-based explanations use concrete instances to make model behavior more comprehensible and relatable, though they may not fully capture the model's overall logic.

Despite noticeable success in demystifying the black-box phenomenon, existing XAI methods face significant challenges when it comes to graph data and GNNs due to the complexity of networks and internal interactions among elements. Specifically, traditional feature attribution methods fail to capture the complex interactions within graphs. Example-based approaches become overwhelmed by the sheer number of potential relationships, struggling to pinpoint the most relevant examples for explanation. Furthermore, the relational nature of graph data necessitates methods that can address both feature attributions and structural explanations, highlighting specific patterns that influence the final outcome. Addressing these challenges requires the development of novel XAI approaches tailored specifically for GNNs.

% summarize of XAI for GNNs
Lately, many methods \cite{yuan2022explainability} have been introduced to address the differences between traditional XAI methods and GNNs, reflecting the absence of a universal solution. These methods tackle the problem from diverse angles, with the majority categorized as post-hoc explanations and emphasizing instance-level explanations and structural analyses. Perturbation methods \cite{ying2019gnnexplainer, luo2020parameterized} are favored due to their benchmark datasets and strong performance. However, they require additional computational resources to train explanation models after the black-box GNN.  In contrast, interpretable models can output predictions with explanations immediately. This benefit has inspired the development of recent self-explainable GNNs  \cite{dai2021towards,zhang2022protgnn}, which rely on similarity-based objective functions. While these architectures achieve promising results on citation graphs  \cite{dgldata}, their generalizability to other datasets remains questionable. the complexity of optimization processes associated with structural similarity measurements can lead to computational challenges, particularly in large-scale graphs.

% summarize of human-centric AI
Human-in-the-Loop (HITL) AI \cite{wu2022survey, mosqueira2023human} and Human-centric AI \cite{taylor2023human}, both connected to XAI, focus on enhancing human interaction with AI systems through clarity and understanding. HITL AI integrates human judgment into the AI operational process, making humans active participants who guide and refine AI decisions. This approach is essential in areas where AI autonomy is limited or sensitive, such as healthcare, legal, and engineering sectors, improving the system's accuracy and trustworthiness by combining human expertise with artificial intelligence. Conversely, Human-centric AI prioritizes the development of AI technologies that resonate with human values and ethics, aiming to create systems that are understandable, equitable, and respectful of human autonomy. It centers on the human experience in the design process, ensuring that AI technologies are not only efficient but also socially responsible and aligned with human goals, thus enhancing usability and inclusiveness. In both paradigms, explainability is key to enabling effective human oversight and ensuring that AI actions are aligned with ethical standards and societal values, thereby fostering trust and acceptance in AI applications.

% summarize this thesis
This thesis addresses the urgent need for a novel XAI framework designed specifically for graph-based machine learning. It aims to develop a framework that seamlessly integrates with existing GNN architectures, promoting adaptability, reusability, and generalization, while enhancing computational efficiency for real-time applications. It also moves beyond individual feature analysis to understand how intricate graph structures influence model predictions. To achieve these objectives, the framework proposes training multiple specialty learners, each concentrating on specific types of interactions. Later, diverse explanation modalities can be generated based on multiple explainers corresponding to trained learners. Inspired by the effectiveness of the example-based explanation approach and the inherent interpretability of the KNN algorithm, the proposed framework presents a novel approach to prediction inference and explanation. The core innovation lies in a concept-focused graph representation learning process, which pays attention to frequently occurring substructures (concepts), forming the basis for interpretable predictors. To further enhance the predictive power and interpretability, the framework implements a unique concept-focused structure similarity algorithm based on the Earth Mover Distance method. The framework further introduces a multifaceted approach to prediction explanation, generating diverse modalities to satisfy different user preferences. The ultimate objective is to provide compact, interpretable, and user-centric insights into model decision-making, which can promote model debugging, debiasing, and improvement. Building upon these advancements, this thesis proposes to enhance human-AI collaboration by incorporating domain-expert knowledge through active verification and multi-level constraints. This collaboration can guide GNNs toward human-aligned representations, reducing biases, increasing trust, and ultimately fostering more responsible use of GNN technology in high-stakes domains. Through various experiments and user studies, this thesis investigates the efficiency and effectiveness of the proposed framework. Finally, it includes a thorough discussion of potential applications that can be built on top of the proposed framework and possible extensions in the latest advanced fields like large language models (LLMs) \cite{zhao2023survey}.

For clarity, the remainder of this thesis is organized as follows:

\cref{chap:background} sets the stage by providing an overview of background concepts, including GNNs, XAI approaches, human-in-the-loop AI and human-centric AI, XAI for GNNs, and LLMs.

\cref{chap:scale} demonstrates how existing XAI methods and fundamental graph algorithms can be effectively adapted to explain GNN predictions. This chapter emphasizes the feasibility of employing established methods and algorithms in the context of GNN XAI, with minimal modifications required. This approach not only validates the adaptability of traditional XAI methods to newer architectures but also underscores the potential of these techniques in elucidating GNN behaviors.

In \cref{chap:concept}, the focus shifts toward structural analysis. This chapter proposes an innovative approach to designing interpretable GNNs without altering the core architecture of the backbone GNNs. This goal is achieved by introducing a novel objective function based on the graph information bottleneck theory. The chapter showcases how structural analysis can be seamlessly integrated into GNN models, enhancing interpretability while preserving predictive capability. Additionally, it presents a unique concept-focused structural similarity algorithm that increases predictive power and interpretability via structure alignments. This progression illustrates the evolution of the framework, highlighting a continuous effort to refine and improve the interpretability of GNN models.

Building upon the previous chapter, \cref{chap:human} recognizes the importance of human involvement in the training and fine-tuning processes. This chapter presents an effective method that allows humans to actively provide domain knowledge and give feedback to models. The method aims to ensure human-model alignment in reference selection. This approach not only leads to more accurate predictions but also makes models more interpretable, fostering human-AI decision-making.

\cref{chap:llm} explores the integration of LLMs into the area of GNNs and empirical applications of the proposed XAI framework. Recent advancements in LLMs open new opportunities and challenges for addressing graph-based machine learning problems. This chapter presents a few promising applications of LLM-GNN integration. It also discusses potential applications of the proposed framework, demonstrating how increased interpretability can drive innovation and responsible usage across various domains.

\cref{chap:conclusion} wraps up the thesis by outlining key contributions, examining the implications of research findings, and proposing avenues for future work.
	\chapter{Background and Fundamentals} \label{chap:background}
\section{Graph Neural Networks: An Overview}
GNNs \cite{zhang2020deep, xia2021graph} represent a class of neural networks specifically designed for handling graphs. Due to the complexity of data structures, graphs require special DL operations compared with images or text. Numerous GNN architectures, training paradigms, and acceleration frameworks have been proposed to address various challenges associated with real-world problems. 

GNNs have found extensive applications across various domains since they can model relational data efficiently. In computer vision, GNNs are effective in tasks like scene graph generation, analyzing point-cloud data from LiDAR scans, and skeleton-based action recognition or pose estimation \cite{to2021real}, leveraging their strength in modeling relationships among objects or points \cite{jiao2022graph}. Particularly in scene understanding, they transform visual elements into graph representations for deeper insight, combining these with textual information for enriched image generation. In recommender systems, GNNs surpass traditional methods by efficiently uncovering hidden patterns in recommendation graphs \cite{wu2020graph, gao2023survey}, significantly enhancing service accuracy. This is evident in their deployment in large-scale web services and e-commerce platforms, where they analyze user behavior for precise product recommendations. Moreover, GNNs are instrumental in social recommendation, modeling complex user interactions and preferences. In the Natural Language Processing (NLP) area, GNNs integrate with pre-trained models and word embeddings to handle syntactical graph representations of sentences and paragraphs, broadening the scope of applications in this field \cite{wu2023graph}. Urban computing also benefits from GNNs \cite{jin2023spatio, xue2022quantifying, li2022graph}, particularly in spatiotemporal problems like taxi demand forecasting, traffic flow analysis, smart parking systems \cite{duong2022towards}, and air quality monitoring \cite{le2023spatiotemporal}, showcasing their versatility in handling dynamic, real-world data sets.

\subsection{GNN architectures}
GNN architectures \cite{zhang2020deep, xia2021graph} have several variants, which can be classified into five main categories: Graph Convolutional Networks (GCNs), Graph Autoencoders, Graph Reinforcement Learning, Recurrent Graph Neural Networks, and Spatiotemporal Graph Neural Networks. This thesis mainly concentrates on explainable methods for GCNs, which are the most essential GNN architecture. 

Most GCN architectures can be represented via message-passing paradigms with three essential functions: propagation, aggregation, and update. A message $m^l_{ij} = \textrm{Message}(h^{l-1}_i, h^{l-1}_j)$, where $h$ denotes representation vectors of nodes at the previous layer $l-1$ passing through an edge between two nodes $j$ and $i$. Received messages at a node $i$ are aggregated as follows: $m^l_i = \textrm{Aggregate}({m^l_{ij}|j \in \mathcal{N}_i})$, where $\mathcal{N}_i$ denotes all incoming neighbors. The next layer's representations are computed via the formula $h^l_i = \textrm{Update}(m^l_i, h^{l-1}_i)$. The last layer's embeddings $h^L$ are utilized in downstream tasks.

\subsection{Fundamental ML Problems with Graphs}
Even though real-world graph applications can be diverse, this section formulates three fundamental classes of ML problems with graphs corresponding to granularity levels. 

\noindent\textbf{Node Classification and Regression.} These problems are fundamental in graph analytics involving assigning outputs to nodes within a graph. In classification, the objective is to label each node with a correct class from a predefined set, using a function $f: V \mapsto \{1,...,C\}$ that maps nodes in $V$ to classes in $C$. The regression task, while similar in methodology, differs in its goal, aiming to assign a continuous value to each node. This is done using a regression model $f: V \mapsto R $ that maps each node in 
$V$ to a numerical score. For both tasks, node embeddings $h^L$ are fed through feed-forward networks, which output either categorical labels or continuous values.

\noindent\textbf{Graph Classification and Regression.} These tasks aim to map input graphs to specific outputs. In Graph Classification, the objective is to assign each graph to one of several predefined classes, represented by a labeling function  $f: G \mapsto \{1,...,C\}$. For Graph Regression, the aim is to map each graph to a continuous score, using a regressor $f: G \mapsto R$. Both tasks involve a similar process where, after applying message-passing operators, a read-out operator is applied to node embeddings $h^L$ to output a single representation vector for the input graph. This vector then serves as the input for a feed-forward network, which outputs either a class label in classification or a prediction score in regression, effectively capturing the structural and feature-based properties of the graph for decision-making.

\noindent\textbf{Link Prediction.} This problem is essential in recommender systems and social networks. Given two nodes $u,v \in \mathcal{V}$, the goal is to predict whether there should be a link between them. Constructing models for this problem is similar to node-level problems except for the objective functions. Regularly, models can be trained via distance-based objectives or contrastive loss functions. 

\section{Explainable Artificial Intelligence}

XAI techniques \cite{dwivedi2023explainable, hassija2024interpreting} offer diverse perspectives for understanding how AI models arrive at their decisions. These techniques can be categorized based on the stage of model construction (Ante-hoc vs. Post-hoc), their scope (Global vs. Local), or the entity they explain (Model-level vs. Instance-level). 

\subsection{Interpretable Models vs. Post-hoc Explanation Methods}

\begin{figure}[ht]
	\centering
	\includegraphics[width=0.9\linewidth]{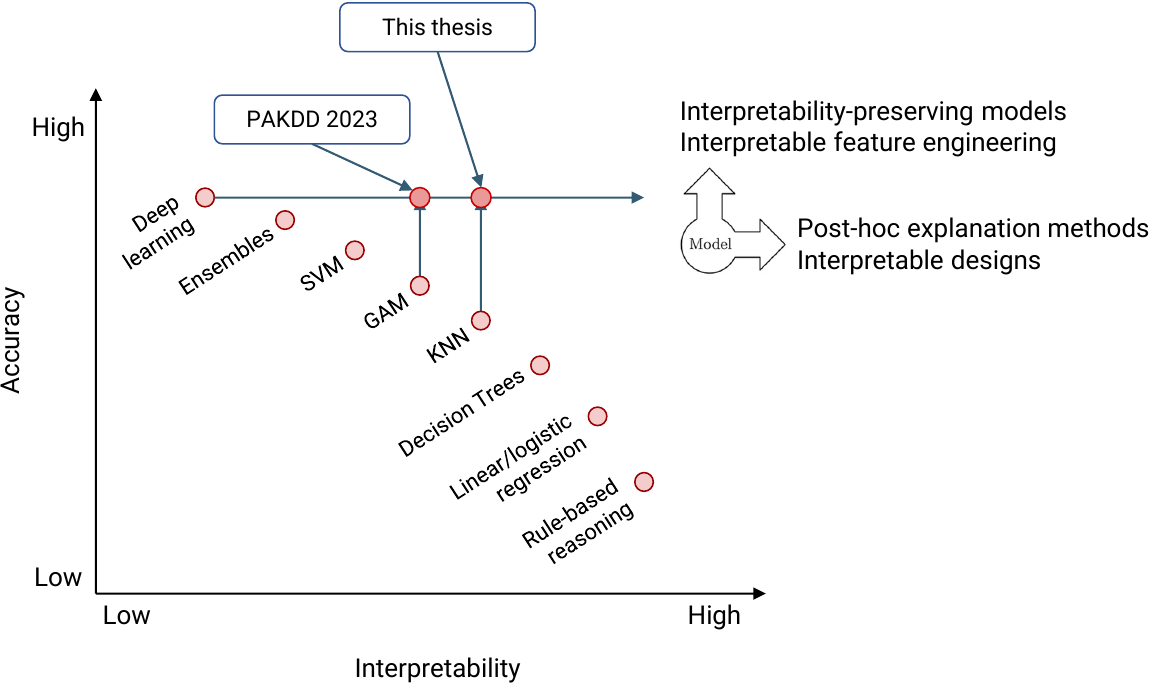}
	\caption{Interpretability vs. Accuracy Trade-off. This figure refers to \cite{nesvijevskaia2021accuracy}. Jang et al., 2023 \cite{jang2023toward} proposed an integration of deep learning and a generalized additive model (GAM).}
	\label{fig:bg_tradeoff}
\end{figure}

A crucial distinction in explainable AI lies between ante-hoc (intrinsically interpretable) and post-hoc approaches based on the stage when interpretable features are considered. The former approach covers a class of models that are inherently designed to be transparent or contain quantitative measurements of feature importance. Representatives of this approach are linear regression and decision tree algorithms. Their simplicity and inherent explainability make them ideal for scenarios requiring high levels of trust and regulatory compliance. However, these models often trade-off model complexity for interpretability, potentially not capturing intricate patterns in complex datasets. On the other hand, post-hoc explanation methods regard training models as black boxes, which usually determine important factors within input-output pairs. Even though the post-hoc approach contains both model-specific and model-agnostic methods, the latter is more prevalent due to its generalizability. This approach is applicable across various models regardless of their internal mechanisms and offers a versatile approach to AI explanations. Methods like SHAP \cite{lundberg2017unified}, LIME \cite{ribeiro2016should}, and InterpretML \cite{nori2019interpretml} make complicated models understandable, although they can be computationally intensive and may offer less precise explanations than those from interpretable models.

\subsection{Global vs. Local and Model-level vs. Instance-level}
Explanation methods in AI can be broadly categorized as model-level or instance-level, often corresponding to global and local explanations respectively.
Model-level explanations provide insight into the inner workings and overall behavior of the AI model itself, offering a comprehensive understanding of its functioning. Global explanation, strongly correlated with the model-level approach, seeks to elucidate the overall behavior and logic of AI models, aiming for a comprehensive understanding of their functioning. In some specific scenarios, this approach aims at providing a global summary of feature importance associated with prediction outcomes. In contrast, local explanation, aligning more with instance-level insights, concentrates on explaining specific decisions or predictions made by the AI model, providing a detailed and focused understanding of individual outcomes. Specifically, instance-level explanations focus on data attributions, examining how individual data points influence specific model predictions, thereby offering a more microscopic view of the model's decision-making process. Explanation methods offer a pathway to demystify the opaqueness of AI models, enabling users to gain a holistic view of the model's rationales or to dissect specific predictions. The choice between these approaches depends on the specific objectives of transparency and understanding, as well as the nature of applications. 

\subsection{Local Explanation Breakdown}

\begin{figure}[ht]
	\centering
	\includegraphics[width=0.7\linewidth]{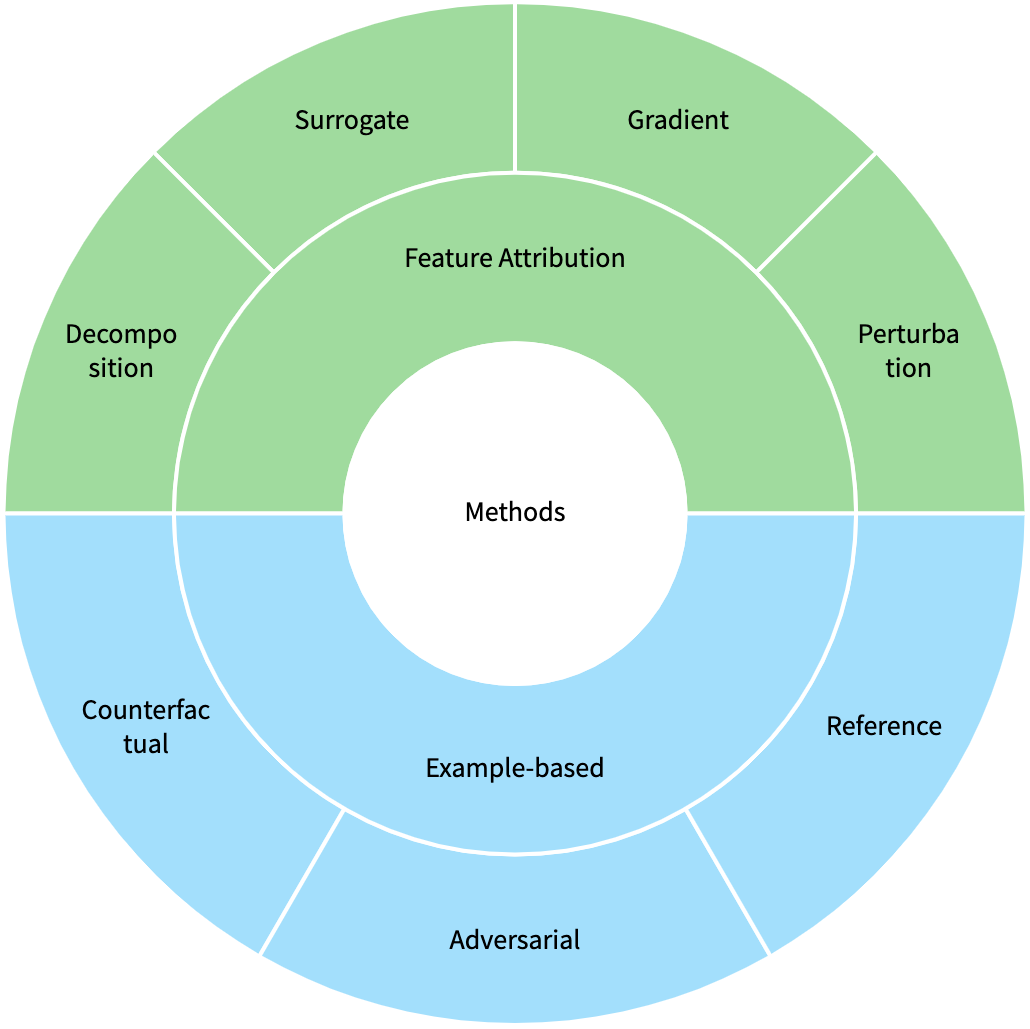}
	\caption{A Breakdown of Local Explanation Methods}
	\label{fig:bg_local}
\end{figure}

Local explanation plays a crucial role in unraveling decision-making processes of AI models for specific data instances. Due to the ubiquity of deep learning models, local explanation gains popularity thanks to its capability of providing interpretable insights into specific factors influencing a model's output for a given input. This focus has led to the development of numerous local explanation techniques \cite{ribeiro2016should, shrikumar2017learning}, which are now widely used across various applications. \cref{fig:bg_local} presents a comprehensive breakdown of these methods.

\noindent\textbf{Feature Attribution.} It offers a way to pinpoint exactly why a model makes certain decisions. These methods focus on determining the importance of each feature in the model's decision-making process. Perturbation techniques, where input features are systematically altered to observe the effect on the model's output, offer insights into feature importance and interaction. Similarly, counterfactual explanations offer valuable insights by exploring how slight modifications to input data can lead to different predictions, thereby highlighting critical decision boundaries. Gradient-based methods \cite{shrikumar2017learning,shapDeep19:online}, often employed in DL models, derive feature importances from computed gradients for input features. In contrast, decomposition-based methods break down model predictions into contributions of individual input features, providing a detailed understanding of feature attributions. Finally, local surrogate methods like LIME \cite{ribeiro2016should} and SHAP \cite{shrikumar2017learning} introduce another approach to measure feature contributions.

\noindent\textbf{Example-based explanation.} Inspired by the human inherent capability of learning from examples, it offers a realistic and intuitive method for understanding AI models through the use of specific data instances. These examples, whether they are influential, representative, or anomalous, offer a tangible perspective into the model's functioning. This approach's main advantage is its intuitive nature; real-world examples are often more understandable than abstract model descriptions. However, they might not comprehensively reflect the model's logic and could be biased towards the selected instances. Influential instances, prototypes, and adversarial examples are typical methods employed in this category.

\section{Human-in-the-loop AI and Human-centric AI}
Both Human-in-the-Loop (HITL) AI \cite{wu2022survey, mosqueira2023human} and Human-centric AI \cite{taylor2023human} are closely related to XAI. In HITL systems, explainability is crucial for enabling effective human interaction and intervention. When humans are involved in the AI decision-making loop, they need clear and comprehensible explanations of the AI's processes and outputs to make informed decisions. In Human-centric AI, explainability contributes to building trust and ensuring that AI systems respect model principles and human values. It enables stakeholders to understand, anticipate, and manage the impacts of AI systems in a way that respects human dignity and agency. Therefore, XAI is a foundational element in both HITL and Human-centric AI, enhancing the effectiveness, ethics, and societal acceptance of AI technologies.

\subsection{Human-in-the-loop AI}
Human-in-the-Loop AI \cite{wu2022survey, mosqueira2023human} refers to a paradigm where human judgment and decision-making are integral parts of an artificial intelligence system. In HITL systems, humans are not just passive recipients of AI-generated outcomes but active participants who guide, evaluate, or modify the AI's decisions. This approach is particularly prevalent in areas where AI's autonomous decision-making is either insufficient or ethically sensitive, such as in medical diagnosis, legal judgments, or complex engineering tasks. By incorporating human expertise and feedback, HITL systems seek to improve accuracy, reliability, and trustworthiness through the strengths of both human intelligence and AI.

\subsection{Human-centric AI}
Human-centric AI \cite{taylor2023human} concentrates on creating AI systems that align with human needs, values, and ethical considerations. It emphasizes designing AI systems that are understandable, fair, and respectful of human autonomy and privacy. In this paradigm, the human experience is central to the development process, ensuring that AI technologies are not just technically proficient but also socially and ethically responsible. Human-centric AI seeks to align AI's capabilities with human goals, emphasizing usability, accessibility, and inclusiveness.

\section{XAI for Graph Neural Works}

\begin{figure}[ht]
	\centering
	\begin{tikzpicture}[
		grow=right,
		sibling distance=1cm,
		every node/.style={
			draw, text width=2.5cm, align=center, rounded corners, thick, draw=DarkBlue, font=\small
		},
		edge from parent path={(\tikzparentnode.south) -- (\tikzchildnode.west)}
		]
		\tikzstyle{level 0}=[level distance=0.75cm]
		\tikzstyle{level 1}=[level distance=2.5cm,sibling distance=3.5cm,edge from parent path={(\tikzparentnode.south) -- ++(0.5,0) |- (\tikzchildnode.west)}]
		\tikzstyle{level 2}=[level distance=5.5cm,edge from parent path={(\tikzparentnode.east) -- ++(0.5,0) |- (\tikzchildnode.west)}]
		%\tikzset{edge from parent/.style={draw,thick,edge from parent fork right}}
		\node [rotate=90,text width=4cm, minimum height=1cm] {XAI Methods for GNNs}
		child {
			node {Interpretable Models}
			child{
				node[minimum width=6.8cm,text width=6.8cm]{
					LPA-GCN \cite{lpa_gcn}, EGNN \cite{li2022egnn}, SEGNN \cite{dai2021towards}, ProtGNN \cite{zhang2022protgnn}, KerGNN \cite{feng2022kergnns}, 
					STExplainer \cite{tang2023explainable}, GDM \cite{nian2024globally}, PGIB \cite{seo2024interpretable}
				}
			}
		}
		child{
			node {Post-hoc Explanation}
			child {
				node[minimum width=6.8cm,text width=6.8cm]{
					\textbf{Model-level Methods} \\
					XGNN \cite{yuan2020xgnn}, PAGE \cite{shin2024page}, GNNInterpreter \cite{wang2022gnninterpreter}, GDM \cite{nian2024globally}
				}
			}
			child {
				node[minimum width=6.8cm,text width=6.8cm]{
					\textbf{Instance-level Methods} \\
					CAM \cite{pope2019explainability}, Grad-CAM \cite{pope2019explainability}, GNNExplainer \cite{ying2019gnnexplainer}, PGExplainer \cite{luo2020parameterized}, GraphMask \cite{schlichtkrull2020interpreting}, SubgraphX \cite{yuan2021explainability}, Excitation BP, GNN-LRP \cite{schnake2020higher}, GraphLIME \cite{huang2022graphlime}, RelEx \cite{zhang2021relex}, PGM-Explainer \cite{vu2020pgm}, ReFine \cite{wang2021towards}, GNES \cite{gao2021gnes}, MATE \cite{spinelli2022meta}}
			}
		};
	\end{tikzpicture}
	\caption{Breakdown XAI methods for GNNs}
	\label{fig:bg_xai_gnn}
\end{figure}
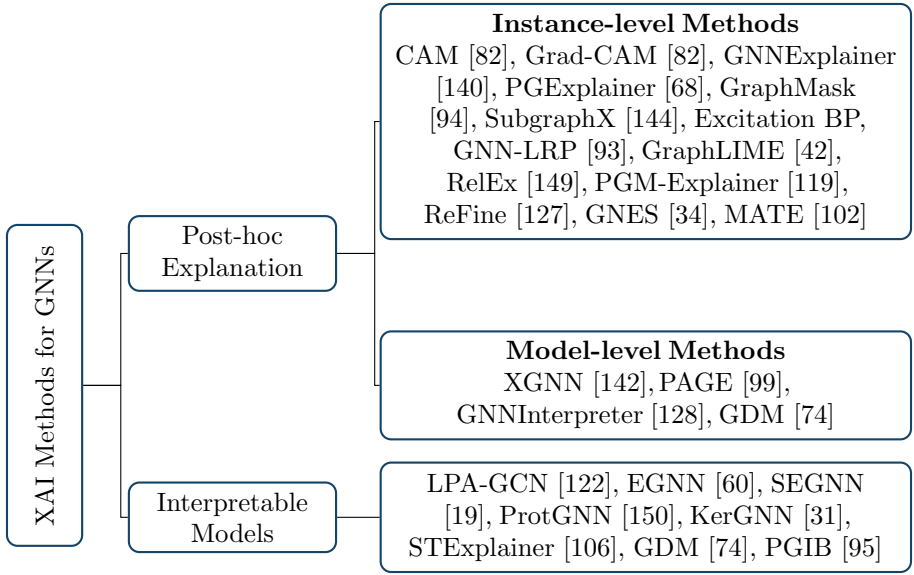

The combination of XAI methods and GNNs represents a promising avenue for research. An outline of the most important elements of explanations in this context is provided in this section.

When it comes to explaining GNN predictions, it is essential to address several questions: why explanations are necessary, what exactly needs to be explained, and for whom and how these explanations should be provided. While recent XAI techniques have explored various facets of explaining GNNs, they often neglect the crucial ``explain to whom" aspect of explanation. In the ``why to explain" context, the primary objectives include enhancing users' comprehension of why specific predictions are generated or facilitating model debugging in cases of model misbehavior. Concerning ``what to explain," explanation methods typically tackle fundamental queries like ``which node features contribute significantly to the prediction?" or ``which patterns have the most substantial impact on prediction scores?"

As presented in \cref{fig:bg_xai_gnn}, explanation techniques for GNNs \cite{yuan2022explainability} can be categorized based on either their granularity or when these explainers are constructed. In terms of granularity, these methodologies fall into two primary categories: instance-level and model-level explanations. Instance-level explanations focus on providing insights into individual input instances, while model-level explanations seek to uncover general patterns that influence predictions across various inputs. Multiple instance-level methods also support model-level explanations by merging groups of instance-level explanations. Given an input graph, explanation methods aim to identify the factors that have the most impact on predictions. In terms of timing, these methods can be divided into two groups: post-hoc methods and intrinsically interpretable models. Post-hoc explainers are developed after the training phase of black-box GNNs has concluded, whereas intrinsically interpretable models are GNN models that are capable of providing explanations based on their trained weights.
	\chapter{Real-time Explanations for GNN Predictions} \label{chap:scale}
\section{Introduction}

%As Graph Neural Networks \cite{zhou2020graph,zhang2020deep} have gained widespread usage in real-world applications, comprehending the rationale behind their predictions has become imperative for evaluating trust, particularly for critical decisions. However, generating explanations for GNN remains challenging for the following reasons. Unlike tabular data, text, and grid-like data, graphs comprise multiple elements, such as graph structures, node features, and edge features. Therefore, identifying the factors that influence a specific prediction can be difficult since it involves intricate interactions among these factors. Additionally, graph datasets are diverse, and each may include a distinct set of components, making it hard to measure their contributions precisely. Furthermore, the complexity of graphs makes it arduous to utilize general XAI tools \cite{ribeiro2016should,lundberg2017unified} to explain GNNs.

As GNNs become increasingly prevalent in practical applications \cite{zhou2020graph,zhang2020deep}, understanding the basis of their predictions is essential for establishing trust, especially in critical decision-making context \cite{rudin2019stop}. However, elucidating GNN predictions presents several challenges because of the following reasons. Different from text, images, and tabular data, graphs are complicated including various elements, such as nodes, edges, and node/edge features. This complexity complicates the identification of the specific factors influencing a given prediction due to the intricate interactions among these elements. Moreover, the diversity of graph datasets, each with unique components, further complicates the precise assessment of their contributions. The inherent complexity of graphs also poses significant difficulties in applying general XAI tools \cite{ribeiro2016should,lundberg2017unified} to GNNs effectively.

%Recently, several approaches have been proposed for explaining GNNs \cite{yuan2022explainability}, highlighting the lack of a single, universal solution within XAI. These methods address the problem from diverse perspectives but mainly focus on structure analyses while neglecting feature attributions. Additionally, most existing approaches belong to the post-hoc explanation category, concentrating on instance-level explanations \cite{yuan2022explainability}. Among them, perturbation methods like GNNExplainer \cite{ying2019gnnexplainer} and PGExplainer \cite{luo2020parameterized} have gained widespread acceptance, primarily because they introduce benchmark datasets for GNN explanation tasks and exhibit exceptional performance. However, these methods necessitate training explanation models for target instances after training black-box GNNs, resulting in additional computational resources and execution time. In contrast, intrinsically interpretable (self-explainable) models such as GAT \cite{velickovic2017graph} can offer explanations for predictions instantaneously. The advantage of this approach inspires the development of novel architectures \cite{dai2021towards,zhang2022protgnn} that leverage structure similarity approaches. These methods often utilize citation graphs \cite{dgldata} for evaluation, and their generalizability to other datasets \cite{palowitch2022graphworld} remains questionable. Additionally, these similarity functions can be computationally intensive in several scenarios, particularly for complex graphs. 

Lately, numerous XAI methods have been introduced for GNNs \cite{yuan2022explainability}, underscoring the absence of a universal solution in this field. These approaches tackle the problem from various angles but predominantly perform structure analyses while often neglecting feature contributions. They fall under the post-hoc approach, particularly instance-level explanations. Perturbation-based techniques \cite{ying2019gnnexplainer, luo2020parameterized} have become increasingly popular due to the initial benchmark datasets for explanation tasks and noticeable results. Nonetheless, explainers necessitate post-hoc training processes, leading to increased computational costs and execution time. In contrast, interpretable models \cite{velickovic2017graph, lpa_gcn} can provide immediate explanations for predictions. This benefit has motivated the development of new architectures \cite{dai2021towards,zhang2022protgnn} that exploit structural similarity approaches. These methods frequently use citation graphs \cite{dgldata} for evaluation, but their generalizability to other datasets \cite{palowitch2022graphworld} is still questionable. Additionally, the employed similarity functions can be computationally demanding, particularly for complex graphs.

%A promising but challenging research problem in the field of XAI is training explainers with a black-box GNN, where each explainer learns to analyze a subset of interactions within an input graph. This approach enables explainers to be as general as post-hoc explanation methods and as rapid as self-explainable models in providing explanations. However, designing an effective training paradigm that allows explainable models to achieve performance equivalent to the black box is not a trivial problem. Explainable models can perform poorly in a standalone training mode, resulting in untrustworthy explanations. One potential solution to this problem is online knowledge distillation \cite{gou2021knowledge}, where a black-box GNN is treated as a teacher, and the explainable models are considered students. The teacher guides the learning processes of students through distillation losses. 

A promising yet difficult research initiative in XAI involves constructing explainers concurrently with the target GNN. In this scenario, an explainer is assigned to analyze a particular type of interactions contained within a computation graph. This method allows explainers to possess the generalizability of post-hoc techniques and simultaneously match the rapid explanatory performance of interpretable models. Nonetheless, designing an efficient training paradigm enabling interpretable components to match the accuracy of their black-box counterparts presents considerable challenges. When trained independently, explainable models may exhibit subpar performance, leading to unreliable explanations. A possible remedy for this issue is training interpretable components and the target black box together based on online knowledge distillation \cite{gou2021knowledge}. Specifically, interpretable components are students, while the target black box acts as a teacher in this paradigm. Moreover, the target GNN provides an additional constraint via its predictive distributions to control the learning process of interpretable components.

\begin{figure}[ht]
	\centering
	\subfloat[Relationships of Approaches]{
		\includegraphics[width=0.46\linewidth]{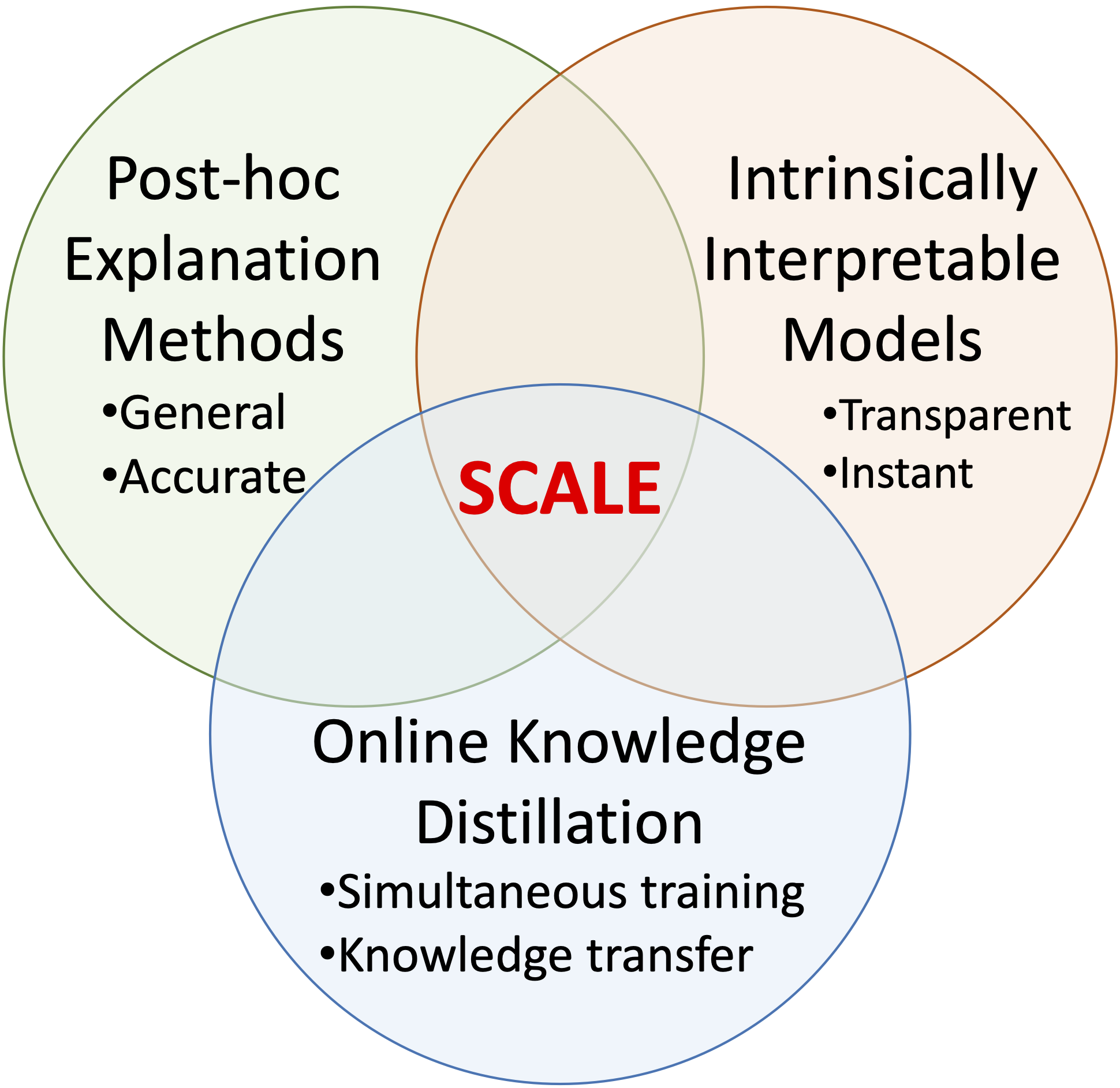}
		\label{fig:venn_SCALE}
	}
	\hspace{-0.3cm}
	\hfill
	\subfloat[Positioning SCALE Against Others]{
		\includegraphics[width=0.46\linewidth]{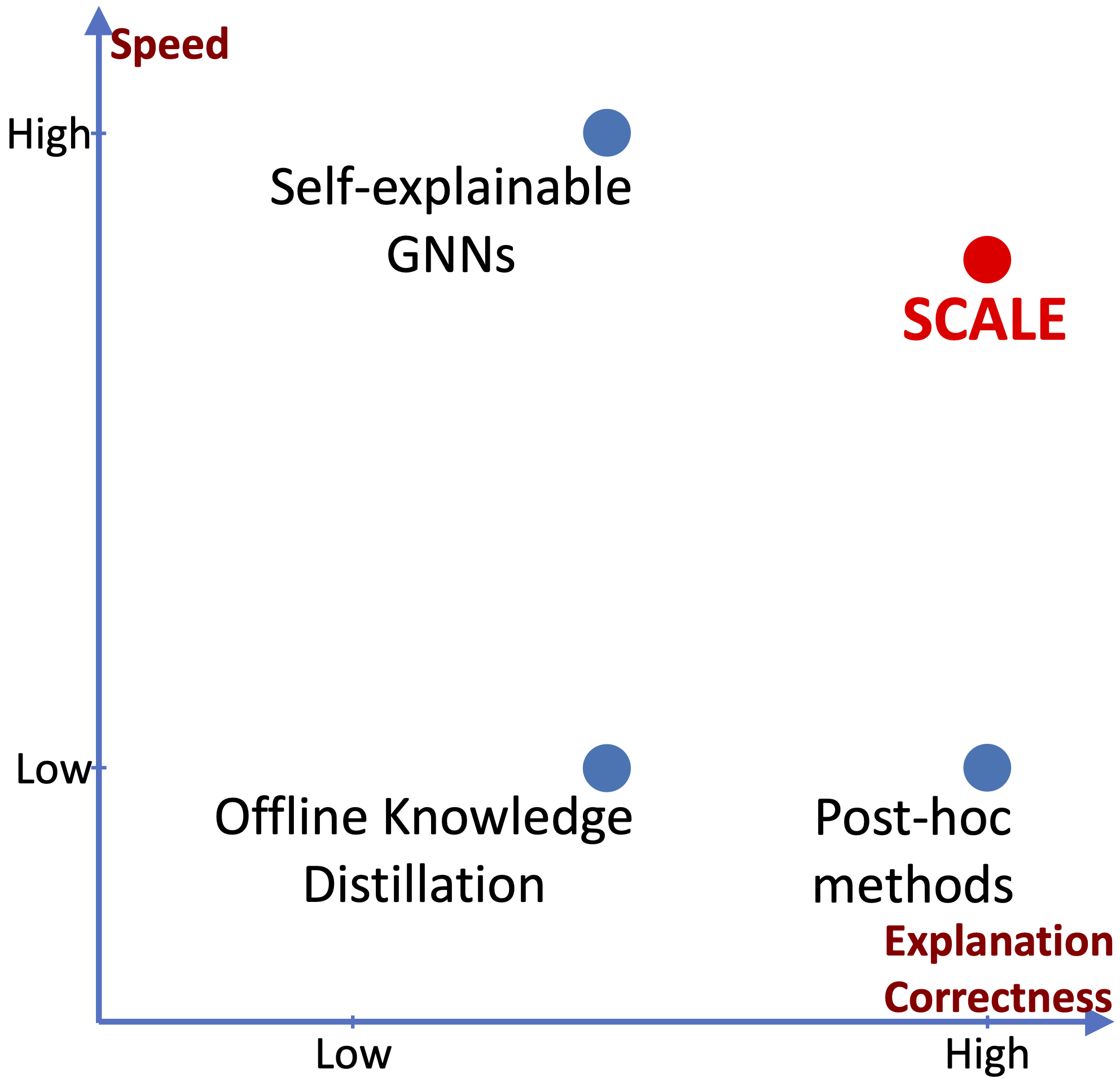}
		\label{fig:position}
	}
%	\caption{SCALE represents the most significant attributes of post-hoc explanation methods and intrinsically interpretable GNN models. Through online knowledge distillation \cite{gou2021knowledge}, SCALE trains multiple specialty learners simultaneously with a black-box GNN, resulting in a framework that offers precise and prompt explanations for GNN predictions. Therefore, it has the capacity to instantly produce highly accurate explanations of GNN predictions.}
	\caption{The proposed framework embodies the benefits of knowledge distillation, interpretable models, and post-hoc methods. Building upon \cite{gou2021knowledge}, the framework constructs and trains specialized interpretable components in parallel with a target black box. This approach allows SCALE to generate correct and rapid explanations for GNN predictions}
	
\end{figure}

%This chapter introduces SCALE, the first explanation framework training multiple \textbf{S}pe\textbf{C}i\textbf{A}lty \textbf{L}earners to \textbf{E}xplain GNNs. As depicted in \cref{fig:venn_SCALE}, SCALE combines the advantages of post-hoc explanation methods, intrinsically interpretable models, and online knowledge distillation. Similar to previous works \cite{ying2019gnnexplainer, luo2020parameterized}, the proposed framework aims to identify the key factors from graph structures, node features, and edge features that contribute the most to predictions. Here, edge features are discarded for simplicity. However, due to complex interactions among these components, creating a single explainer to analyze the attributions takes time and effort. Therefore, SCALE trains multiple specialty learners, each focusing on different aspects of a prediction. A specialty learner is a model that learns from a black-box GNN to capture a subset of interactions. As shown in \cref{fig:overview_a}, SCALE trains structural importance and feature transformation learners simultaneously with a black-box GNN using an online knowledge distillation paradigm. This paradigm enables the black-box GNN to guide the learners to achieve comparable predictive performance. 

This chapter introduces SCALE, a novel XAI framework designed to elucidate a GNN's predictions by constructing multiple interpretable components concurrently with the target GNN. As illustrated in \cref{fig:venn_SCALE}, the proposed framework inherits the significant features of knowledge distillation, interpretable models, and post-hoc methods. Similar to previous studies \cite{ying2019gnnexplainer, luo2020parameterized}, this framework aims to identify influential elements that significantly drive outcomes from node/edge features and networks. Given the complex interactions among graph elements, designing a unique explainer to identify influences presents a significant challenge. Consequently, this work proposes to break the original explanation problem into solvable sub-problems, where each corresponds to determine contributions of only a subset of elements. Particularly, it designs and trains several interpretable components or specialized learners in parallel with a target GNN via an online knowledge distillation approach, where each learner is tasked to focus on only a specific aspect of the target model. This approach allows the target black box to assist interpretable components to obtain a high level of predictive performance while being able to provide instant explanations. For simplicity, the proposed framework employs two learners: feature transformation and structural importance, as depicted in \cref{fig:overview_a}. 

%During the explanation phase, instant explainers output predictions with explanations at various levels using specialized procedures. First, learnable masks of a structural importance learner are employed to eliminate unimportant edges and nodes for a graph-level prediction. Second, SCALE customizes the random walk with restart (RWR) algorithm, which has been widely applied to recommender systems and search engines \cite{brin1998anatomy,chiang2013exploring,park2017comparative,wang2020personalized}, to provide structural explanations for node-level predictions. Here, a target node that requires an explanation is considered the query vertex. As the random walker travels through neighboring nodes and edges, it includes them in the explanation. Additionally, DeepLIFT \cite{shrikumar2017learning,shapDeep19:online} is utilized in the feature attribution module on top of a trained multilayer perceptron (MLP) learner to provide instance-level feature attributions and overall attribution summaries. DeepLIFT is selected due to its efficiency in measuring feature attributions in deep neural networks.

At inference, interpretable components infer predictions and generate various explanations through specialized algorithms. Initially, a trained mask of a structural importance learner is used to discard insignificant edges for graph classifications. Next, SCALE adapts the random walk with restart algorithm, commonly utilized in recommender systems and search engines \cite{brin1998anatomy,chiang2013exploring,park2017comparative,wang2020personalized}, to generate explanatory graphs for node classification problems. In this algorithm, an explained node is treated as the start vertex. As neighbors are visited by the random walker, they are incorporated into the explanation. Furthermore, the feature attribution module leverages DeepLIFT \cite{shrikumar2017learning,shapDeep19:online} on a trained MLP to generate summaries of feature contributions and specific attributions of an individual prediction. This library is chosen for its effectiveness and high performance in measuring feature attributions in DNNs.

%Evaluating the efficiency of GNN explanation methods is difficult due to the absence of ground-truth explanations in most graph datasets. The approach proposed by Ying et al. \cite{ying2019gnnexplainer} and Luo et al. \cite{luo2020parameterized} is adopted for quantitative evaluation. Specifically, it formulates structural explanations as binary classification tasks by including influential nodes and edges in explanations. Then, SCALE is compared with baselines on the correctness of structural explanations and execution performance, both quantitatively and qualitatively. Additionally, a user study is conducted to assess the usefulness of explanations and their influences on the decision-making processes. Next, the feature attribution module's efficiency is evaluated via a real-world dataset with intelligible features. Since ground-truth information is unavailable, the results of this analysis are compared with a state-of-the-art method (Zhang et al., 2020 \cite{zhang2020gcn}) for evaluation purposes. Furthermore, a series of ablation studies are conducted to gain a better understanding of the algorithms in SCALE. Extensive experiments and analyses conclude that SCALE is highly effective in both explanation capability and execution performance.

Assessing the effectiveness of XAI methods presents significant challenges since ground-truth explanations are usually unavailable. In the graph domain, Ying et al. \cite{ying2019gnnexplainer} and Luo et al. \cite{luo2020parameterized} proposed to generate synthetic data with predefined ground-truth patterns, which are then utilized for quantitative evaluations. This approach conceptualizes the construction of explanatory graphs as binary classification tasks by predicting which edges should be retained. The proposed framework employs this approach for comparative evaluations against baselines regarding the accuracy of explanatory graphs and computation time. Moreover, this research includes a user study that aims to determine explanation usefulness and how explanations aid in user comprehension of model behavior. The feature attribution module is then assessed with a data mining task utilizing a practical dataset consisting of understandable features. In the absence of ground-truth data, the outcomes of this analysis are compared with those from a leading research (Zhang et al., 2020 \cite{zhang2020gcn}) for validation. Additionally, ablation studies are performed to enhance the understanding of the proposed algorithms. Comprehensive experiments and analyses reveal that the proposed framework excels in both explanatory power and execution time.

The research presented in this chapter, including the proposed method and experimental results, was published in \cite{bui2023generating}. This work's contributions are detailed as follows:
\begin{itemize}
%	\item SCALE leverages multiple specialty learners to provide instant, accurate explanations for message-passing-based GNN architectures. 

\item The proposed framework utilizes multiple specialty learners to generate immediate and precise explanations for message-passing-based GNNs.

%	\item The proposed framework demonstrates versatility and efficiency. It offers rapid explanations for input graphs, highlighting its suitability for practical applications.

\item It is both versatile and efficient, providing quick explanations for input graphs, which underscores its practicality.
	
%	\item This research pioneers the application of RWR for node-level prediction explanations. This innovation enables the identification of distinct contributions of neighbors to a node's output.

\item This work is the first to apply the random walk with restart algorithm to explain node-level predictions, providing neighbors' contributions to a node's prediction.

%	\item Extensive experimentation and ablation studies validate the effectiveness and efficiency of the proposed framework. The results demonstrate SCALE's superior explanation capability and execution performance compared to existing approaches.

\item Comprehensive experiments and analyses confirm the efficiency and effectiveness of the framework, showcasing its superior explanatory capabilities and execution performance in comparison to existing methods.

\end{itemize}

%Here is the structure of this chapter. \cref{related_work} provides an overview of related work in the field of GNN explainability. \cref{method} delves into the detailed methodology of the proposed framework. \cref{exp_setups} outlines the experimental setups, and \cref{exp_results} presents the experimental findings. \cref{prototype} demonstrates a system prototype to illustrate the framework's practical applications. A comprehensive discussion in \cref{discussion} covers limitations, potential for improvement, and broader applications of the research. Finally, \cref{conclusion_part} concludes the chapter and presents avenues for future research.

This chapter is structured as follows: \cref{related_work} provides a comprehensive overview of related work. \cref{method} details the methodology. \cref{exp_setups} describes experimental setups, while \cref{exp_results} presents the findings from these experiments. In \cref{prototype}, a system prototype is demonstrated to showcase the practical applications of the framework. The discussion in \cref{discussion} addresses the limitations, potential improvements, and broader implications of the research. Finally, \cref{conclusion_part} concludes the chapter and suggests directions for future research.

\section{Related Work} \label{related_work}
\subsection{Post-hoc Explanation Methods}

Most current XAI methods for GNNs operate post-hoc, regarding GNNs as black boxes. Yuan et al. \cite{yuan2022explainability} recently conducted an extensive study of these methods. A significant focus within this field is an instance-level approach, which encompasses four primary categories: surrogate, decomposition, gradient-based, and perturbation. While the first three categories modify current XAI methods for GNNs \cite{baldassarre2019explainability, schnake2020higher}, perturbation methods, initially proposed by Ying et al. \cite{ying2019gnnexplainer}, remain a dynamic area of research, with multiple following research papers \cite{luo2020parameterized, schlichtkrull2020interpreting, yuan2021explainability} contributing to the field. These methods aim to extract critical graph patterns using either edge pruning or MCTS \cite{swiechowski2023monte} algorithms. However, they frequently suffer from overfitting due to the extensive size of perturbed samples and often neglect feature attributions. Furthermore, post-hoc methods are unable to provide immediate explanations for GNNs' predictions due to the computational burden. Conversely, SCALE incorporates a special training paradigm consisting of multiple specialized learners and a black-box teacher, delivering precise and real-time explanations without necessitating retraining. Moreover, post-hoc methods often utilize K-hop sampling to convert node-level tasks into graph-level ones, which may be inefficient when dealing with graphs including small cycles. In contrast, SCALE employs distinct algorithms to elucidate both node-level and graph-level predictions.

\subsection{Self-explainable Graph Neural Networks}
%Intrinsically interpretable (or self-explainable) models offer a promising solution to the performance problems of post-hoc explanation methods. These models can provide explanations instantly based on their trained weights, avoiding the need for additional computation. Graph Attention Network (GAT) \cite{velickovic2017graph} is an example of a self-explainable model that uses attention matrices to construct explanations. GCN-LPA \cite{wang2020unifying} integrates label propagation into GCN \cite{kipf2016semi} and replaces the normalized adjacency matrix with a learnable matrix, offering a novel approach to self-explanation. Lately, two novel models based on similarity functions \cite{dai2021towards,zhang2022protgnn} have been proposed. Zhang et al. \cite{zhang2022protgnn} model the feature and label similarity in GNN execution to extract prototypes for explanations.
% However, these methods are optimized for node classification datasets with the homophily property and may not perform well on datasets without this property, leading to less trustworthy explanations. Additionally, the explanation procedures in these models are either not discussed or too simplistic for general datasets. Even though \cite{zhang2022protgnn} can explain GNNs using prototypes extracted during training, it is significantly slow. In contrast, the proposed method aims to offer instant explanations and can be applied to all message-passing GNN architectures, regardless of the datasets' properties.

Models that are inherently interpretable, also known as self-explainable, present a promising alternative to the performance issues associated with post-hoc explanation methods. They can generate explanations immediately using their interpretable elements, eliminating additional computational steps. For instance, GAT \cite{velickovic2017graph} exemplifies an interpretable model by utilizing attention matrices to produce explanations. Another example is the GCN-LPA \cite{wang2020unifying}, which employs a trainable adjacency matrix to capture the propagation flow of labels and features across nodes, thereby providing an alternative approach. Recently, similarity-based GNNs have been introduced \cite{dai2021towards,zhang2022protgnn}. In their work, Dai et al. \cite{dai2021towards} model the similarity of labels and features during the execution of GNNs to derive prototypes for explanations.
Nonetheless, this approach is tailored for graph data exhibiting the homophily property and may underperform on datasets lacking this attribute, thus generating less reliable explanations. Furthermore, the explanation mechanisms are either not thoroughly discussed or overly simplistic for broader datasets. Although the model proposed by \cite{zhang2022protgnn} can elucidate GNNs through prototypes extracted during training, it suffers from substantial computation load due to the subgraph discovery procedure. Conversely, the proposed method in this work seeks to deliver immediate explanations and is applicable to diverse message-passing architectures, irrespective of the graph characteristics.

\subsection{Knowledge Distillation Methods for GNNs}
%Knowledge distillation (KD) \cite{gou2021knowledge}, initially introduced as a model compression technique \cite{hinton2015distilling}, has become increasingly popular for developing interpretable student models \cite{alharbi2021learning}. Various KD approaches have been proposed for GNNs to construct smaller models that outperform pre-trained teacher models in prediction accuracy \cite{deng2021graph,joshi2021representation,yang2021extract,li2022egnn}. 
%However, as graph-free KD \cite{deng2021graph} is expensive, most of these methods typically use the same datasets as the original model. Even though a few approaches for explaining GNNs have been proposed \cite{yang2021extract,li2022egnn}, they are primarily optimized to obtain high predictive accuracy on datasets with the homophily phenomenon. Additionally, their explanation solutions are too straightforward, making it challenging to achieve significant results on other datasets \cite{ying2019gnnexplainer,luo2020parameterized}. Furthermore, all methods implement offline knowledge distillation paradigms that require additional computational resources. 

Since its introduction as a model compression method, knowledge distillation (KD) \cite{hinton2015distilling} has gained significant traction for constructing interpretable models \cite{alharbi2021learning}. Different KD strategies have been developed for GNNs to build more compact models that surpass the predictive power of teacher ones \cite{deng2021graph,joshi2021representation,yang2021extract,li2022egnn}. Due to the high cost associated with graph-free KD \cite{deng2021graph}, most models are usually trained with the same datasets as the original ones. Despite the proposal of several techniques for elucidating GNNs' predictions \cite{yang2021extract,li2022egnn}, these methods primarily focus on improving predictive performance while neglecting the importance of explanation construction. Their explanation mechanisms are often overly simplistic, posing challenges for delivering meaningful results on different datasets \cite{ying2019gnnexplainer,luo2020parameterized}. Moreover, current methods employ offline knowledge distillation paradigms, causing extra costs and increasing execution latency.

%This chapter presents an online knowledge distillation paradigm for training specialty learners to explain GNNs. The proposed framework includes multiple algorithms that provide node-level and graph-level explanations from different perspectives, making it more effective than existing methods. Additionally, this approach does not require additional computational resources and can provide high-quality explanations for different datasets. 

This chapter introduces an online knowledge distillation algorithm designed to train specialized learners to explain GNNs. The proposed framework incorporates distinct algorithms to generate feature attributions and explanatory graphs for node and graph classifications from various angles, differentiating itself from existing methods. Furthermore, this approach eliminates extra computational load while achieving high-quality explanations across datasets.

\section{Research Approach} \label{method}

\subsection{Problem Formulation}
%A GNN predicts $\hat{y}$ by operating on a trained model $f$ and an input graph $G_c$, called the computation graph. To explain $\hat{y}$, analysis can focus on $G_c$ by identifying essential subgraphs and crucial node features. For simplicity, edge features are ignored here. Formally, SCALE constructs an explanation for a prediction $\hat{y}$ as ($G_s, \Phi_x$), where $G_s$ is a subgraph of $G_c$ and $\Phi_x = \{\phi_1, \phi_2,...,\phi_d\}$ represents the contribution of node features to the prediction $\hat{y}$. Additionally, $G_s$'s edges contain importance scores corresponding to their contribution to $\hat{y}$. 

A Graph Neural Network outputs an outcome $\hat{y}$ via a trainable projector $f$ and a given input graph $G_c$. To explain its decision for $\hat{y}$, an analysis is conducted on $G_c$ to determine influential patterns and significant features. To simplify the process, this work discards features corresponding to edges. Formally, the proposed framework generates an explanation for the outcome $\hat{y}$ as ($G_s, \Phi_x$), wherein $G_s$ is a substructure of $G_c$ and $\Phi_x = \{\phi_1, \phi_2,...,\phi_d\}$ denotes features' contributions to the outcome $\hat{y}$. Moreover, each edge in $G_s$ carries a real value that indicates the quantitative influence of this edge on $\hat{y}$.

\subsection{Framework Overview}

\begin{figure}[h!]
	\centering
	\subfloat[Model Training]{
		\includegraphics[width=0.9\linewidth]{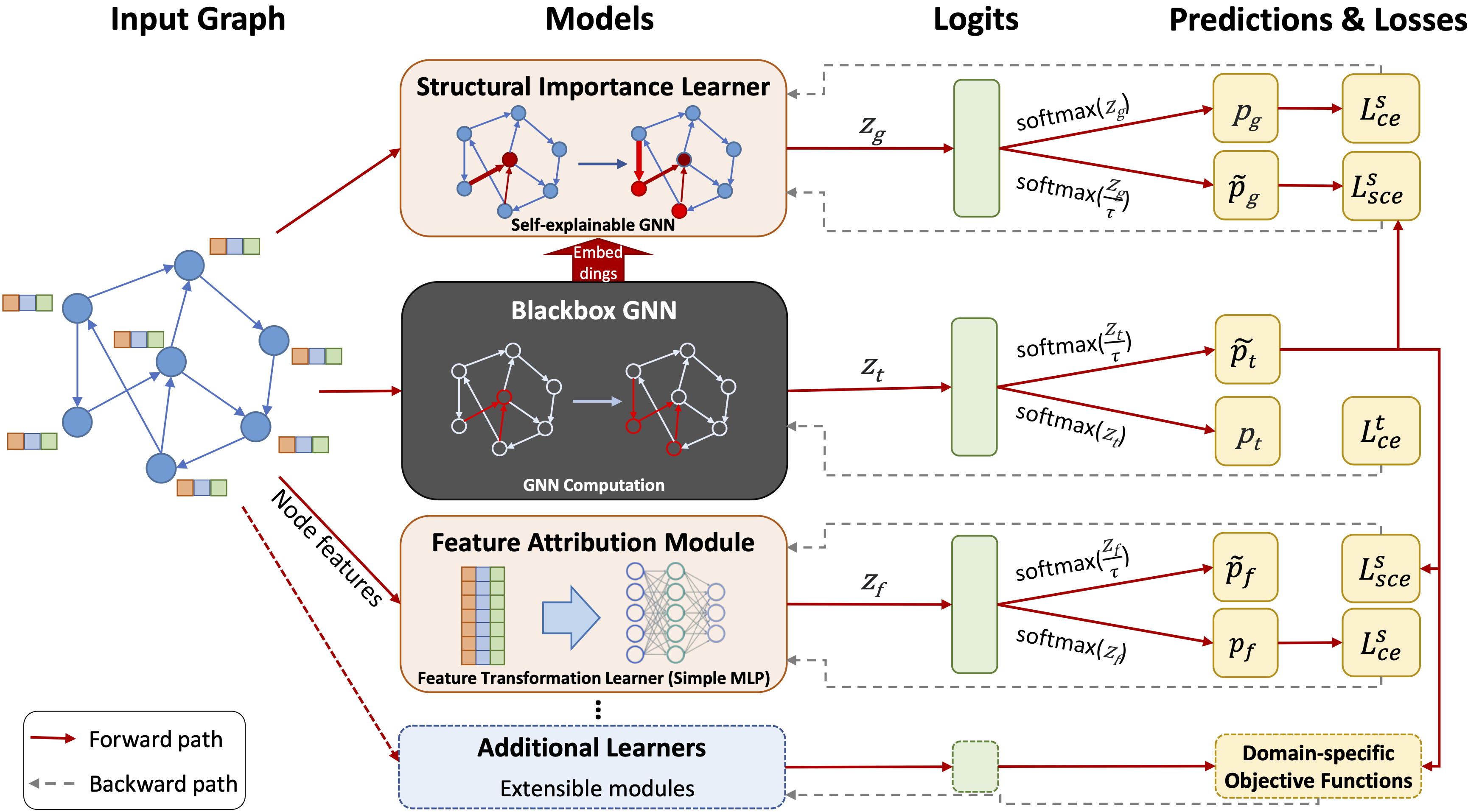}   \label{fig:overview_a}
	}
	\vspace{-0.4cm}
	\hfil
	\subfloat{
		\centering
		\resizebox{0.7\linewidth}{0.5cm}{%
			\begin{tikzpicture}
				\draw[-] (5,0) -- (20,0) node[]{};
			\end{tikzpicture}%
		}
	}
	\vspace{-0.4cm}
	\hfil
	\setcounter{subfigure}{1}
	\subfloat[Inference \& Explanation]{
		\includegraphics[width=0.7\linewidth]{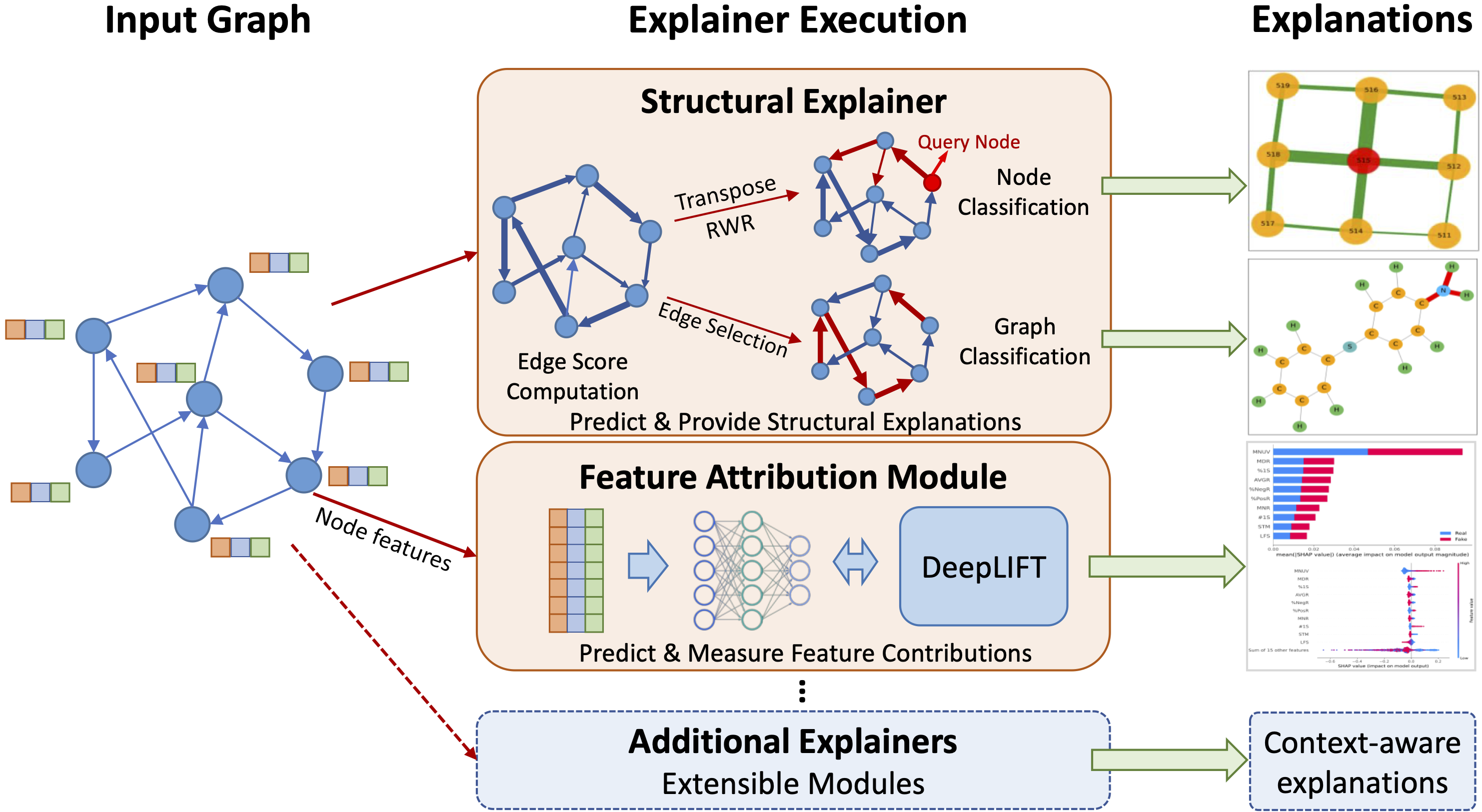}
		\label{fig:overview_b}
	}
%	\caption{An illustration of SCALE Architecture. The upper figure demonstrates the training process of SCALE, while the bottom one presents how SCALE provides explanations based on trained learners. Additional learners and explainers can be implemented depending on particular requirements.}
	\caption{Framework Overview. The upper portion of the paradigm demonstrates the training procedure, whereas the bottom part shows how explanations and predictions are inferred. Depending on specific needs, additional learners and explainers can be incorporated.}
	\label{fig:overview}
\end{figure}
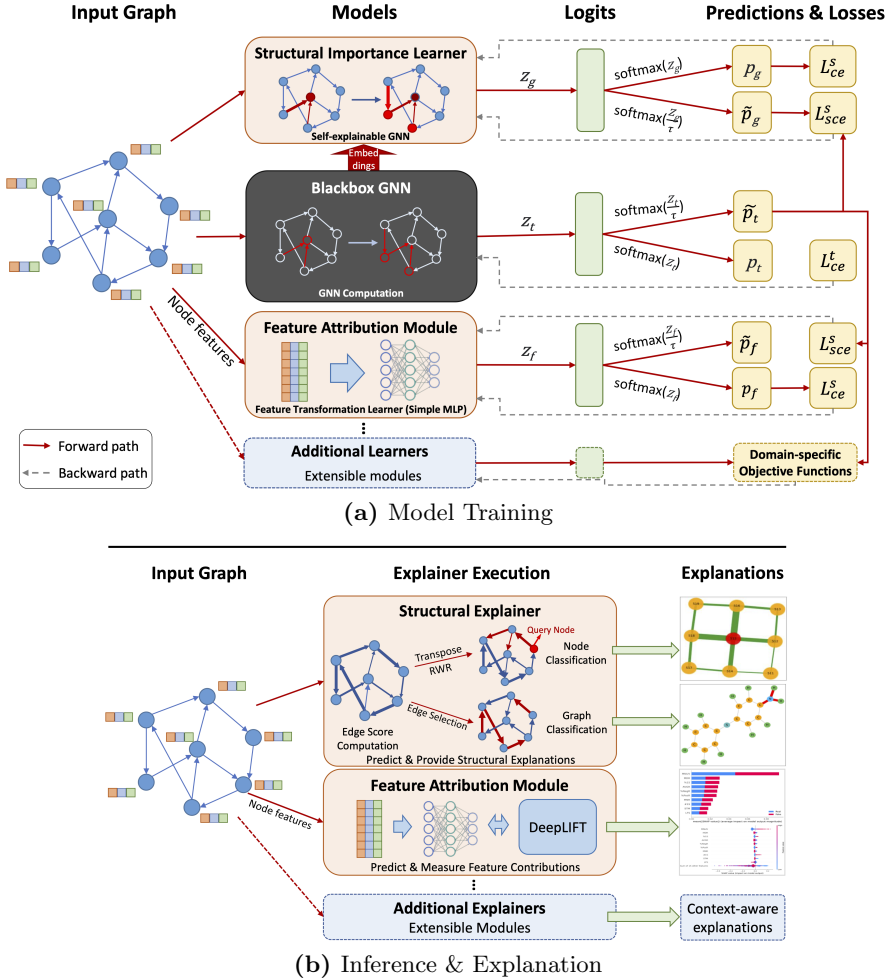

%SCALE is developed based on observations of existing GNN explanation methods. First, most post-hoc explanation methods aim at finding important factors that affect particular predictions from graph structures and node features. Therefore, they design explainable models based on prior knowledge of a pre-trained GNN, such as node embeddings or predicted labels. However, post-hoc computations hinder them from providing instant explanations. Even though intrinsically interpretable models can provide instant explanations without retraining, they are mainly based on model-specific approaches evaluated with simple datasets, thus lacking generalization to broader scenarios. Moreover, all methods mainly concentrate on structural explanations while overlooking feature attributions.

The proposed framework is developed by analyzing current XAI methods for GNNs. Initially, most post-hoc methods seek to identify significant graph patterns driving specific predictions derived from graph structures. Consequently, these methods create explainable models utilizing pre-existing knowledge from a trained GNN, such as predicted labels or node embeddings. Nonetheless, the burden of post-hoc computations prevents them from offering immediate explanations. Interpretable models are usually model-specific methodologies, even though they can deliver instant explanations without the need for retraining. Therefore, the generalizability and adaptability to diverse scenarios of these approaches remain questionable. Additionally, all these approaches predominantly focus on structure analyses, while neglecting feature contributions.

%SCALE is a model-agnostic framework that offers real-time interpretability for a range of GNN architectures without requiring the retraining of explainers. It functions as an adapter with multiple transformation engines that enable message-passing-based GNNs to be converted into explainable versions. The framework leverages an online knowledge distillation paradigm to train multiple specialty learners concurrently with a black-box GNN, enabling them to explain different aspects of an input graph. After training, the system executes multiple instant explainers corresponding to trained learners to provide real-time predictions and explanations. For clarity, here are brief descriptions of learners and explainers.

SCALE is a model-agnostic framework designed to provide instant explanations across various GNNs without requiring the explainers to be retrained. Acting as an adapter, it incorporates various transformation engines to produce explainable versions of message-passing-based architectures. The framework employs a special training approach based on online knowledge distillation to enable a black-box GNN to share its knowledge with multiple specialty learners. This approach allows these learners to focus on different interactions within graphs. During inference, the system deploys several instant explainers associated with the trained learners, delivering instant interpretable predictions.  For clarification, definitions are provided below.

% paraphrased
\begin{itemize}
	\item \textbf{Learner}: A learner is a trained agent that, during training, is assisted by a GNN teacher to imitate specific interactions within the teacher model.
	\item \textbf{Explainer}: An explainer represents an agent that generates explanations for individual predictions derived from a learner.
\end{itemize}

The subsequent subsections discuss the training and inference procedures and the computational complexity analysis.

\subsection{Training Algorithm}
%Existing explanation models, such as those proposed in \cite{ying2019gnnexplainer,luo2020parameterized}, are students trained with hard labels provided by the original GNNs. However, soft targets provide significantly more information in training than hard versions due to their higher entropy values \cite{hinton2015distilling,cho2019efficacy}. As a result, they can help student networks learn better by reducing overfitting to training data \cite{shen2021label} and being more robust to adversarial noise \cite{zhou2021rethinking}. Motivated by this advantage, this work adopts an online knowledge distillation paradigm using \cite{hinton2015distilling} to train black-box GNNs and learners concurrently.

The separation of explainers and learners in post-hoc XAI methods for GNNs \cite{ying2019gnnexplainer, luo2020parameterized} is not obvious since learners are trained after the target GNN (the one that needs explanations). Specifically, the model's predictions are utilized as ``hard'' labels to train explanation learners. Nonetheless, ``soft'' labels are considerably more informative in the training process compared to the hard versions since they provide higher entropy values \cite{hinton2015distilling, cho2019efficacy} compared to the hard ones. Consequently, they aid learners in training more effectively by diminishing overfitting to the training data \cite{shen2021label} and enhancing robustness against adversarial noise \cite{zhou2021rethinking}. Inspired by this benefit, this work employs an online knowledge distillation approach, as outlined by \cite{hinton2015distilling}, to simultaneously train the target GNN and their corresponding learners.

%A hypothesis is that explanations of a black-box model can be viewed from different perspectives, each requiring a different level of detail. To capture these distinct aspects, the proposed approach trains multiple learners that are subsequently used to construct explainers. For simplicity, this work implements two learners: one for structural explanations and the other for node feature attributions. As shown in \cref{fig:overview_a}, these learners are trained together with a black-box GNN model. In this training scheme, the learners act as students, while the black-box GNN serves as the teacher to guide their learning. To ensure the students' computations do not influence the teacher's performance during training, their computational paths are separated from the teacher's. The teacher model is trained with a cross-entropy loss function as follows:

One hypothesis is that explanations of a model's predictions can be interpreted in multiple ways and at varying levels of detail. To address these diverse facets, the proposed framework incorporates multiple interpretable components tailored for specific purposes, which are employed to generate explanations at inference. As illustrated in \cref{fig:overview_a}, the framework employs a special training paradigm based on online knowledge distillation. This approach enables a black-box GNN model (teacher) to impart predictive knowledge to the learners (students).  To simplify the process, this work implements two types of learners: one specializes in structure analysis, while the other assesses the contributions of node features. During training, the backward computations of learners are isolated from those of the teacher to prevent them from influencing the teacher's performance. The teacher GNN is trained using a cross-entropy loss function, as detailed below:

\begin{equation}
	\mathcal{L}_{ce}^t = -\frac{1}{N} \sum_{i=1}^N y_i \cdot \textrm{log}(\textrm{softmax}(z^t_i)),
	\label{eq:cross_entropy}
\end{equation} % para
where $N$ represents the size of the training set, $y_i$ denotes the $i^{th}$ label vector, and $z^t_i$ indicates the $i^{th}$ predictive vector. Learners are trained with the following objective function.

\begin{equation}
	\mathcal{L}^s = \mathcal{L}_{ce}^s + \lambda \mathcal{L}_{sce}^s,
	\label{combined_student_loss}
\end{equation} % para
where $\mathcal{L}_{sce}^s$ denotes the soft cross entropy objective, while $\mathcal{L}_{ce}^s$ is the same as one in \cref{eq:cross_entropy}. Based on practical results, soft cross-entropy is preferred over KL divergence. The amount of distilled knowledge is controlled by $\lambda$. $\mathcal{L}_{ce}^s$ and $\mathcal{L}_{sce}^s$ are detailed in the following equation. 

\begin{equation}
	\begin{aligned}
		\mathcal{L}_{ce}^s &= -\frac{1}{N} \sum_{i=1}^{N} y_i \cdot \textrm{log}(\textrm{softmax}(z^s_i)), \\
		\mathcal{L}_{sce}^s &= -\frac{1}{N} \sum_{i=1}^N \textrm{softmax}(\frac{z^t_i}{\tau}) \cdot \textrm{log}(\textrm{softmax}(\frac{z^s_i}{\tau})),
	\end{aligned}
\end{equation} % para
where $z^s_i$ denotes the $i^{th}$ predictive distribution of a student model, while $\tau$ represents a temperature term scaling predictive distributions.

%The weights of the black-box GNN change over time, causing instability in node embeddings and probability distributions. This phenomenon can make it challenging for student models to mimic the behaviors of their teacher. However, batch normalization \cite{ioffe2015batch} operators can alleviate this issue. First, prior work \cite{ioffe2015batch} demonstrates that batch normalization can accelerate training by reducing the internal covariate shift, stabilize training by adding noise to inputs, and prevent the model from sticking to a local optimum. Second, these operators provide a beneficial initialization for student training, as they help align the teacher decision boundary around the training data \cite{balestriero2022batch}. Therefore, batch normalization is applied to multiple fully connected layers in student networks to reduce the effect of weight updating in the teacher network. With batch normalization, the prediction accuracy of students can be stable in experiments, leading to reliable explanations. The proposed joint training procedure, presented in \cref{on_the_fly_algorithm}, leverages the objective functions above to train both the black-box model and learners simultaneously.

The instability of node embeddings and predictive distributions, caused by the updates of GNN weights, presents a challenge for student models attempting to replicate the teacher's behavior. This issue can be mitigated by employing batch normalization operators \cite{ioffe2015batch}. Prior research \cite{ioffe2015batch} has shown that batch normalization not only speeds up training by minimizing internal covariate shift but also stabilizes it by introducing noise to the inputs, thereby preventing the model from getting stuck in local optima. Furthermore, these operators facilitate a favorable initialization for training a student by aiding in aligning the teacher's decision boundary with the training data \cite{balestriero2022batch}. Consequently, several layers within student networks employ batch normalization to minimize the weight updates' effects on the teacher network. By using this special operator, the predictive performance of student models remains stable, resulting in consistent and reliable explanations. The training algorithm is detailed in \cref{on_the_fly_algorithm}.

\begin{algorithm}[ht]
	\caption{Training Algorithm}
	\begingroup
	\raggedright
	\textbf{Input}: Training dataset $\mathbb{T}$, \#Epochs $T$ \\
	\textbf{Output}: $f, g$ \\
	\endgroup
	\begin{algorithmic}[1]
		\FOR{$ epoch = 0 \rightarrow T$}
		\STATE{$f$ = train($\mathbb{T}$, $\mathcal{L}_{ce}^t$)} \COMMENT{Train a target GNN $f$}
		\STATE{$g$ = distill($\mathbb{T}$, $f$, $\mathcal{L}^s$))} \COMMENT{Share knowledge with students}
		\ENDFOR
	\end{algorithmic}
	\label{on_the_fly_algorithm}
\end{algorithm}

\subsection{Structural Explanations}

%Structural explanations are the main focus of most existing GNN explanation methods \cite{yuan2022explainability}. They aim to specify important nodes and edges to a prediction. For simplicity, GCN \cite{kipf2016semi} is selected as the black-box GNN model. This selection enables the description of structural importance learners (self-explainable GNNs) for node and graph classification problems. The simplest form of a layer-wise propagation rule is as follows:

Structural explanations (or structure analyses in some specific scenarios) are the primary emphasis of many current XAI methods for GNNs \cite{yuan2022explainability}, which seek to identify the crucial graph patterns influencing predictions. To simplify the process, GCN \cite{kipf2016semi} is chosen as the study object of this work. This choice facilitates the description of structural importance learners for both graph and node classification tasks. The simplest matrix form of the message-passing paradigm in GCN is as follows:

\begin{equation}
	f(H^l, A) = \sigma(A H^l W^l),
	\label{basic_gcn}
\end{equation}
wherein $A$ represents an input graph's adjacency matrix, $H^l$ denotes node embeddings , $W^l$ indicates a weight matrix, and $\sigma$ denotes a non-linear function. 

\noindent\textbf{Graph Classification.}
An interpretable GNN is designed by incorporating a trainable matrix $M$ into \cref{basic_gcn} as detailed below:
\begin{equation}
	f(H^l, A, M) = \sigma((A \odot M)H^l W^l).
	\label{graph_classification_mask}
\end{equation}

% paraphrase
Referred to \cite{luo2020parameterized}, $M$ is initialized through an MLP. Specifically, node embeddings of sources and targets are concatenated, resulting in edge embeddings. These edge embeddings are subsequently inputted into the MLP model to calculate probabilities that indicate the likelihood of these edges being retained or pruned. The following equation details the process.

\begin{equation}
	m_{ij} = \textrm{sigmoid}(\textrm{MLP}([h_i, h_j])),
\end{equation}

%\noindent where $h_i$ and $h_j$ are representation vectors of the source and target nodes of an edge. A batch normalization layer follows each layer in the MLP model to mitigate the covariate shift problem caused by the weight updates of the black-box model. The last layer outputs probabilities $m_ij$ serving as filters for messages passing on edges. A self-explainable GNN model is trained with the black-box GNN using \cref{on_the_fly_algorithm}, as illustrated in \cref{fig:overview_a}. In the inference and explanation phase, edges with the highest probability values in $M$ are selected based on threshold values depending on particular datasets to construct a structural explanation for a prediction. Here, the probability corresponds to the likelihood that an edge influences the model prediction.

\noindent where vectors $h_i$ and $h_j$ correspond to the embeddings of two vertices of an edge, respectively. Each layer of the MLP model incorporates a batch normalization layer to mitigate the covariate shift issue arising from updates in the teacher's weights. The final layer acts as a selective gate, which keeps an edge $e_{ij}$ with the probability $m_{ij}$. 

As shown in \cref{fig:overview_a}, the interpretable GNN is jointly trained with a black-box teacher via \cref{on_the_fly_algorithm}. At the explanation phase, an explainer agent selects only edges having values $m$ higher than a context-aware threshold to construct an explanatory graph for an individual prediction. These probabilities reflect the extent to which an edge is likely to affect the prediction outcome.

%Most existing methods use the same approach as graph classification problems to explain a node's prediction. Specifically, a subgraph including the target node is extracted and fed to an explainable model to obtain a structural explanation. Even though this approach can select and filter out uninfluential edges and neighbors, it cannot specify the exact contributions of these factors to a prediction. 
\noindent\textbf{Node Classification.} Most contemporary methodologies transform node-level problems into graph-level ones to elucidate a prediction via subgraph sampling. Specifically, a substructure rooted at the target node is generated and processed through an explainer to produce an explanatory graph. Despite the capability of discarding irrelevant nodes and associated edges, this approach often fails to precisely quantify the influences of each input element on the outcome.

%This work follows the intuition that influential nodes are commonly found within close proximity. Therefore, if a neighbor is important to the target node's prediction, the edge between them should receive a high importance score. RWR is an efficient method for calculating relevance scores between two nodes. Inspired by its efficiency in recommender systems \cite{chiang2013exploring,park2017comparative,wang2020personalized}, a customized version is implemented to provide structural explanations for node-level predictions. Let us recall the formula of the algorithm as follows:

This research operates on the premise that ``nearby neighbors are more valuable than a faraway relative''. Consequently, if a neighbor significantly impacts the target node's outcome, its associated connection should be assigned a larger contribution score compared to those with less influence. Random Walk with Restart (RWR) is a proficient algorithm for determining node importance and relevance in a graph. Inspired by its success \cite{chiang2013exploring,park2017comparative,wang2020personalized}, RWR is extended to measure neighbor influences and generate explanatory graphs for node-level problems. The original algorithm is detailed as follows:

\begin{equation}
	r_{t+1} = (1 - d)r_0 + d \hat{\mathcal{A}}_c r_t.
	\label{alg:sc_rwr_or}
\end{equation}

%where $\hat{\mathcal{A}}_c$ is a column-wise normalized transition matrix, $r_t$ is a probability distribution vector at time $t$, $r_0$ is the initial probability distribution vector, and $d$ is the probability that random walkers go to a new state. 

In \cref{alg:sc_rwr_or}, $r_t$ denotes a distribution of nodes at time $t$, and $r_0$ signifies the initial distribution. $\hat{\mathcal{A}}_c$ represents a transition matrix that has been normalized column-wise, while $d$ indicates the likelihood of random walkers transitioning to a new state.

%A target node can be considered a query node in the RWR algorithm for explaining node-level predictions, wherein its corresponding element in $r_0$ is initialized as 1, while others are set to 0. The transition matrix $\hat{\mathcal{A}}_c$ is the transposed version of a trainable adjacency matrix $\hat{\mathcal{A}}$ taken from a structural importance learner, as presented in \cref{node_self_exp}. 

Within the RWR algorithm, the node needed an analysis of neighbor influences can be treated as an initial state of the walker, wherein the respective element in $r_0$ is set to 1, while all other elements are initialized to 0. $\hat{\mathcal{A}}_c$ is the transpose of a trained adjacency matrix $\hat{\mathcal{A}}$, which is detailed in \cref{node_self_exp}.

\begin{equation}
	f(H^l, \hat{\mathcal{A}}) = \sigma(\hat{\mathcal{A}} H^l W^l)
	\label{node_self_exp}
\end{equation}

\noindent Given an edge between two nodes $i$ and $j$ with corresponding representation vectors $h_i$ and $h_j$ taken from the black-box GNN, its respective value in $\hat{\mathcal{A}}$ is measured as follows:
\begin{equation}
	\hat{a}_{ij} = \textrm{softmax}(\textrm{MLP}([h_i, h_j])),
\end{equation}
where the softmax operator is to perform row-wise normalization on $\hat{\mathcal{A}}$.

%A converged vector $r$ tells us the importance of nodes in the input graph to the target node's prediction. An explanation visualization includes the target node and top $k$ nodes selected by ordering probabilities in $r$. Note that these $k$ vertices can include neighbors in multiple hops. The complete procedure is described in \cref{querying_rwr}.

Once reaching the stationary state, $r$ reveals the influence of vertices on the target prediction. An explanatory graph consists of the target node and its $k$ most significant neighbors, determined by ranking the values in $r$. It is important to note that these $k$ nodes may encompass neighbors located several hops away.  \cref{querying_rwr} details the entire process.

\begin{algorithm}[ht]
	\caption{Influence Analysis of a Node's Prediction}
	\begingroup
	\raggedright
	\textbf{Input}: A target node $v$ 
	\\\hspace{0.92cm} Trained Matrix $\hat{\mathcal{A}}$, 
	\\\hspace{0.92cm} \#Iteration $T$, 
	\\\hspace{0.92cm} Walking probability $d$,
	\\\hspace{0.92cm} \#Nodes $k$ in explanatory graph\\
	\textbf{Output}: An explanation of $v$'s prediction \\
	\endgroup
	\begin{algorithmic}[1]
		\STATE{$\hat{\mathcal{A}}_c$ = transpose($\hat{\mathcal{A}}$)} 
		\STATE{$\mathcal{P}_V$ = RWR($v, \hat{\mathcal{A}}_c, T, d$)} \COMMENT{Influence scores of nodes}
		\STATE{$\mathcal{P}_E$ = $\textrm{diag}(\mathcal{P}_V) \cdot \hat{\mathcal{A}}$} \COMMENT{Influence scores of edges}
		\STATE{$\mathcal{R}_V$ = top\_k($\mathcal{P}_V, k$)} \COMMENT{$k$ most influential neighbors}
		\STATE{visualize($\mathcal{P}_V, \mathcal{P}_E, \mathcal{R}_V$)} \COMMENT{Present explanatory graph}
	\end{algorithmic}
	\label{querying_rwr}
\end{algorithm}

\subsection{Feature Attribution Analysis}
%In many real-world scenarios, meaningful node features significantly benefit machine learning models. Masking techniques proposed by Ying et al. \cite{ying2019gnnexplainer} cannot clarify the exact contributions of features to predictions. Due to the complexity of message-passing patterns, it is difficult to utilize general XAI tools \cite{ribeiro2016should, lundberg2017unified} to measure feature attributions. Furthermore, examining feature attributions using transformation matrices of a pre-trained GNN is challenging. This is because ignoring graph structure drastically reduces model accuracy, leading to inconsistent results.

Node features considerably enhance the performance of ML models in various practical applications. However, the masking method introduced by Ying et al. \cite{ying2019gnnexplainer} falls short of elucidating the precise influences of these features on predictions. The intricate nature of message-passing operations hinders the adoption of conventional XAI tools \cite{ribeiro2016should, lundberg2017unified} to the measurement of feature attributions with graph data. Additionally, executing this procedure on the transformation matrices of a GNN poses challenges, as disregarding the graph-based operations substantially diminishes model accuracy, leading to inconsistent outcomes.

%To solve the problems above, this research constructs a feature attribution module to output attributions for specific predictions and a contribution summary for a group of them. This module consists of a feature transformation learner, a simple MLP model trained with the black-box GNN using \cref{on_the_fly_algorithm}. In training, \cref{on_the_fly_algorithm} allows the black-box teacher to guide the MLP student to achieve approximate predictive performance. DeepLIFT \cite{shrikumar2017learning,shapDeep19:online} is then executed on top of the MLP model in the explanation phase to produce feature attributions for predictions. In reality, several methods, such as \cite{tsang2020does}, can be integrated into SCALE since examining feature attributions is of great interest to the XAI research community. DeepLIFT is selected since it is an effective method for decomposing feature contributions in deep learning models and is very fast to compute.

To address the aforementioned issues, this work develops a component designed to measure feature attributions for individual predictions and summarize feature contributions for a group of predictions. This component incorporates a feature transformation learner, implemented based on an MLP network, which is assisted by the black-box GNN in training via \cref{on_the_fly_algorithm}. Specifically, \cref{on_the_fly_algorithm} enables the MLP student to boost its capability of capturing feature importances by approximating the black-box predictive distributions. During inference, an explainer agent integrates DeepLIFT \cite{shrikumar2017learning,shapDeep19:online} with the trained MLP model to generate attributional scores of features for predictions. In fact, alternative methods like \cite{tsang2020does} can substitute DeepLIFT for the measurement of feature attributions. However, DeepLIFT is chosen for its effectiveness and efficiency in approximating feature contributions in DL models.

\subsection{Example-based Explanation} 

%Example-based approaches have been extensively studied in XAI for computer vision and NLP problems, as evidenced by the work of Jeyakumar et al. \cite{jeyakumar2020can}. However, these methods have been relatively underexplored in the context of GNNs. The complexity of graph structures makes it challenging for users to comprehend predictions based solely on similar or counterfactual samples. Presenting reference examples with other explanation types may enhance users’ understanding of predictions. To obtain example-based explanations, the following function is employed:

Example-based XAI methods have been thoroughly investigated in computer vision and NLP domains \cite{jeyakumar2020can}. Nonetheless, these techniques have received less attention in the field of GNNs. The intricate nature of graph networks poses difficulties for users to understand predictions through only reference samples. Furthermore, defining similarity metrics for reference selections is also a challenging task. Combining example-based explanations with other interpretable elements can improve user understanding. The following function is utilized to generate example-based explanations.

\begin{equation}
	\begin{aligned}
		\mathbb{E} &= \textrm{Example}(\mathcal{G}, \mathbf{I}) \\
		&= \argmax_{g' \in \mathbb{T}_\mathbf{I}} \textrm{Similarity} (e_\mathcal{G}, e_{g'}),
	\end{aligned}
\end{equation}
%where $\mathbb{E}$ is a set of graphs closely related to the input graph based on a chosen metric. $\mathbf{I}$ is an indicator that determines whether the returned samples belong to the same or a different class. $\mathbb{T}_\mathbf{I}$ is a subset of the training dataset corresponding to the indicator $\mathbf{I}$. Next, the embedding vector of a graph is denoted by $e$. Graph similarity can be measured using distance-based metrics, such as cosine distance.

\noindent where $\mathbb{E}$ represents a graph set that is highly similar to the input graph according to a selected metric. The indicator $\mathbf{I}$ identifies whether the retrieved samples are in either a separate class or the same one. The subset $\mathbb{T}_\mathbf{I}$ refers to the portion of the training data associated with $\mathbf{I}$. Additionally, $e$ represents a graph's representation vector. For simplicity, graph similarity can be quantified based on distance-based metrics, such as Cosine or Gaussian distance. Investigating a more sophisticated and interpretable similarity metric remains an avenue for future research.

\subsection{Computational Complexity Analyses}
%The computational cost scales linearly with the number of learners denoted as $N$. Specifically, the training time is approximately $N+1$ times longer than that of a black-box GNN. Additionally, approximately $N+1$ times the GPU memory space is required to store the models. Methods to accelerate these limitations are explored in \cref{discussion}. Additionally, the number of parameters in a self-explainable GNN is equal to the number of weights in the black-box GNN plus the number of parameters of an MLP used to compute edge weights. The computational cost of the feature attribution module depends on the size of the student MLP model. Graphs are loaded into the GPU only once and shared among models to reduce memory consumption and share them among models. 

\noindent\textbf{Training.} The number of learners $N$ directly influences the computational cost with a linear scaling. Particularly, the demanded computational resources and training time are approximately $N+1$ greater than that of the original GNN. \cref{discussion} discusses methods to mitigate these shortcomings. Furthermore, the size of an interpretable GNN is equal to the summation of the black-box GNN's parameters and those of the MLP network used for measuring edge importance. The feature attribution procedure's computational cost linearly scales with the size of the feature transformation learner. Graph datasets are transferred to the GPU only once and utilized collectively by models to minimize memory footprints.

%Inference can be performed using either the black-box model or explainable modules, as both achieve comparable accuracy. An explainable module provides not only a prediction but also an accompanying explanation. Therefore, the computation cost includes both the cost of executing the predictive model and generating the explanation. For instance, \cref{querying_rwr} incurs computation time in constructing structural explanations for node-level predictions while finding feature importances involves executing DeepLIFT. The time complexity of an example-based explanation function can be reduced from $O(N)$, where $N$ corresponds to the number of graphs in a training subset, to $O(K)$, where $K \ll N$, through the use of clustering techniques. In practice, Faiss \cite{johnson2019billion}, an efficient similarity search library, is integrated into the proposed framework.

\noindent\textbf{Inference \& Explanation.} Both the black-box model and explainable ones can provide predictions, as they achieve comparative predictive performance. However, explainable models have their unique features, which are interpretable predictions with diverse explanation modalities. Therefore, their computational cost encompasses the cost of executing the predictive component and explanation engine. The explanation generation time varies for algorithms. For example, in \cref{querying_rwr}, the most burden operation is performing random walk iterations, which scale linearly with the number of iteration steps. Similarly, DeepLIFT computation consumes the majority of computational costs required for feature attribution measurement. The example-based explanation's time complexity can be decreased from $O(N)$, where $N$ is the total number of reference graphs, to $O(K)$, where $K \ll N$, by employing clustering techniques. Practically, the proposed framework incorporates a lightning-fast library \cite{johnson2019billion} for reference retrievals.

\section{Experimental Settings}\label{exp_setups}

\subsection{Objectives} % modified
%The primary objective of this research was to demonstrate SCALE's correctness and execution efficiency. First, quantitative comparisons were conducted between SCALE and selected baselines to showcase the distinguishing features of the proposed framework. Specifically, the aim was to establish that SCALE outperforms post-hoc explanation methods in both correctness and execution efficiency while also surpassing intrinsically interpretable models in explanation correctness. Second, qualitative assessments of the framework were performed by comparing it with two state-of-the-art post-hoc explanation methods \cite{ying2019gnnexplainer,luo2020parameterized}, highlighting SCALE's superior quality of explanations.  Third, a user study was conducted to assess the user perception of structural explanations for node classification, allowing us to understand the strengths and weaknesses of the method compared to counterparts in structural contribution measurement. Fourth, the feature attribution module was evaluated by confirming its results on the Amazon dataset with those obtained from a data mining-based method \cite{zhang2020gcn}. The results demonstrated that SCALE provided more comprehensive information on structural explanations and multi-level feature contributions than baselines. Finally, several ablation studies were conducted to validate the framework's efficiency from different aspects.

The primary goal was to validate the accuracy and efficiency of the proposed framework. Initially, quantitative analyses were performed to compare SCALE against chosen baselines, highlighting its unique attributes. The next objective was to prove that the proposed framework surpasses post-hoc methods in accuracy and execution speed and exceeds interpretable models in explanation accuracy. Subsequently, a qualitative comparison between SCALE with two leading post-hoc explanation techniques \cite{ying2019gnnexplainer,luo2020parameterized} was undertaken, underscoring the quality of its explanations. Additionally, a user study was implemented to measure user comprehension of structure analyses for node classification, offering insights into the framework's advantages and drawbacks relative to other approaches. Furthermore, the feature attribution module was assessed by comparing analyses on the Amazon dataset with findings derived from a data mining-based approach \cite{zhang2020gcn}. Results from these experiments indicated that SCALE delivered more detailed information on structural explanations and feature attributions than its counterparts. Lastly, several ablation studies were carried out to assess the framework's capability from various perspectives.

\subsection{Datasets} %modified

%Experiments were conducted using six node classification and two graph classification datasets, as outlined in \cref{tab:dataset}. Except for the Amazon and Cora datasets, these datasets have been commonly used to perform functionally grounded evaluations \cite{doshi2017towards} of GNN explanation methods. 

As detailed in \cref{tab:dataset}, experiments were carried out with two graph classification and six node classification datasets.  Except for the Cora and Amazon datasets, the other ones are employed for functional-grounded evaluations \cite{doshi2017towards} of GNN explanations.

\begin{table}[ht]
	\centering
	\caption{Dataset Statistics. K denotes a thousand. In experiments, edge numbers vary for the first four datasets.}
	\begin{tabular}{c|c|c|c|c|c}
		\hline
		& \#graphs & \#nodes & \#edges & \#feat. & \#labels \\
		\hline
		BA-Shapes     & 1    & 700   & - & 10 & 4 \\
		BA-Community  & 1    & 1400  & - & 10 & 8 \\
		Tree-Cycle    & 1    & 871   & - & 10 & 2 \\
		Tree-Grid     & 1    & 1231  & - & 10 & 2 \\
		Amazon        & 1    & 11.9K & 351.2K & 25 & 2\\
		Cora          & 1    & 2708  & 10K    & 1433 & 7 \\
		\hline
		BA-2motifs    & 1K   & 25K    & 51.4K & 10 & 2 \\
		Mutag         & 4.3K & 131.5K & 266.9K & 14 & 2 \\
		\hline
		
	\end{tabular}
	\vspace{0.1cm}
	\label{tab:dataset}
\end{table}

%Four synthetic graphs with ground-truth explanations, provided by \cite{ying2019gnnexplainer}, were utilized to evaluate the correctness and quality of structural explanations provided by methods. In each synthesis, the number of nodes was kept constant while the number of edges varied. Specifically, BA-Shapes (BA-S) was constructed by attaching 80 five-node houses to a 300-node BA graph. BA-Community (BA-C) was created by joining two BA-Shapes graphs. Similarly, Tree-Cycle (Tree-C) and Tree-Grid (Tree-G) were generated by randomly attaching cycle motifs and 3-by-3 grids to nodes in 8-level balanced binary trees, respectively. 

\noindent\textbf{Node Classification.} Referred to  \cite{ying2019gnnexplainer},  four synthetic graphs were generated with ground-truth explanations. Based on these datasets, methods were compared on the accuracy and quality of the structural explanations. In each synthetic graph, the number of edges varied, while the number of nodes remained fixed. Particularly, the BA-Shapes (BA-S) dataset was generated by connecting 80 five-vertex houses to a 300-vertex BA network. Joining two BA-Shapes graphs resulted in a BA-Community (BA-C) graph. Likewise, Tree-Grid (Tree-G) and Tree-Cycle (Tree-C) were created by randomly connecting 3-by-3 grids and cycle patterns to vertices in balanced binary trees with eight levels, correspondingly.

%Since the node features of synthetic graphs had no semantic meaning, the study leveraged Amazon dataset \cite{rayana2015collective, dou2020enhancing} to evaluate the feature attribution module. The dataset consisted of three graphs with links established based on mutual information between users. The goal was to identify fraudulent users based on given product reviews. Specifically, the research involved conducting experiments with all three graphs and selecting a graph of users reviewing the same products that produced the highest recall score to measure node feature attributions.

Given that synthetic datasets did not consist of semantic node features, the Amazon dataset \cite{rayana2015collective, dou2020enhancing} was employed to assess the feature attribution module. This dataset was used to identify fraudulent users through their product reviews. It included different graphs where nodes were users and edges were constructed based on mutual information between them. Initially, experiments were conducted on each graph. Based on practical results, the graph, wherein edges represented co-review behaviors, was chosen to evaluate feature attributions since the model yielded the highest recall score.

%Cora dataset was utilized for a user study on the impact of structural explanations on user perception of model decisions. This dataset is constructed from nodes representing papers and citation connections among them. Nodes belong to seven clusters corresponding to different paper categories. The detailed procedure and observations are described in \cref{sec:user_test}.

The Cora dataset was employed in a user study examining how structural explanations affect users' comprehension of predictions. The dataset comprises nodes that symbolize papers, along with the citation links between them. These nodes are categorized into seven clusters, each representing a different category of papers. The comprehensive methodology and findings are elaborated in \cref{sec:user_test}.

%This study utilized one synthetic graph and one real-world dataset for evaluations. Specifically, BA-2motifs (BA-2m) \cite{luo2020parameterized} consists of 1000 graphs with two classes constructed by adding specific motifs to BA graphs, where half contain 5-node house motifs and the other half include 5-node cycle motifs. Additionally, the Mutag dataset contains 4337 graphs classified into two classes based on their mutagenic effects. The dataset also includes ground-truth edge labels pointing out crucial subgraphs linked to mutagenic effects.

\noindent\textbf{Graph Classification.} This research employed both synthetic and practical datasets for evaluation purposes.  The BA-2motifs (BA-2m) dataset \cite{luo2020parameterized} includes 1000 graphs divided into two classes, created by connecting particular graph patterns to BA graphs. Half of these graphs feature 5-vertex cycle motifs, while the other half comprise 5-vertex house patterns. The Mutag dataset consists of 4337 graphs categorized into two groups according to their mutagenic effects. This dataset provides ground-truth edge labels that indicate accurate patterns associated with these mutagenic effects.

\subsection{Baselines}
%This thesis demonstrates the correctness of explanations and the execution performance of SCALE through a comparative analysis of experimental results with five baseline methods. Specifically, baselines are categorized into intrinsically interpretable models and perturbation methods. Evaluation of these methods incorporated quantitative and qualitative metrics, enabling us to assess their strengths and limitations thoroughly.

The research validates the explanation correctness and performance efficiency of the proposed framework by comparing experimental results with six baselines. These baselines are divided into two groups: interpretable models and post-hoc methods. The assessment of these methods involved both quantitative and qualitative analyses, allowing for a comprehensive evaluation of their respective strengths and weaknesses.

\noindent\textbf{Intrinsically Interpretable Models} use internal model weights to explain predictions directly. Four baseline models were selected as follows:
\begin{itemize}
	%	GCN \cite{kipf2016semi} is not an intrinsically interpretable model. Therefore, the normalized adjacency matrix was replaced with a trainable matrix analogous to \cref{graph_classification_mask,node_self_exp} for graph and node classification, respectively. Learnable adjacency matrices were then used to provide structural explanations. 
	\item \textbf{GCN-MLP:} GCN \cite{kipf2016semi} lacks interpretability. The original adjacency matrix was substituted with a trainable matrix similar to \cref{graph_classification_mask,node_self_exp} for the purposes of graph and node classification, respectively. These learnable adjacency matrices were subsequently employed to offer structural explanations.
	
	%	\item \textbf{GAT \cite{velickovic2017graph}} can be regarded as a self-explainable GNN since attention heads capture node interactions. In this study, three attention heads were employed for each layer, and an averaging of all heads across layers was conducted to derive explanations.
	\item \textbf{GAT \cite{velickovic2017graph}} can be considered an interpretable model since explanatory graphs can be generated from its attention heads. In this research, each GAT layer contains three attention heads, which are averaged to output a unique importance matrix.
	
	%	\item\textbf{SEGNN \cite{dai2021towards}} is a self-explainable GNN based on a similarity module that calculates structure distances between an unlabelled node and K-nearest labeled neighbors. In this implementation, K was determined in a manner to avoid out-of-memory issues while maximizing the recall scores.
	\item \textbf{SEGNN \cite{dai2021towards}} is an interpretable model that employs a similarity component to compute structure similarities between a target vertex and its closest labeled neighbors. In this model, the number of nearest nodes was selected to prevent out-of-memory issues while maximizing recall scores.
	
	%	\item\textbf{EGNN \cite{li2022egnn}} is a self-explainable model based on an offline KD paradigm \cite{hinton2015distilling}, which eliminates non-crucial messages from 2-hop neighbors through two distinct masking layers. For the construction of explanations, neighbor scores were aggregated and the top K nodes were selected to optimize recall scores.
	\item \textbf{EGNN \cite{li2022egnn}} is an interpretable model employing an offline KD paradigm \cite{hinton2015distilling}, whose objectives are to filter out unimportant messages in a 2-hop subgraph using two different masking layers. For explanation generation, influential scores of nodes were aggregated, and the top K vertices were chosen to maximize recall scores.
	
\end{itemize}

\noindent\textbf{Post-hoc Methods} require additional training processes to measure the influence of elements in an input graph on an outcome. The proposed framework is compared with two fundamental techniques that share the same approach.
\begin{itemize}
	%	was the first work that trained edge masks to determine crucial patterns based on an information theory approach. However, it has to retrain a mask for each target instance, thus making it less effective for inductive settings and large-scale graphs.
	\item \textbf{GNNExplainer \cite{ying2019gnnexplainer}} was the pioneering work that employed the information theory and trained an edge mask to identify significant patterns of an input graph. It necessitates retraining the mask matrix for each explanation, hindering it from being adopted in inductive scenarios and large-scale graphs.
	%	 shared the same approach as GNNExplainer \cite{ying2019gnnexplainer} but initialized masks using embedding vectors from the pre-trained model. Moreover, target instances share trainable weights.
	\item \textbf{PGExplainer \cite{luo2020parameterized}} employed a similar methodology to \cite{ying2019gnnexplainer} but initiated the mask matrix using node embeddings of the black-box GNN. Additionally, trainable weights are shared among target instances.	
\end{itemize}

%The explanation querying algorithm is compatible with a wide range of pre-trained GNNs provided that these models include normalized adjacency matrices representing interactions between nodes. When integrated with GCN-MLP and GAT, the proposed approach yields two distinct variants: SCALE-GCN-MLP and SCALE-GAT. These integrations demonstrate the approach's versatility and its potential for widespread adoption across various GNN applications.
\noindent\textbf{Executing \cref{querying_rwr} on GCN-MLP and GAT:} \cref{querying_rwr} can be applied to various GNNs, as long as these models incorporate edge weights representing node interactions or influences. When applied to GAT and GCN-MLP, this algorithm results in two subsequent models: SCALE-GAT and SCALE-GCN-MLP. These combinations highlight the versatility of the proposed algorithm and its potential for broad adoption across different GNN architectures.

\subsection{Quantitative Evaluation Metrics for Explanatory Graphs} 
%The approach of \cite{ying2019gnnexplainer,luo2020parameterized} was followed in formulating explanations as binary classification problems, where edges within pre-defined ground-truth motifs were assigned positive labels, and all others were considered negative. While previous methods used the AUC score as the primary evaluation metric, precision and recall scores were utilized for several reasons. First, the AUC score is unsuitable for evaluating \cref{querying_rwr}. Second, an investigation of how the ratio of true positive and false positive edges varied across different scenarios was desired. Finally, precision and recall scores offered more comprehensive information to assess the efficacy of explanation methods. In many cases, explanation methods can achieve high recall scores by including all ground-truth edges but still obtain low precision scores due to numerous false positive edges. Therefore, a good explanation method must provide subgraphs that include all ground-truth edges and contain as few wrong edges as possible, resulting in high scores in both metrics.

Following the approach of \cite{ying2019gnnexplainer,luo2020parameterized}, explanations were formulated as binary classification tasks. With this approach, edges that were part of pre-defined ground-truth patterns were labeled as 1, while all other edges were marked as 0. Contrary to previous methods that primarily relied on the AUC score for evaluation, precision, and recall metrics were employed for several important reasons. Firstly, the AUC score was deemed inappropriate for assessing \cref{querying_rwr}. Secondly, there was a need to analyze how the ratio of true positive to false positive edges varied across different scenarios. Lastly, precision and recall scores provided a more detailed evaluation of the effectiveness of explanation methods. It is noteworthy that explanation methods can often achieve high recall scores by including all ground-truth edges, yet still attain low precision scores due to the presence of numerous false positive edges. Consequently, an effective explanation method should yield subgraphs that encompass all ground-truth edges while minimizing the inclusion of incorrect edges, thereby achieving high precision and recall scores.

\begin{equation}
	\begin{aligned}
		\textrm{Precision} &= \frac{\textrm{True Positive}}{\textrm{True Positive} + \textrm{False Positive}} \\
		\textrm{Recall} &= \frac{\textrm{True Positive}}{\textrm{True Positive} + \textrm{False Negative}}
	\end{aligned}
\end{equation}

\subsection{Configurations}

%The approach of \cite{ying2019gnnexplainer, luo2020parameterized}, which used the 8:1:1 (train/validation/test) splitting strategy, was adopted for experiments. For fair comparisons, attempts were made to contact the authors of GNNExplainer, PGExplainer, and SEGNN to request evaluation scripts for all datasets. As responses were not received, evaluation scripts were implemented to the best extent possible based on their public source codes. Baseline models were trained on datasets according to the methods described in their respective published papers and source codes. Additionally, Youden's J Statistic was utilized to determine selection thresholds in baselines that output edge selection probabilities. According to \cite{ying2019gnnexplainer, luo2020parameterized}, explained instances were manually selected, regardless of which set they belonged to in training. 

Following the configurations in \cite{ying2019gnnexplainer, luo2020parameterized}, the splitting strategy 8:1:1 was employed. To ensure fair comparisons, evaluation scripts for baselines were developed as accurately as possible based on the publicly available source codes. Specifically, the baseline models were trained on datasets following the methodologies and configurations outlined in respective publications and source codes. Furthermore, Youden's J Statistic was applied to define the edge pruning thresholds for baselines that produce selection probabilities on edges. Following the procedures in \cite{ying2019gnnexplainer, luo2020parameterized}, the instances to be explained were manually chosen, irrespective of their categorization in the datasets.

\begin{table}[ht]
	\centering
	\caption{Hyper-parameters Used in Training}
	\begin{tabular}{c|c|c|c|c|c}
		\hline
		& \thead{MLP\\Layers} & \thead{GCN\\Layers} & \thead{Hidden\\Size} & $\lambda$ & \thead{Num.\\ Epochs}\\
		\hline
		Amazon & 2 & 2 & 32 & 0.1 & 200\\
		BA-Shapes & 3 & 6 & 32 & 0.1 & 1000\\ 
		BA-Community & 3 & 6 & 64 & 0.1 & 1000 \\ 
		Tree-Cycle & 3 & 6 & 64 & 0.1 & 1000\\
		Tree-Grid & 3 & 6 & 64 & 0.1 & 1000\\
		\hline
		BA-2motifs & 3 & 4 & 64 & 4 & 200 \\
		Mutag & 3 & 4 & 64 & 4 & 200 \\
		\hline
	\end{tabular}
	\vspace{0.1cm}
	\label{tab:hyper}
\end{table}

%SCALE modules were trained with hyper-parameter settings presented in \cref{tab:hyper}. The hidden size represents the dimension of transformation matrices in GNN-based models and denotes the first layer's dimension in MLP models. The sizes of the last layers in MLP models depend on the model roles, which can be 1 in mask initialization or 2 in classification tasks. Furthermore, the hidden size was reduced by half after each layer. For instance, $[64,32,2]$ denotes an MLP with three layers, with the first layer containing 64 hidden units.

\cref{tab:hyper} detailed the hyper-parameters utilized for the training paradigm in the proposed framework. In GNN models, the hidden size denotes the dimension of linear layers, while in MLP networks, it indicates the size of the first layer. The dimensions of the final layers in MLP networks vary depending on their function: it may be 1 for edge pruning tasks or 2 for classification ones. Additionally, the hidden size was halved after each subsequent layer. For example, an MLP with three layers represented as $[64, 32, 2]$ has 64 units in the first layer.

%Practically, the learning rate was set to 0.01 and $\tau$ to 2 for all experiments. The probability $d$ was established at 0.9 for Tree-Grid and 0.55 for other datasets. The magnitude of $\lambda$ correlated with the amount of knowledge distilled from the teacher to the student during training. Ablation studies were conducted to examine the impacts of $\lambda$ and $d$ on explanation correctness.

In experiments, the learning rate was fixed at 0.01, and $\tau$ was set to 2 for all experimental trials. The parameter $d$ in \cref{querying_rwr} was determined to be 0.55 for all datasets, except for the Tree-Grid dataset which is 0.9. The value of $\lambda$ was linked to the extent of knowledge transferred from the teacher model to a student model in the training process. This work conducted ablation studies to investigate the effects of $\lambda$ and $d$ on the accuracy of explanations.

%SCALE and baselines were executed five times on each dataset using a machine equipped with an NVIDIA Tesla V100 16GB GPU, and the average results were reported. SCALE's models were implemented based on PyTorch v1.10.2 and DGL v0.9.0. Except for PGExplainer which used TensorFlow v2.9.1, baseline models utilized the same PyTorch version. The DeepLIFT execution employed the PyTorch API as provided by \cite{shapDeep19:online}.

For each dataset, all models were run five times on a system powered by an NVIDIA Tesla V100 16GB GPU, and the mean results were reported. PyTorch v1.10.2 was used for the implementation of all models, except for PGExplainer, which was implemented using TensorFlow v2.9.1. DGL v0.9.0 was used for implementing GNN models in the proposed framework.  For the execution of DeepLIFT, the PyTorch API as detailed by \cite{shapDeep19:online} was utilized.

\begin{sidewaystable}
	\begin{minipage}[c][\textheight]{\linewidth}
		\centering
%		\caption{Quantitative Comparison of SCALE and baselines on structural explanation correctness. SCALE is superior to baselines, especially intrinsically interpretable models. Explanation results are significantly improved when executing \cref{querying_rwr} on GCN-MLP and GAT. Here, P is short for the precision score, and R denotes the recall score.}
		\caption{A Comparison of the correctness of structural explanations. Compared to baselines, especially self-explainable models, SCALE has outstanding performance. Executing \cref{querying_rwr} on GCN-MLP and GAT yields considerably better explanation results compared to the simple approach. P and R stand for precision and recall scores, respectively.}
		\begin{tabular}{c|c|c|c|c|c|c|c|c|c|c|c|c}
			\hline
			& \multicolumn{2}{c|}{BA-Shapes} & \multicolumn{2}{c|}{BA-Community} & \multicolumn{2}{c|}{Tree-Cycle} & \multicolumn{2}{c|}{Tree-Grid} & \multicolumn{2}{c|}{BA-2motifs} & \multicolumn{2}{c}{Mutag}\\
			\cline{2-13}
			& P & R & P & R & P & R & P & R & P & R & P & R \\
			\hline
			GCN-MLP & 94.03 & 52.50 & 65.51 & 92.08 & 68.75 & 85.56 & 48.73 & 69.13 & 21.66 & \textbf{100} & 13.71 & 66.67 \\
			GAT & 89.35 & 99.03 & 72.19 & 97.36 & 68.16 & 75.83 & 54.71 & 30.14 & - & - & - & - \\
			SEGNN & 98.55 & 49.38 & \underline{97.39} & 46.46 & 69.79 & 82.78 & 76.35 & 73.33
			& - & - & - & - \\
			EGNN & 56.55 & \underline{99.44} & 42.73 & 93.89 & 62.17 & 66.67 & 77.85 & 63.03
			& - & - & - & - \\
			\hline
			GNNExplainer & 80.44 & \textbf{100} & 59.27 & \textbf{100} & 73.92 & 87.34 & 75.18 & 53.79 & 26.46 & \underline{94.18} & 14.66 & 71.49 \\
			PGExplainer & 96.85 & \textbf{100} & 54.97 & 97.07 & 99.25 & \underline{99.57} & 91.05 & \underline{87.81} & \underline{95.58} & \textbf{100} & \underline{50.64} & \underline{99.37} \\
			\hline
			
			SCALE & \textbf{98.90} & \textbf{100} & \textbf{99.17} & \textbf{100} & \textbf{99.45} & \textbf{100} & \textbf{97.11} & \textbf{91.00} & \textbf{96.25} & \textbf{100} & \textbf{66.18} & \textbf{99.72} \\ 
			SCALE-GCN-MLP & \underline{98.87} & \textbf{100} & 85.75 & \underline{98.61} & \underline{99.36} & \textbf{100} & \underline{94.90} & 84.31 & - & - & - & - \\
			SCALE-GAT & 98.63 & \textbf{100} & 91.89 & 94.44 & 83.59 & 79.02 & 84.11 & 64.84 & - & - & - & - \\
			\hline
		\end{tabular}
	\vspace{0.1cm}
	\label{tab:quant_compare}
	\end{minipage}
\end{sidewaystable}

\section{Experimental Results}\label{exp_results}

\subsection{Comparison on Explanation Accuracy}

%The first comparison of methods is based on the accuracy of their structural explanations. \cref{tab:quant_compare} leads to the following observations. SCALE is superior to all baselines in explaining both node and graph predictions. Specifically, it achieves outstanding precision and recall scores in node classification datasets and outperforms state-of-the-art methods GNNExplainer and PGExplainer. GCN-MLP, GAT, and EGNN achieve high recall scores on BA-based datasets since ground-truth motifs only need at most 2-hop traversals. SEGNN performs poorly on these datasets since the number of sampling hops cannot be larger than one due to out-of-memory errors. In tree-based datasets, self-explainable models are inferior to SCALE due to the ineffectiveness of their explanation procedures. Moreover, perturbation-based methods follow sampling-then-choosing approaches, which can lead to multiple false edges being included in explanations. Conversely, SCALE expands explanation motifs from target nodes until meeting vertex thresholds corresponding to ground-truth motifs based on \cref{querying_rwr}, thereby achieving high precision scores. SCALE outperforms baselines on Mutag, with precision score gains of 15.54\% compared to PGExplainer and 51.52\% compared to GNNExplainer. Furthermore, its performance is comparable to that of PGExplainer on the BA-2motifs dataset. 

The first experiment focuses on the comparison of the correctness of structural explanations provided by different methods. The insights drawn from \cref{tab:quant_compare} are as follows: SCALE surpasses all baseline methods in both node and graph tasks on the accuracy metric. Notably, it obtains remarkable precision and recall values in node classification tasks, outperforming post-hoc methods like GNNExplainer and PGExplainer. EGNN, GAT, and GCN-MLP exhibit high recall values on BA-related node classification tasks because the ground-truth patterns require at most 2-hop traversals. In contrast, SEGNN's performance is significantly low on these tasks due to its inability to handle sampled subgraphs with a hop size of more than one, encountering out-of-memory issues. In tree-related datasets, interpretable models are less effective than SCALE due to the shortcomings in their explanation methodologies. Additionally, post-hoc methods, employing sampling-then-pruning strategies, tend to include numerous wrong edges in generated explanations. SCALE, on the other hand, extends explanatory graph patterns from target nodes until reaching thresholds that align with ground truths, thus attaining high precision scores. On the Mutag dataset, SCALE surpasses baselines with a precision improvement of 51.52\% over GNNExplainer and 15.54\% over PGExplainer. On the BA-2motifs dataset, its performance is on par with PGExplainer.

\begin{table}[ht]
	\centering
%	\caption{A Comparison of Execution Time among Methods. Results are measured in seconds. }
	\caption{A Comparative Analysis of Method Execution Times Recorded in Seconds}
	\begin{tabular}{c|c|c|c|c|c|c}
		\hline
		& BA-S & BA-C & Tree-C & Tree-G & BA-2m & Mutag \\
		\hline
		GCN-MLP & 0.16 & 0.22 & 0.20 & 0.89 & 0.42 & 2.49 \\
		GAT & 0.16 & 0.20 & 0.18 & 1.11 & - & -\\
		SEGNN & 0.24 & 0.26 & 0.33 & 1.59 & - & -\\
		EGNN & 13.52 & 19.60 & 15.43 & 23.05 & - & -\\
		\hline
		GNNExpl. & 40.79 & \underline{40.77} & \underline{34.11} & \underline{155.35} & \underline{107.42} & 630.42 \\
		PGExpl. & \underline{29.33} & 167.89 & 55.61 & 515.16 & 183.4 & \underline{153.2}\\
		\hline
		SCALE & \textbf{1.58} & \textbf{1.62} & \textbf{2.17} & \textbf{5.81} & \textbf{1.53} & \textbf{6.70} \\
		\hline
	\end{tabular}
	\label{tab:time_compare}
\end{table}

%Our second objective is to demonstrate SCALE's superior running performance. As shown in \cref{tab:time_compare}, SCALE outperforms post-hoc explanation methods by a significant margin in all experiments, with performance gains of up to 94x compared to GNNExplainer and 120x compared to PGExplainer. Even though SCALE is slightly slower than self-explainable baselines in certain scenarios, the gaps are negligible. Furthermore, this minor drawback is acceptable, considering SCALE's exceptional explanation scores compared with these methods.

The experiment is to illustrate the superior runtime performance of SCALE. As indicated in \cref{tab:time_compare}, the proposed framework significantly surpasses post-hoc methods in settings, with performance improvements reaching up to 120 times that of PGExplainer and 94x compared to GNNExplainer. Although other interpretable models are marginally faster than SCALE in some settings, these differences are minimal. Additionally, this slight drawback is justified by SCALE's outstanding explanation quality relative to these models.

\subsection{Qualitative Comparison on Explanatory Graphs}

\begin{figure}[ht]
	\centering
	\setlength\tabcolsep{3pt}
	\renewcommand{\arraystretch}{1}
	\begin{tabular}{r p{1.7cm}p{1.7cm}p{1.7cm}p{1.7cm}p{1.7cm}p{1.7cm}}
		\textbf{GT} & 
		\multicolumn{1}{m{1.7cm}}{\includegraphics[width=1.7cm]{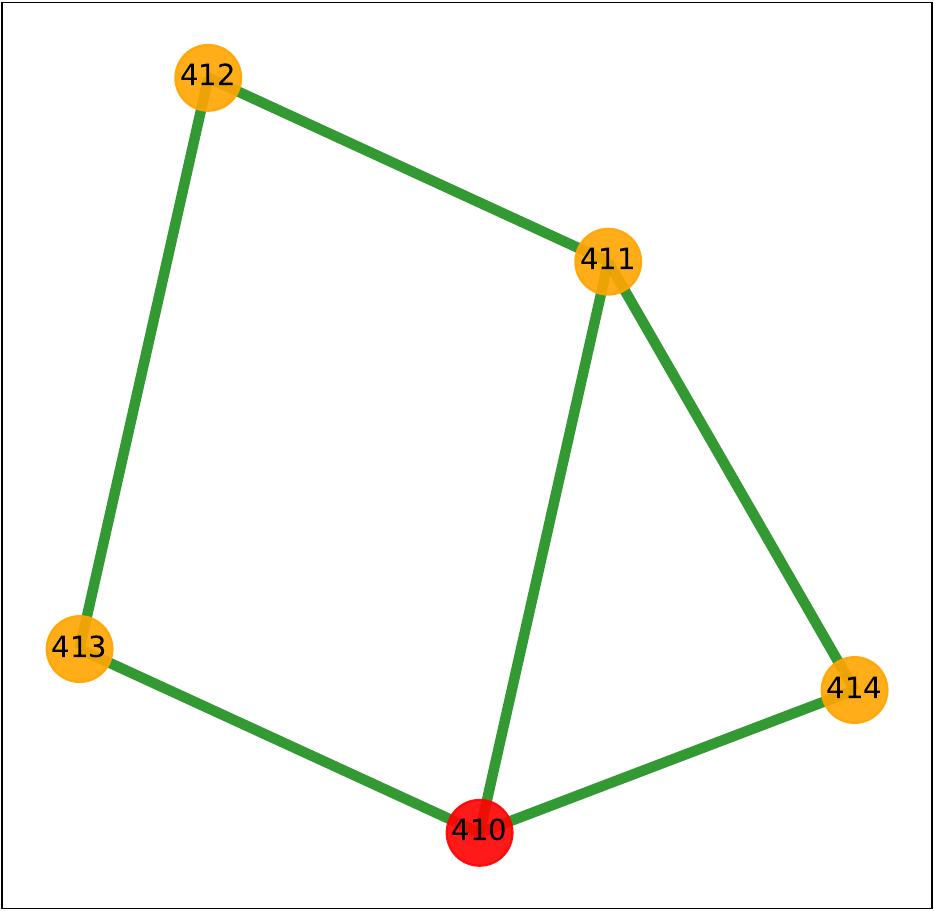}} & 
		\multicolumn{1}{m{1.7cm}}{\includegraphics[width=1.7cm]{scale/figures/graphs/ba_shape_410_real.pdf}} & 
		\multicolumn{1}{m{1.7cm}}{\includegraphics[width=1.7cm]{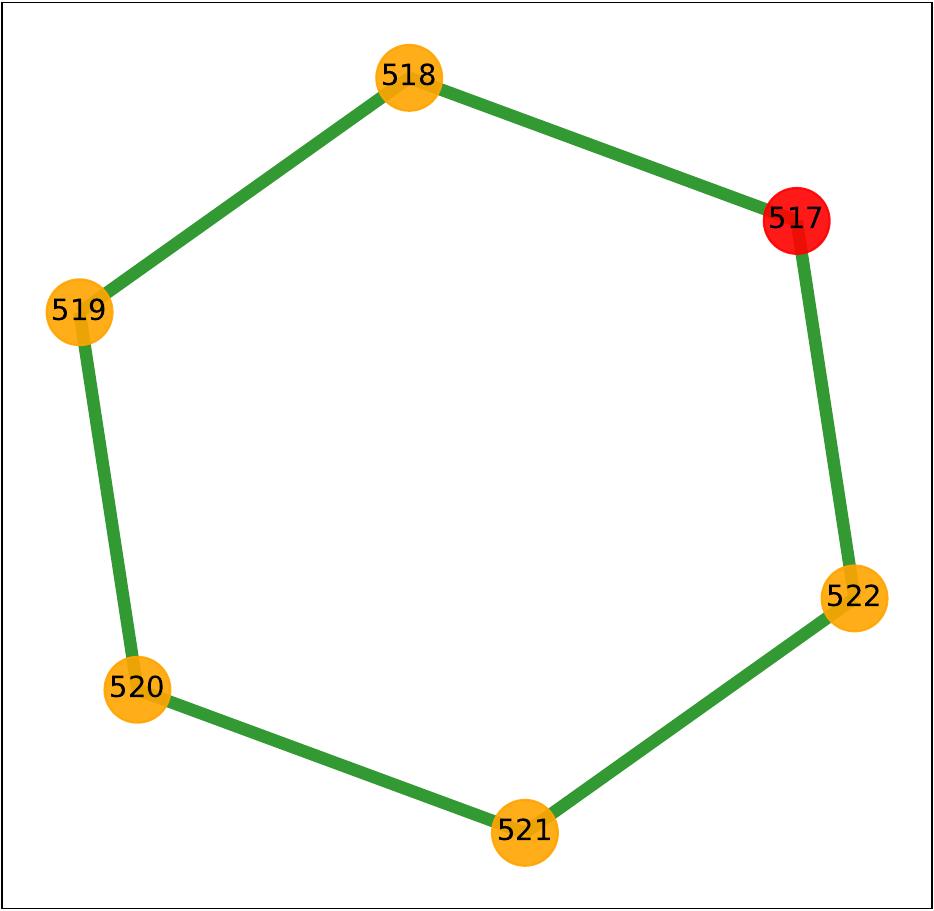}} & 
		\multicolumn{1}{m{1.7cm}}{\includegraphics[width=1.7cm]{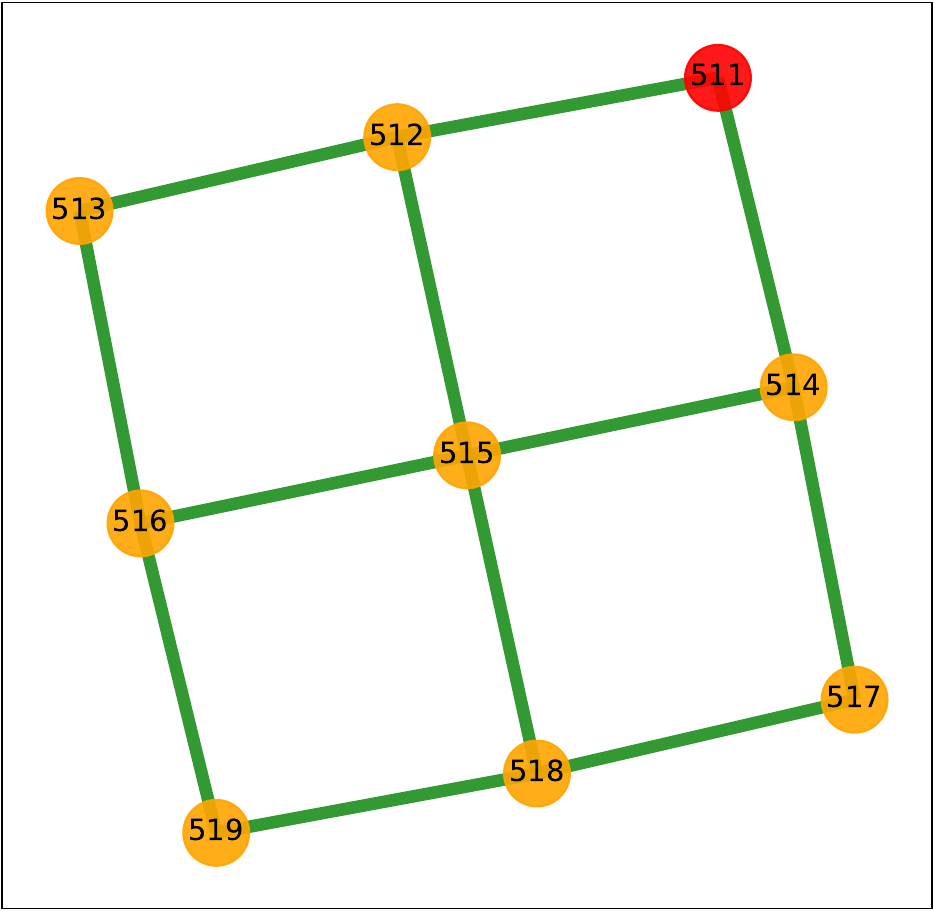}} & 
		\multicolumn{1}{m{1.7cm}}{\includegraphics[width=1.7cm]{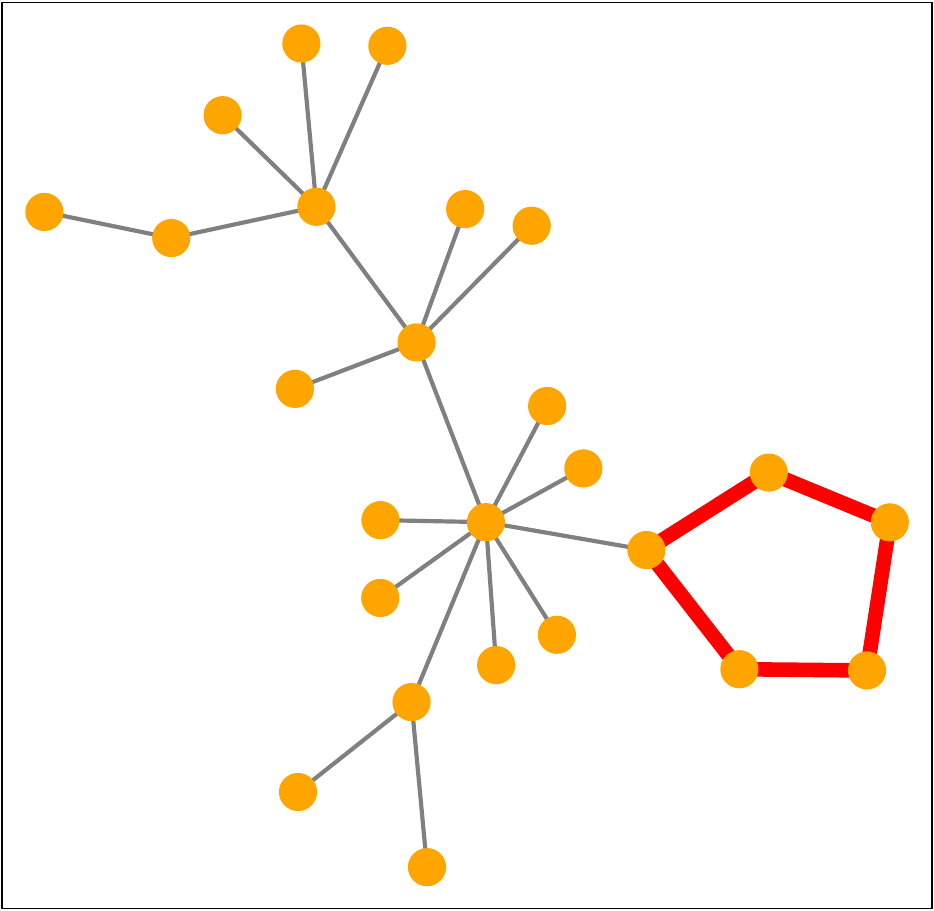}} & 
		\multicolumn{1}{m{1.7cm}}{\includegraphics[width=1.7cm]{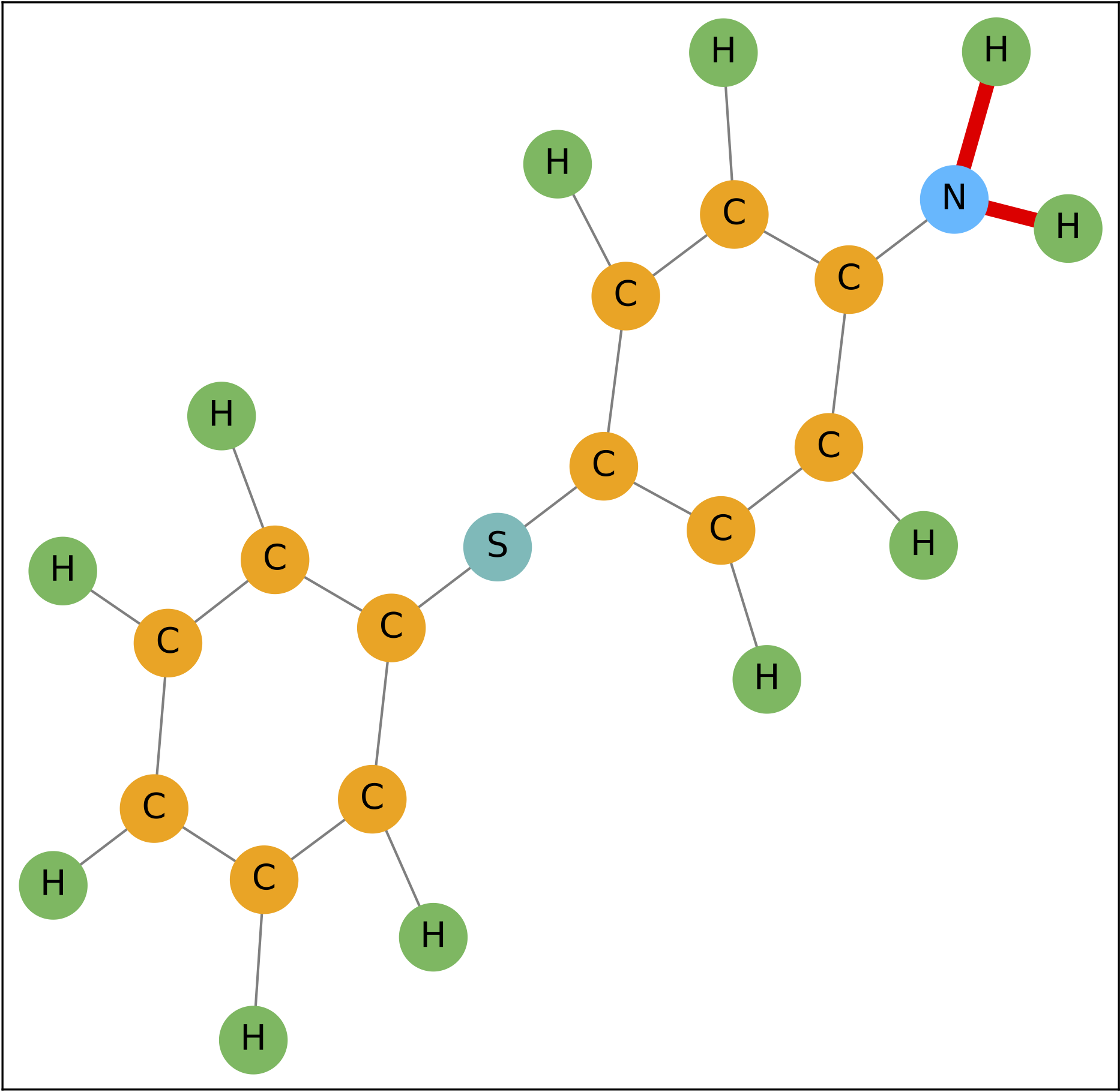}}
		\\
		\textbf{GX} & 
		\multicolumn{1}{m{1.7cm}}{\includegraphics[width=1.7cm]{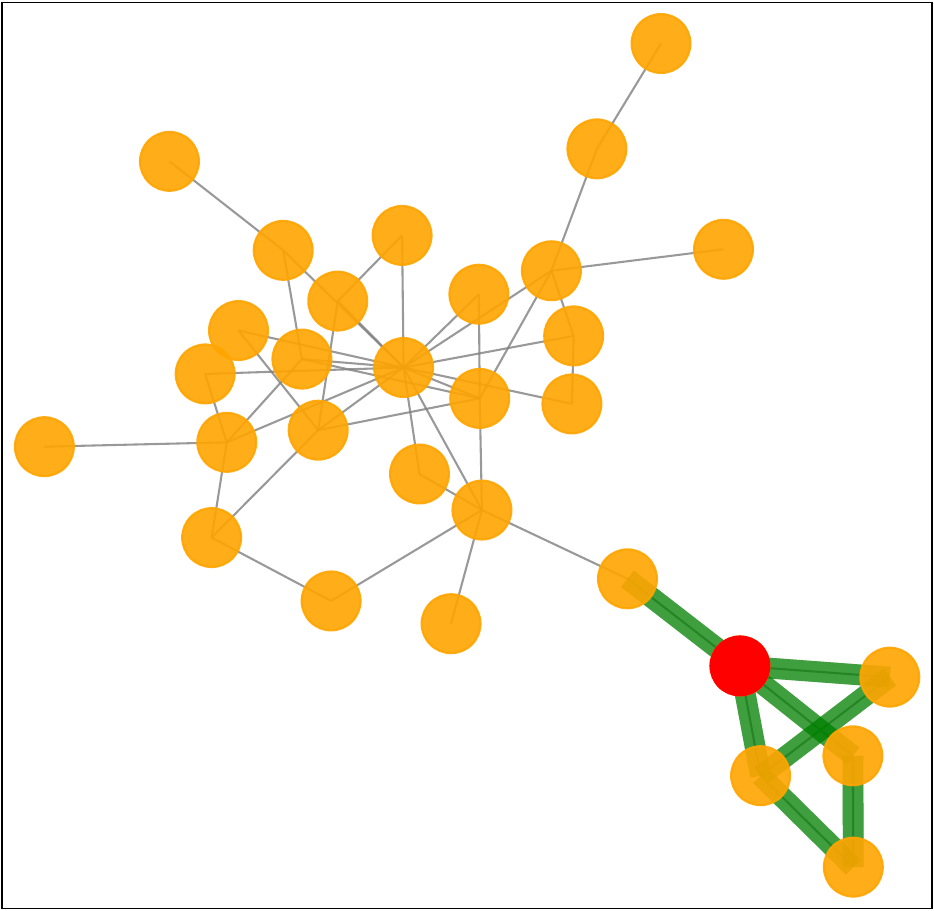}} & 
		\multicolumn{1}{m{1.7cm}}{\includegraphics[width=1.7cm]{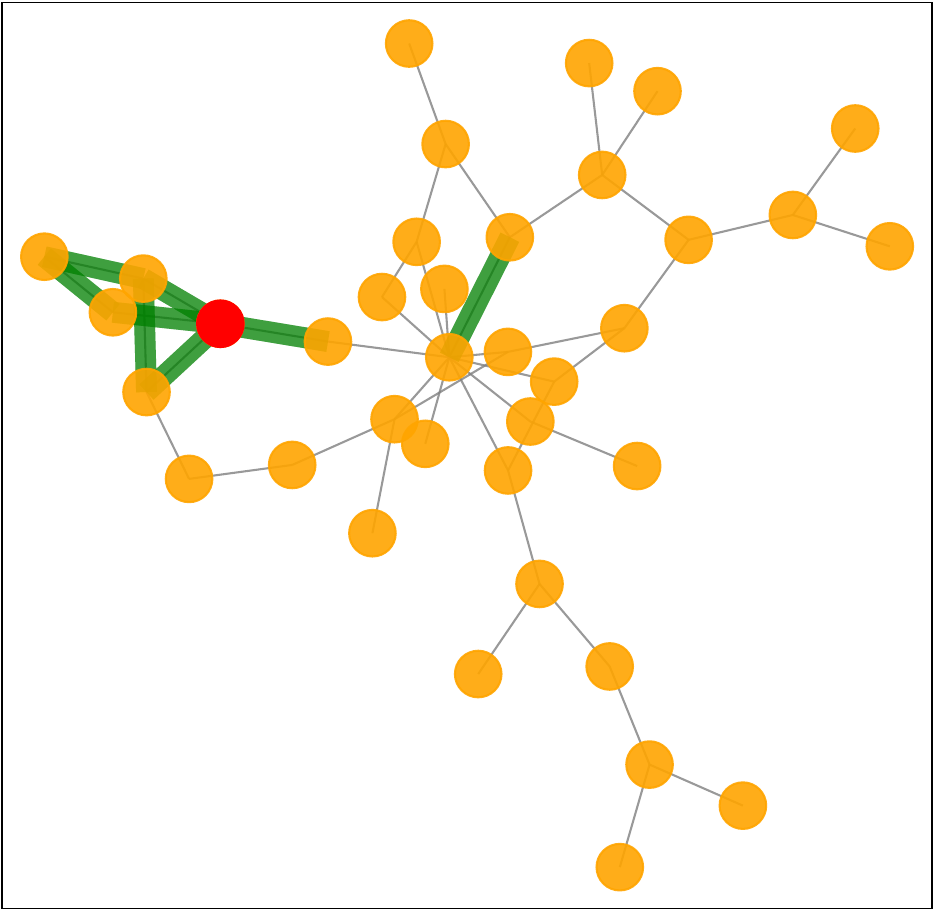}} & 
		\multicolumn{1}{m{1.7cm}}{\includegraphics[width=1.7cm]{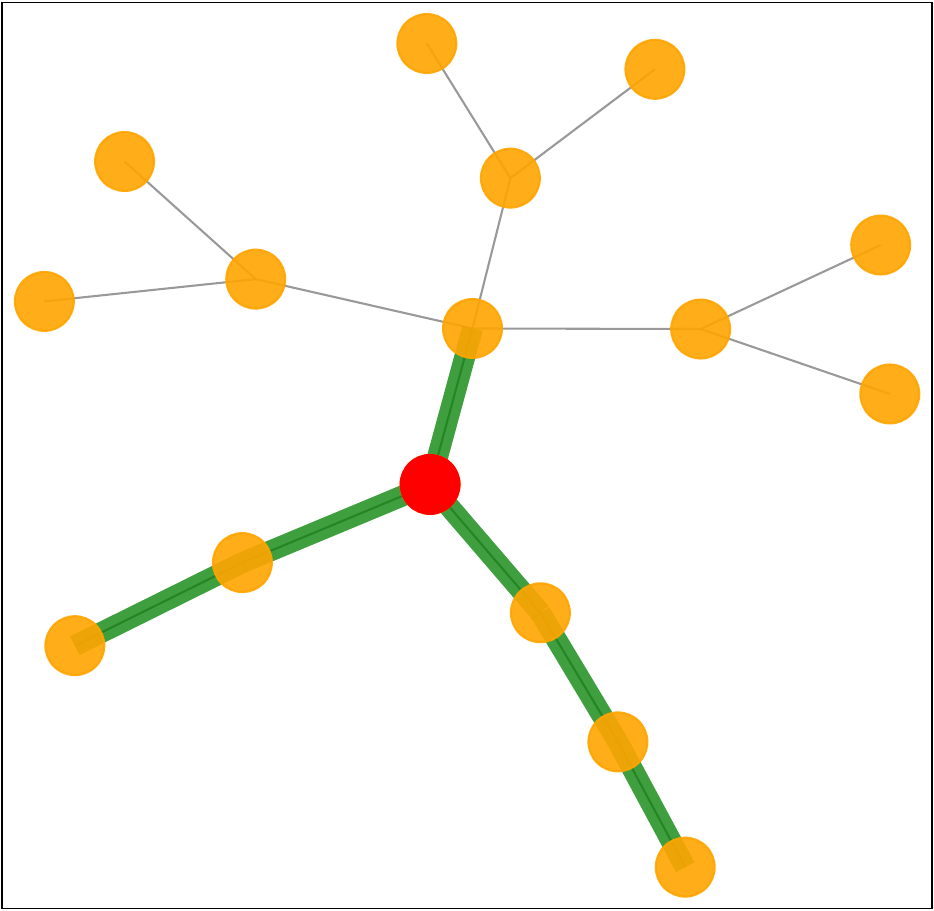}} & 
		\multicolumn{1}{m{1.7cm}}{\includegraphics[width=1.7cm]{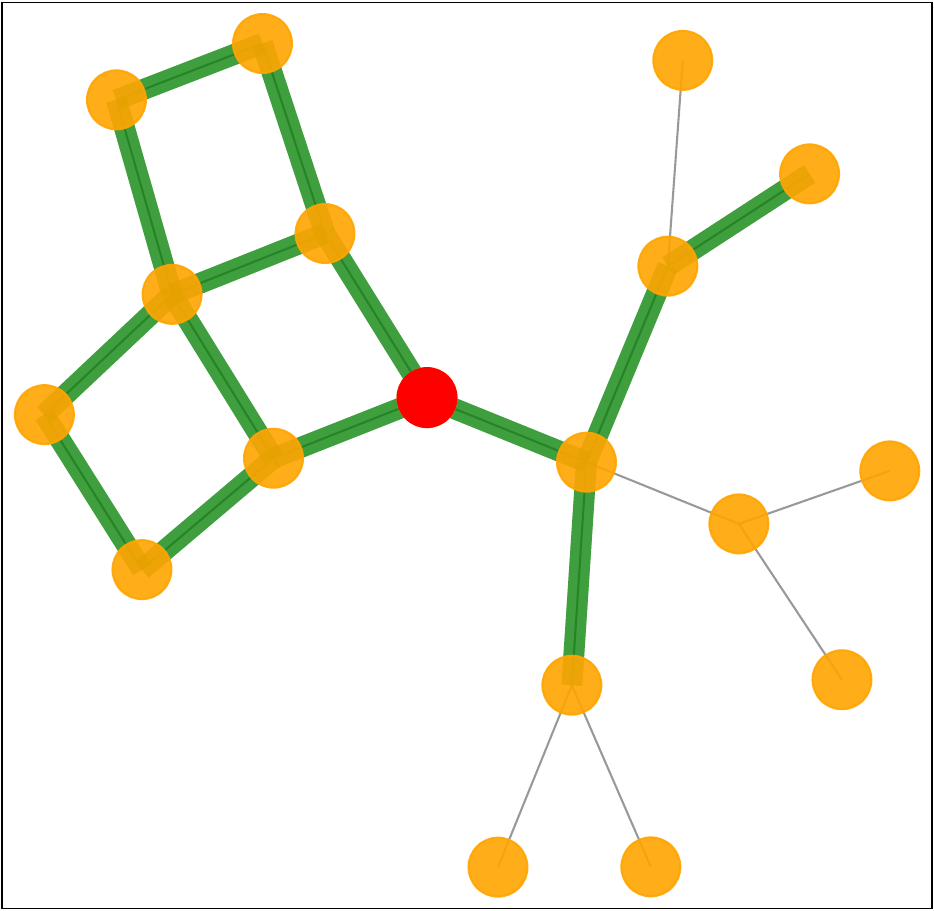}} & 
		\multicolumn{1}{m{1.7cm}}{\includegraphics[width=1.7cm]{scale/figures/graphs/pg_ba2motif_5.pdf}} & 
		\multicolumn{1}{m{1.7cm}}{\includegraphics[width=1.7cm]{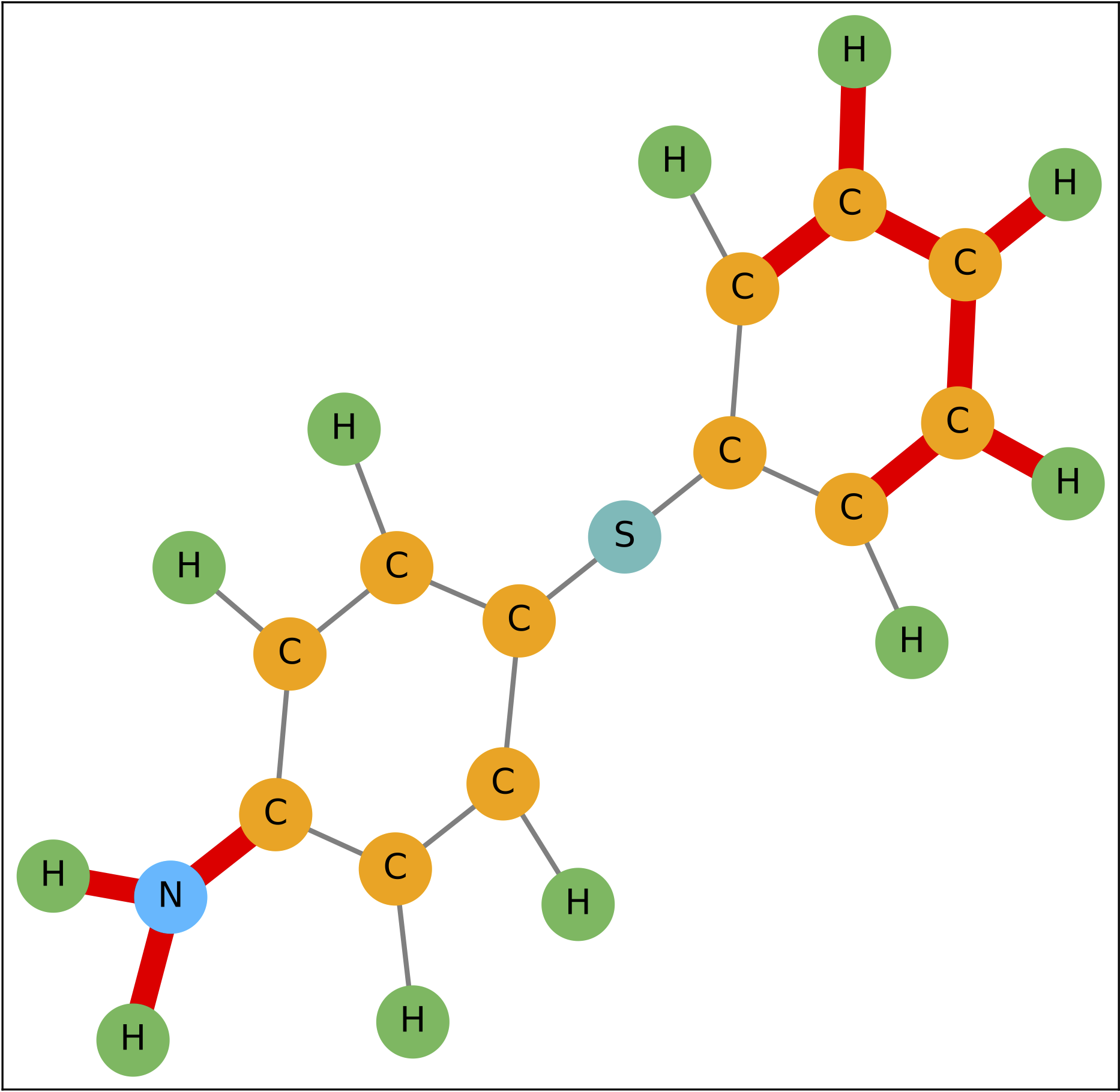}}
		\\
		
		\textbf{PX} & 
		\multicolumn{1}{m{1.7cm}}{\includegraphics[width=1.7cm]{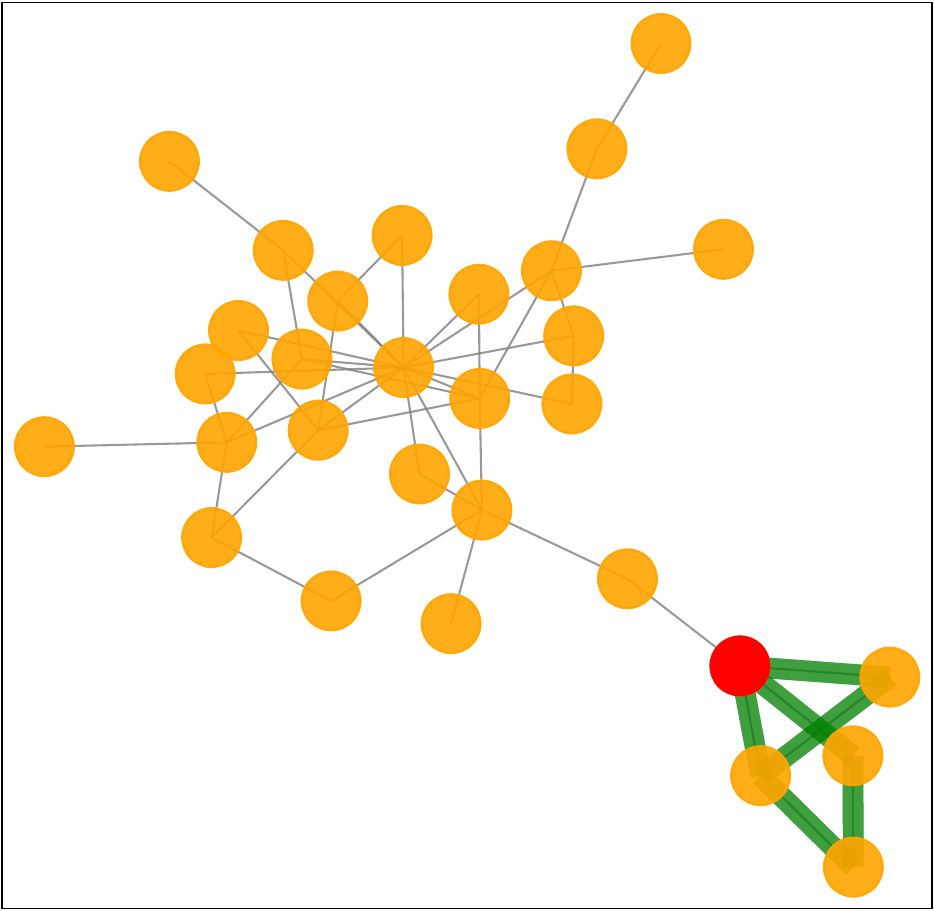} } & 
		\multicolumn{1}{m{1.7cm}}{\includegraphics[width=1.7cm]{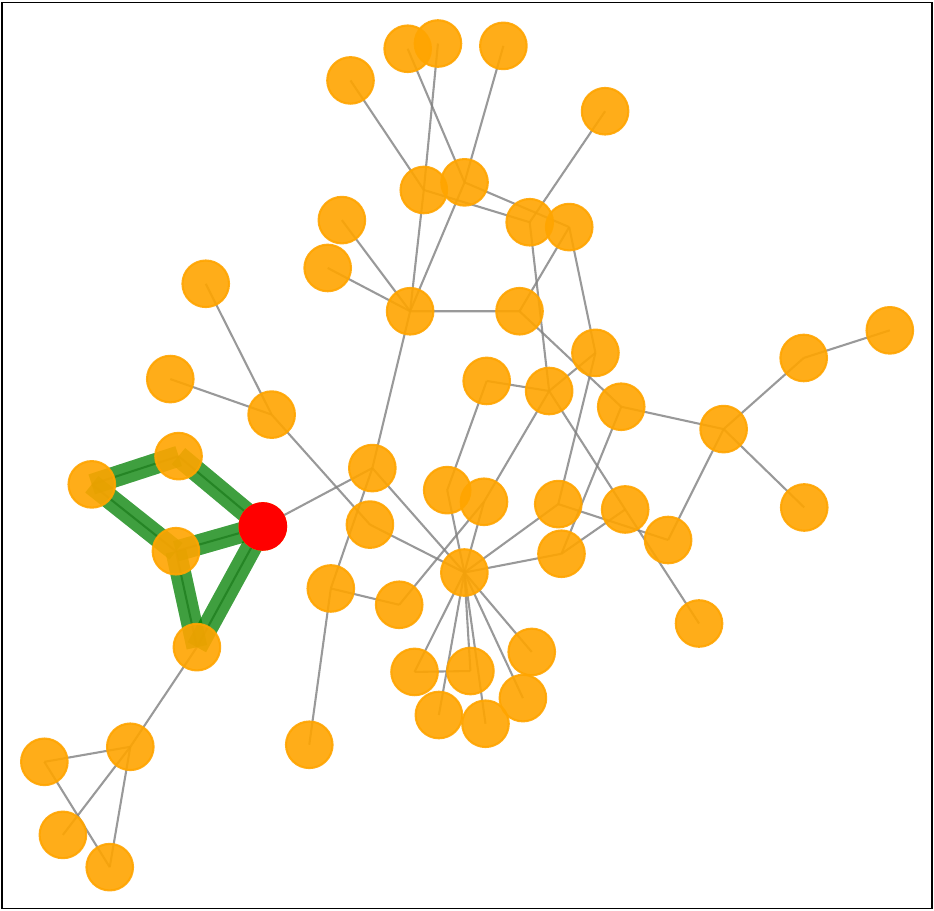} } & 
		\multicolumn{1}{m{1.7cm}}{\includegraphics[width=1.7cm]{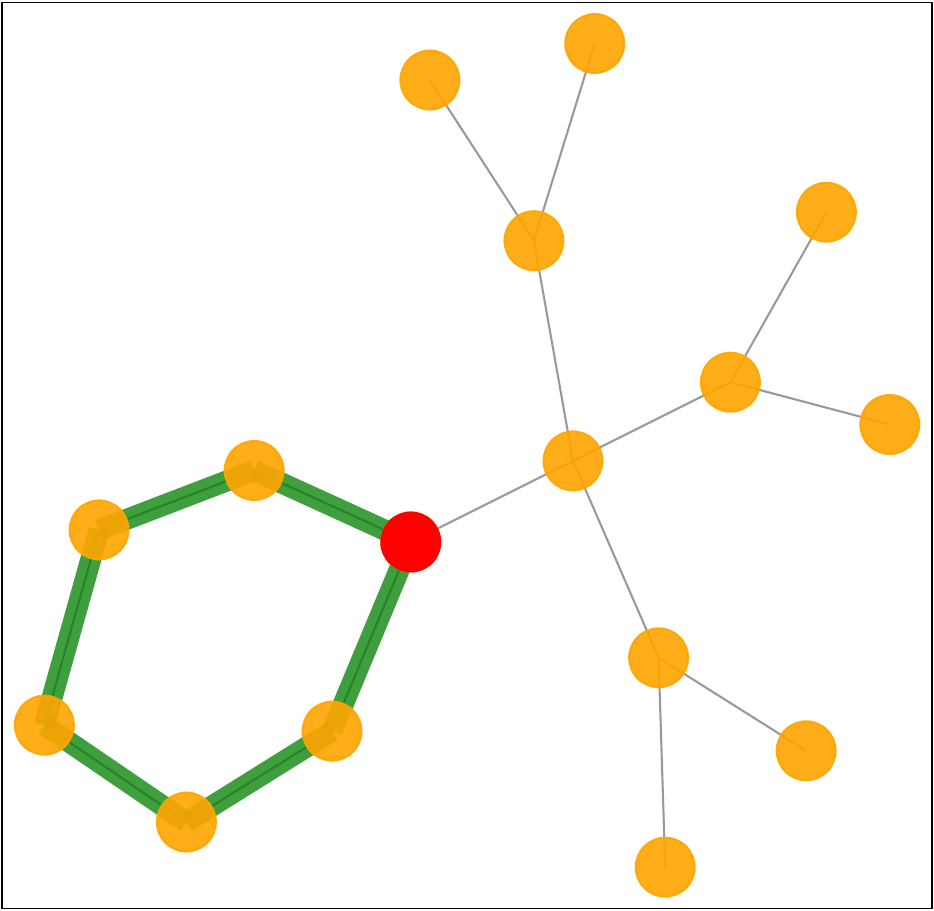} } & 
		\multicolumn{1}{m{1.7cm}}{\includegraphics[width=1.7cm]{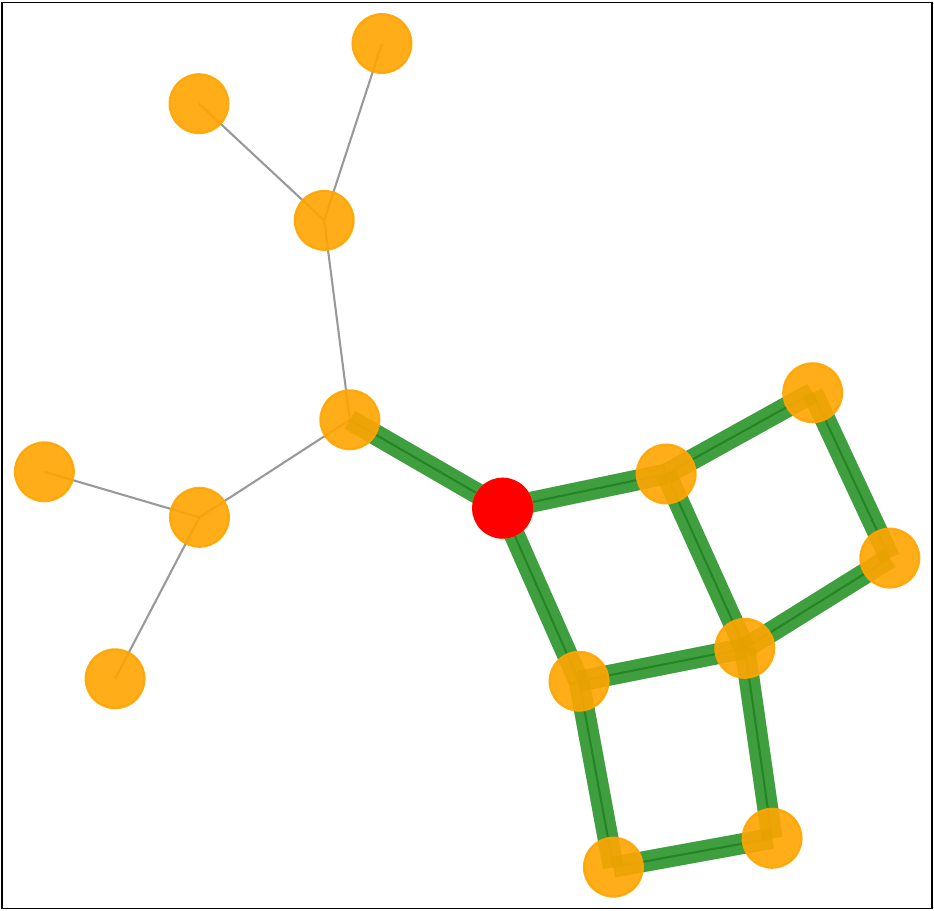} } & 
		\multicolumn{1}{m{1.7cm}}{\includegraphics[width=1.7cm]{scale/figures/graphs/pg_ba2motif_5.pdf} } & 
		\multicolumn{1}{m{1.7cm}}{\includegraphics[width=1.7cm]{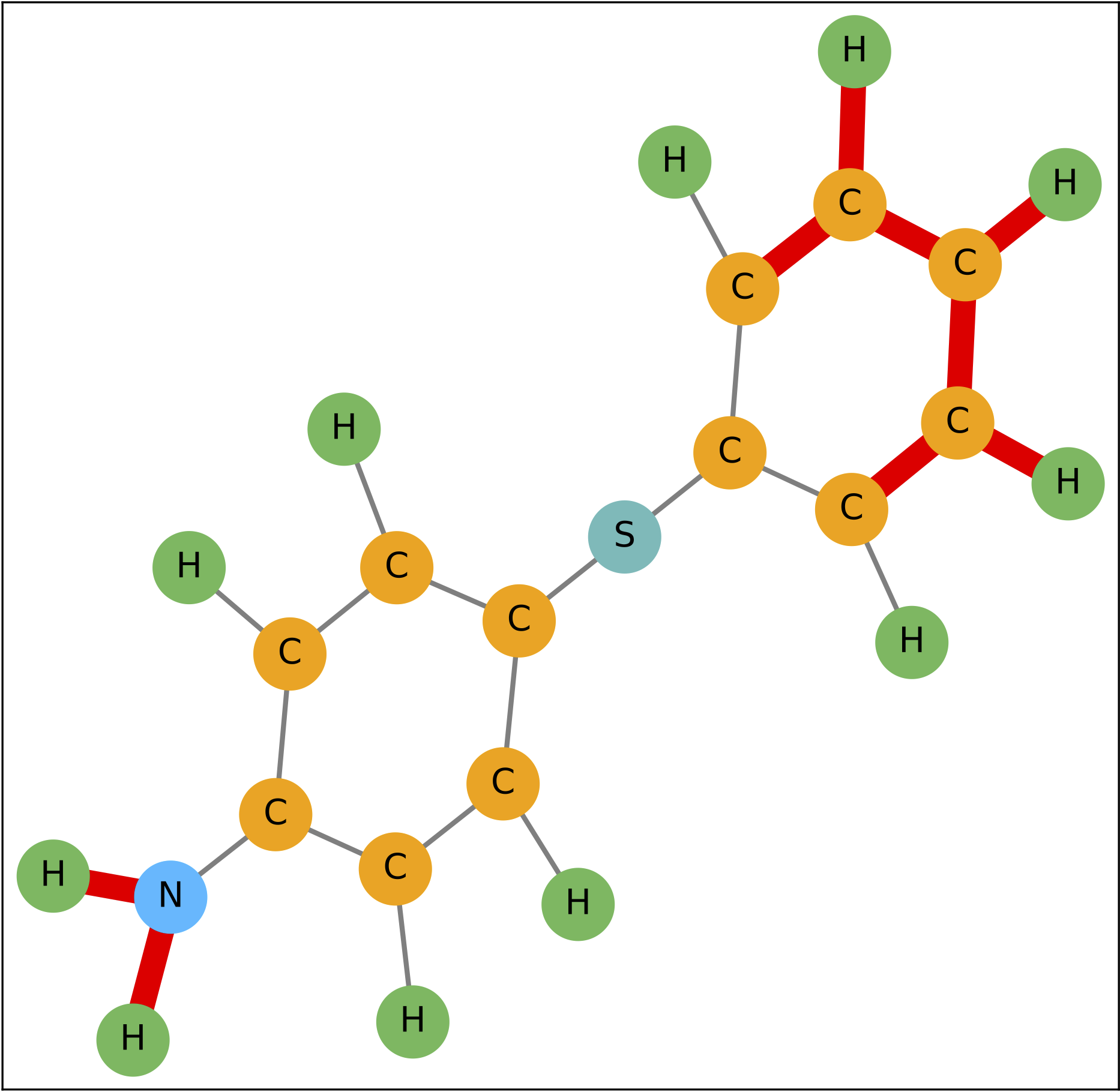}}
		\\
		
		\textbf{SC} & 
		\multicolumn{1}{m{1.7cm}}{\includegraphics[width=1.7cm]{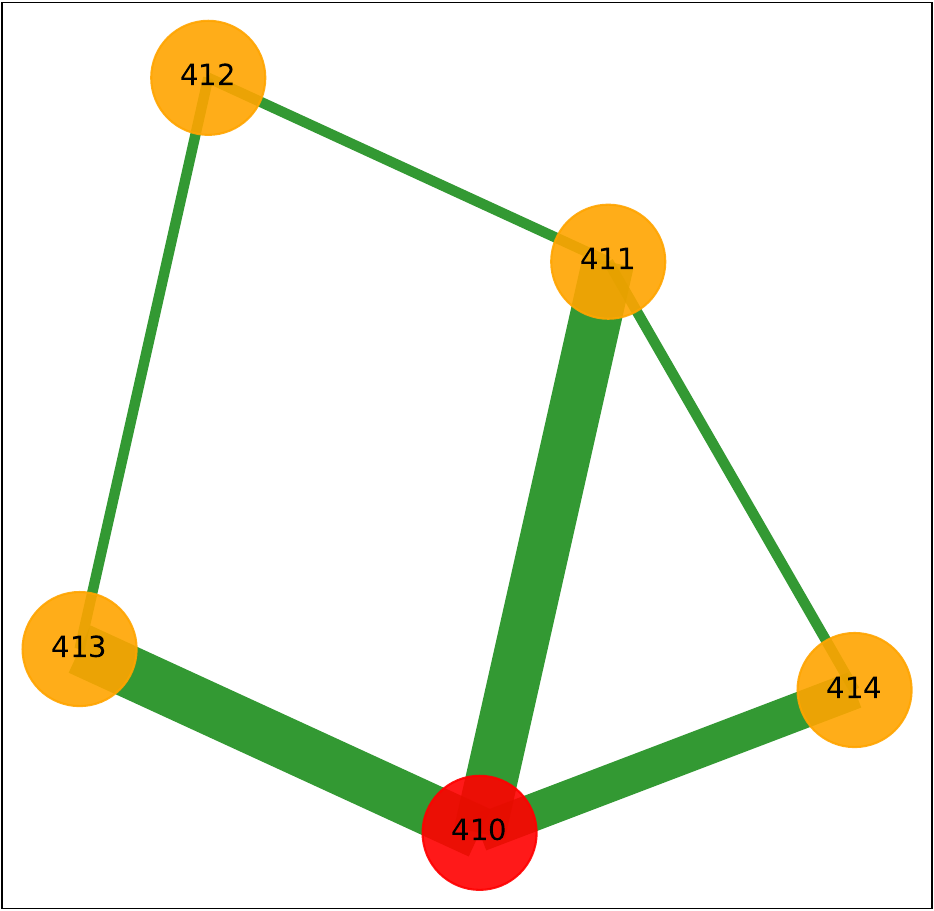}} & 
		\multicolumn{1}{m{1.7cm}}{\includegraphics[width=1.7cm]{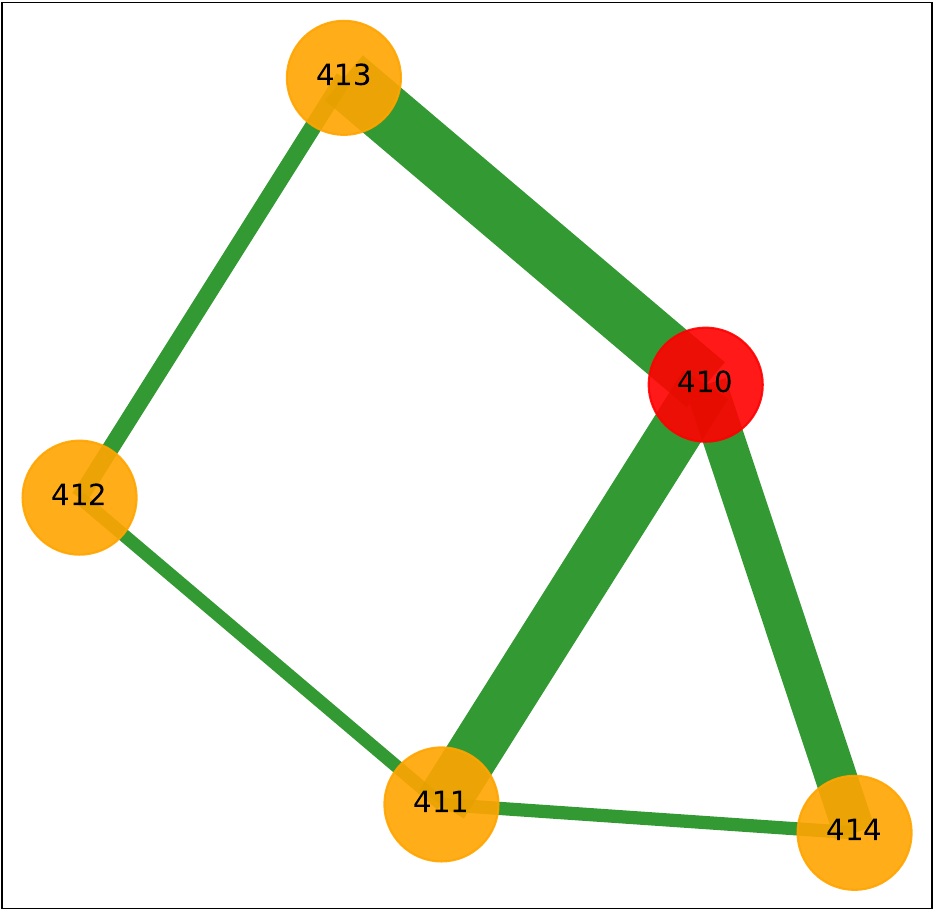}} & 
		\multicolumn{1}{m{1.7cm}}{\includegraphics[width=1.7cm]{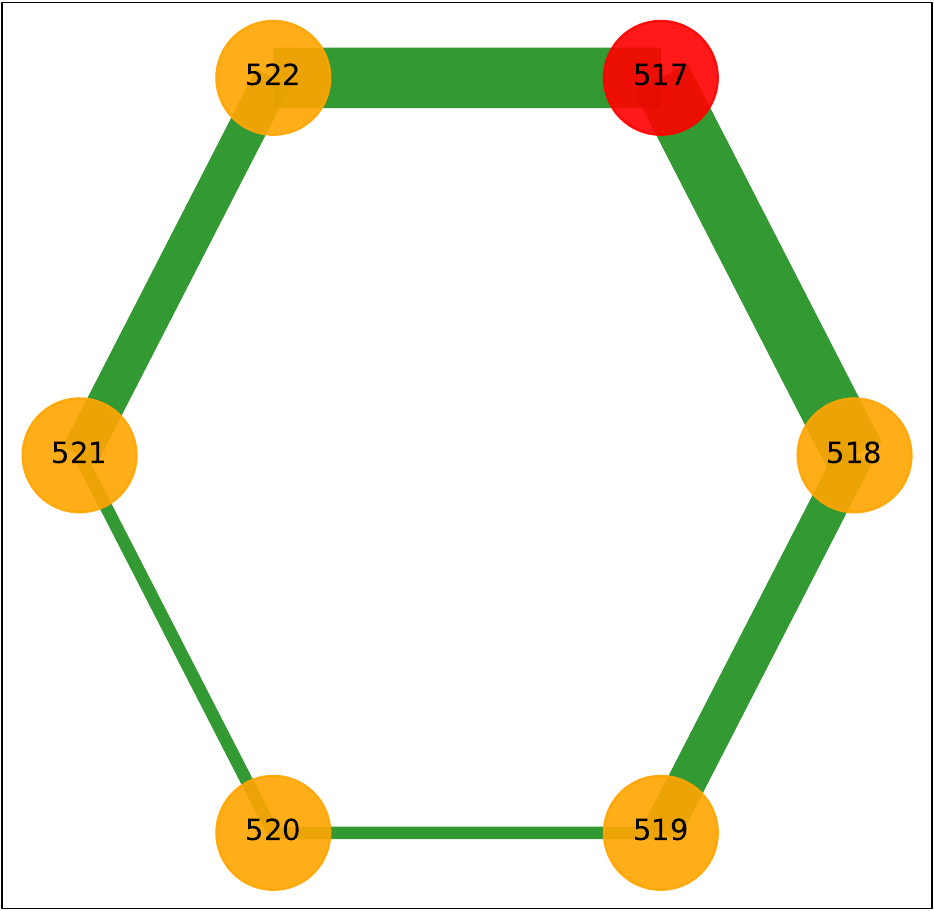}} & 
		\multicolumn{1}{m{1.7cm}}{\includegraphics[width=1.7cm]{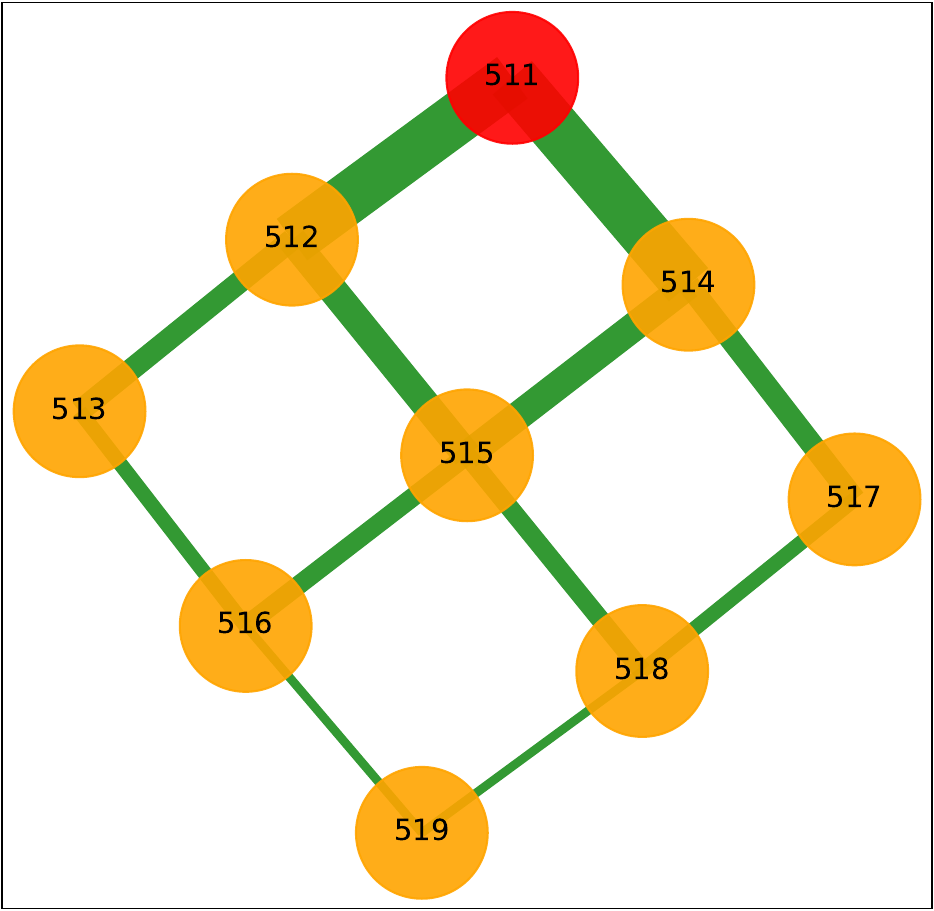}} & 
		\multicolumn{1}{m{1.7cm}}{\includegraphics[width=1.7cm]{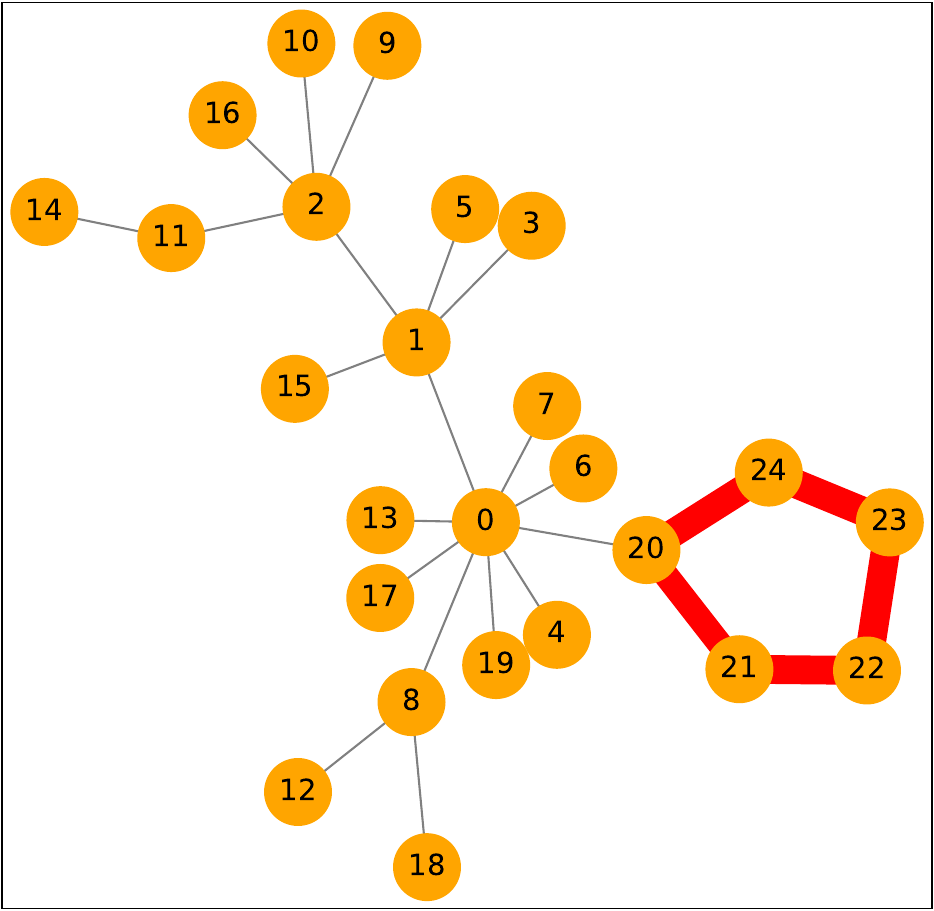}} & 
		\multicolumn{1}{m{1.7cm}}{\includegraphics[width=1.7cm]{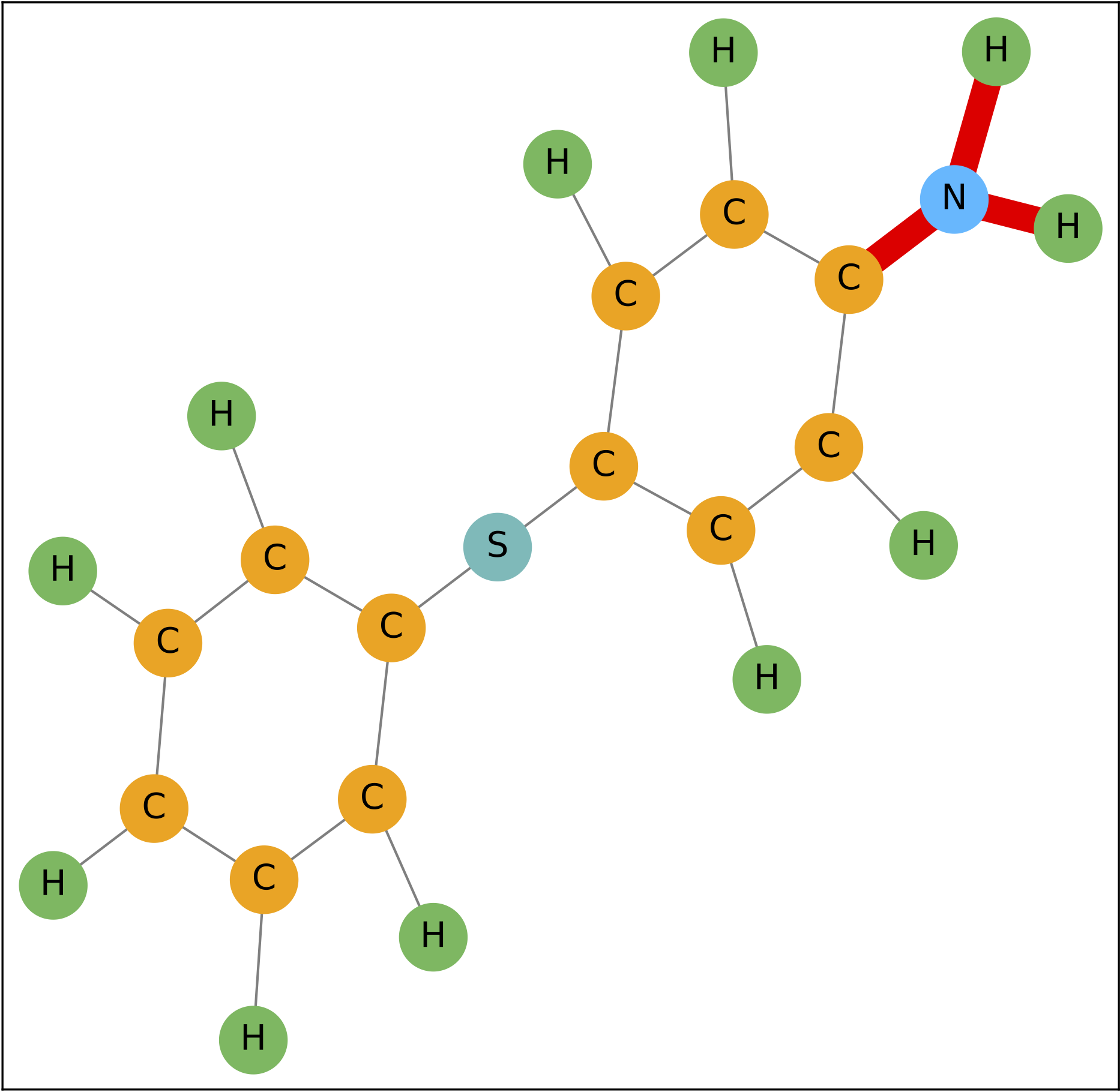}} 
		\\
		& \multicolumn{1}{c}{\scriptsize{(a) BA-S}} & 
		\multicolumn{1}{c}{\scriptsize{(b) BA-C}} & 
		\multicolumn{1}{c}{\scriptsize{(c) Tree-C}} & 
		\multicolumn{1}{c}{\scriptsize{(d) Tree-G}} & 
		\multicolumn{1}{c}{\scriptsize{(e) BA-2m}} & 
		\multicolumn{1}{c}{\scriptsize{(f) Mutag}} 
	\end{tabular}
	% paraphrased
	\caption{A Visualization of Ground Truth (GT) Explanations and Those Provided by GNNExplainer (GX), PGExplainer (PX), and SCALE (SC). Edges chosen for explanations are highlighted in red and green, with thicker edges signifying greater importance to the predictions. Only SCALE can differentiate edge importance in node classification scenarios.}
	\label{fig:qualitative_comparison}
\end{figure}

%visualizes explanations provided by SCALE, GNNExplainer, and PGExplainer. Like the two baselines, SCALE highlights crucial edges in graph classification explanations. However, SCALE achieves higher precision scores, resulting in fewer false positive edges in its explanations. Even though GNNExplainer and PGExplainer can highlight impactful edges in node classification explanations, they cannot differentiate the contributions of these edges since edge weights only represent selection probabilities. Conversely, SCALE visualizes edges with different widths, representing the probability that a random walker travels through these edges. This feature makes explanations highly intuitive, with thicker edges indicating greater importance to target nodes. Additionally, edges starting from direct neighbors within the same community received higher scores than those of distant neighbors. 

The visualization of explanations provided by GNNExplainer, PGExplainer, and SCALE is presented in \cref{fig:qualitative_comparison}. Similar to the other methods, SCALE identifies essential patterns and highlights them in graph classification explanations. However, its higher precision scores result in fewer false positive edges. While the compared counterparts can also highlight significant edges in explanations of node classification tasks, they fail to differentiate the neighbor contributions since the edge weights merely indicate selection probabilities. In contrast, SCALE represents edges with varying widths, corresponding to the probability that a random walker will traverse these edges in its paths. This feature enhances the comprehension of explanations, with thicker edges signifying a larger influence on the target nodes. Furthermore, edges originating from adjacent neighbors are assigned higher scores compared to those from distant neighbors.

\begin{figure}[ht]
	\centering
	\setlength\tabcolsep{3pt}
	\begin{tabular}{r p{2cm}p{2cm}p{2cm}}
		\textbf{Tree Grid} &
		\multicolumn{1}{m{2cm}}{\includegraphics[width=2cm]{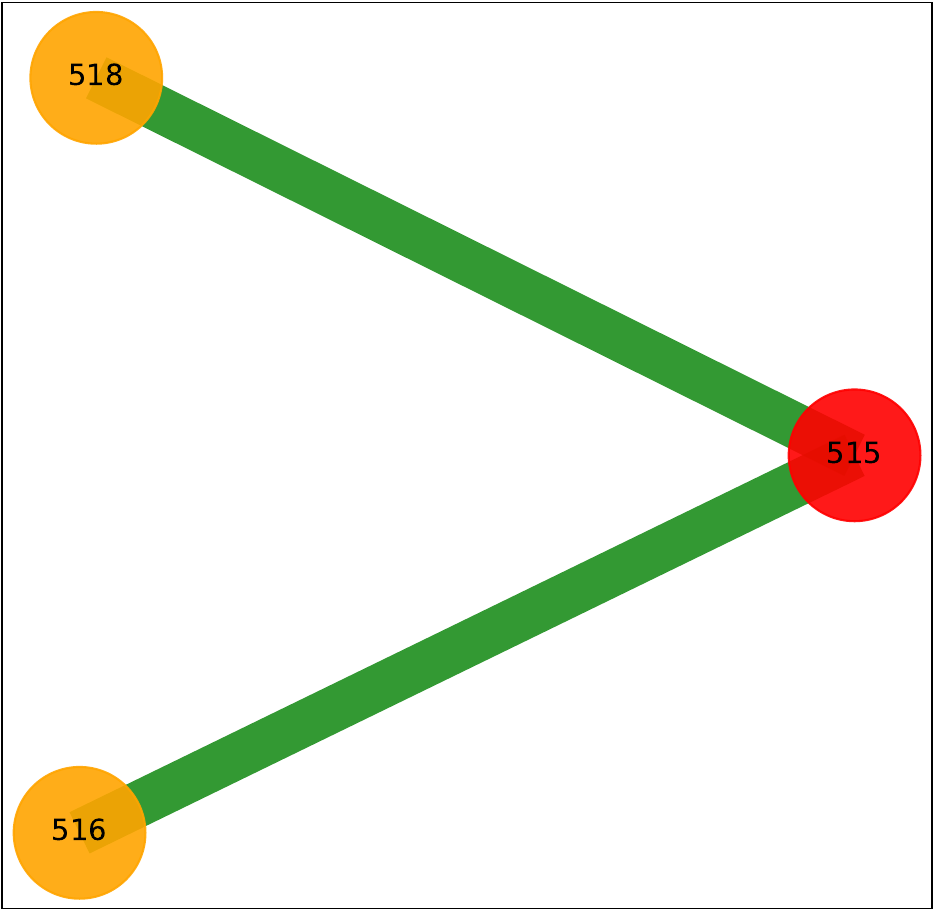} } & 
		\multicolumn{1}{m{2cm}}{\includegraphics[width=2cm]{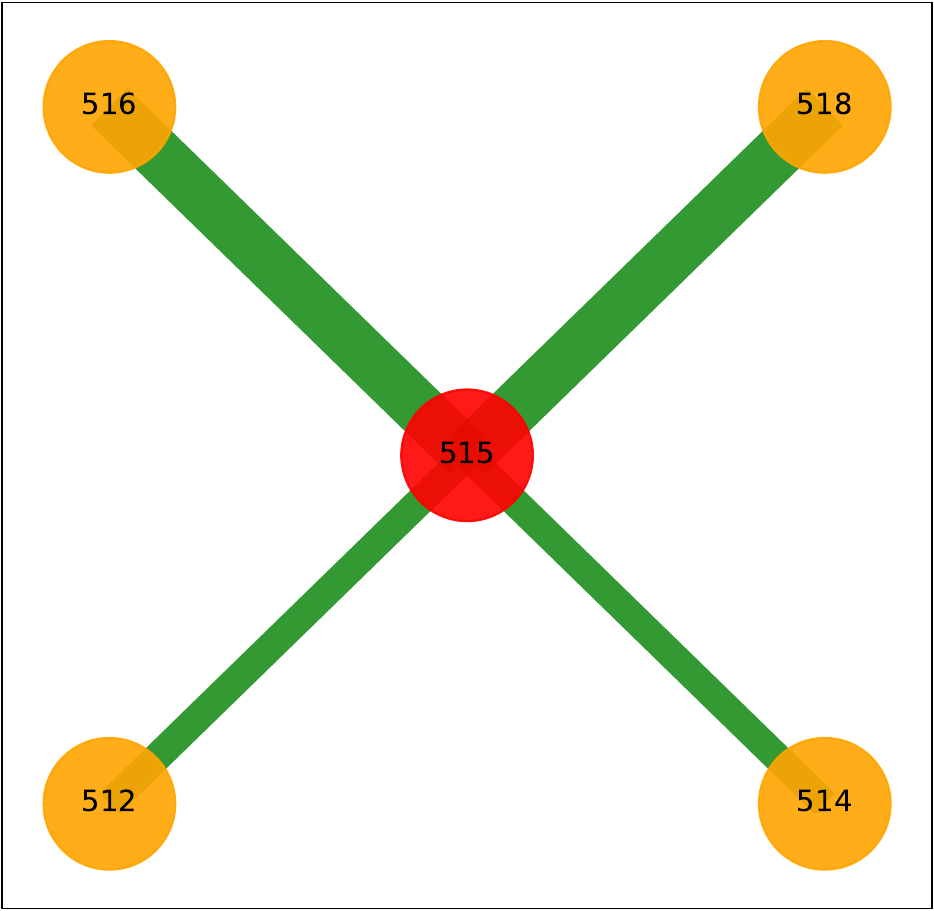} } & 
		\multicolumn{1}{m{2cm}}{\includegraphics[width=2cm]{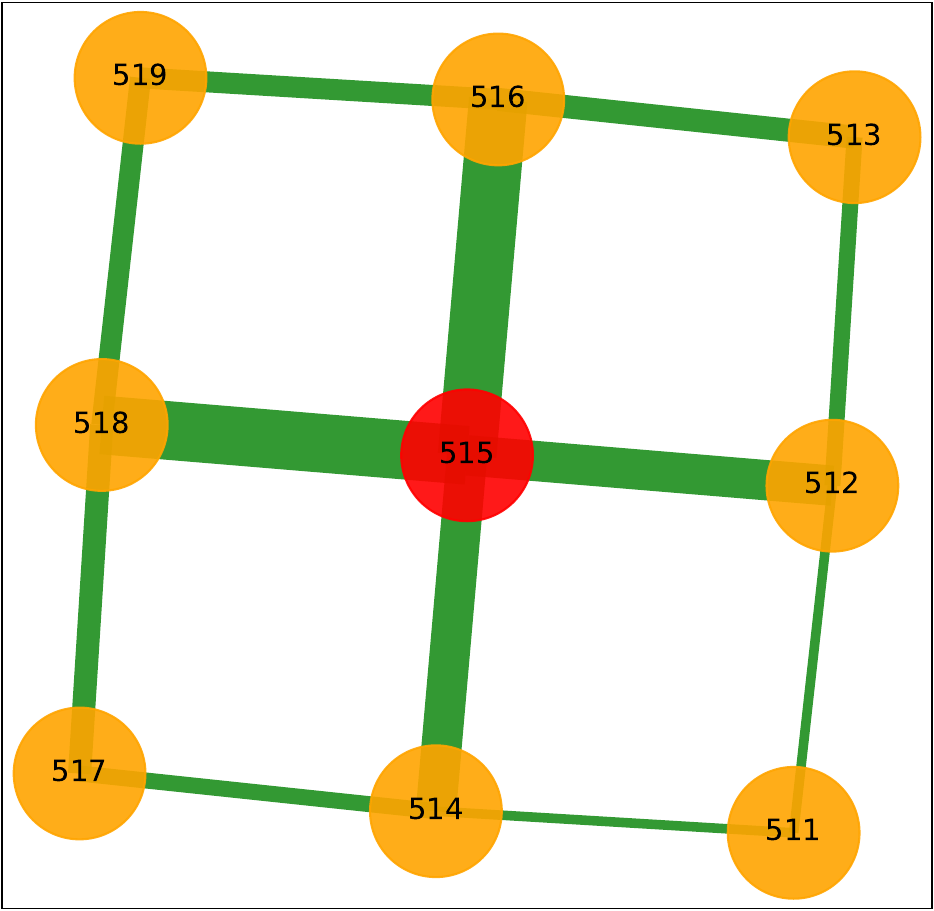} }
		\\
		
		& \multicolumn{1}{c}{\scriptsize{(a) K = 2}} & 
		\multicolumn{1}{c}{\scriptsize{(b) K = 4}} & 
		\multicolumn{1}{c}{\scriptsize{(c) K = 8}} \vspace{0.1cm} \\
		
		\textbf{Tree Cycle} & 
		\multicolumn{1}{m{2cm}}{\includegraphics[width=2cm]{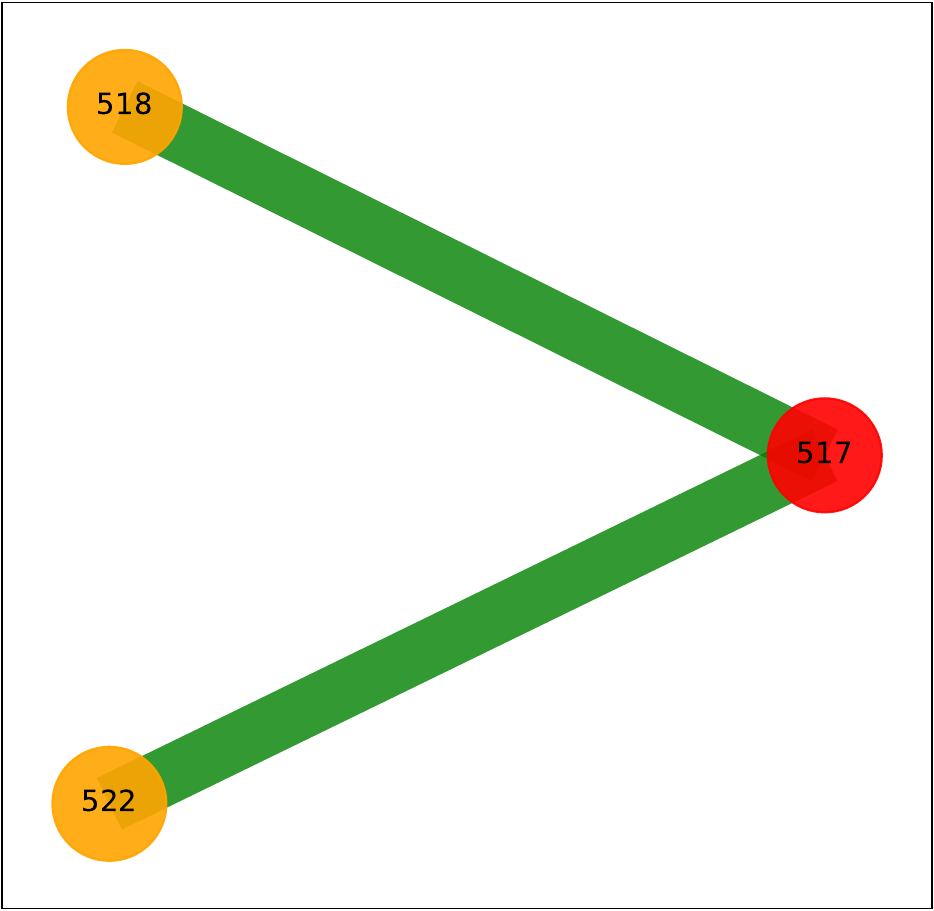}} & 
		\multicolumn{1}{m{2cm}}{\includegraphics[width=2cm]{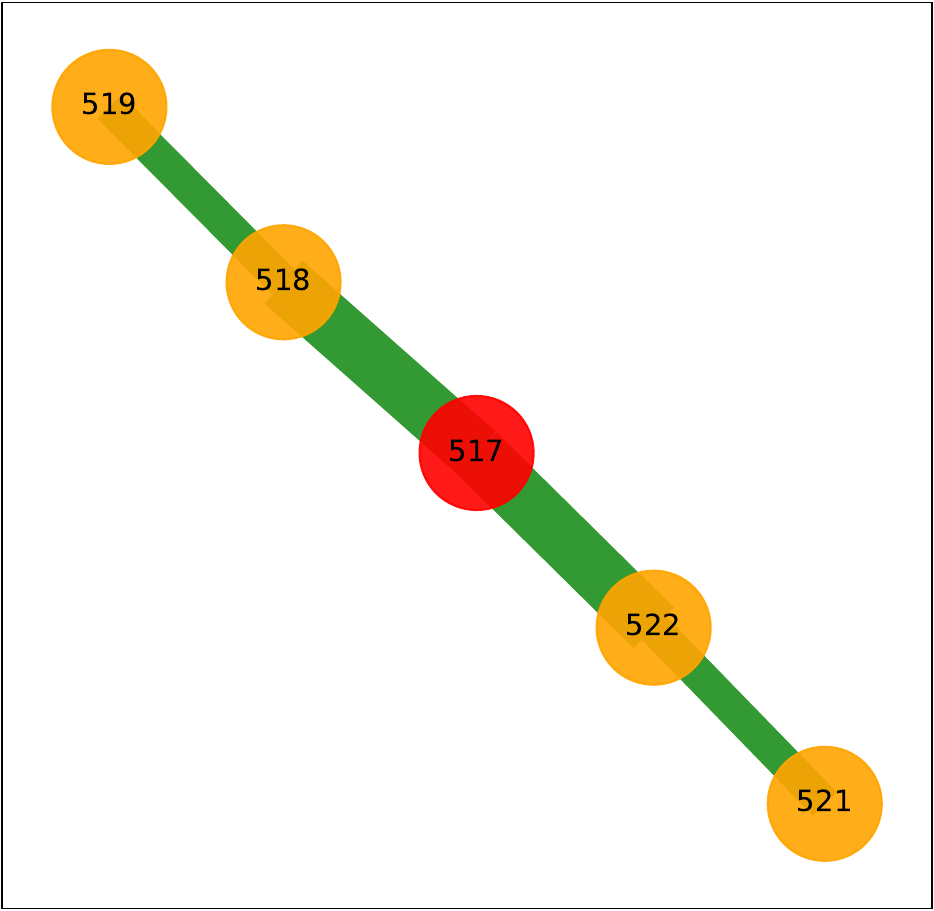}} & 
		\multicolumn{1}{m{2cm}}{\includegraphics[width=2cm]{scale/figures/graphs/tree_cycle_517_l3.pdf}} 
		\\
		& \multicolumn{1}{c}{\scriptsize{(c) K = 2}} & 
		\multicolumn{1}{c}{\scriptsize{(d) K = 4}} & 
		\multicolumn{1}{c}{\scriptsize{(f) K = 5}} 
	\end{tabular}
	
%	\caption{Multi-level Expansion of Structural Explanations for Node-level Predictions. K denotes the number of top influential nodes sorted by importance scores.}
	\caption{Expansion of Node Prediction Explanations at Multiple Levels. The variable K represents the number of the most influential nodes, ranked according to their importance scores.}
	\label{fig:multi_level}
\end{figure}

%Multi-level expansion of explanations is critical in various scenarios, such as recommender systems. However, adjusting the visibility threshold in GNNExplainer and PGExplainer to present explanations on multiple levels may result in outputs with multiple disconnected components, given the independence of edge selections. Furthermore, edges originating from direct (1-hop) neighbors may have lower selection probabilities than those from 2-hop or 3-hop neighbors. The maximum number of hops is also predefined at the sampling step before training explainable models, limiting the extent to which explanations can be expanded. In contrast, SCALE's explanations are more intuitive, as presented in \cref{fig:multi_level}. Close neighbors receive higher scores than distant ones, and explanations can be expanded without limitations by adjusting the visibility threshold or displaying influential neighbors layer by layer.

In various contexts, such as recommender systems, expanding explanatory subgraphs at different levels is essential. Even though GNNExplainer and PGExplainer can generate multi-level explanations by modifying the selection threshold of edge weights, this approach may provide explanatory patterns with several disconnected components. Additionally, 1-hop neighbors' edges may exhibit lower probabilities compared to distant neighbors' links. The predefined hop value further restricts the dynamic expansion potential of explanations. Conversely, SCALE's explanations, as illustrated in \cref{fig:multi_level}, are more intuitive. Immediate neighbors are assigned higher scores than distant ones. Furthermore, SCALE supports the dynamic expansion of explanations by simply adjusting the visibility threshold or inspecting essential nodes layer by layer.

% modified
\subsection{Effectiveness of RWR}

%This study investigated the effectiveness of \cref{querying_rwr} by applying it to adjacency matrices provided by GCN-MLP and GAT. Naively selecting important edges for node classification of target nodes proved inefficient, especially in complicated graphs. Therefore, the explanation results given by this approach on the original GCN-MLP and GAT were inaccurate in Tree-Cycle and Tree-Grid datasets where ground-truth motifs were complicated. However, the integration of \cref{querying_rwr} on these models significantly improved explanation correctness, as presented in the last two rows of \cref{tab:quant_compare}. Nevertheless, the improved results still fell short of those achieved by SCALE. The correctness of edge weights, which represent the influences of neighbors, heavily impacted the RWR procedure and the quality of explanations. The proposed online KG paradigm enables a self-explainable GNN to capture interactions between nodes efficiently, resulting in more accurate edge weights.

This experiment examined the efficacy of \cref{querying_rwr} by applying it to the learnable adjacency matrix of GCN-ML and the aggregated version of attention heads in GAT. Naive selections of influential edges for node classification of target nodes were found to be inefficient, particularly in complex graphs. Consequently, the explanatory graphs generated by this approach for predictions of GAT and GCN-MLP were less accurate in Tree-Grid and Tree-Cycle datasets, where ground-truth patterns are intricate. As shown in the final two rows of \cref{tab:quant_compare}, incorporating \cref{querying_rwr} into these models considerably enhanced the accuracy of explanations. Despite this improvement, the results still did not match the performance of SCALE. The magnitude of edge weights, which reflect the influence of neighbors, significantly affected the RWR algorithm and the quality of explanatory graphs. The training paradigm proposed in this chapter enables an interpretable GNN to effectively measure edge importance in message-passing operations, leading to more precise edge weights.

\subsection{User Comprehension of Structural Explanations} \label{sec:user_test}
%This study aimed to evaluate the user perception of structural explanations for node classification, addressing two questions: (1) How effective are explanations in improving user comprehension of predictions? (2) What additional information can be integrated to enhance comprehension? 
\noindent\textbf{Objectives.} This experiment sought to evaluate user comprehension of explanatory graphs for node classification, focusing on two primary questions: (1) How effectively do these explanations enhance user understanding of predictions? (2) What information can be integrated to further augment comprehension?

% Procedure: how to conduct the test and hire people?  Goal of this test?
%This test was organized as a color prediction competition, in which participants competed with each other to win a prize of a 30\$ coupon gift for each task. A hypothesis was that the contest format would incentivize participants to try their best to win the prize. Additionally, this format also prevented participants from using their prior knowledge of GNN explanation methods. Here is the competition procedure:
\noindent\textbf{Procedure.} The examination was structured as a color prediction game where players could win a \$30 coupon for each challenge. It was hypothesized that the competitive nature would motivate participants to exert maximum effort to get the reward. Furthermore, this format mitigated the likelihood of participants relying on their pre-existing understanding of GNN explanation techniques. The following outlines the competition's procedure:

\begin{itemize}
%	\item \textbf{Step 1:} Nodes located near the decision boundaries of the Cora graph were randomly selected. Then, subgraphs with structural explanations generated by various methods were then presented. Nodes in each subgraph were colored corresponding to predicted labels, except the target node.
%	\item \textbf{Step 2:} Participants had to predict the correct color for a target node based on the colors of its neighbors and a provided explanation. Each player answered the same number of questions for fair comparisons.
%	\item \textbf{Step 3:} Participants' predictions were compared to the actual model predictions to measure accuracy scores. The participant with the highest score won a task in the competition.
		\item \textbf{Step 1:} A random selection of nodes near the decision boundaries within the Cora graph was chosen. These nodes were then incorporated into subgraphs containing explanatory graphs constructed by different methods. Within each subgraph, all nodes except the target node were colored according to predicted labels.
		\item \textbf{Step 2:} Players were asked to determine the target color by analyzing the colors of the node's neighbors and presented explanations. To ensure fairness, each participant played with the same set of queries.
		\item \textbf{Step 3:} Players' predictions were evaluated by comparing them to the model's predictions. The one who most consistently aligned with the model's predictions was declared the winner of the game.
\end{itemize}

%paraphrased
\noindent This study adopted the forward simulation approach from \cite{doshi2017towards}. The game has four independent rounds, as follows:

\begin{itemize}
	\item \textbf{W/O Explanation:} Edges weights are invisible to participants. It was hypothesized that attendants would choose a color for a target node randomly from the colors of its neighbors.
	\item \textbf{GNNExplainer:} Participants were shown with subgraphs containing both node colors and selection probabilities of edges determined by GNNExplainer. The objective was to evaluate the effectiveness of GNNExplainer's explanations in assisting users with their predictions. Prior to the task, all attendants were briefed on the meaning of edge weights.
	\item \textbf{PGExplainer:} Players were presented with subgraphs containing both node colors and PGExplainer's selection probabilities of edges. The step-by-step process was the same as for the GNNExplainer setting.
	\item \textbf{SCALE:} Attendants were presented with SCALE's explanations for predictions. Detailed descriptions were given to clarify the differences in meaning between SCALE's edge weights and those of the other two methods to avoid biases and confusion.
	
\end{itemize}

\noindent In each round, the moderator accepted only the first 33 submissions sorted by submission time. The game was promoted within several research communities, resulting in 132 submissions from 41 unique participants.

\begin{figure}[ht]
	\centering
	\includegraphics[width=0.6\linewidth]{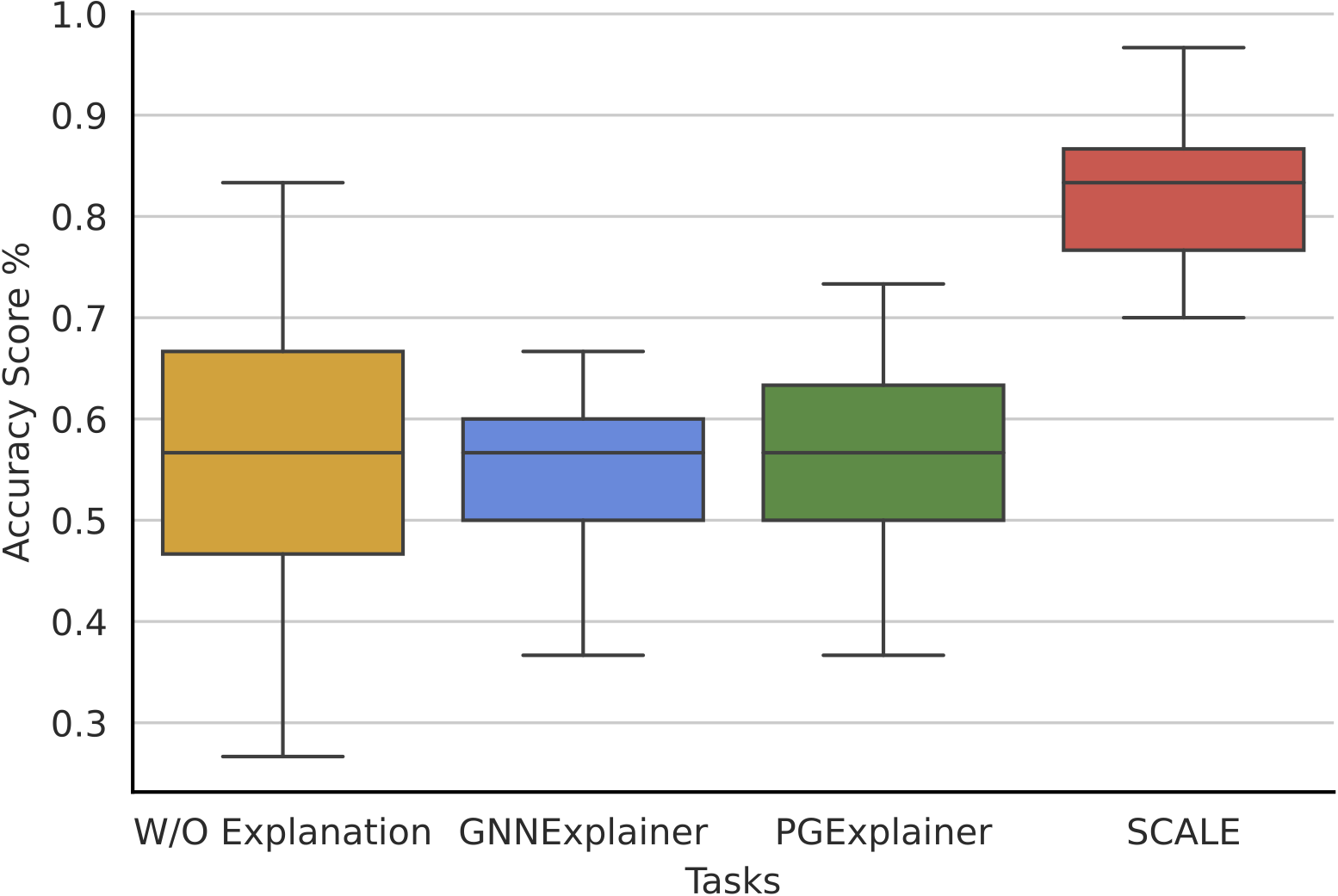}
%	\caption{User Prediction Accuracy in Four Tasks. SCALE's explanations improve user understanding of model predictions, leading to outstanding accuracy scores.}
	\caption{Performance of User Predictions. Quantitative information and interactive explanations significantly boost user performance.}
	\label{fig:user_test}
\end{figure}

%The results confirmed the hypothesis that participants' decisions would be entirely random in the first task. This randomness resulted in significantly varying prediction accuracies, as illustrated in \cref{fig:user_test}. Despite players' understanding of the meaning of edge weights in GNNExplainer and PGExplainer tasks, the weights sometimes confused their decisions. This confusion arose because multiple edges could have the same weights, making it challenging for participants to select the correct colors. Although the average accuracy scores were almost identical for both GNNExplainer and PGExplainer tasks, PGExplainer caused more confusion in explanation weights. The observations indicated that PGExplainer's reparameterization trick \cite{luo2020parameterized} led the selection probabilities to approach 1 in most cases, making it difficult for participants to differentiate edge influences. As a result, user prediction accuracies varied more in the PGExplainer task than in the GNNExplainer task. In the SCALE task, participants aggregated the influence values of neighbors and selected the color corresponding to the highest value most of the time. This strategy helped them achieve the highest accuracy scores in the final task, outperforming the results in other tasks. This result emphasized the helpfulness of SCALE's explanations on user decisions.
\noindent\textbf{Observations.} The findings confirmed that players' decisions in the initial task would be completely arbitrary. This arbitrariness led to substantial variations in accuracies, as demonstrated in \cref{fig:user_test}. Even though participants comprehended the significance of edge weights in both PGExplainer and GNNExplainer tasks, these values occasionally perplexed their predictions. This confusion emerged because several edge weights could possess identical values, complicating the selection of the correct colors. Despite the nearly identical average prediction accuracies for the two tasks, PGExplainer induced more confusion regarding these weights. An observation was that the reparameterization trick \cite{luo2020parameterized} caused weight values to converge towards 1 in most instances, thereby hindering participants' ability to discern edge importance. Consequently, user prediction performance fluctuated more in the PGExplainer task compared to the GNNExplainer one. In the SCALE task, participants grouped up the influence values of neighboring nodes by colors and predominantly chose the color associated with the highest value. This approach enabled players to obtain the best prediction scores in the final task, surpassing the outcomes of other tasks. These results underscored the effectiveness of the proposed framework in generating useful explanations to aid user decisions.

%After the competition, participants provide constructive feedback. Most reported that selection probabilities \cite{ying2019gnnexplainer,luo2020parameterized} on edges were confused in many situations since edges of distant neighbors sometimes have even higher values. Some players recommended incorporating contributions of node features to aid in tasks since node features can significantly influence predicted classes. Participants also suggested integrating multiple information sources on the same screen to reduce cognitive load and enhance comprehension.

\noindent\textbf{User Suggestions.} Following the game, users gave us constructive feedback. The majority noted that the selection probabilities \cite{ying2019gnnexplainer,luo2020parameterized} for edges were often perplexing, as distant neighbors exhibited even higher values sometimes. Several participants suggested including the contributions of node features to assist with tasks, given that these elements can substantially impact predictions. Additionally, participants suggested that displaying different information on the same page could alleviate cognitive load and improve understanding.

\subsection{An Assessment of Feature Attribution Component}

%This study aims to evaluate the efficiency of the feature attribution module using the Amazon dataset, which contains intelligible node features. Each node in the Amazon graph has 25 statistical properties representing the reviewing behaviors of users for products on the Amazon platform. Fraudsters or attackers are users who try to cheat the recommender system to promote particular products while also attempting to be like benign users as much as possible. Since ground-truth explanations do not exist, a comparison was made between feature attributions provided by SCALE and the insights discovered by Zhang et al. \cite{zhang2020gcn}.

This experiment sought to evaluate the efficacy of the feature attribution function using the Amazon dataset, which includes comprehensible node features. A vertex in this graph is characterized by twenty-five statistical properties that reflect users' reviewing behaviors for products on the Amazon website. Fraudulent users are those who attempt to deceive the recommendation engine to boost the ranking of specific products while striving to resemble normal users as closely as possible. In the absence of ground-truth explanations, a comparison was conducted between the insights extracted from the framework's generated attributions and those uncovered by Zhang et al. \cite{zhang2020gcn}.

\begin{figure}[ht]
	\centering
	\includegraphics[width=0.6\linewidth]{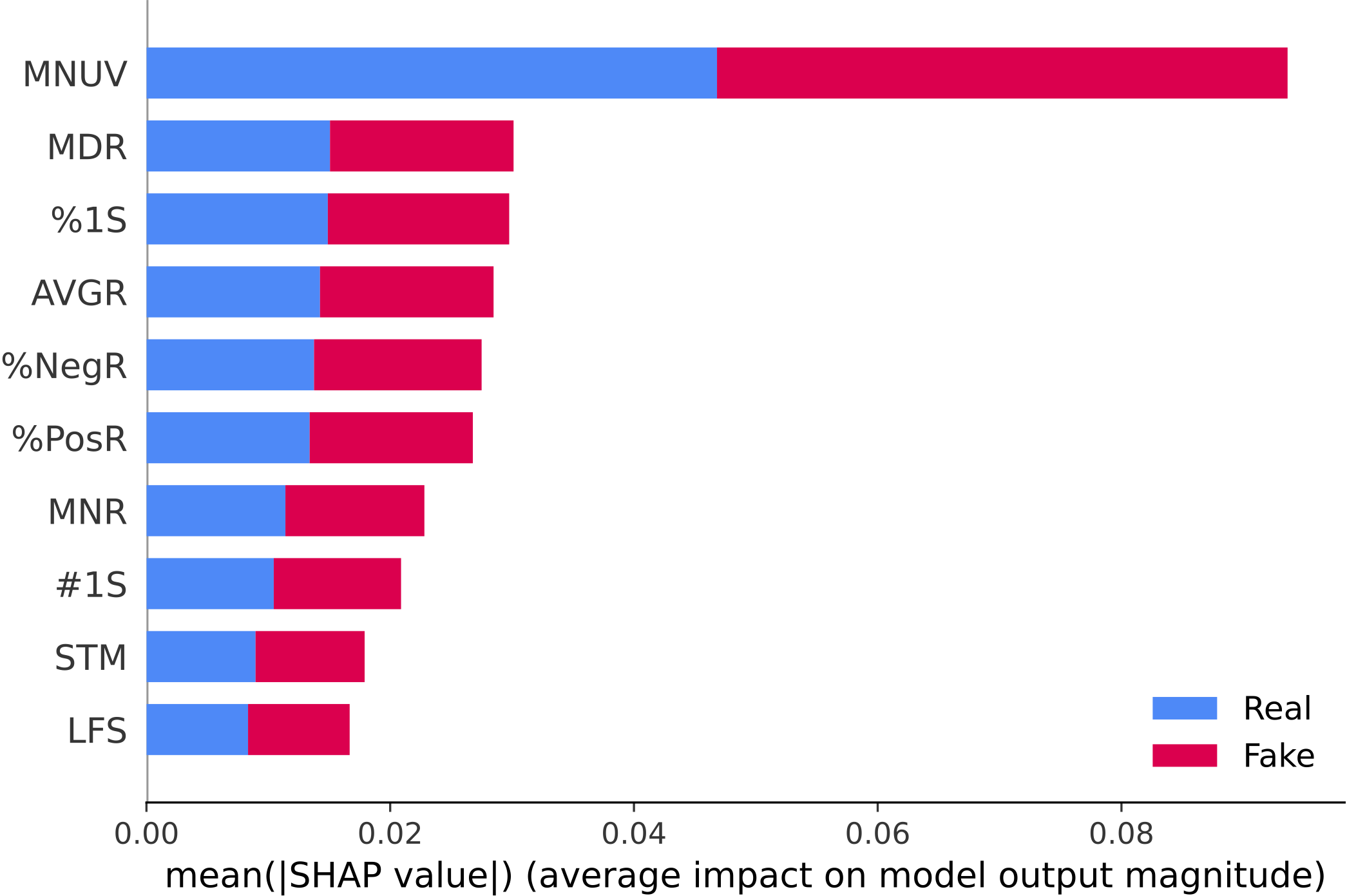}
%	\caption{An Overall Summary of Feature Attributions for Node Classification on Amazon Dataset. It summarizes the average impact of features on predictive probabilities. The longer the bar, the more influential a feature is.}
	\caption{Feature Attribution Summary on Amazon Dataset with the Node Classification Task. This summary presents the average feature importance, with longer bars indicating more influential features.}

	\label{fig:feature_importance}
\end{figure}

\begin{figure}[ht]
	\centering
	\subfloat[Fraudulent User Class]{
		\includegraphics[width=0.7\linewidth]{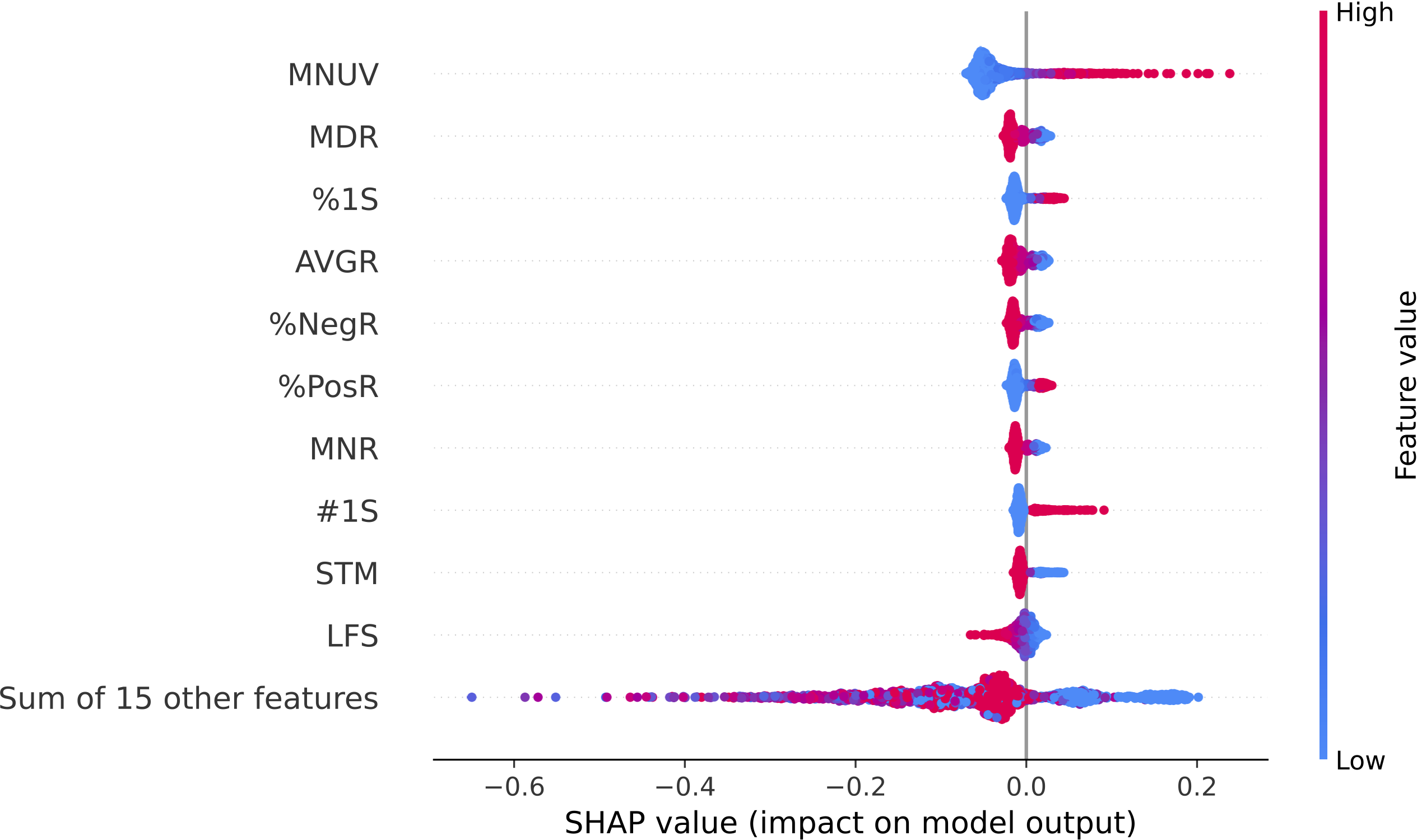}
		\label{fig:feat_a}
	}
	\hfil
	\subfloat[Benign User Class]{
		\includegraphics[width=0.7\linewidth]{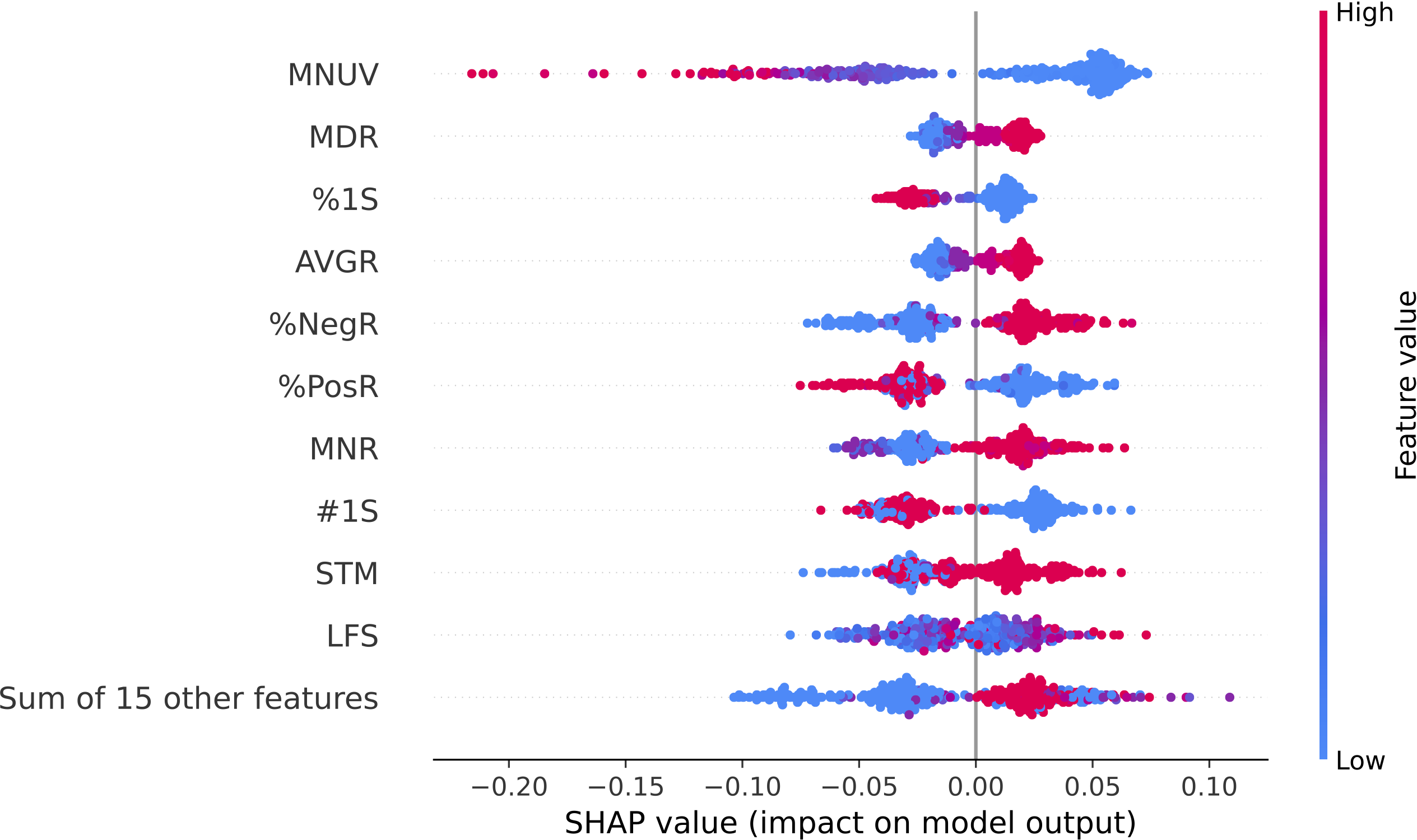}
		\label{fig:feat_b}
	}
%	\caption{A Summary of Value-impact Relationships for Each Class in Amazon Dataset. Y-axes are feature names, and X-axes are feature impact values. The color bar in the bottom figures represents the magnitude of values (redder - bigger, bluer - smaller). All figures have the same order of features. }
	\caption{An overview of each class's value-impact relationships from the Amazon dataset. X-axes are feature impact values, while Y-axes are feature names. The magnitude of values is demonstrated by the color bar (larger is represented by a red bar, and smaller by a blue bar).}
	\label{fig:feature_importance_2}
\end{figure}

\begin{figure}[ht]
	\centering
	\hfil
	% \vspace{-0.2cm}
	\subfloat[Fraudulent User Class]{
		\includegraphics[width=0.9\linewidth]{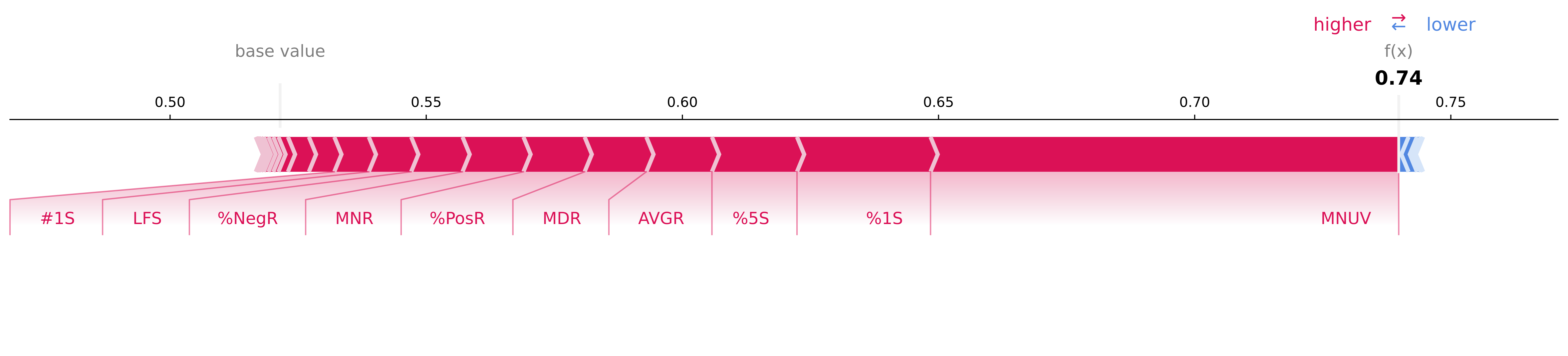}
		\label{fig:example_real}
	}
	\hfil
	% \hspace{-0.3cm}
	\subfloat[Benign User Class]{
		\includegraphics[width=0.9\linewidth]{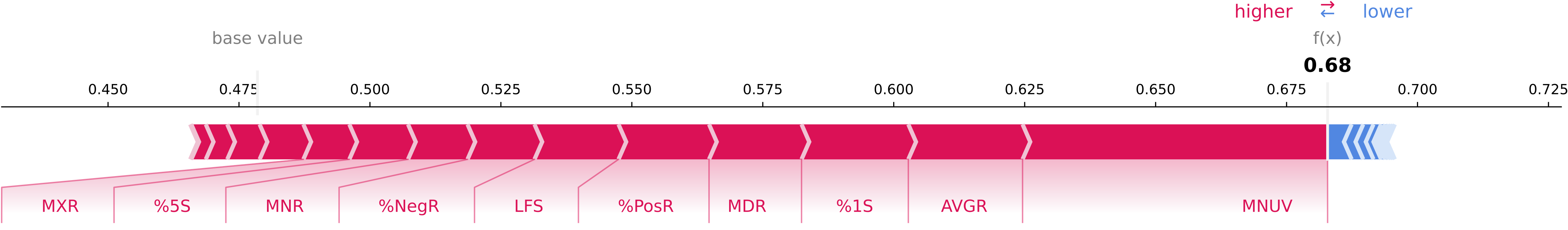}
		\label{fig:example_fake}
	}
	\hfil
%	\caption{Examples of Feature Attributions of Predictions for Each User Class. Red color bars present how much each feature increases the predictive probability, while blue color bars demonstrate the impact of features in the opposite direction.}
	\caption{A Visualization of Feature Contributions for Each Class. The red bars show the magnitude of positive contributions to prediction, whereas the blue ones demonstrate the negative impacts.}
	\label{fig:instance_explanations}
\end{figure}

%This section's figures present an overview of feature contributions and examples of instance-level explanations. Specifically, \cref{fig:feature_importance} summarizes the overall importance of features, while \cref{fig:feature_importance_2} provides further insight into the relationships between feature values and their impacts on predictive probabilities. Additionally, \cref{fig:instance_explanations} visualizes two examples of instance-level explanations for each class. The SHAP values in the figures represent the magnitude of marginal probability contributions. Here are descriptions of some features presented in these figures: \textbf{MNUV} - Minimum number of unhelpful votes; \textbf{MDR} - Median of ratings; \textbf{\%1S} - Ratio of 1-star votes; \textbf{AVGR} - Average of ratings; \textbf{\%NegR }- Ratio of negative ratings; \textbf{STM} - Sentiment of feedback; \textbf{LFS} - Length of feedback. Please refer to \cite{rayana2015collective, dou2020enhancing} for the complete description.

The figures in this section provide a comprehensive summary of feature contributions and include examples of instance-level explanations. Particularly, \cref{fig:feature_importance} illustrates the overall significance of the features, whereas \cref{fig:feature_importance_2} delves deeper into the connections between feature values and their effects on predictive probabilities. \cref{fig:instance_explanations} showcases two examples of explanations for individual predictions with each class. The SHAP values depicted in these figures indicate the extent of marginal probability contributions. Below are descriptions of a few features highlighted in these figures:\textbf{MNUV} - Minimum number of unhelpful votes; \textbf{MDR} - Median of ratings; \textbf{\%1S} - Ratio of 1-star votes; \textbf{AVGR} - Average of ratings; \textbf{\%NegR }- Ratio of negative ratings; \textbf{STM} - Sentiment of feedback; \textbf{LFS} - Length of feedback. For a full description, please check publications \cite{rayana2015collective, dou2020enhancing}.

%The presented reports led to the following observations. The minimum number of unhelpful votes plays a crucial role in model predictions, indicating clear differences between the voting patterns of the two user classes. As \cref{fig:feature_importance_2} shows, fraudulent users receive numerous negative votes from others since high \textbf{MNVU} values correspond to large probability contributions. Additionally, a large number/ratio of low-star ratings and feedback with negative sentiment increases the likelihood of a user being fraudulent. Conversely, regular users give fair ratings and reviews with neutral or positive sentiments. Furthermore, feedback summaries of fake reviews are usually shorter than those of real reviews. These observations align with findings by Zhang et al. \cite{zhang2020gcn} that attackers typically give a high rating to a target item (promoted one) and low ratings to other regular items. Therefore, the proposed method for examining feature contributions is effective and accurate.

The reported analyses led to several key insights. The minimum number of unhelpful votes significantly influences model predictions, revealing distinct voting patterns between two user classes. As illustrated in \cref{fig:feature_importance_2}, fake users get a large number of negative votes from others, with high \textbf{MNUV} values. Furthermore, a high number or proportion of low-star ratings and feedback exhibiting negative sentiment enhances the likelihood of a user being identified as fraudulent. In contrast, genuine users typically provide a balanced proportion of ratings and reviews with neutral or positive sentiments. Additionally, the length of feedback from fake reviews tends to be shorter compared to those of authentic ones. These observations are consistent with the results presented by Zhang et al. \cite{zhang2020gcn}, which indicate that attackers generally assign high ratings to a promoted item while giving low values to other genuine items. In conclusion, the proposed method for calculating feature contributions proves to be accurate and effective.

\subsection{Ablation Studies}

%A series of ablation studies were conducted to study different aspects of SCALE. First, the selection of appropriate keep-going probabilities for specific scenarios was investigated. Subsequently, the relationship between explanation correctness and changes in model accuracy was studied. Different KD settings were also evaluated to examine the efficiency of the online KD paradigm. Finally, an investigation was conducted into how distilled knowledge impacts explanations in different classification tasks.

Ablation studies were carried out to examine various aspects of the proposed framework. Initially, the investigation focused on determining suitable walking probabilities for particular scenarios. Following this, the relationship between the accuracy of explanations and variations in model accuracy was analyzed. Various KD settings were assessed to evaluate the efficiency of the proposed training paradigm. Lastly, the impact of distilled knowledge on explanation quality across tasks was explored.

\pgfplotsset{width=6.5cm,height=5cm}
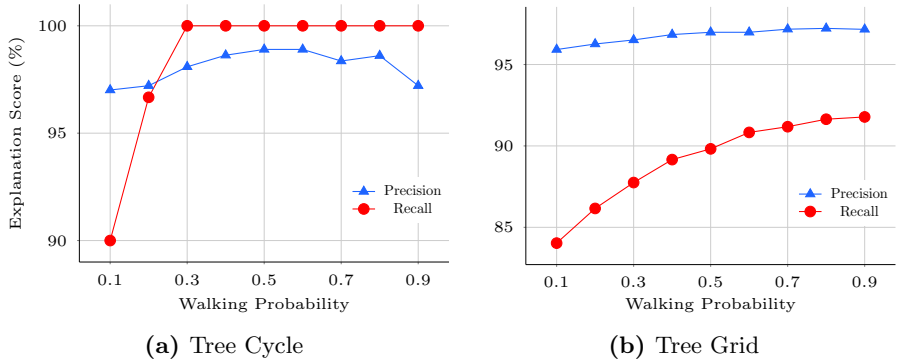
\begin{figure}[ht!]
	\centering
	\subfloat[Tree Cycle]{
		\begin{tikzpicture}
			\begin{axis}[
				grid=both,
				grid style={line width=.03pt, draw=gray!40},
				major grid style={line width=.03pt,draw=gray!40},
				mark size=2pt,
				legend style={draw=none,fill=white,at={(1.01,0.33)},anchor=north east,legend columns=1,nodes={scale=0.5, transform shape},column sep=0.08cm},
				x label style={at={(axis description cs:0.5,-0.1)},anchor=north},
				y label style={at={(axis description cs:-0.12,0.5)},anchor=south},
				xtick={0.1,0.3,0.5,0.7,0.9},
				xlabel=Walking Probability,ylabel=Explanation Score (\%)]
				\addplot[color={rgb,255:red,30;green,100;blue,255},mark=triangle*] coordinates{
					(0.1,97.01)
					(0.2,97.21)
					(0.3,98.09)
					(0.4,98.63)
					(0.5,98.9)
					(0.6,98.9)
					(0.7,98.36)
					(0.8,98.61)
					(0.9,97.21)
				};
				\addlegendentry{Precision}
				
				\addplot[color={red},mark=*] coordinates{
					(0.1,90.0)
					(0.2,96.67)
					(0.3,100)
					(0.4,100)
					(0.5,100)
					(0.6,100)
					(0.7,100)
					(0.8,100)
					(0.9,100)
				};
				\addlegendentry{Recall}
			\end{axis}
		\end{tikzpicture}
	}
	\hfil
	\hspace{-0.5cm}
	\subfloat[Tree Grid]{
		\begin{tikzpicture}
			\begin{axis}[
				grid=both,
				grid style={line width=.03pt, draw=gray!40},
				major grid style={line width=.03pt,draw=gray!40},
				mark size=2pt,
				legend style={draw=none,fill=white,at={(1.01,0.33)},anchor=north east,legend columns=1,nodes={scale=0.5, transform shape},column sep=0.08cm},
				x label style={at={(axis description cs:0.5,-0.09)},anchor=north},
				y label style={at={(axis description cs:-0.09,0.5)},anchor=south},
				xtick={0.1,0.3,0.5,0.7,0.9},
				xlabel=Walking Probability,ylabel=\empty]
				\addplot[color={rgb,255:red,30;green,100;blue,255},mark=triangle*] coordinates{
					(0.1,95.92)
					(0.2,96.26)
					(0.3,96.51)
					(0.4,96.84)
					(0.5,96.98)
					(0.6,96.98)
					(0.7,97.17)
					(0.8,97.22)
					(0.9,97.16)
				};
				\addlegendentry{Precision}
				
				\addplot[color={red},mark=*] coordinates{
					(0.1,84.03)
					(0.2,86.16)
					(0.3,87.75)
					(0.4,89.16)
					(0.5,89.82)
					(0.6,90.83)
					(0.7,91.18)
					(0.8,91.64)
					(0.9,91.78)
				};
				\addlegendentry{Recall}
			\end{axis}
		\end{tikzpicture}
	}
	\hfil
	
%	\caption{Relationships Between Walking Probability and Explanation Scores. A reasonable probability should be between 0.5 and 0.9.}
	\caption{Impacts of Walking Probability on Explanation Graphs. A plausible value should fall within the range of 0.5 to 0.9.}
	\label{fig:jumping_scores}
\end{figure}

%Experiments on Tree-Cycle and Tree-Grid datasets led to the following observations. Random walkers tend to restart more frequently with small probabilities, whereas exploring new states with larger values. As illustrated in \cref{fig:jumping_scores}, small keep-going probabilities cause low precision and recall scores, especially when multiple hops are required to complete the motifs. In the Tree-Cycle dataset, the precision score gradually improves as the keep-going probability increases to 0.6 but decreases when the probability exceeds this value. For the Tree-Grid dataset, precision and recall scores correspond to the magnitude of the keep-going probability due to the complexity of grid motifs, which requires long walks to traverse all nodes in ground-truth motifs. Therefore, a large probability is appropriate for the Tree-Grid case, while a value between 0.5 and 0.6 is better for the Tree-Cycle dataset. In practice, a reasonable value can range from 0.5 to 0.9 depending on the characteristics of graphs in particular scenarios.

\noindent\textbf{Effects of Walking Probability on Explanatory Graphs.} The experiments conducted on Tree-Grid and Tree-Cycle datasets yielded several insights. Random walkers exhibit a tendency to explore new states with higher probabilities while restarting frequently with low values. As shown in \cref{fig:jumping_scores}, the low values result in inferior accuracy results, particularly when long walking paths are necessary to fully retrieve the ground-truth patterns. In the Tree-Cycle scenario, the precision score steadily increases as the walking probability rises to 0.6 but diminishes when the value surpasses this threshold. For the Tree-Grid scenario, accuracy values are strongly correlated with the magnitude of the walking probability since ground-truth patterns are intricate, necessitating multiple walking steps to retrieve all ground-truth nodes. Consequently, a probability between 0.5 and 0.6 is optimal for the Tree-Cycle case, whereas a high value is suitable for the Tree-Grid scenario. In practical applications, a suitable probability can be selected between 0.5 and 0.9 based on the specific characteristics of networks.

\pgfplotsset{width=7cm,height=5cm}
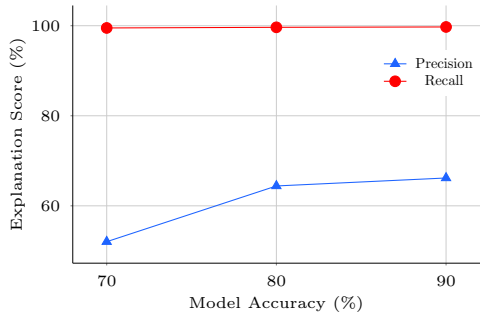
\begin{figure}[ht!]
	\centering
	\begin{tikzpicture}
		\begin{axis}[
			grid=both,
			grid style={line width=.03pt, draw=gray!40},
			major grid style={line width=.03pt,draw=gray!40},
			mark size=2pt,
			legend style={draw=none,fill=white,at={(1.01,0.83)},anchor=north east,legend columns=1,nodes={scale=0.5, transform shape},column sep=0.08cm},
			x label style={at={(axis description cs:0.5,-0.09)},anchor=north},
			y label style={at={(axis description cs:-0.09,0.5)},anchor=south},
			xtick={70,80,90},
			xlabel=Model Accuracy (\%),ylabel=Explanation Score (\%)]
			\addplot[color={rgb,255:red,30;green,100;blue,255},mark=triangle*] coordinates{
				(70.000000,52.03)
				(80.000000,64.41)
				(90.000000,66.18054054)
			};
			\addlegendentry{Precision}
			
			\addplot[color={red},mark=*] coordinates{
				(70.000000,99.50)
				(80.000000,99.64)
				(90.000000,99.72)
			};
			\addlegendentry{Recall}
		\end{axis}
	\end{tikzpicture}
	\caption{Relationships between Model Accuracy and Explanatory Graphs. They are strongly correlated with each other.}
	\label{fig:model_explanation_score}
\end{figure}

%Most existing GNN explanation methods assume that pre-trained models are extremely accurate. This study explores the influence of model accuracy on explanation correctness within the Mutag dataset.  As \cref{fig:model_explanation_score} depicts, the precision score increases significantly as the model accuracy improves from 70\% to 80\%, resulting in fewer false positive edges in explanations. When the model accuracy increases to 90\%, the precision score only improves slightly, suggesting that 80\% accuracy is sufficient for extracting influential subgraphs in this dataset. These results lead to the conclusion that explanations are more relevant and accurate as the model accuracy increases.
\noindent\textbf{Effects of Model Accuracy on Explanatory Graphs.} Current post-hoc XAI methods for GNNs operate under the assumption that black-box models possess high levels of accuracy. This experiment seeks to investigate the relationships between model accuracy and the quality of explanatory graphs using the Mutag scenario. As illustrated in \cref{fig:model_explanation_score}, as the model accuracy rises from 70\% to 80\%, the precision notably improves, leading to a reduction in unimportant edges in the explanatory graphs. However, when model accuracy reaches 90\%, the precision shows only a slight improvement, indicating that an 80\% accuracy level is adequate for identifying influential patterns in this dataset's graphs. These findings suggest that as model accuracy enhances, the relevance and accuracy of explanatory graphs improve correspondingly.

%This experiment examines the impact of distilled knowledge, namely embedding vectors and predictive distributions of a black-box GNN, on the correctness of structural explanations. This experiment was conducted on the Mutag dataset and compared the results of four settings on the self-explainable GNN as follows:
\noindent\textbf{Effectiveness of Knowledge Distillation}
This study investigates the effect of distilled knowledge, specifically embedding vectors and predictive distributions, on the quality of explanatory graphs. Based on experiments with the Mutag dataset, the study compares the outcomes of four different configurations on the structural learner outlined below:

\begin{itemize}
%	\item \textbf{Naive}: The learner uses neither embedding vectors nor predictive distributions in training.
%	\item \textbf{Embed}: The learner uses only embedding vectors to initialize learnable masks and sets $\lambda = 0$ in $\mathcal{L}^s$.
%	\item \textbf{KDL}: The learner does not use embedding vectors to initialize learnable masks.
%	\item \textbf{Joint}: The learner uses both components in training.
	\item \textbf{Naive}: Learners do not utilize predictive distributions or node embeddings during training.
	\item \textbf{Embed}: Learners utilize only node embeddings for initializing trainable masks and fixes $\lambda$ as zero in $\mathcal{L}^s$.
	\item \textbf{KDL}: Learners do not utilize node embeddings for initializing trainable masks.
	\item  \textbf{Joint}: Learners incorporate both predictive distributions and embeddings in training.
\end{itemize}

\pgfplotsset{width=6.5cm,height=5cm}
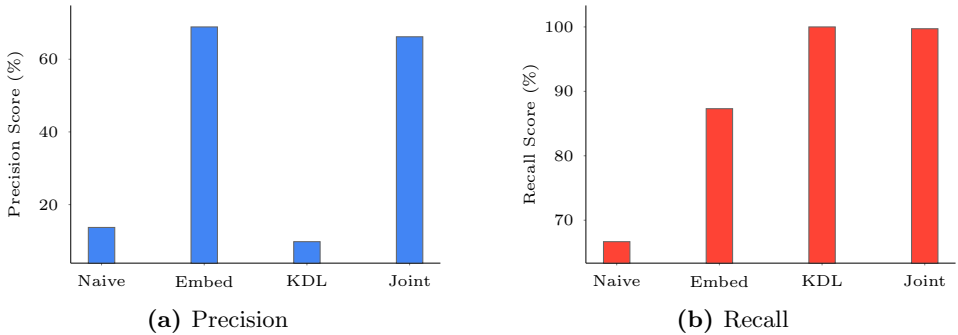
\begin{figure}[ht!]
	\centering
	\subfloat[Precision]{
		\begin{tikzpicture}
			\begin{axis}[
				% x=2cm,
				ybar,
				legend cell align=left,
				area legend,
				bar width=10pt,
				ylabel={Precision Score (\%)},
				y label style={at={(axis description cs:-0.1,0.5)},anchor=south},
				symbolic x coords={Naive, Embed, KDL, Joint},
				]
				\addplot[fill={rgb,255:red,66;green,133;blue,244},draw opacity=0.6] coordinates{(Naive, 13.71) (Embed, 68.89) (KDL, 9.80) (Joint, 66.18)};
				\legend{}
			\end{axis}
		\end{tikzpicture}
	}
	\hfill
	\subfloat[Recall]{
		\begin{tikzpicture}
			\begin{axis}[
				% x=2cm,
				ybar,
				legend cell align=left,
				area legend,
				bar width=10pt,
				ylabel={Recall Score (\%)},
				y label style={at={(axis description cs:-0.1,0.5)},anchor=south},
				symbolic x coords={Naive, Embed, KDL, Joint},
				]
				\addplot[fill={rgb,255:red,254;green,67;blue,53},draw opacity=0.6] coordinates{(Naive, 66.67) (Embed, 87.31) (KDL, 100) (Joint, 99.72)};
				\legend{}
			\end{axis}
		\end{tikzpicture}
	}
	\hfill
%	\caption{Explanation Correctness in Different Settings of the Online Knowledge Distillation Paradigm. SCALE provides the most accurate explanations when initializing a learnable mask with a black-box model's embedding vectors and using the KD loss for training learners.}
\caption{The Impact of Knowledge Distillation on Explanation Correctness. The framework achieves the highest correctness in explanations when it initializes a learnable mask based on the black box GNN's embeddings and is assisted by its distilled knowledge.}
	\label{fig:online_kg_ab}
\end{figure}

%Each setting was five times and reported the average results. As presented in \cref{fig:online_kg_ab}, precision and recall scores were high in the \textbf{Joint} case but significantly low in the \textbf{Naive} case. Furthermore, the explanation model achieved good precision in the \textbf{Embed} scenario, even better than the \textbf{Joint} case. In this case, many true positive edges were not selected in explanations, causing high precision and low recall. Conversely, the \textbf{KDL} setting showed high recall and low precision scores since the learner could not learn to eliminate unimportant edges. These results demonstrated that embedding vectors from a black-box GNN is essential for mask initialization, and distilled information from the predictive distribution of the black-box GNN enhanced the learning process of the self-explainable GNN. 

Each configuration was executed five times, and the mean results were reported. As shown in \cref{fig:online_kg_ab}, precision and recall are at high levels in the \textbf{Joint} configuration but significantly low in the \textbf{Naive} one. Additionally, the explainer obtained high precision in the \textbf{Embed} case, surpassing even the \textbf{Joint} one. An observation was that many influential edges were not included in explanatory graphs, causing low recall but high precision. In contrast, the \textbf{KDL} setting exhibited high recall but low precision because the learner was unable to discard unimportant edges. These findings indicate that embedding vectors are crucial for initializing the mask matrix, and distilled knowledge from predictive distributions is beneficial to the learning process of structural importance learners.

%This study aims to assess the effect of the balancing factor $\lambda$ in the joint objective function of student models on explanation correctness. Results from the Tree-Cycle and BA-2motifs datasets are presented in \cref{fig:balancing_factor}, as they exhibited clear variations in precision/recall scores.

\noindent\textbf{Effects of Balancing Factor.} This analysis sought to evaluate the impacts of $\lambda$ on the training performance of student models using experimental results from the BA-2motifs and Tree-Cycle datasets, as they showed distinct trends in accuracy scores. These results are reported in \cref{fig:balancing_factor}.

\pgfplotsset{width=9cm,height=5cm}
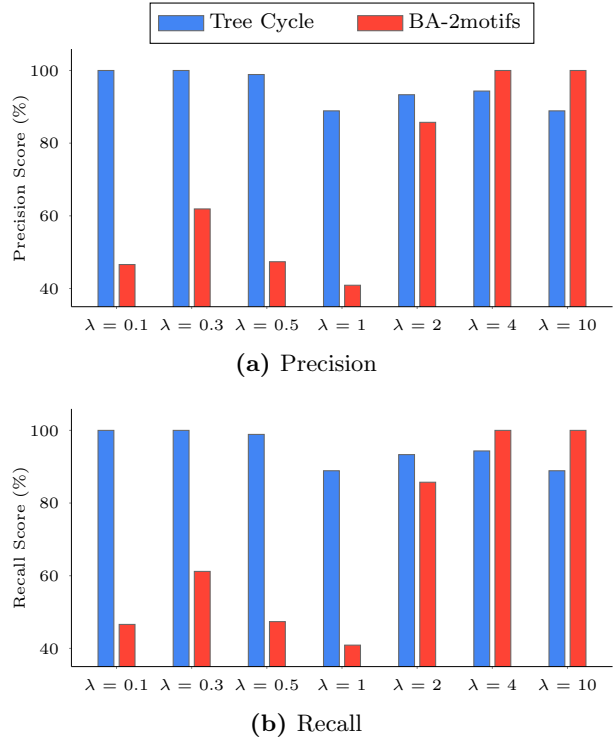
\begin{figure}[ht!]
	\centering
	\subfloat[Precision]{
		\begin{tikzpicture}
			\begin{axis}[
				x=1cm,
				ybar,
				legend cell align=left,
				area legend,
				bar width=6pt,
				ylabel={Precision Score (\%)},
				y label style={at={(axis description cs:-0.06,0.5)},anchor=south},
				symbolic x coords={$\lambda$ =  0.1,$\lambda$ =  0.3,$\lambda$ =  0.5,$\lambda$ =  1,$\lambda$ =  2,$\lambda$ =  4,$\lambda$ =  10},
				legend style={font=\scriptsize, at={(0.5,1.02)}, anchor=south,legend columns=-1,/tikz/every even column/.append style={column sep=0.5cm}},
				]
				\addplot[fill={rgb,255:red,66;green,133;blue,244},draw opacity=0.6] coordinates{($\lambda$ =  0.1, 100) ($\lambda$ =  0.3, 100) ($\lambda$ =  0.5, 98.89) ($\lambda$ =  1, 88.89) ($\lambda$ =  2, 93.33) ($\lambda$ =  4, 94.34) ($\lambda$ =  10, 88.89)};
				% {rgb,255:red,254;green,67;blue,53}
				\addplot[fill={rgb,255:red,254;green,67;blue,53},draw opacity=0.6] coordinates{($\lambda$ =  0.1, 46.6) ($\lambda$ =  0.3, 61.9) ($\lambda$ =  0.5, 47.37) ($\lambda$ =  1, 40.9) ($\lambda$ =  2, 85.73) ($\lambda$ =  4, 100) ($\lambda$ =  10, 100)};
				
				\legend{Tree Cycle,BA-2motifs}
			\end{axis}
		\end{tikzpicture}
	}
	\hfil
	\subfloat[Recall]{
		\begin{tikzpicture}
			\begin{axis}[
				x=1cm,
				ybar,
				legend cell align=left,
				area legend,
				bar width=6pt,
				ylabel={Recall Score (\%)},
				y label style={at={(axis description cs:-0.06,0.5)},anchor=south},
				symbolic x coords={$\lambda$ =  0.1,$\lambda$ =  0.3,$\lambda$ =  0.5,$\lambda$ =  1,$\lambda$ =  2,$\lambda$ =  4,$\lambda$ =  10}
				]
				\addplot[fill={rgb,255:red,66;green,133;blue,244},draw opacity=0.6] coordinates{($\lambda$ =  0.1, 100) ($\lambda$ =  0.3, 100) ($\lambda$ =  0.5, 98.89) ($\lambda$ =  1, 88.89) ($\lambda$ =  2, 93.33) ($\lambda$ =  4, 94.34) ($\lambda$ =  10, 88.89)};
				\addplot[fill={rgb,255:red,254;green,67;blue,53},draw opacity=0.6] coordinates{($\lambda$ =  0.1, 46.6) ($\lambda$ =  0.3, 61.19) ($\lambda$ =  0.5, 47.37) ($\lambda$ =  1, 40.9) ($\lambda$ =  2, 85.73) ($\lambda$ =  4, 100) ($\lambda$ =  10, 100)};
				\legend{}
			\end{axis}
		\end{tikzpicture}
	}
	\caption{Effect of $\lambda$ on Explanatory Graphs. The framework achieves high explanation correctness with a value less than or equal to 1 for node classification and greater than or equal to 2 for graph classification.}
	\label{fig:balancing_factor}
\end{figure}

%The distilled knowledge from a black-box GNN is crucial for training a self-explainable GNN model for graph classification datasets. Experiments with SCALE showed that $\lambda \geq 2$ produced better results. Even though the self-explainable model achieved high prediction accuracy, it could not provide accurate explanations when $\lambda \leq 1$. Furthermore, a slight decrease in explanation correctness was observed when $\lambda \geq 4$. This change suggests that too much information from the black box model is not helpful for the explainable model.

Extracting insights from a black-box GNN is essential for creating an interpretable GNN model tailored to graph classification tasks. Experimental results demonstrated that a $\lambda$ value of 2 or higher yielded superior outcomes. While the interpretable model provided accurate predictions, it failed to deliver precise explanations when $\lambda \leq 1$. Additionally, a minor decline in explanation accuracy was noted when $\lambda$ reached 4 or above. This trend indicates that an excessive amount of information from the black-box model can be detrimental to the performance of the explainable model.

%In contrast, for node classification problems, $\lambda$ values smaller than one allowed SCALE to provide accurate explanations. The explanation correctness gradually increased as the balancing factor decreased when $\lambda \le 1$. These results suggested that student models cannot handle excessive distilled knowledge from the teacher model. 

Conversely, SCALE was able to generate accurate explanations with $\lambda$ values below one for node classification tasks. The correctness of these explanations progressively improved as the balancing factor decreased when $\lambda \le 1$.  These findings indicate that learners struggle to process an excessive amount of distilled knowledge from the black-box GNN.

\section{System Prototype and Demonstration} \label{prototype}
\subsection{System Design}

\begin{figure}[ht!]
	\centering
	\includegraphics[width=\linewidth]{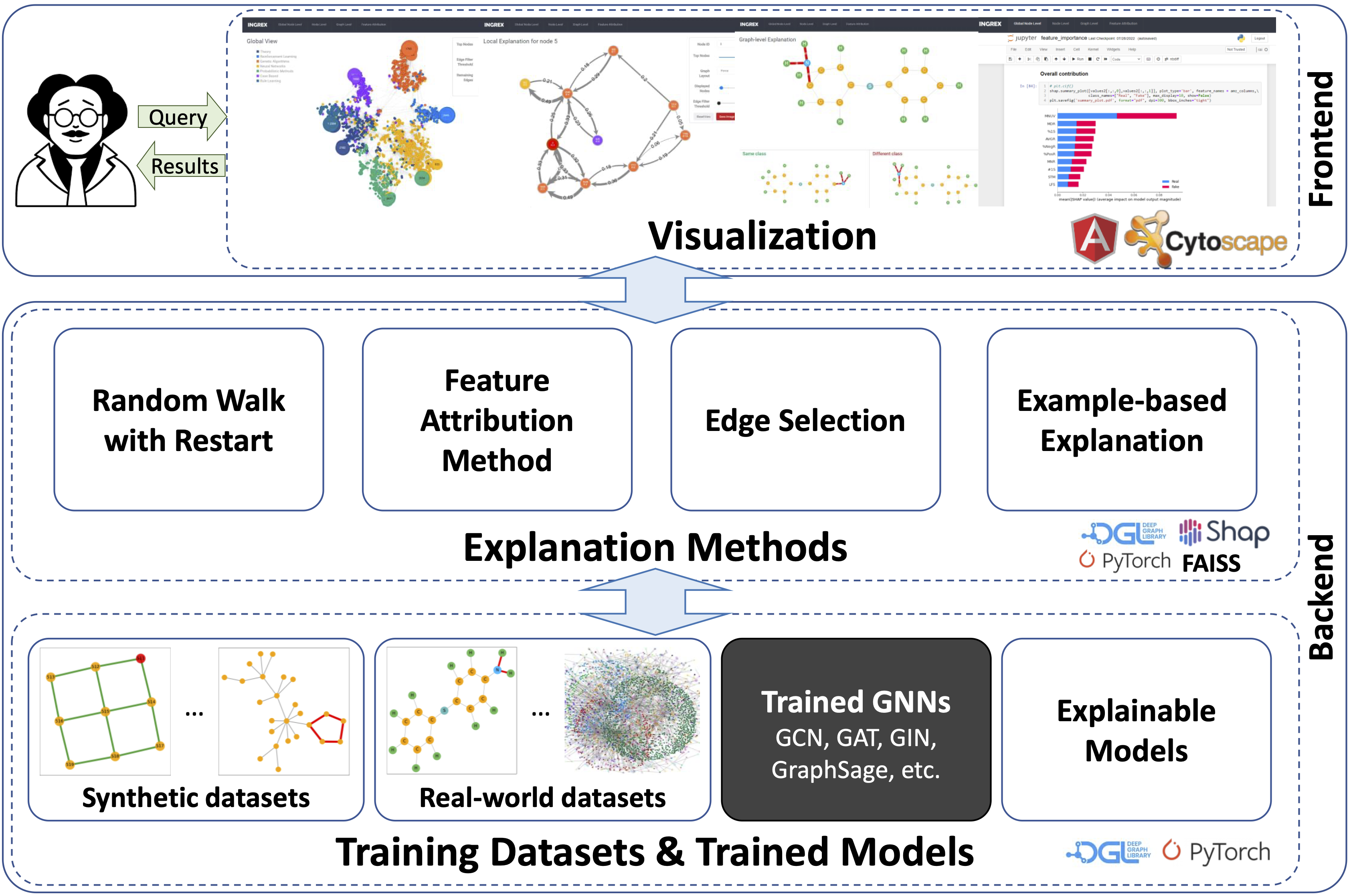}
	\caption{A Prototype System Architecture}
	\label{fig:application_system}
\end{figure}

%Based on algorithms proposed in \cref{method}, a simple prototype system was designed to enhance the visualization and utility of explanations. It is essential to emphasize that there is no silver bullet for XAI in GNNs. Therefore, multiple explanation methods allow users to confirm prior hypotheses and extract insights from explanations. Additionally, the demo highlights the potential and limitations of the proposed methods. The system is demonstrated in \cref{fig:application_system}. 

As illustrated in \cref{fig:application_system}, a prototype system was developed based on \cref{method} to showcase the visualization and potential applications of explanation functions. Since there is no universal solution in the XAI domain, especially for GNNs, providing various explanation modalities through different approaches enables users to validate their hypotheses and derive insights from explanations. Furthermore, the demonstration underscores both the potential and shortcomings of the proposed framework.

%A Python web server was implemented using Flask. The backend consists of two main components: trained models and explanation methods. Currently, the system manages datasets and models as files and loads them into memory when needed. Explanation methods correspond to algorithms described in \cref{method}. The Python web server allows us to integrate several libraries for model execution and explanation methods, such as DGL\cite{dgldata}, PyTorch, NetworkX, SHAP\cite{shap_doc}, and Faiss \cite{johnson2019billion}. DGL and PyTorch are used for module training and inference. SHAP provides feature attributions of predictions. Faiss, a billion-scale similarity search, is utilized for providing example-based explanations.

\noindent\textbf{Backend.} A Flask web server was developed to be a controller. The backend is composed of two primary elements: explanation methods and trained models. The system stores datasets and models as files, loading them into memory as required. The explanation methods refer to \cref{method}. The backend facilitates the integration of external libraries for model construction and explanation generation, including PyTorch,  DGL\cite{dgldata}, NetworkX, Faiss\cite{johnson2019billion}, and SHAP\cite{shap_doc}.  PyTorch and DGL are employed for model construction and execution, SHAP outputs feature attributions, and Faiss is leveraged for retrieving comparative examples.

%A web application for receiving user requests and visualizing explanations was implemented using Angular and CytoscapeJS. Angular was selected due to its efficiency in building cross-platform web applications and its large developer community. CytoscapeJS \cite{Cytoscape} is a Javascript library of Cytoscape, a powerful open-source platform for visualizing complex networks and providing useful UI/UX features. Similar to Angular, CytoscapeJS allows additional plugins to be integrated.

\noindent\textbf{Frontend.} A web application designed for processing user requests and visualizing explanations was developed based on CytoscapeJS \cite{Cytoscape} and Angular. Angular was chosen for its effectiveness in creating cross-platform applications and its developer community. CytoscapeJS, a JavaScript library derived from Cytoscape, offers robust capabilities for visualizing complex networks and delivering helpful UI/UX features. Like Angular, CytoscapeJS supports the integration of additional plugins.

%Explanations provided by the system are best suited for ML practitioners, model developers, and domain experts. ML practitioners and model developers can use explanation results for detecting abnormalities in datasets and trained models, ultimately leading to improved predictive accuracy.  Domain experts can gain insights from explanations or confirm results with their hypotheses.

\noindent\textbf{Users.} The framework's explanations are primarily designed for ML developers, practitioners, and domain specialists. These explanations aid ML developers and practitioners in identifying anomalies within datasets and trained models, which can enhance predictive accuracy. Additionally, domain specialists can leverage these explanations to validate their hypotheses or derive insights. 

%When a user requests an explanation for a particular prediction, the frontend web application sends a REST request to the backend server. The backend server parses the request and executes a corresponding explanation method based on an explainable model and a training dataset. After that, it returns an explanation result to the web application to show to the user. After being loaded into the memory, models, and datasets are retained to reduce the execution time of future requests.

\noindent\textbf{Operational Flows.} When a user seeks clarification regarding a specific prediction, the web application initiates a REST request directed to the backend. Upon receiving this request, the backend interprets it and employs an appropriate explanation method associated with constructed explainers. Subsequently, the server delivers an explanation back to the frontend application for visualization. To minimize the execution time for subsequent requests, datasets and models are preserved in memory once they are loaded.

\subsection{Demonstration Scenarios}

\noindent\textbf{Scenario 1: Structural Explanations of Node Classification}

\noindent\textit{Input.} An input graph $G = (V, E, X)$ includes of a vertex set $V$, an edge set $E$, and a feature matrix $X$ associated with nodes. The objective is to quantify neighboring influences that significantly drive a specific node-level outcome.

\begin{figure}[ht!]
	\centering
	\includegraphics[width=\linewidth]{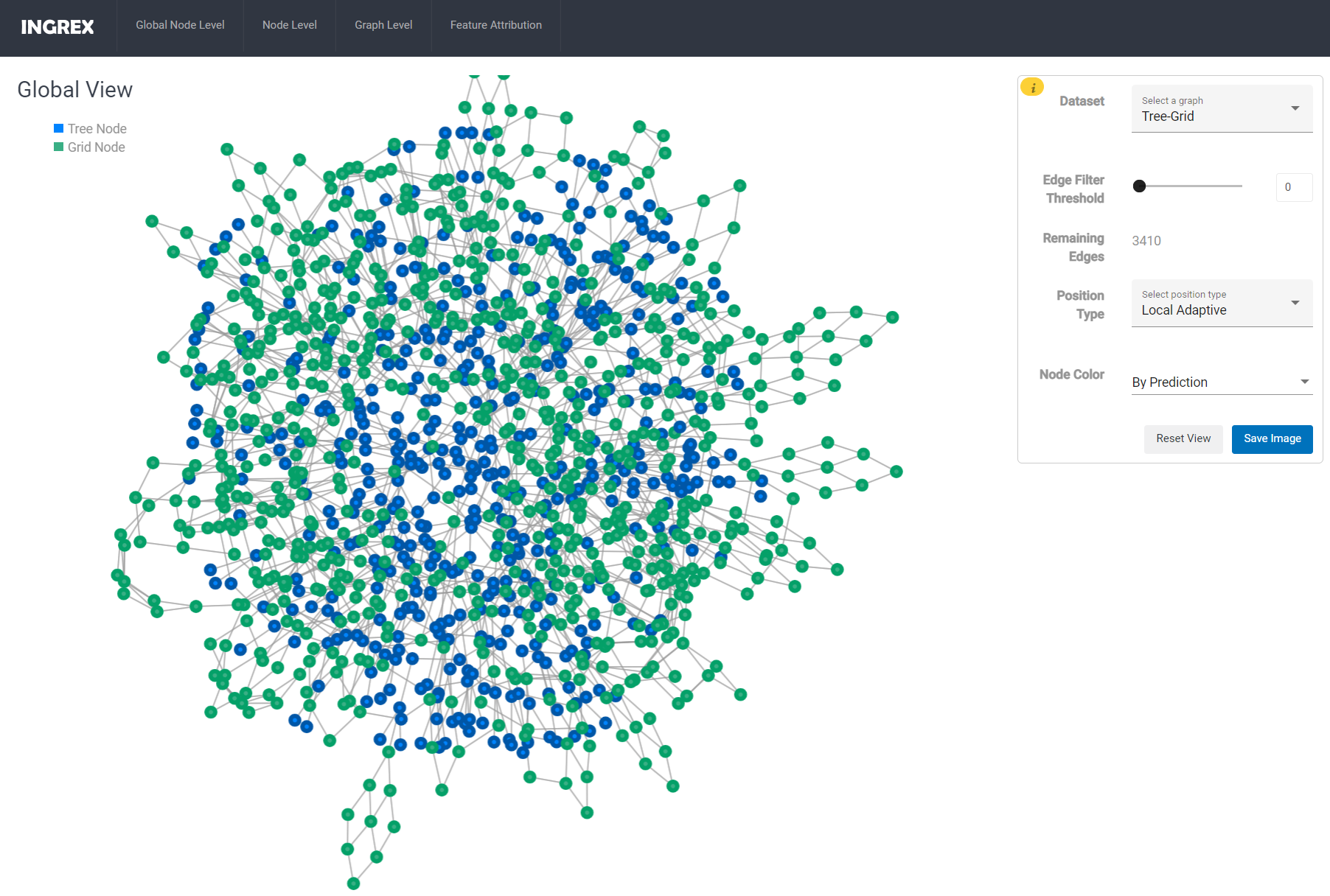}
	\caption{A Demonstration of Graph Embeddings in Two-dimensional Space}
	\label{fig:global_graph}
\end{figure}

%\noindent\textit{Process.} First, the system visualizes the input graph based on either node embeddings or adaptive neighborhood positioning, as presented in \cref{fig:global_graph}. This visualization allows users to select target nodes and examine local explanations. Once a target node is selected, the system's backend parses the explanation request and executes \cref{querying_rwr} to identify crucial edges and neighbors that contribute to the target node. The system then generates a subgraph containing these elements providing a transparent and effective means of understanding the underlying factors driving the model prediction.

\noindent\textit{Process.} As illustrated in \cref{fig:global_graph}, the system presents an input graph in two-dimensional space via either adaptive neighborhood positioning or node embeddings. The interface enables users to choose specific nodes and perform local analyses. Upon selecting a target node, the system's backend processes the request and runs \cref{querying_rwr} to quantify neighbors' contributions. Consequently, the system extracts a subgraph rooted at this node and includes quantitative information on adjacent influences, thereby offering a clear and effective way to comprehend the factors driving the model's outcome.

%\noindent\textit{Output.} \cref{fig:node_ex} presents the web interface of an explanation, while \cref{fig:node_structure} demonstrates two explanations for two example nodes in the Cora dataset \cite{dgldata}. By leveraging these explanations, users can gain insight into the contributions of k-hop neighbors to a target node's prediction. Furthermore, the explanations can also shed light on the reasons for incorrect predictions, especially when nodes reside near class boundaries and are influenced by cross-class edges.

\begin{figure}[ht!]
	\centering
	\includegraphics[width=\linewidth]{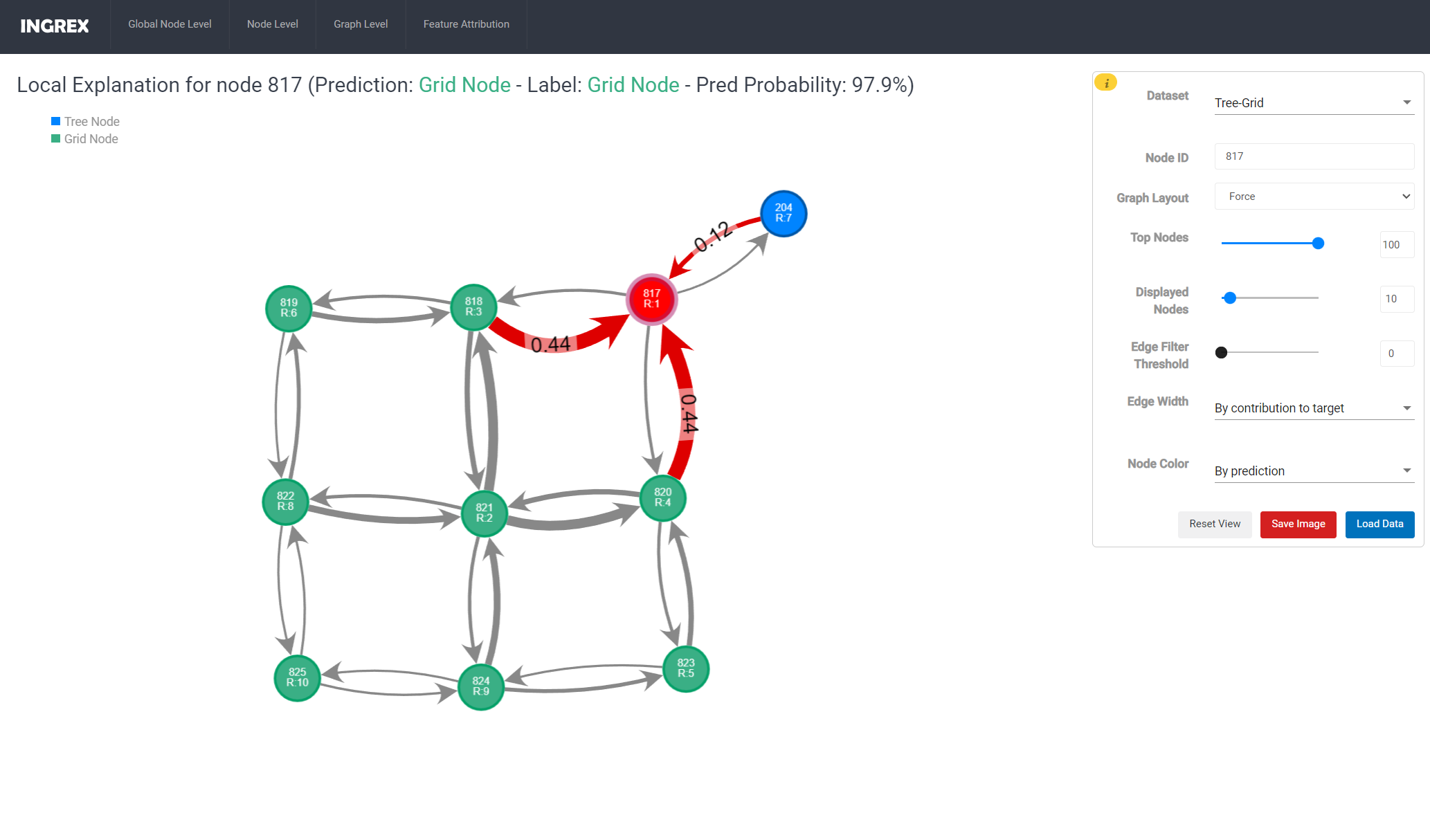}
	\caption{A Visualization of an Explanation for a Node Prediction}
	\label{fig:node_ex}
\end{figure}

\begin{figure}[ht!]
	\centering
	\subfloat{
		\includegraphics[width=0.7\linewidth]{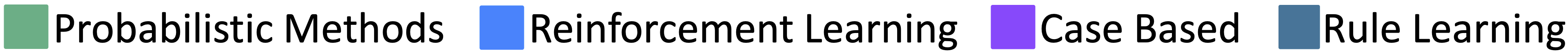}
	}
	\setcounter{subfigure}{0}
	\hfil
	\subfloat[True Prediction]{
		\includegraphics[width=0.4\linewidth]{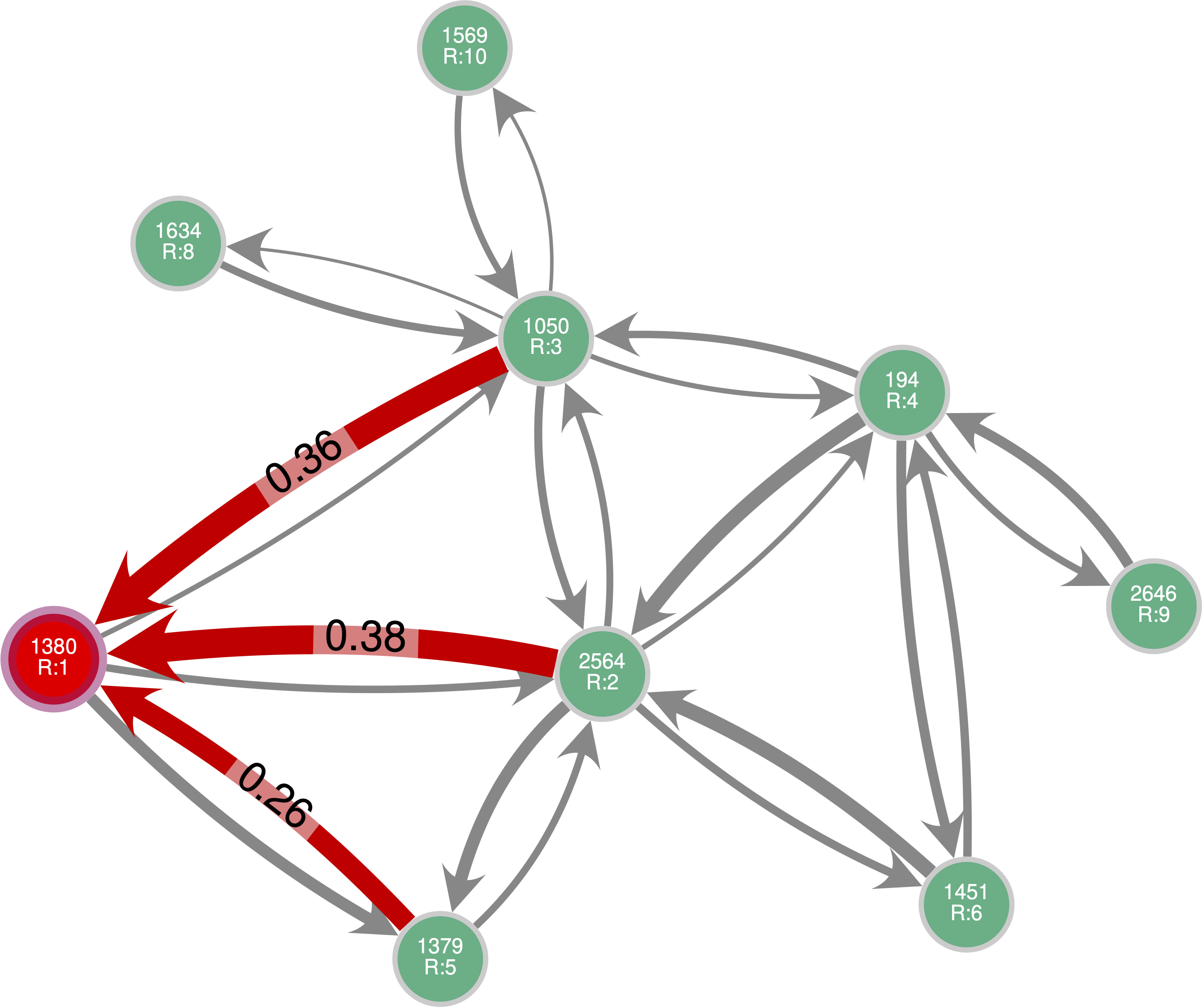}
	}
	\subfloat[Wrong Prediction]{
		\includegraphics[width=0.4\linewidth]{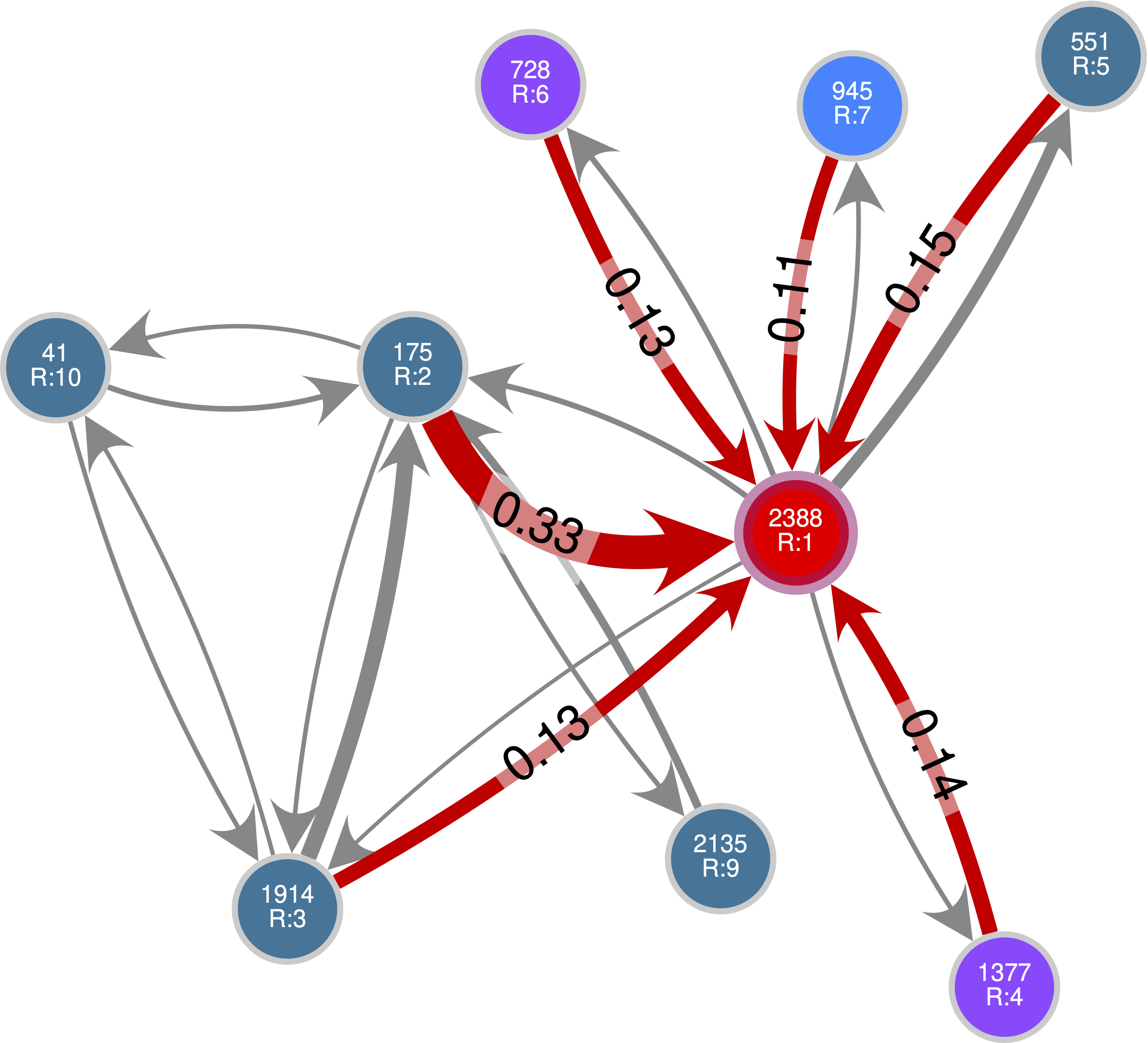}
	}
%	\caption{Structural Explanations of Node-level Predictions in Cora Dataset. Node colors illustrate different classes. The red node is the explained one. Red edges highlight contributions of 1-hop neighbors with corresponding ratios. Contributions can be viewed on multiple levels (k-hop neighbors).}
	\caption{Explanation Graphs of Node Classification on Cora Dataset. The colors of nodes represent classes, with the red node indicating the one being explained. Red edges emphasize the influences of 1-hop neighbors, along with their respective ratios.}
	\label{fig:node_structure}
\end{figure}

\noindent\textit{Output.} The web interface for this type of explanation is depicted in \cref{fig:node_ex}, while \cref{fig:node_structure} illustrates local analyses for two examples taken from the Cora dataset \cite{dgldata}. By analyzing local structures, users can understand the influences of k-hop neighbors on the node's outcome. Moreover, these explanations can elucidate the reasons behind inaccurate classifications, particularly when nodes are positioned near decision boundaries and are affected by transboundary edges.

\noindent\textbf{Scenario 2: Feature Attributions of Node Classification}

%The goal is to find the overall contributions of node features to model predictions and specific feature attributions of a certain node-level prediction. The Amazon dataset is utilized for this scenario.

\noindent\textit{Input.} The objective is to quantify the exact attributions of node features to individual predictions and to provide summaries of their contributions to overall model outcomes. The analyses are conducted on the Amazon dataset.

%The system integrates Jupyter Notebook into the web application to facilitate direct user interaction with feature attribution methods. With the multitude of available attribution frameworks \cite{shap_doc,nori2019interpretml}, this integration enables users to select a method that aligns with their preferences. Several experiments were conducted with methods offered by SHAP \cite{shap_doc} on an MLP student guided by a GNN model teacher in training. 

\noindent\textit{Process.} The web application incorporates Jupyter Notebook, allowing users to interact directly with feature attribution methods. Given the diverse range of available frameworks \cite{shap_doc,nori2019interpretml}, this integration permits users to choose a method that best suits their needs. Numerous analyses were performed based on methods provided by SHAP \cite{shap_doc} on an MLP, which was trained under the guidance of a black-box GNN.

%\cref{fig:instance_explanations_demo} illustrates the web interface of the system integration, including a contribution summary and an instance explanation. Global summarizations provide users with a comprehensive understanding of feature contributions at an overall level, whereas local attributions offer detailed insights into the precise feature influences on particular predictions.

\begin{figure}[ht!]
	\centering
	\includegraphics[width=0.98\linewidth]{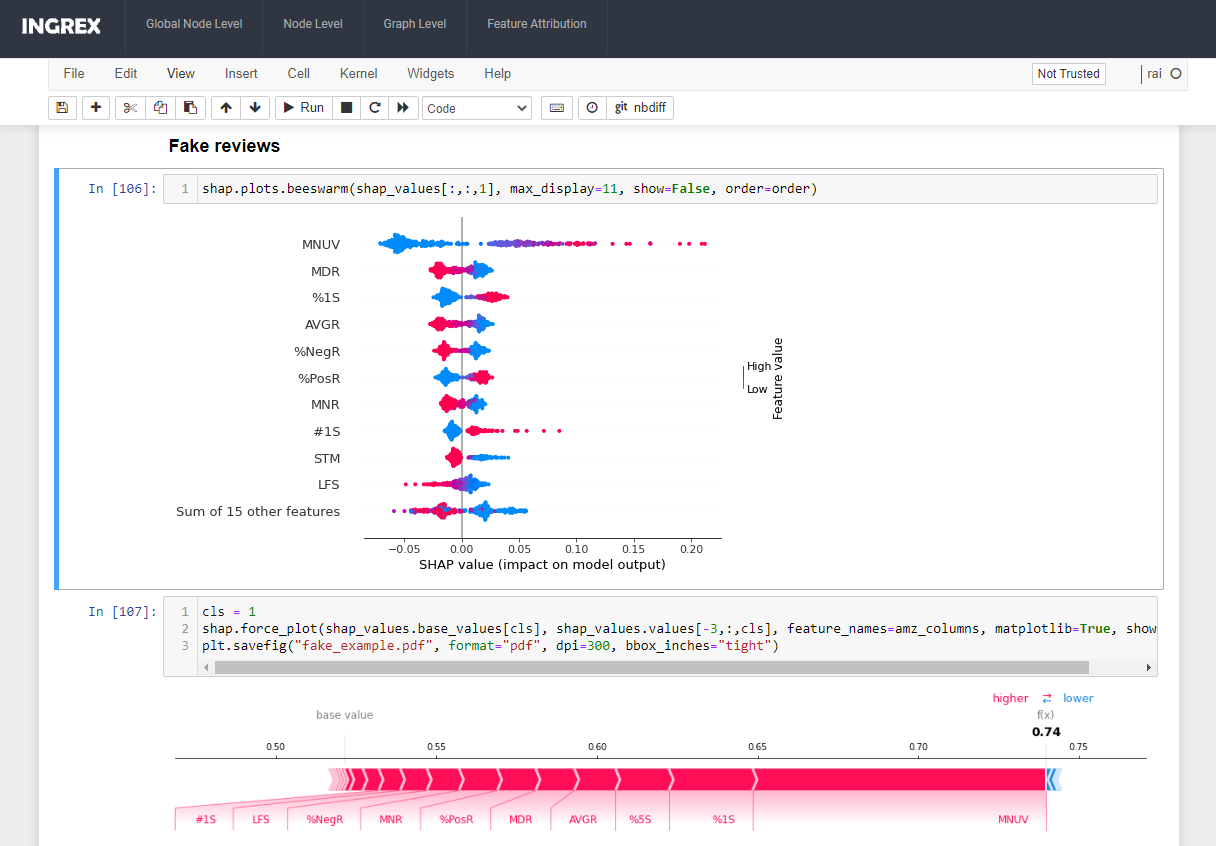}
	\caption{Interactive Explanations of Feature Attributions}
	\label{fig:instance_explanations_demo}
\end{figure}

\noindent\textit{Output.} \cref{fig:instance_explanations_demo} depicts the integration module's interface, which features an instance explanation and a summary of contributions. Global summaries deliver a broad overview of feature contributions, enabling users to grasp their overall impact. Conversely, local attributions furnish in-depth insights into the specific influences of features on individual predictions.

\noindent\textbf{Scenario 3: Explaining Graph-level Predictions}

% Given an input graph, the goal is to find crucial motifs in the input graph that contribute to a specific prediction. The system allows users to upload a text file, including an input graph's data, in a pre-determined format.

\noindent\textit{Input.} Given a target graph, the objective is to identify essential patterns that influence a particular outcome. The system enables users to upload graph data via a text file, adhering to a specified format.

%The backend executes an edge selection procedure on the mask matrix $M$ of trained self-explainable GNNs. Next, the system returns the structural explanation with an example-based explanation, which includes explanations of samples from the same and different classes.

\noindent\textit{Process.}  The backend performs an edge pruning process on the trained matrix $M$. Subsequently, the system generates structure analyses for the target graphs and their comparative references.

\begin{figure}[ht!]
	\centering
	\includegraphics[width=\linewidth]{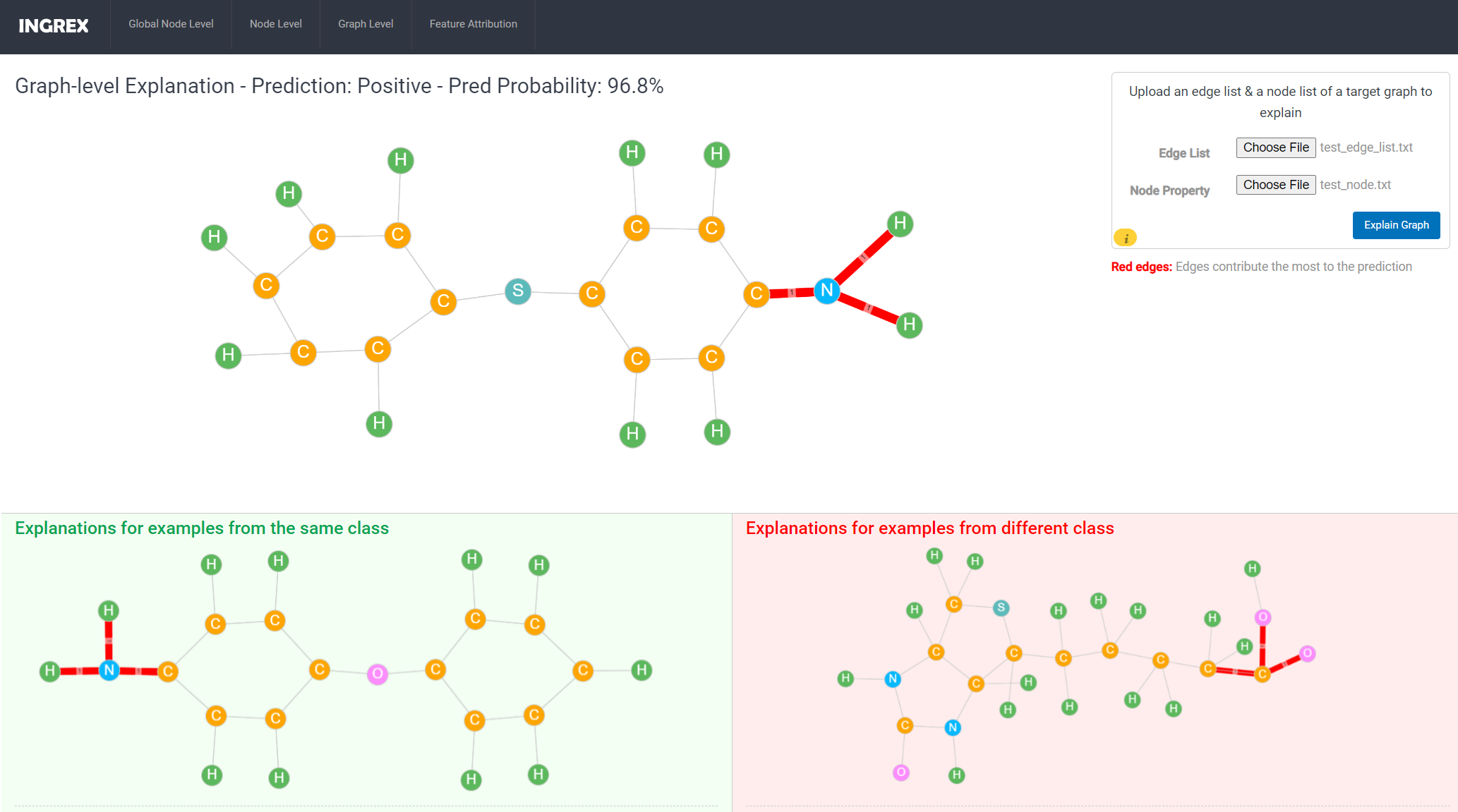}
	\caption{Graph Classification Explanation. Both highlighted structures and example-based explanations are presented.}
	\label{fig:graph_ex_wd}
\end{figure}

\begin{figure}[ht!]
	\centering
	\hfil
	\subfloat[Target Instance]{
		\includegraphics[width=0.31\linewidth]{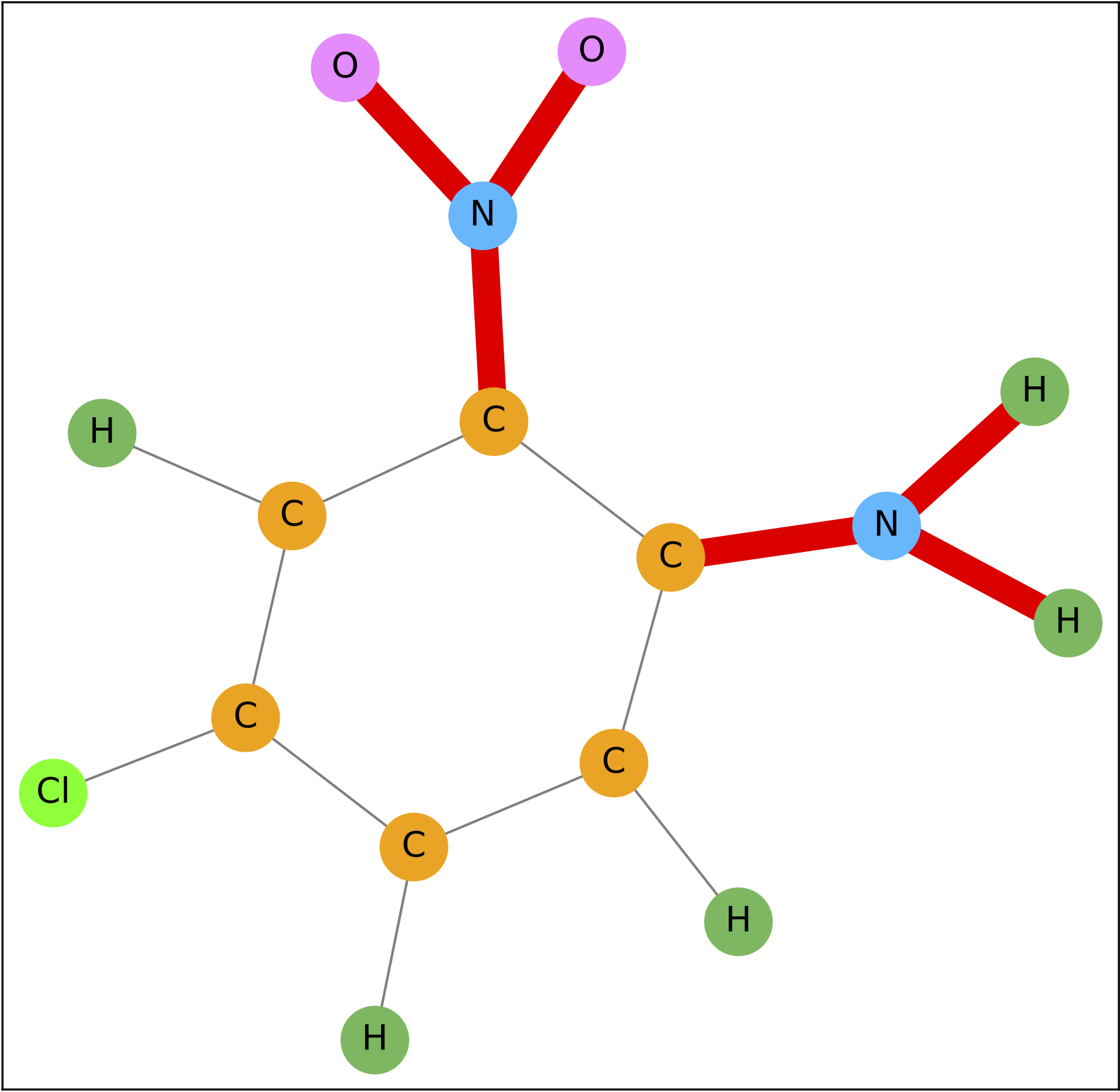}
	}
	\subfloat[Same Class]{
		\includegraphics[width=0.31\linewidth]{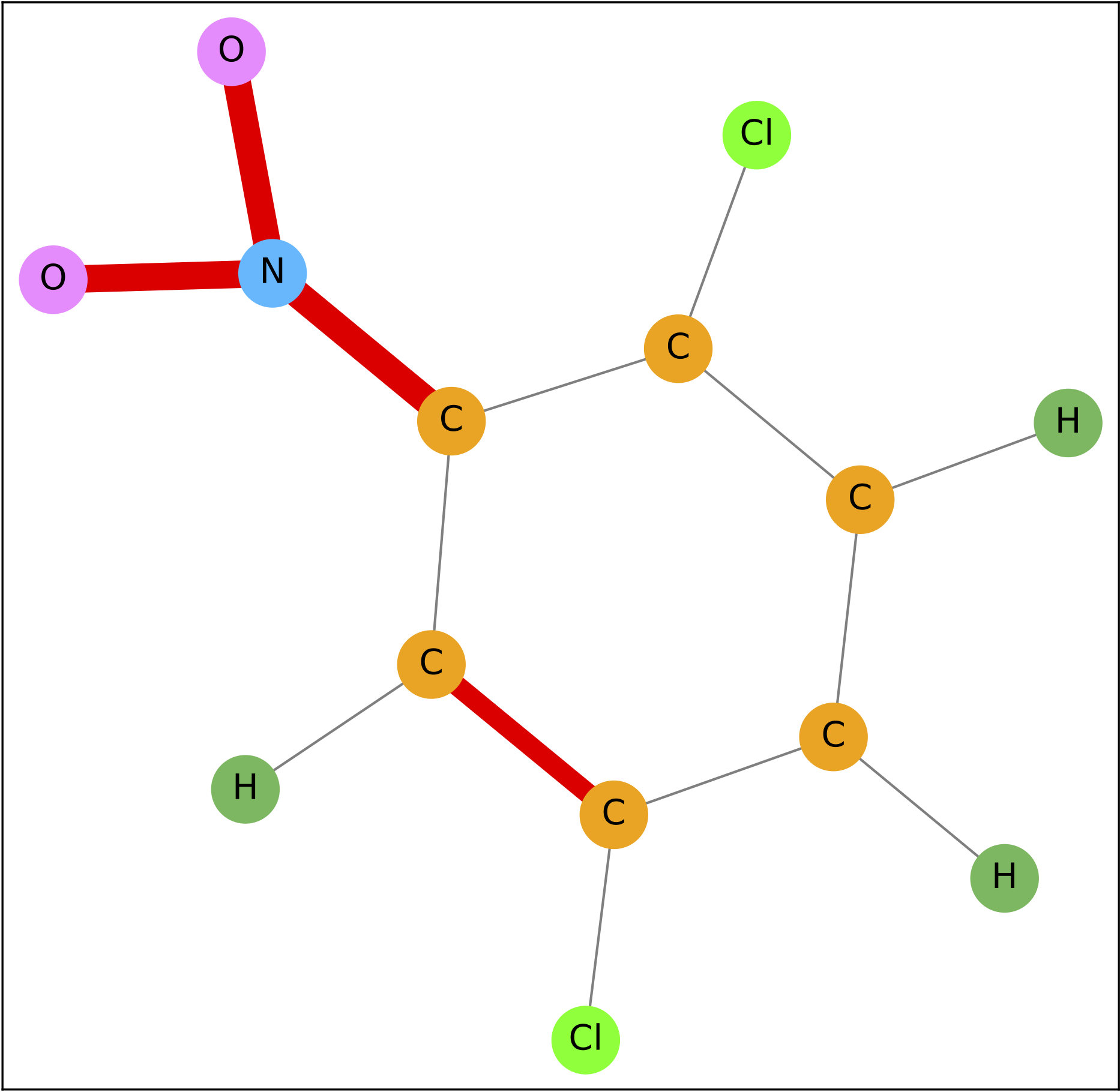}
	}
	\subfloat[Different Class]{
		\includegraphics[width=0.31\linewidth]{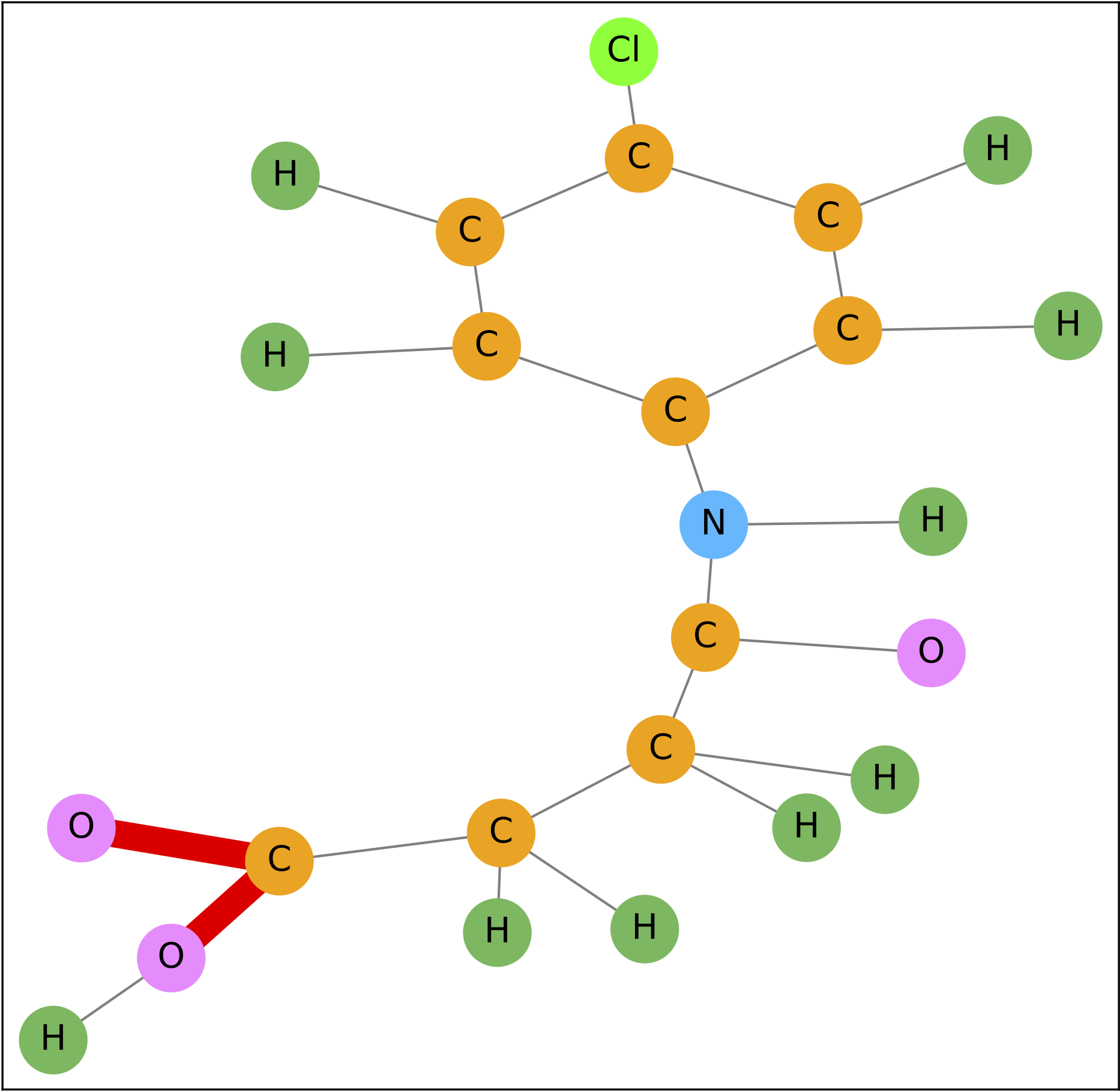}
	}
	\caption{A Comparative Explanation of Graph Classification. Influential edges are highlighted in red.}
	\label{fig:graph_explanation}
\end{figure}

% \cref{fig:graph_explanation} presents structural explanations highlighting crucial edges for the predictions of a Mutag graph and reference samples. \cref{fig:graph_ex_wd} illustrates the web interface of these explanations. Highlighted edges first provide users with information about important motifs. Comparing explanations of samples in different classes gives them more insights into the target graph's prediction.

\noindent\textit{Output.} \cref{fig:graph_explanation} visualizes influential edges that are crucial to the prediction alongside comparative references. The web interface for this explanation is depicted in \cref{fig:graph_ex_wd}. Initially, the highlighted edges offer users insights into essential patterns. Comparative analyses enable users to gain a deeper understanding of the prediction outcomes for the target graph.

\section{Discussion} \label{discussion}
\subsection{Limitations and Future Improvements}
%Even though SCALE offers several advantages over existing methods, it also has some limitations. First, it does not consider the explanation problem in link prediction. However, this problem can be resolved by employing the same approach for explaining node-level predictions. Second, learners and a black-box GNN model are trained in a single thread. In practice, distributed settings could further accelerate this training process. Third, the RWR algorithm is executed in a naive way in experiments that will suffer from long execution time for enormous real-world graphs. Since RWR is widely adopted in many large-scale systems, several techniques, such as \cite{tong2006fast}, can accelerate this algorithm. Fourth, structural explanations for graph classification only show selected influential edges without detailed importance scores. Game-theoretic methods, like those in \cite{lundberg2017unified}, could enhance explanation quality by computing exact edge contributions. Additionally, this chapter does not study interactions among graph structures and node features due to their complexity. Integrating methods such as \cite{tsang2020does,lpa_gcn} into SCALE can enhance its explanation capability. Finally, as the number of learners increases with different explanation aspects, combining their outputs is a promising approach that allows the explanation framework to match users' preferences.

Despite the numerous benefits compared to existing methods, the proposed framework also has certain limitations. First, the link prediction problem has not been considered in this work. However, this shortcoming can be addressed by updating the proposed techniques with slight modifications. Second, a black-box GNN and its learners are trained sequentially, but in real-world scenarios, using a distributed setup could significantly speed up the process. Third, this work only implements the RWR algorithm in a basic manner in experiments, which results in long execution times for large graphs. Various acceleration techniques, such as one proposed by Tong et al. \cite{tong2006fast}, can be applied to solve this issue. Fourth, the explanation graphs in graph classification merely highlight selected patterns without providing detailed importance scores. Integrating game-theoretic methods \cite{lundberg2017unified} could improve explanation quality by accurately determining neighbor influences. Moreover, this chapter does not explore the interactions between features and structures due to their complicated nature. Incorporating methods like \cite{tsang2020does,lpa_gcn} into the framework could enhance its explanatory power. Lastly, the framework can increase the diversity of its provided explanations by designing new context-aware learners, aligning well with users' preferences.

\subsection{Improvements and Extensions}
%Since this system is currently in a prototype form, there is plenty of room for improvements and extensions. First, in-memory libraries can be integrated into the system to optimize model and data management. The integration can enhance the system's efficiency by reducing the time required to access and manipulate data. Second, cache and load-balancing techniques can be employed to further accelerate the system's performance. These techniques can increase the speed and scalability of the system, allowing it to handle larger and more complex datasets. Third, the system is open to integrating additional explanation methods, which expand the range of interpretability techniques available to users. Fourth, a text generation module can be designed on top of explanation methods to personalize explanations for various groups of users. This module can increase the accessibility of the system and enable users with varying levels of technical expertise to understand its output. Finally, combining the results of different explanation methods into one view further facilitates user comprehension of predictions. These potential improvements and extensions can enhance the system's functionality, efficiency, and interpretability.

Since the current framework exists in a prototype stage, there is significant potential for improvements and expansions. First, incorporating in-memory libraries could optimize data management and inference, thus increasing efficiency by reducing data access and execution times. Second, adopting caching and load-balancing methods could further enhance system performance, improving its speed and scalability to handle more complex datasets. Third, the system could integrate novel methods, broadening the range of explanation features available to users. Fourth, developing a text generation component could personalize the explanation experience, making predictions more understandable to users with different levels of knowledge. Finally, users could benefit from diverse visualizations and explanation modalities with interactive supports. These promising enhancements could significantly improve the framework's functionality, efficiency, and interpretability.

\subsection{Potential Applications} 
%SCALE can be applied to numerous applications and systems since it instantly provides accurate explanations. For instance, recommender systems \cite{wu2020graph,fan2019graph} can easily integrate SCALE into their engines to increase the transparency of their systems through explanation functions. Furthermore, SCALE can provide valuable insights and accelerate research in several domains, such as human action recognition \cite{sun2022human}, bioinformatics \cite{zhang2021graph}, and to name a few. Additionally, SCALE's feature attribution module can be particularly useful in scenarios containing intelligible node/edge features. Therefore, SCALE can aid in mitigating the difficulties associated with extracting insights from input features in graph analytics problems.

The proposed framework can be utilized across various applications and systems due to its capability to generate precise predictions with explanations promptly. For example, recommender systems \cite{wu2020graph,fan2019graph} can incorporate SCALE to enhance system transparency through explanatory functions. Moreover, SCALE can accelerate research in several fields through insights provided by explanations, such as bioinformatics \cite{zhang2021graph} and human action recognition \cite{sun2022human}. Additionally, the feature attribution module is advantageous in graph datasets with meaningful node/edge features. Thus, the framework can help alleviate the challenges related to extracting actionable information from not only graph structure but also features in analytical tasks.

\section{Conclusion} \label{conclusion_part}
%This chapter presented SCALE, the first explanation framework that trained multiple specialty learners to explain GNNs, which is challenging due to the complexity of examining attributions of factors in an input graph. The objective was to formulate explanation problems as general as post-hoc GNN explanation methods while achieving the explanation speed of self-explainable models. SCALE identified influential factors affecting model predictions from graph structures and node features. It provided explanations with more detailed information compared to existing methods. To achieve these goals, specialty learners were trained simultaneously with a black-box GNN model based on online knowledge distillation. After training, the framework provided predictions and explanations instantly using several explainers that examined different contributions of factors. Specifically, structural explanations for graph-level and node-level predictions were provided through edge masking and random walk with restart procedures. Additionally, node feature attributions at different levels were obtained by executing a fast feature attribution method on top of a trained MLP learner. Extensive experiments and ablation studies demonstrated SCALE's capabilities and superior performance. 

This chapter introduced the first framework that constructed multiple specialized components to elucidate GNNs, addressing the inherent complexity of analyzing factor attributions within an input graph. The aim was to design an XAI framework that was as broad as the post-hoc approach while matching the inference speed of interpretable models. The proposed framework determined key factors influencing model predictions by examining both features and graph structures, offering more detailed explanations than current methods. In training, a target GNN-assisted interpretable components or specialized learners based on a special knowledge distillation paradigm. At inference, the framework delivered predictions with immediate explanations generated by constructed explainers. Particularly, it delivered structure analyses through edge pruning and RWR procedures. Moreover, it obtained node feature attributions at different granularity by applying an efficient approximation algorithm. Comprehensive experiments and analyses highlighted the proposed framework's capabilities and outstanding performance.

%This chapter explored the promising potential of the example-based explanation approach, demonstrating how comparative insights can enhance understanding. The subsequent chapter will build upon this foundation, shifting the focus to concept-focused graph structure similarity and further refining the approach to develop interpretable GNNs.

This chapter also investigated the potential of the example-based explanation approach, showing how comparative insights can enhance user comprehension. The following chapter will expand on this foundation, focusing on concept-based graph structure similarity and further refining the methodology to develop interpretable GNNs.

	\chapter{Interpretable GNNs via Concept Matching Model} \label{chap:concept}
\section{Introduction} 
% Intro of GNN & % Importance & challenges of Interpretable GNNs
Explaining the inner workings of GNNs presents significant challenges since the complex nature of graphs causes intricate interactions during message-passing processes. Subgraph extraction is a fundamental approach to pattern recognition in graph problems, and it plays a crucial role in XAI methods for GNNs \cite{yuan2022explainability}. By identifying and isolating frequent substructures that are critical to the network's decision-making, these methods reveal how specific subgraphs or patterns within the larger graph influence the model's output, providing valuable insights into its reasoning.

% How do others do? Both post-hoc & self-interpretable 
%Various post-hoc methods and interpretable GNNs have been proposed lately. The post-hoc approach \cite{ying2019gnnexplainer,luo2020parameterized,yuan2021explainability} treats GNNs as black boxes and identifies minimal substructures contributing to specific predictions. However, the trustworthiness of explanations provided by these methods is questionable, particularly in high-stakes decision-making scenarios \cite{rudin2019stop}. In contrast, interpretable GNNs \cite{dai2021towards,zhang2022protgnn} offer transparent architectures capable of generating explanations through internal weights or components. Nevertheless, existing methods often prioritize model accuracy and algorithmic evaluation over comprehensive explanation generation with a thorough assessment of user perception of explanations. Comprehensive explanations often necessitate the integration of multiple information sources, each offering unique insights to enhance user understanding of the predictions. 

Recent advancements in XAI have introduced numerous post-hoc methods and explainable GNNs. Post-hoc methods \cite{ying2019gnnexplainer,luo2020parameterized,yuan2021explainability} consider GNNs as opaque entities and focus on identifying key substructures essential for specific outcomes. The reliability of the explanations these methods provide is often debated, especially in contexts requiring transparent decision-making, as highlighted by Rudin et al., 2019 \cite{rudin2019stop}. On the other hand, interpretable GNNs \cite{dai2021towards,zhang2022protgnn} feature designs that inherently facilitate interpretability by leveraging their internal mechanisms or elements. Despite these innovations, current approaches generally concentrate on model performance and algorithmic precision over the diversity of explanation generation and a critical evaluation of how users perceive these explanations. Truly comprehensive explanations typically require merging various data sources, each providing distinct perspectives that collectively deepen user comprehension of the model’s predictions.

% motivation
%The motivation for this work stems from humans' inherent cognitive abilities in learning from examples.  For instance, a child can generalize the concept of "elephant" from a few pictures. Applying this concept to ML, Vinyals et al. \cite{vinyals2016matching} underline the importance of combining parametric and non-parametric models by predicting using reference examples. In XAI, recent studies \cite{cai2019effects} have emphasized the significance of example-based explanations, as they enable users to better understand model behaviors, even in cases where the models make incorrect predictions. Additionally, concept-based explanations \cite{ghorbani2019towards} have shown a promising direction in enhancing the comprehension of model behaviors. These insights emphasize the need to explore and leverage example-based explanations to enhance the interpretability and understanding of complex ML models.

The motivation for this research is rooted in the natural cognitive capabilities of humans to learn from past examples. For example, a child can easily deduce the general concept of a ``cat" or a ``dog" from just a few images. In the context of machine learning, Vinyals et al., 2016 \cite{vinyals2016matching} highlight the critical role of integrating parametric and non-parametric models to predict outcomes based on references. In the XAI field, recent investigations, such as those by Cai et al., 2019 \cite{cai2019effects}, have underlined the value of example-based explanations. This approach to explanations provides different insights into model decisions, especially when errors occur in predictions. Furthermore, concept-based explanations \cite{ghorbani2019towards} offer a promising avenue for improving the understanding of model behaviors. These findings underscore the importance of further investigations on example-based explanations to improve the overall explainability of complex models like GNNs.

%This chapter proposes a novel interpretable \textbf{G}NN architecture based on a \textbf{CON}cept-matching model named \textbf{CONG}, designed to achieve high prediction accuracy and interpretability. The architecture consists of five key components: a GNN encoder, a concept discovery module, a concept-based prediction function, a concept corpus, and an explanation construction module. The GNN encoder is responsible for encoding the input graphs' structural information and relational dependencies, enabling the model to uncover hidden patterns and correlations in the data. Subsequently, the concept discovery module extracts salient concepts from input graphs, representing frequent substructures that capture abstract information related to groups of outcomes. This module is trained using the graph information bottleneck theory \cite{yu2020graph} with modified constraints. All concepts derived from training graphs are managed in an in-memory concept corpus. During inference, the concept discovery module first identifies minimal sufficient substructures from an input graph. The non-parametric concept-based prediction function looks up references from the corpus and makes a prediction using an attention mechanism. Finally, an explanation construction module leverages retrieved concepts and attention scores to generate multiple explanations for various scenarios and user preferences. Extensive experiments and a thorough user study validate the performance of the proposed approach in model prediction and explanation generation.

This research introduces an innovative approach to interpretable GNNs called \textbf{CONG}, which incorporates a concept-matching model to simultaneously enhance predictive performance and model interpretability. The architecture is comprised of five principal modules: a graph encoder, a concept discovery module, an interpretable prediction function, a concept corpus, and an explanation module. The graph encoder is tasked with capturing the structural and relational dependencies within input graphs, thus revealing underlying patterns and associations. Following this, the concept discovery module identifies significant concepts that represent common substructures, encapsulating generic information pertinent to specific outcome groups. This module operates under a training paradigm extended from the graph information bottleneck theory \cite{yu2020graph}. It extracts and stores all concepts from training graphs in an in-memory concept repository. In the inference phase, this module pinpoints essential substructures in an input graph. Interpretable prediction functions then refer to concepts in the corpus and utilize an attention mechanism for making predictions. Subsequently, the explanation module uses these concepts and reference scores to craft multiple explanations tailored to different situations and user needs. Comprehensive testing and an in-depth user study affirm the effectiveness of this model in both prediction and explanation capability.

%The research presented in this chapter, including the proposed method and experimental results, was published in \cite{bui2023toward,bui2024toward}. The chapter's remainder is as follows. \cref{related_work} describes related work. \cref{method} presents the methodology. Experiments are reported in \cref{con_exp_setups,exp_results}. The paper is concluded in \cref{conclusion}.

The findings presented in this chapter, including the proposed method and experimental results, were published in \cite{bui2023toward,bui2024toward}. The remainder is structured as follows. \cref{related_work} includes a literature review of related works. \cref{method} details the methodology employed. The experimental setups and results are explored in \cref{con_exp_setups,exp_results}. The chapter concludes with \cref{conclusion}.

\section{Related Work} \label{sec:cm_related_work}

\subsection{Subgraph Discovery and Graph Retrieval}
Subgraph discovery aims to identify meaningful patterns within larger graphs, providing insights into component relationships. Traditional approaches involve graphlet decomposition \cite{ahmed2017graphlet}, domain-specific pattern recognition \cite{degen2008art}, sampling-based strategies \cite{huang2021towards}, or clustering algorithms \cite{Theodoridis_Koutroumbas_2009}. Other methods include frequent subgraph mining \cite{yan2002gspan} or dense subgraph discovery \cite{fang2022densest}. Recently, GNN explanation methods have given rise to multiple subgraph recognition methods \cite{ying2019gnnexplainer,luo2020parameterized,yu2020graph}.

Graph Retrieval involves retrieving similar graphs of a query graph from a large collection. Structural similarity is typically measured using graph matching, which can be exact or approximate. Exact matching methods \cite{bonnici2013subgraph} are prevalent in domains with deterministic connections like biology or chemistry. For domains with complex and uncertain graphs, approximate matching \cite{jun2017sequential} is more suitable. Traditional methods are based on graph edit distance (GED) or Monte Carlo approaches. With the help of GNN encoders, multiple neural matching methods \cite{li2019graph,lou2020neural,roy2022interpretable} have been proposed lately.

\subsection{Explanation Methods based on Subgraphs}

Existing GNN explanation methods \cite{yuan2022explainability} primarily concentrate on discovering essential subgraphs from inputs contributing to certain model behaviors. These methods can be categorized into instance-level and model-level approaches \cite{yuan2022explainability}. Instance-level methods focus on extracting subgraphs from an input graph leading to a specific prediction, while the model level aims to generate patterns associated with groups of predictions. Instance-level explanations have received significant attention, resulting in numerous publications \cite{ying2019gnnexplainer, luo2020parameterized,  schlichtkrull2020interpreting, yuan2021explainability}. However, existing publications overlook the importance of user perception assessment and mostly concentrate on algorithmic evaluation.

\subsection{Measuring Similarity in Graph Structures}
%Graph structure similarity measurement is an essential problem. Several methods \cite{nikolentzos2017matching}\cite{togninalli2019wasserstein}\cite{vincent2021semi} leverage the family of Wasserstein distance to solve the graph similarity problem. Most of them utilize the graph similarity metric to design graph kernels for leveraging conventional ML algorithms, such as SVM, to solve downstream tasks. Practically, high-order Wasserstein metrics are computationally intensive compared to EMD \cite{cuturi2013sinkhorn}. Lately, Vincent et al. \cite{vincent2022template} proposed to add one layer on top of GNNs, which computes structure similarity between an input graph and templates using Fused Gromov-Wasserstein distance. Unlike us, it learned template structures due to the burden of template selection costs. Furthermore, none of the methods discuss model interpretability and the importance of weighting node contributions.

Measuring graph structure similarity is critical in interpretable GNNs. Various methods, including those by Nikolentzos et al., 2017 \cite{nikolentzos2017matching}, Togninalli et al., 2019 \cite{togninalli2019wasserstein}, and Vincent et al., 2021 \cite{vincent2021semi}, utilize the Wasserstein distance family to address the issue of graph similarity. These methods often employ graph similarity to create graph kernels that enable the use of traditional ML algorithms, such as Support Vector Machines, for downstream tasks. However, higher-order Wasserstein metrics can be computationally demanding when compared to the Earth Mover's Distance (EMD) \cite{cuturi2013sinkhorn}. More recently, Vincent et al., 2022 \cite{vincent2022template} introduced an additional layer to GNNs, which calculates the structural similarity between an input graph and templates through the Fused Gromov-Wasserstein distance. This approach contrasts with others as it involves learning template structures to mitigate the costs associated with template selection. However, these methods do not typically consider model interpretability or the significance of weighting contributions from individual nodes.

\subsection{Interpretable Graph Neural Networks}
Interpretable GNNs aim to provide transparent and understandable explanations for their predictions and behaviors. Techniques such as attention mechanisms, label or feature propagation, and prototypes enhance GNN interpretability. Graph Attention Network \cite{velickovic2017graph} employed attention layers to capture the relevance of neighboring nodes. Wang et al. \cite{wang2020unifying} proposed combining label propagation with GCN \cite{kipf2016semi}, offering a novel solution for self-explanation. Recent studies by Zhang et al. \cite{zhang2022protgnn} and Dai et al. \cite{dai2021towards} integrated similarity modules with GNN encoders to improve prediction accuracy and interpretability. However, the method proposed by Dai et al. faced challenges with slow training and did not adequately address the construction of explanations. Furthermore, current methodologies focus on predictive performance while neglecting the importance of evaluation on user perception of explanations.

\section{Concept Matching Model} \label{sec:cm_method}

\begin{figure}[ht]
	\centering
	\includegraphics[width=\linewidth]{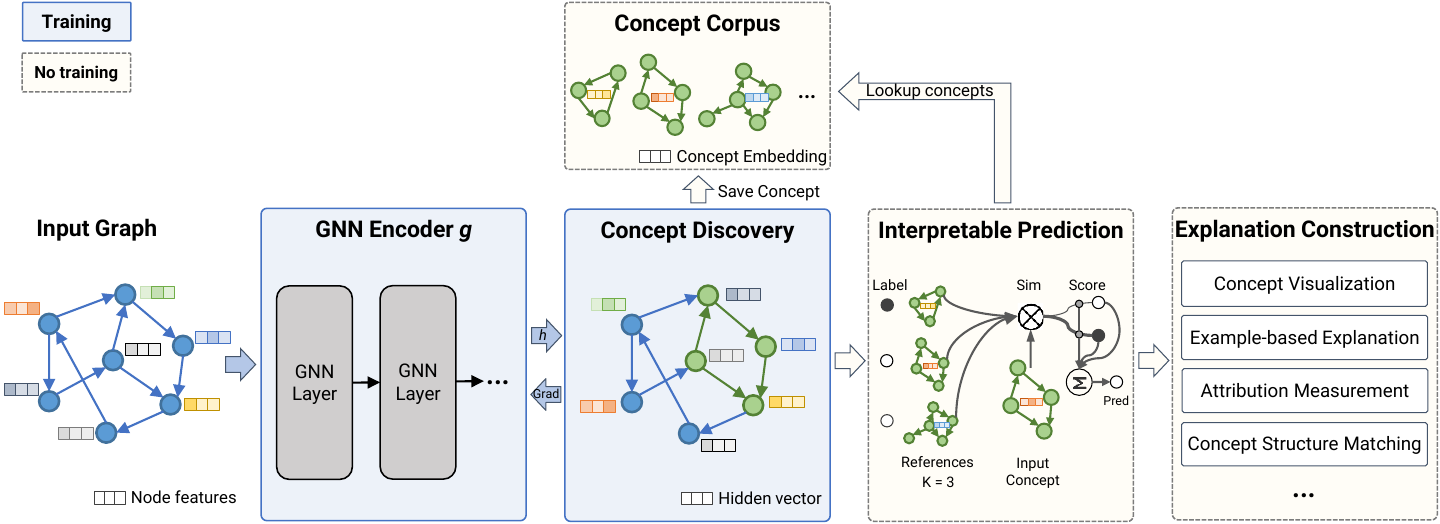}
	\caption{An Overview of CONG. It contains two trainable and three non-trainable components represented in blue and yellow boxes, respectively.}
	\label{fig:cm_overview}
\end{figure}

\subsection{Problem Statement}
Let $\mathcal{D} = \{(\mathcal{G}_1, Y_1),...,(\mathcal{G}_N, Y_N)\}$ be a dataset of $N$ samples, wherein each includes a graph $\mathcal{G}$ with a real-value outcome or class $Y$. An $i^{th}$ graph $\mathcal{G}_i = \{V, E, \mathbf{A}, \mathbf{X}_v\}$ comprises a vertex set $V$, an edge set $E$, an adjacency matrix $\mathbf{A}$, and a node feature matrix $\mathbf{X}_v$. 

% modified
Let $\tilde{\mathcal{G}}$ be a subgraph of a graph $\mathcal{G}$. Based on a GNN encoder $g$, a function $s$ identifies essential subgraphs $\tilde{\mathcal{G}}$. Each extracted subgraph $\tilde{\mathcal{G}}$ is considered a concept, and a set of concepts is regarded as a concept corpus $\mathcal{R}$. Subsequently, a concept-oriented predictor $P$ derives a predicted outcome $\hat{Y}$ for $\mathcal{G}_i$ using a subset of reference concepts $\mathcal{R}_{\tilde{\mathcal{G}}} = \{r_1, r_2,..., r_K\}$, where $\mathcal{R}_{\tilde{\mathcal{G}}} \subset \mathcal{R}$. From $\mathcal{R}_{\tilde{\mathcal{G}}}$, $\tilde{\mathcal{G}}$, and $P$, a series of explanations $\mathcal{E} = \{e_1, e_2, .., e_M\}$ is produced.

\subsection{Overview}

%As presented in \cref{fig:cm_overview}, the proposed model comprises five essential components. The GNN encoder encodes input graphs into the embedding space. Based on graph representations, a concept discovery module extracts concepts from input graphs, representing frequent substructures. Concepts derived from training graphs are managed in a concept corpus. During inference, the non-parametric prediction function looks up reference concepts and makes a prediction based on an attention mechanism. Finally, an explanation module, consisting of multiple functions implemented based on discovered concepts, generates multiple explanations for various scenarios. A step-by-step execution process is presented in \cref{alg:cm_concept_matching}.

As depicted in \cref{fig:cm_overview}, the presented paradigm features five critical components. A GNN encoder transforms input graphs into embeddings. Utilizing these representations, a concept discovery module identifies and extracts frequent substructures as concepts from input graphs. These concepts, once derived from the training graphs, are organized into a concept corpus. In the inference phase, a non-parametric predictor grounds its decisions on the proximity between inputs and retrieved concepts. Additionally, an explanation module, which comprises several functions leveraging identified concepts and similarity scores, produces varied explanations for different scenarios. The entire process is sequentially detailed in \cref{alg:cm_concept_matching}.

\begin{algorithm}[ht]
	\caption{An Overall Algorithm}
	\begingroup
	\raggedright
	\textbf{Input}: Initialized GNN $g$, $\phi$, $\theta$, 
	\\ \hspace{0.92cm} Training dataset  $\mathcal{D}$, Num. of epochs $T$ \\
	% \\\hspace{0.92cm} Keep-going probability $d$,
	\textbf{Output}: Trained $g$ and $s$, Concept corpus $\mathcal{R}$  \\
	\endgroup
	\begin{algorithmic}[1]
		% \COMMENT{Training Phase}
		\FOR{$i = 1$ to $T$}
		\STATE{Execute $g$ and $s$} \COMMENT {\cref{sec:cm_gnn,sec:cm_con_dis}}
		\STATE{Train $g$ and $s$ via \cref{eq:cm_overall_loss}} \COMMENT {Update model weights}  
		\STATE{Update $\mathcal{R}$ via \cref{eq:cm_corpus}} \COMMENT {\cref{sec:cm_corpus}}
		\ENDFOR
		% \COMMENT{Inference Phase}
		\STATE{Execute the non-parametric predictor} \COMMENT {\cref{sec:cm_cong_func}}
		\STATE{Generate explanations} \COMMENT {\cref{sec:cm_explanation}}
	\end{algorithmic}
	\label{alg:cm_concept_matching}
\end{algorithm}

\subsection{GNN Encoder} \label{sec:cm_gnn}

A GNN encoder function is the fundamental input block in the proposed paradigm, designed to accommodate diverse GNN architectures. Specifically, it transforms input graphs into low-dimensional representations that capture both the structural and attribute features of the graph. A GNN encoder function is represented as follows: 

\begin{equation}
	\begin{aligned}
		\mathbf{H}^l &= \text{GNN}(\mathcal{G}, \mathbf{A}, \mathbf{H}^{l-1}), \\
	\end{aligned}
	\label{eq:gnn}
\end{equation}
where $l$ denotes the current GNN layer, and $\mathbf{H}$ represents a representation matrix of all nodes. The initial $\mathbf{H}^0$ is $\mathbf{X}_v$. 

\subsection{Concept Discovery} \label{sec:cm_con_dis}
A ``concept" refers to an abstract idea representing a category or mental construct. In this research, concepts refer to frequent substructures that signify specific outcomes and facilitate the creation of interpretable representations.

The framework adopts the information bottleneck principle to extract concepts from graphs by adding an additional constraint to control the informative representation learning process. First, let us recall the definition of graph information bottleneck (GIB) from \cite{yu2020graph}, which is represented as:
\begin{equation}
	\begin{aligned}
		\max_Z \quad & I(\hat{Y}, \tilde{\mathcal{G}}) \\
		\text{s.t.} \quad & I(\mathcal{G}, \tilde{\mathcal{G}}) \leq I_c,
	\end{aligned}
\end{equation}
where $I_c$ is the constraint for mutual information (MI) between $\mathcal{G}$ and $\tilde{\mathcal{G}}$. The goal is finding minimal sufficient substructures maximizing $I(\hat{Y}, \tilde{\mathcal{G}})$. An additional constraint for MI between $\mathcal{G}$ and $\hat{Y}$ is implemented to ensure that the trained model does not overly focus on only specific subgraphs. These modifications result in the following optimization problems:

\begin{equation}
	\begin{aligned}
		\max_{\tilde{\mathcal{G}} \subset \mathcal{G}} \, & I(\hat{Y}, \tilde{\mathcal{G}}) \\ 
		\quad \text{s.t.} \, &  I(\mathcal{G}, \tilde{\mathcal{G}}) \leq I_{c_1}, \\ 
		\, & I(\hat{Y}, \mathcal{G}) \geq I_{c_2},
	\end{aligned}
	\label{eq:optimization}
\end{equation}
where $I_{c_1}$ and $I_{c_2}$ serve as MI constraints. By applying Lagrange multipliers to \cref{eq:optimization}, it becomes unconstrained, as follows:
\begin{equation}
	\max_{\tilde{\mathcal{G}} \subset \mathcal{G}} \quad I(\hat{Y}, \tilde{\mathcal{G}}) - \alpha I(\mathcal{G}, \tilde{\mathcal{G}}) + \beta I(\hat{Y}, \mathcal{G}).
	\label{eq:uncon_optim}
\end{equation}

\cref{eq:uncon_optim} consists of three terms. Solving the first and the third terms are similar. The first term, measuring the MI between $\tilde{\mathcal{G}}$ and $\hat{Y}$, can be expanded as:

\begin{equation}
	\begin{aligned}
		I(\hat{Y}, \tilde{\mathcal{G}}) &= H(\hat{Y}) - H(\hat{Y}|\tilde{\mathcal{G}}) \\
		&= H(\hat{Y}) + \int p(\hat{y}, \tilde{\mathcal{G}})\,\text{log}\,p(\hat{y} | \tilde{\mathcal{G}})\,d\hat{y}\,d\tilde{\mathcal{G}}
	\end{aligned}
	\label{eq:ig}
\end{equation}
The entropy term $H(Y)$ can be discarded. It is proved in \cite{yu2020graph} that the second term in \cref{eq:ig} can be relaxed using a variational approximation $q_{\phi}(\hat{y}|\tilde{\mathcal{G}})$, as represented in the following equation:

\begin{equation}
	I(\hat{Y}, \tilde{\mathcal{G}}) \geq \frac{1}{N} \sum_{i=1}^N \text{log}\,q_{\phi}(\hat{y}_i | \tilde{\mathcal{G}}_i) =: -\mathcal{L}_{cls}(q_{\phi}(\hat{y}|\tilde{\mathcal{G}}), y),
	\label{eq:iq_app}
\end{equation}
where $\mathcal{L}_{cls}$ is the cross-entropy loss function between $\hat{y}_i$ and $\tilde{\mathcal{G}}_{i}$, and $y$ is the ground-truth label for corresponding graphs. \cref{eq:ig} and \cref{eq:iq_app} can also be applied to $I(\hat{Y}, \mathcal{G})$.

In practice, minimizing $\mathcal{L}_{cls}$ is equivalent to increasing the predictive capability of the subgraph about the graph outcome. Consequently, maximizing $I(\hat{Y}, \mathcal{G})$ and $I(\hat{Y}, \tilde{\mathcal{G}})$ is analogous to minimizing their respective objective functions, which can be combined into a single form as follows:

\begin{equation}
	\mathcal{L}_{cls}(\phi, \mathcal{G}, \tilde{\mathcal{G}}) = \mathcal{L}_{cls}(q_{\phi}(\hat{y}|\tilde{\mathcal{G}}), y) + \beta \mathcal{L}_{cls}(q_{\phi}(\hat{y}|\mathcal{G}), y).
	\label{eq:cls}
\end{equation}

%The most difficult part is minimizing $I(\mathcal{G}, \tilde{\mathcal{G}})$, the second term in \cref{eq:uncon_optim}, due to the discrete nature of graphs. Following \cite{yu2020graph}, the Donsker-Varadhan representation of the KL-divergence is applied to $I(\mathcal{G}, \tilde{\mathcal{G}})$, then the following equation is obtained:

The greatest challenge lies in reducing $I(\mathcal{G}, \tilde{\mathcal{G}})$, the second element of \cref{eq:uncon_optim}, which is complicated by the discrete characteristics of graphs. Referred to \cite{yu2020graph}, this research approximates this term via the Donsker-Varadhan representation of KL-divergence, as follows:

\begin{equation}
	\begin{aligned}
		I(\mathcal{G}, \tilde{\mathcal{G}}) = \sup_{f_{\theta}}\,
		& \mathbb{E}_{\mathcal{G},\tilde{\mathcal{G}} \in p(\mathcal{G}, \tilde{\mathcal{G}})}[f_{\theta}(\mathcal{G}, \tilde{\mathcal{G}})] \\
		& - \text{log}\,\mathbb{E}_{\mathcal{G} \in p(\mathcal{G}),\tilde{\mathcal{G}} \in p(\tilde{\mathcal{G}})} [e^{f_{\theta}(\mathcal{G}, \tilde{\mathcal{G}})}],
	\end{aligned}
	\label{eq:graph_sim}
\end{equation}
where $f_{\theta}(\mathcal{G}, \tilde{\mathcal{G}})$ outputs a real value for two given graphs. The function $f_{\theta}$ is a function measuring the similarity between two graphs. The objective is to maximize the similarity score between closely related graphs while minimizing the value between unrelated ones. This objective is formulated as follows:

%The goal is to maximize the score between two close graphs and minimize the score between irrelevant ones. This objective leads us to solve the following optimization problem:

\begin{equation}
	\begin{aligned}
		\max_{\theta}\quad \mathcal{L}_{sc} (\theta, \mathcal{G}, \tilde{\mathcal{G}}) = &\frac{1}{N} \sum_{i=1}^N f_{\theta}(\mathcal{G}_i, \tilde{\mathcal{G}}_i) \\
		&- \text{log}\,\frac{1}{N} \sum_{i=1, j \neq i}^N e^{f_{\theta}(\mathcal{G}_i, \tilde{\mathcal{G}}_{j})}.
	\end{aligned}
	\label{eq:max_graph_sim}
\end{equation}

Practically, $f_{\theta}$ is a function that processes the graph embeddings of $\mathcal{G}$ and $\tilde{\mathcal{G}}$. These embeddings are merged together prior to being fed into an MLP layer to derive a similarity score. \cref{eq:cls,eq:graph_sim,eq:max_graph_sim} are combined into a tractable bi-level optimization problem via an approximation for $I(\mathcal{G}, \tilde{\mathcal{G}})$ as follows:

\begin{equation}
	\begin{aligned}
		\min_{\tilde{\mathcal{G}},\phi}\quad & \mathcal{L}(\tilde{\mathcal{G}}, \phi, \theta^{*}) = \mathcal{L}_{cls} + \alpha \mathcal{L}_{sc} \\
		\text{s.t.}\quad &\theta^{*} = \argmax_{\theta} \mathcal{L}_{sc}.
	\end{aligned}
	\label{eq:objective}
\end{equation}

%\cref{eq:objective}'s optimization involves an inner loop followed by the minimization of the overall objective function. It also requires a differentiable function for $\tilde{\mathcal{G}}$ generation. Node-based and edge-based methods are proposed as follows.

The optimization process of \cref{eq:objective} aims to minimize the overall loss, while optimizing the similarity score function via an inner loop. Additionally, this process necessitates the use of a differentiable function to generate $\tilde{\mathcal{G}}$. As solutions, both node-based and edge-based approaches are introduced.

\noindent \textbf{Node-based Concept Discovery:} As shown in \cref{eq:node_based}, the node embeddings of a GNN encoder are fed to an MLP model, followed by a softmax operator. Before executing the softmax operator, the reparameterization trick \cite{luo2020parameterized} is applied to encourage selection probabilities to be discrete. 

\begin{equation}
	\begin{aligned}
		\mathbf{S} &= \text{softmax}(\text{MLP}(\mathbf{H}^l)) \\
	\end{aligned}
	\label{eq:node_based}
\end{equation}
Using $\mathbf{S}$, the graph embeddings for $\mathcal{G}$ and $\tilde{\mathcal{G}}$ can be calculated based on \cref{eq:graph_emb}. 

\begin{equation}
	\begin{aligned}
		h^n_{\tilde{\mathcal{G}}} &= \mathbf{S}^T \mathbf{H}^l \\
		h^n_{\mathcal{G}} &=  \mathbf{1}^T \mathbf{H}^l \\
	\end{aligned}
	\label{eq:graph_emb}
\end{equation}

The subgraph $\tilde{\mathcal{G}}$ can be constructed in two ways. The first one is obtaining nodes whose values are close to 1 from the first column of $\mathbf{S}$, assuming that this assignment matrix is well-trained and its values saturate to 0/1. The other approach is less aggressive, which performs the following calculation: $\mathbf{A}^n_{\tilde{\mathcal{G}}} = \mathbf{S}^T \mathbf{A}$. In the second one, a node is in the subgraph if most of its neighbors are selected.

\noindent \textbf{Edge-based Concept Discovery:} In this approach, each edge is assigned a score $m_{ij}$ representing whether it is selected. The reparameterization trick is also applied to $m_{ij}$.

\begin{equation}
	\begin{aligned}
		m_{ij} &= \sigma (\text{MLP}(\text{CONCAT}(h_i, h_j))) \\
		h_{ij} &= m_{ij} (h_i + h_j)
	\end{aligned}
	\label{eq:edge}
\end{equation}

\cref{eq:edge} presents how to calculate the edge score and representation, where ${ij}$ represents an edge between two nodes $i$ and $j$, and $\sigma$ denotes a non-linear function. Next, graph embeddings for $\mathcal{G}$ and $\tilde{\mathcal{G}}$ and the extracted subgraph's adjacency matrix  $\mathbf{A}^e_{\tilde{\mathcal{G}}}$ in \cref{eq:edge_graph_emb} are defined. Also, $\mathbf{M}$ is the matrix formulated by multiple elements $m$, and $\odot$ is the Hadamard product.

\begin{equation}
	\begin{aligned}
		\mathbf{A}^e_{\tilde{\mathcal{G}}} &= \mathbf{M} \odot \mathbf{A} \\
		h^e_{\tilde{\mathcal{G}}} &= \sum_{ij} h_{ij} \\
		h^e_{\mathcal{G}} &= h^n_{\mathcal{G}} \\
	\end{aligned}
	\label{eq:edge_graph_emb}
\end{equation}

%\noindent\textbf{Connectivity Loss:} The following connectivity loss functions are defined for the two discovery methods above to further encourage the concept discovery model to extract the smallest possible subgraph.

\noindent\textbf{Connectivity Loss:}  The connectivity loss functions outlined below are specified for the node-based and edge-based discovery methods to further promote the model's ability to extract the smallest feasible substructures.

\begin{equation}
	\begin{aligned}
		\mathcal{L}^n_{con} &= || \text{Norm}(\mathbf{S}^T \mathbf{A} \mathbf{S}) - \mathbf{I}_2||_F \\
		\mathcal{L}^e_{con} &= \sum_{ij} m_{ij} - B
	\end{aligned}
	\label{eq:connectivity}
\end{equation}
%In \cref{eq:connectivity}, $\mathcal{L}^n_{con}$ and $\mathcal{L}^e_{con}$ denote connectivity losses for node-based and edge-based approaches, respectively. In the edge-based loss, $B$ is a predefined budget smaller than the number of edges. The node-based loss is similar to \cite{yu2020graph}, which includes a $2\times2$ identity matrix $\mathbf{I}_2$.

In \cref{eq:connectivity}, $\mathcal{L}^n_{con}$ and $\mathcal{L}^e_{con}$ represent regularization terms for network connectivity corresponding to methods based on nodes and edges. The regularization in the node-based approach aligns with \cite{yu2020graph}, which incorporates a $2\times2$ identity matrix $\mathbf{I}_2$. For the approach based on edges, $B$ specifies a budget that is less than the total edge count.

Integrating \cref{eq:objective} with \cref{eq:connectivity} results in the final objective function as follows:

\begin{equation}
	\begin{aligned}
		\min_{\tilde{\mathcal{G}},\phi}\quad & \mathcal{L}(\tilde{\mathcal{G}}, \phi, \theta^{*}) = \mathcal{L}_{cls} + \alpha \mathcal{L}_{sc} + \lambda \mathcal{L}_{con} \\
		\text{s.t.}\quad &\theta^{*} = \argmax_{\theta} \mathcal{L}_{sc}.
	\end{aligned}
	\label{eq:cm_overall_loss}
\end{equation}

\subsection{Concept Corpus Management} \label{sec:cm_corpus}
After training a GNN encoder and a concept extraction model, the framework executes these modules on graphs in a training set to extract concepts. Two levels of indices are constructed based on concept representation vectors with a k-centroid approach to efficiently manage concepts in memory for inferences, as follows:
\begin{equation}
	\begin{aligned}
		\mathcal{I}_{\mathcal{D}} &= \text{build\_index}( \{h_{\tilde{\mathcal{G}}} \}_{i=1}^{|\mathcal{R}|}, K_c), \\
		\mathcal{I}_c &=\text{build\_class\_index}( \{h^c_{\tilde{\mathcal{G}}} \}_{i=1}^{|\mathcal{R}_c|}, K_c).
	\end{aligned}
	\label{eq:cm_corpus}
\end{equation}

$\mathcal{I}_{\mathcal{D}}$ and $\mathcal{I}_c$ represent indices for the whole concept repository and for a subset corresponding to a class $c$. $|\mathcal{R}|$ and $|\mathcal{R}_c|$ represent the number of concepts of the whole corpus and a class, respectively. The number of centroids for clustering is indicated by $K_c$. In practice, Faiss \cite{johnson2019billion} is utilized to implement indexing functions. 

\subsection{Graph Structure Similarity} \label{sec:cm_structure}

While Euclidean distance is a valuable measure for interpretable predictions, incorporating graph structure similarity can provide a supplementary perspective. This is particularly useful in cases where the Euclidean-based strategy selects references that users find difficult to understand. Although graph edit distance is a traditional approach to measure structure similarity, its exponential time complexity, specifically $O(2^{|V|+|E|})$, poses a significant challenge for practical applications that require efficient computation. This research proposes to address this computational problem via the optimal transport theory with EMD \cite{rubner2000earth}, a metric for measuring distances between two sets of weighted objects.

Let $\mathcal{V}_q = \{(v_q^1, w_q^1),..., (v_q^N, w_q^N)\}$ and $\mathcal{V}_r = \{(v_r^1, w_r^1),...,(v_r^N, w_r^N)\}$ be vertex-weight pairs of a query graph and a reference graph. Let $d_{ij}$ be a Euclidean distance between ($v_q^i$, $v_r^j$) and $\mathbf{D} = (d_{ij}) \in \mathbb{R}^{N \times N}$ be the ground distance matrix. The transport flow between $\mathcal{V}_q$ and $\mathcal{V}_r$ is denoted by $\mathbf{T} = (t_{ij}) \in \mathbb{R}^{N \times N}$, with $t_{ij}$ indicating the transport cost from $v_q^i$ to $v_r^j$. The goal is to determine the optimal transport flow $\mathbf{T}^*$ that minimizes the cost function, as follows:

\begin{equation}
	\begin{aligned}
		\text{COST} & (\mathcal{V}_q, \mathcal{V}_r, \mathbf{T}) = \sum_{i=1}^N \sum_{j=1}^N d_{ij} t_{ij} \\
		\text{s.t} & \quad t_{ij} \geq 0, \quad \sum_{j=1}^{N} t_{ij} \leq w_q^i \quad \sum_{i=1}^{N} t_{ij} \leq w_r^j, \\
		& \quad \sum_{i=1}^N \sum_{j=1}^N t_{ij} = \text{min} \biggl(\sum_{i=1}^N w_q^i, \sum_{j=1}^N w_r^j\biggr).
	\end{aligned}
	\label{eq:emd}
\end{equation}

Weights are normalized such that $\sum_{i=1}^N w_q^i = \sum_{j=1}^N w_r^j = 1$. The optimal transport matrix $\mathbf{T}^*$ is obtained via the Sinkhorn algorithm \cite{cuturi2013sinkhorn}. The distance or structural similarity between two graphs is then defined as:
\begin{equation}
	d_{\textrm{sc}}(\mathcal{V}_q, \mathcal{V}_r) = \sum_{i=1}^N \sum_{j=1}^N d_{ij} t^*_{ij}, \quad
	s_{\textrm{sc}}(\mathcal{V}_q, \mathcal{V}_r) = \sum_{i=1}^N \sum_{j=1}^N s_{ij} t^*_{ij},
\end{equation}
where $d_{ij}$ is Euclidean distance and $s_{ij} = \textrm{exp}(-d_{ij})$ is Gaussian similarity. 

%A naive node weighting method is uniform initialization, where $w_i = 1 / N$. However, weighting nodes by their contributions enhances prediction accuracy and structural correspondence comprehension. Intuitively, nodes appeared in a concept signifying specific outcomes merit higher weights. $\mathbf{S}$'s first column ($\mathbf{S}_0$) comprises probabilities indicating that vertices are included in $\mathcal{G}_s$. Intuitively, nodes in $\mathcal{G}_s$ should have higher probability values than others. Row-wise normalized probabilities are used for calculating the importance weight $w_i$ of a node $i$ as follows: $w_i = s_i / \sum_{j=0}^{N} s_j$, where $s_i$ and $s_j$ are rows $i$ and $j$'s values in $\mathbf{S}_0$.  Note that, outputs of GNN encoder and the concept discovery module can be reused to initialize the procedure's parameters.

A straightforward approach to node weighting involves uniform initialization, setting each node weight, $w_i$, at $1 / N$. However, assigning weights based on node contributions can improve both prediction accuracy and understanding of structural relationships. Nodes that are crucial for specific outcomes within a concept generally warrant increased weights. The first column of matrix $\mathbf{S}$ ($\mathbf{S}_0$) contains probabilities that reflect the likelihood of vertices being part of $\mathcal{G}_s$. Logically, nodes within $\mathcal{G}_s$ are assigned higher probability values compared to others. The importance weight $w_i$ for node $i$ is determined using row-wise normalized probabilities, calculated as $w_i = s_i / \sum_{j=0}^{N} s_j$, where $s_i$ and $s_j$ represent the respective values from rows $i$ and $j$ in $\mathbf{S}_0$. It is important to note that the outputs from the GNN encoder and the concept discovery module can be leveraged to initialize the parameters for this procedure.

\subsection{Concept-based Prediction Function} \label{sec:cm_cong_func}
Given $\mathcal{G}$, $\tilde{\mathcal{G}}$, a representation vector $h_{\tilde{\mathcal{G}}}$, a set of reference concepts $\mathcal{R}_{\tilde{\mathcal{G}}} = \{r_1, r_2,..., r_K\}$, a set of representation vectors $H^r = \{ h_1^r, h_2^r,..., h^r_K \}$ of references, and a set of ground-truth labels $Y = \{y_1, y_2, ..., y_K\}$, the goal is to find a function $P$ assigning a label $\hat{y}$ for $\mathcal{G}$. This goal raises two following questions. How to determine a set of references? How to infer the prediction?

\noindent\textbf{Reference set construction:} This work proposes three simple yet effective strategies for reference construction based on the KNN and k-centroids algorithms, as represented in \cref{eq:refset}. Given a concept embedding $h_{\tilde{\mathcal{G}}}$ and a corpus index $\mathcal{I}$ or class indices $\mathcal{I}_c$, the KNN algorithm returns $K$ most similar concepts $\mathcal{R}_{\tilde{\mathcal{G}}}$ to an input graph. Similarly, the k-centroids algorithm retrieves $K_c$ central points of each class with $\mathcal{I}_c$.

\begin{equation}
	\begin{aligned}
		\mathcal{R}_{\tilde{\mathcal{G}}} &= \text{KNN}(\mathcal{I}, h_{\tilde{\mathcal{G}}}, K) \\
		\mathcal{R}_{\tilde{\mathcal{G}}} &= \{\text{KNN\_Class}(\mathcal{I}_c, h_{\tilde{\mathcal{G}}}, K)\}_{c=1}^C \\
		\mathcal{R}_{\tilde{\mathcal{G}}} &= \{\text{K\_Centroids}(\mathcal{I}_c, h_{\tilde{\mathcal{G}}}, K_c)\}_{c=1}^C \\
	\end{aligned}
	\label{eq:refset}
\end{equation}

%Naively, structural similarity can be calculated between $\mathcal{G}$ and all graphs in training data. However, this approach results in enormous computational costs due to the complexity of \cref{eq:emd}. The following two-stage approach can reduce the computational burden. First, a Euclidean-based strategy is executed to select $\alpha \times K$ number of candidates with the smallest Euclidean distances, where $\alpha > 1$. Second, the structural similarities are calculated between $\mathcal{G}$ and $\alpha \times K$ candidates from the first state and re-rank them based on new scores. The top $K$ candidates with the highest structural similarities are chosen as references.

%\noindent\textbf{Two-stage Reference Selection for Structural Similarity:} Initially, one might compute structural similarity between graph $\mathcal{G}$ and all graphs in the training dataset. However, this method incurs substantial computational costs owing to the complexity outlined in \cref{eq:emd}. To alleviate this, a two-stage approach is suggested. The first phase involves a Euclidean-based method to identify $\alpha \times K$ graphs with the least Euclidean distances, where $\alpha > 1$. In the second phase, structural similarities are determined between $\mathcal{G}$ and these selected $\alpha \times K$ candidates from the initial phase, followed by a re-ranking based on these newly calculated scores. Ultimately, the top $K$ graphs exhibiting the highest structural similarities are selected as reference points.

\noindent\textbf{Two-stage Reference Selection for Structural Similarity:} Direct computations of structural similarities between an input graph and all graphs from a training dataset might be costly and intensive due to the time complexity of \cref{eq:emd}. A more efficient, two-stage strategy is proposed to address this issue. Initially, an Euclidean-based function is implemented to shortlist $\alpha \times K$ graphs ranked by their distances to the input, where $\alpha > 1$ represents the multiplier for an expanded candidate pool. Subsequently, in the second stage, the focus shifts to assessing structural similarities between the input graph and these pre-selected $\alpha \times K$ graphs. This step is followed by a re-ranking process using the similarity scores computed in this phase. The final selection comprises only $K$ graphs that show the highest levels of structural similarity, serving as references.

%\noindent\textbf{Prediction Inference:}  The interpretable non-parametric predictor $P$ takes similarity scores between an input graph $\mathcal{G}$ and references in $\mathcal{R}_\mathcal{G}$ as parameters, as presented in \cref{eq:non_param}.

\noindent\textbf{Non-parametric Predictor:} As defined in \cref{eq:non_param}, a predictor $P$ utilizes similarity scores derived from reference selection strategies as its parameters for inferring predictions.

\begin{equation}
	\begin{aligned}
		P(\hat{y} | h_{\tilde{\mathcal{G}}}, H^r) &= \sum_{i=1}^K a(h_{\tilde{\mathcal{G}}},  h_i^r) y_i, \\
		a(h_{\tilde{\mathcal{G}}},  h_i^r) &= \text{softmax} (sim(h_{\tilde{\mathcal{G}}},  h_i^r)),
	\end{aligned}
	\label{eq:non_param}
\end{equation}
where $y_i$ denotes the ground-truth label expressed in a one-hot encoding vector and $sim$ is a function gauging the closeness or similarity between two vectors. $sim$ can be Gaussian similarity based on Euclidean distance or structural similarity $s_{sc}$ presented in \cref{sec:cm_structure}.

\subsection{Explanation Construction Module} \label{sec:cm_explanation}

In practice, let $P(\hat{y}_{user} | \mathcal{E}, \mathcal{U})$ represent the probability that a user can guess the model prediction correctly given explanation $\mathcal{E}$ and uncertainty factors $\mathcal{U}$. These factors include emotions, experiences, personal traits, cognition, and many others that are beyond the scope of this research. This work only focuses on how to maximize the user understanding of model predictions via explanation modalities. Lai et al. \cite{lai2019human} found a direct correlation between the amount of context information given to users and the accuracy of their predictions.

One significant challenge of interpretable GNNs is their limited capability to explain their decision-making process to users directly. This work introduces an explanation construction module as an intermediary to maximize the benefits of interpretable components in the architecture. This module systematically arranges information and prepares clear explanations that are easily comprehensible to users. Specifically, it incorporates several explanation functions to provide insights into the model's predictions, presented below:

\begin{enumerate}[label=(\arabic*)]
	\item \textbf{Concept visualization} allows users to visually explore key substructures within input graphs. This function is built upon the concept discovery module.
	\item \textbf{Finding similar graphs/concepts:} Example-based explanations are employed using reference strategies, providing insights by comparing and contrasting instances.
	\item \textbf{Reference Concept Attribution:} Measurement attributions of decisive references identify influential concepts contributing to predictions. This function takes outputs of the concept-based prediction layer as its inputs.
	\item \textbf{Concept Structure Matching Visualization} aids the interpretation capability by visualizing the mapping assignment between two graphs. 
\end{enumerate}

%The proposed module stands out from existing methods by generating different explanations. It offers a comprehensive and multi-faceted understanding of predictions by integrating diverse information types into unique explanations. This approach caters to different users' preferences, whether they prefer concept visualization to gain insights into the graph structure or find example-based explanations more intuitive. Additionally, attribution measurements of reference concepts can provide users with quantitative information, enhancing their understanding of model decisions. These functions offer a holistic and customizable explanation experience, ensuring a wide range of users' preferences are satisfied and facilitating effective interpretation and trust in the GNN model.

The introduced explanation module distinguishes itself from current methodologies by producing varied explanations. It provides a detailed and multifaceted comprehension of predictions by merging various types of information into distinctive explanations. This strategy meets the diverse preferences of users, accommodating those who favor concept visualization for a deeper understanding of the graph structure as well as those who find explanations based on examples more intuitive. Furthermore, the attribution measurements of reference concepts furnish users with quantitative information, augmenting their grasp of model decisions. These features deliver a comprehensive and adaptable explanatory experience, satisfying a broad spectrum of user preferences and promoting effective interpretation and confidence in GNN models.

\subsection{Computational Complexity}
%\noindent\textbf{Training.}  The training costs involve resources for training the GNN encoder and concept discovery module. The optimization process of \cref{eq:max_graph_sim} significantly increases the training time per epoch. Compared to training a backbone GNN alone, training the model takes approximately twice as long per epoch. Considering interpretation benefits, the additional costs are acceptable. 

\noindent\textbf{Training.} The training expenses encompass resources needed for GNN encoding and concept discovery. The optimization of \cref{eq:max_graph_sim} notably extends the training duration per epoch. When contrasted with training a standalone GNN, the training time per epoch is roughly doubled. However, given the advantages in interpretability, these extra costs are acceptable.

%\noindent\textbf{Inference \& Explanation:} The costs incurred during inference consist of executing the pre-trained GNN encoder, the concept discovery module, and the interpretable predictor. As for explanations, their cost is negligible since they rely on the results of the lower components. The interpretable predictor's main cost arises from the reference lookup process. Assuming $\epsilon$ is Euclidean distance computational cost, the Euclidean-based reference selection's complexity is reduced from $O(\epsilon M)$ to $O(\epsilon K)$ via vector storage like \cite{johnson2019billion}, where $K$ is the number of references and $K \ll M$. Adding up the cost for \cref{eq:emd} computation, which is approximately $O(N^2)$ via \cite{cuturi2013sinkhorn}, the two-stage reference selection's complexity is nearly $O(K(\epsilon + N^2))$.

\noindent\textbf{Inference \& Explanation:} The expenses associated with inference include running a pre-trained encoder, concept discovery, and a predictor. The cost of generating explanations is minimal as they only reuse outcomes from these underlying components. The primary expense for the interpretable predictor stems from reference strategies. With $\epsilon$ representing the computational cost of calculating a Euclidean distance, the complexity of a Euclidean-based reference strategy is reduced from $O(\epsilon M)$ to $O(\epsilon K)$, thanks to vector storage methods like those described in \cite{johnson2019billion}, where $K$ represents the number of references and is significantly smaller than $M$. Considering the computational cost for \cref{eq:emd}, which is roughly $O(N^2)$ based on the Sinkhorn approximation algorithm \cite{cuturi2013sinkhorn}, the overall complexity of the two-stage reference selection approximates $O(K(\epsilon + N^2))$.

\section{Experimental Setups} \label{con_exp_setups}
\subsection{Research Questions}
Extensive experiments were conducted on graph classification datasets at various scales to answer the following research questions.
\begin{enumerate}[label=\textbf{RQ\arabic*:},leftmargin=1.3cm]
	\item Is the proposed framework superior to baselines in predictive performance?
	\item How to visualize explanations generated by the proposed approach?
	\item What is the performance of node-based versus edge-based concept discovery approaches?
	\item Why does the proposed method provide more accurate predictions than GIB, a similar model?
	\item How do reference selection strategies affect the concept-based prediction function?
	\item How do explanations help users understand predictions?
\end{enumerate}

\subsection{Baselines}
Four well-known GNN architectures GCN \cite{kipf2016semi}, GraphSage \cite{hamilton2017inductive}, GIN \cite{xu2018powerful}, and GAT \cite{velickovic2017graph} were selected as baselines.

\begin{itemize}
	\item \textbf{GCN \cite{kipf2016semi}} was the very first GNN model, which leverages spectral graph convolutions to propagate information between nodes.
	\item \textbf{GraphSage \cite{hamilton2017inductive}} was an inductive learning framework for scalable graph representation learning, which leverages graph convolutions and neighborhood sampling to generate node embeddings, enabling effective generalization to unseen nodes.
	\item \textbf{GIN \cite{xu2018powerful}} utilized multiple graph convolution layers with learnable aggregation functions to generate node embeddings that are permutation invariant. It generalized the WL test to achieve maximum discriminative power.
	\item \textbf{GAT \cite{velickovic2017graph}} incorporates attention mechanisms to capture important information from neighboring nodes during information aggregation. By dynamically weighing the importance of neighboring nodes, it allows effective and adaptive learning of node representations.
\end{itemize}

Each GNN backbone model included two GNN layers, a hidden layer, and a prediction layer. The second group of baseline models was designed based on GIB \cite{yu2020graph}, denoted with the GIB prefix. Similarly, the proposed concept discovery method was applied to the four backbone GNNs, making another group of models labeled with a \textit{CONG} prefix. The final two model groups were created by combining the concept-based prediction function with trained concept embeddings, denoted as $\text{CONG}^+$ and $\text{CONG}^\dagger$ corresponding to Euclidean-based and EMD-based similarity metrics.

\subsection{Datasets}
\begin{table}[ht]
	\centering
	\caption{Dataset Statistical Information}
	\begin{tabular}{c|c|c|c|c|c}
		\toprule
		\thead{\textbf{Dataset} \\ \textbf{Name} } & \thead{\textbf{\#Num.} \\ \textbf{Graphs}} & \thead{\textbf{\#Avg} \\ \textbf{Nodes}} & \thead{\textbf{\#Avg} \\ \textbf{Edges}} & \thead{\textbf{\#Num.} \\ \textbf{Features}} & \thead{\textbf{\#Num.} \\ \textbf{Classes}}  \\
		\midrule
		Mutag & 188 & 17.93 & 19.79 & 7 & 2\\
		Proteins & 1113 & 39.06 & 72.82 & 29 & 2\\
		IMDB-Binary & 1000 & 19.77 & 96.53 & 271 & 2 \\
		DD & 1178 & 284.32 & 715.66 & 89 & 2 \\
		Twitter & 6940 & 21.10 & 20.10 & 768 & 3 \\
		\bottomrule
	\end{tabular}
	\label{tab:data_stat}
\end{table}

This work selected five famous graph classification datasets: Mutag \cite{rupp2012fast}, Proteins \cite{borgwardt2005protein}, IMDB-Binary (IMDB) \cite{rossi2015network}, DD\cite{rossi2015network}, and Graph-Twitter (Twitter) \cite{yuan2022explainability}. Data statistic information is presented in \cref{tab:data_stat}.

\subsection{Implementations and Configurations}
%All datasets were obtained from \cite{morris2020tudataset} except for the Twitter dataset. The training datasets were prepared based on 10-fold cross-validation and an 8:1:1 (train/validation/test) splitting strategy. Node features in IMDB and DD datasets were one-hot vectors corresponding to node degrees. The Proteins dataset's node features were normalized using standard normalization. 

The Twitter dataset was loaded via \cite{dig}, while other datasets were downloaded from \cite{morris2020tudataset}. Training data were arranged using a 10-fold cross-validation approach and divided using an 8:1:1 ratio for training, validation, and testing. For the IMDB and DD datasets, node features comprised one-hot vectors linked to node degrees, while node features for the Proteins dataset were subjected to standard normalization.

% model training
%All models were trained with Torch v.2.0.1 and DGL v1.1.0. Hyper-parameters were selected based on \cite{yu2020graph}. For instance, models were trained with 100 epochs with a learning rate of 0.01 and decayed with a factor of 0.5 after 50 epochs. The Adam optimizer was employed with a weight decay of 0.001 for the L2 penalty. The optimization process for \cref{eq:max_graph_sim} involved 20 inner loops. Regularization terms ($\alpha$, $\beta$, and $\lambda$) were set to 0.1, 1, and 0.1, respectively. The number of hidden units in all layers was 32, except for models trained with the Twitter dataset, which used 128 units. GAT models utilized 8 attention heads and ReLU activation. GraphSage models used Mean aggregators for all datasets except for the Twitter dataset, which employed GCN aggregators. Hyper-parameters for training the neural graph matching followed \cite{roy2022interpretable}.

This research implemented models based on PyTorch v.2.0.1 and DGL v1.1.0. The selection of hyper-parameters followed guidance from \cite{yu2020graph}. Models were typically trained over 100 epochs with an initial learning rate of 0.01, which was halved after 50 epochs. The training utilized the Adam optimizer, incorporating a 0.001 weight decay for the L2 penalty. The optimization of \cref{eq:max_graph_sim} included 20 inner loops. Regularization coefficients ($\alpha$, $\beta$, and $\lambda$) were established at 0.1, 1, and 0.1, correspondingly. Hidden layers in most models featured 32 units, except ones for the Twitter dataset, which contained 128 units. GAT models employed 8 attention heads along with a ReLU function. For all datasets except Twitter, GraphSage models utilized Mean aggregators, whereas ones for Twitter used GCN aggregators. 

% corpus implementation
%Experiments were conducted on a machine with one NVIDIA Tesla V100 16GB GPU. Indexing functions for corpus management were implemented with Faiss v.1.7.4. The number of centroids $K_c$ in indexing functions depended on the number of training graphs and did not affect the prediction accuracy of $\text{CONG}^+$. For instance, this number was set to 3 in the Mutag dataset and 5 for the others. For the KNN and KNN\_Class strategies, the number of neighbors was defined as 10 and 3, respectively. 

Experiments were conducted on a machine with one NVIDIA Tesla V100 16GB GPU. Indexing functions for corpus management were implemented with Faiss v.1.7.4. The selection of $K_c$ in these functions was based on the number of training graphs and did not influence the predictive performance $\text{CONG}^+$. For example, this research set $K_c = 3$ for the Mutag dataset and $K_c = 5$ for other datasets. In retrieval strategies KNN and KNN\_Class, retrieval sizes were set at 10 and 3, respectively. It is worth noting that the retrieval size in KNN\_Class corresponds to the number of references from each class.

\section{Experimental Results} \label{sec:cm_ exp_results}

\subsection{Accuracy Comparison Among Methods}
%The first experiment's goal was to compare the proposed architecture to baselines in graph classification. The results in \cref{tab:cm_acc_comp} demonstrate CONG's effectiveness in improving classification accuracy by reducing structure redundancies. Notably, \textit{CONG} outperforms GIB, a similar approach. Furthermore, $\text{CONG}^+$ is superior to other methods in all datasets, affirming the correctness of the concept discovery process and the concept-based non-parametric predictor. 

The initial experiment aimed to evaluate the proposed architecture against baseline methodologies in graph classification tasks. As shown in \cref{tab:cm_acc_comp}, the results underscore the efficacy of the proposed framework in enhancing prediction accuracy by minimizing structural redundancies. Significantly, \textit{CONG} surpasses GIB, which employs a comparable training strategy. Moreover, $\text{CONG}^+$ exceeds the performance of baselines across all datasets, affirming the validity of concept discovery and interpretable prediction procedures.

\begin{sidewaystable}
	\begin{minipage}[c][\textheight]{\linewidth}
	\caption{A Comparison of Models on Prediction Accuracy. Notably, all variants of CONG outperform baselines in all settings. }
	\centering
		\begin{tabular}{c|c|c|c|c|c}
			\toprule
			\textbf{Method}                                & \textbf{Mutag} & \textbf{Proteins} & \textbf{IMDB} & \textbf{DD} & \textbf{Twitter} \\
			\midrule
			GCN & 0.718 $\pm$ 0.094 & 0.714 $\pm$ 0.051 & 0.710 $\pm$ 0.049 & 0.715 $\pm$ 0.040& 0.642 $\pm$ 0.017 \\
			GraphSage & 0.730 $\pm$ 0.096 & 0.694 $\pm$ 0.049 & 0.715 $\pm$ 0.051 & 0.743 $\pm$ 0.038 & 0.636 $\pm$ 0.021 \\
			GIN & 0.862 $\pm$ 0.096 & 0.750 $\pm$ 0.052 & 0.726 $\pm$ 0.029 & 0.699 $\pm$ 0.035 & 0.651 $\pm$ 0.013 \\
			GAT & 0.750 $\pm$ 0.112 & 0.672 $\pm$ 0.120 & 0.726 $\pm$ 0.034 & 0.699 $\pm$ 0.035 & 0.664 $\pm$ 0.019 \\
			\hline
			GIB + GCN & 0.772 $\pm$ 0.089 & 0.731 $\pm$ 0.044 & 0.726 $\pm$ 0.054 & 0.765 $\pm$ 0.026 & 0.513 $\pm$ 0.043 \\
			GIB + GraphSage & 0.750 $\pm$ 0.089 & 0.699 $\pm$ 0.046 & 0.720 $\pm$ 0.054 & 0.772 $\pm$ 0.037 & 0.546 $\pm$ 0.081 \\
			GIB + GIN & 0.841 $\pm$ 0.100 & 0.721 $\pm$ 0.023 & 0.702 $\pm$ 0.057 & 0.729 $\pm$ 0.036 & 0.630 $\pm$ 0.023 \\
			GIB + GAT & 0.771 $\pm$ 0.104 & 0.684 $\pm$ 0.050& 0.717 $\pm$ 0.045 & 0.698 $\pm$ 0.100 & 0.505 $\pm$ 0.010 \\
			\hline
			CONG + GCN & 0.772 $\pm$ 0.124 & 0.718 $\pm$ 0.040 & 0.724 $\pm$ 0.032 & 0.752 $\pm$ 0.055 & 0.643 $\pm$ 0.016 \\
			CONG + GraphSage & 0.756 $\pm$ 0.108 & 0.726 $\pm$ 0.048 & 0.724 $\pm$ 0.047 & 0.763 $\pm$ 0.032 & 0.667 $\pm$ 0.011\\
			CONG + GIN & \underline{0.878 $\pm$ 0.068} & \underline{0.756 $\pm$ 0.047} & 0.732 $\pm$ 0.043 & 0.718 $\pm$ 0.038 & 0.636 $\pm$ 0.022 \\
			CONG + GAT & 0.787 $\pm$ 0.079 & 0.731 $\pm$ 0.033 & 0.732 $\pm$ 0.042 & 0.751 $\pm$ 0.064 & \underline{0.684 $\pm$ 0.021} \\
			\hline
			$\text{CONG}^+$ + GCN & 0.792 $\pm$ 0.085 & 0.722 $\pm$ 0.027 & 0.713 $\pm$ 0.039 & \underline{0.785 $\pm$ 0.025}  & 0.650 $\pm$ 0.017 \\
			$\text{CONG}^+$ + GraphSage & 0.808 $\pm$ 0.095 & 0.713 $\pm$ 0.045 & 0.721 $\pm$ 0.031 & \textbf{0.789} $\mathbf{\pm}$ \textbf{0.041} & 0.672 $\pm$ 0.013  \\
			$\text{CONG}^+$ + GIN & \textbf{0.888} $\mathbf{\pm}$ \textbf{0.062} & \textbf{0.767} $\mathbf{\pm}$ \textbf{0.033} & \textbf{0.748} $\mathbf{\pm}$ \textbf{0.048} & 0.740 $\pm$ 0.078 & 0.633 $\pm$ 0.018 \\
			$\text{CONG}^+$ + GAT & 0.798 $\pm$ 0.085 & 0.713 $\pm$ 0.035 & \underline{0.746 $\pm$ 0.040} & 0.777 $\pm$ 0.028  & \textbf{0.687} $\mathbf{\pm}$ \textbf{0.017} \\
			\hline
			$\text{CONG}^\dagger$ + GCN & 0.846 $\pm$ 0.079 & 0.706 $\pm$ 0.051 & 0.702 $\pm$ 0.032 & \textbf{0.785} $\pm$ \textbf{0.032} & \underline{0.678 $\pm$ 0.030} \\
			$\text{CONG}^\dagger$ + GraphSage & \underline{0.862 $\pm$ 0.071} & 0.739 $\pm$ 0.039 & \underline{0.728 $\pm$ 0.046} & \underline{0.778 $\pm$ 0.022} & 0.658 $\pm$ 0.011 \\
			$\text{CONG}^\dagger$ + GIN & \textbf{0.877} $\pm$ \textbf{0.068} & \textbf{0.757} $\pm$ \textbf{0.039} & 0.724 $\pm$ 0.038 & 0.750 $\pm$ 0.036 & 0.631 $\pm$ 0.028 \\
			$\text{CONG}^\dagger$ + GAT & 0.798 $\pm$ 0.090 & 0.738 $\pm$ 0.066 & \textbf{0.729} $\pm$ \textbf{0.032} & 0.769 $\pm$ 0.036 & \textbf{0.683} $\pm$ \textbf{0.015} \\
			\bottomrule
		\end{tabular}
	\label{tab:cm_acc_comp}
	\end{minipage}
\end{sidewaystable} 

\subsection{Interpretation Analysis}

\begin{figure}[ht]
	\centering
	\begin{tabular}{ccc}
		\thead{\textbf{Input Graph}} & \thead{\textbf{Extracted Concept}} &  \thead{\textbf{Matching Concepts}} \\
		
		\multicolumn{1}{m{4.2cm}}{\includegraphics[clip,trim=0.5cm 0.8cm 1cm 0.6cm,width=3.5cm]{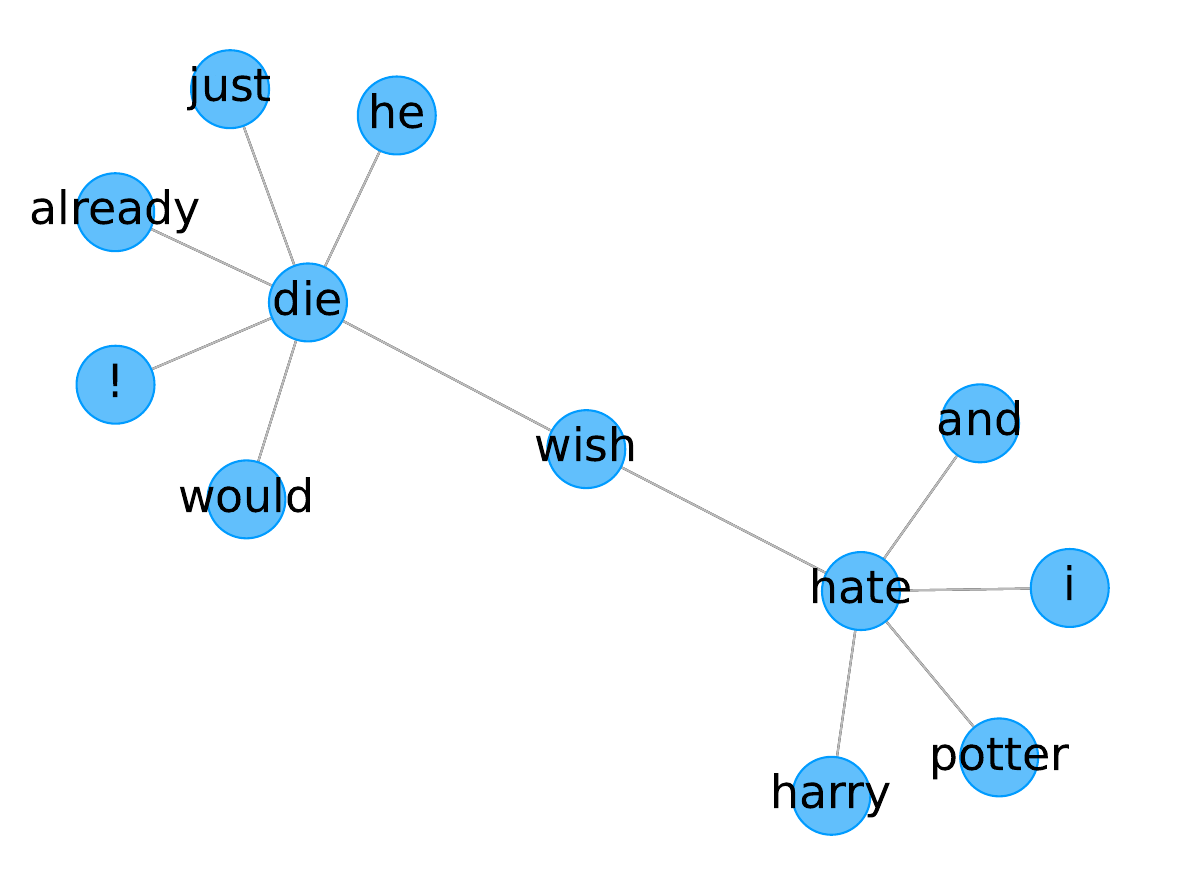} \newline 
			\makecell{\small I hate harry potter and wish \\ he would just die already !}}
		
		& \multicolumn{1}{m{4.2cm}}{\includegraphics[clip,trim=0.5cm 0.5cm 0.9cm 0.6cm,width=3.5cm]{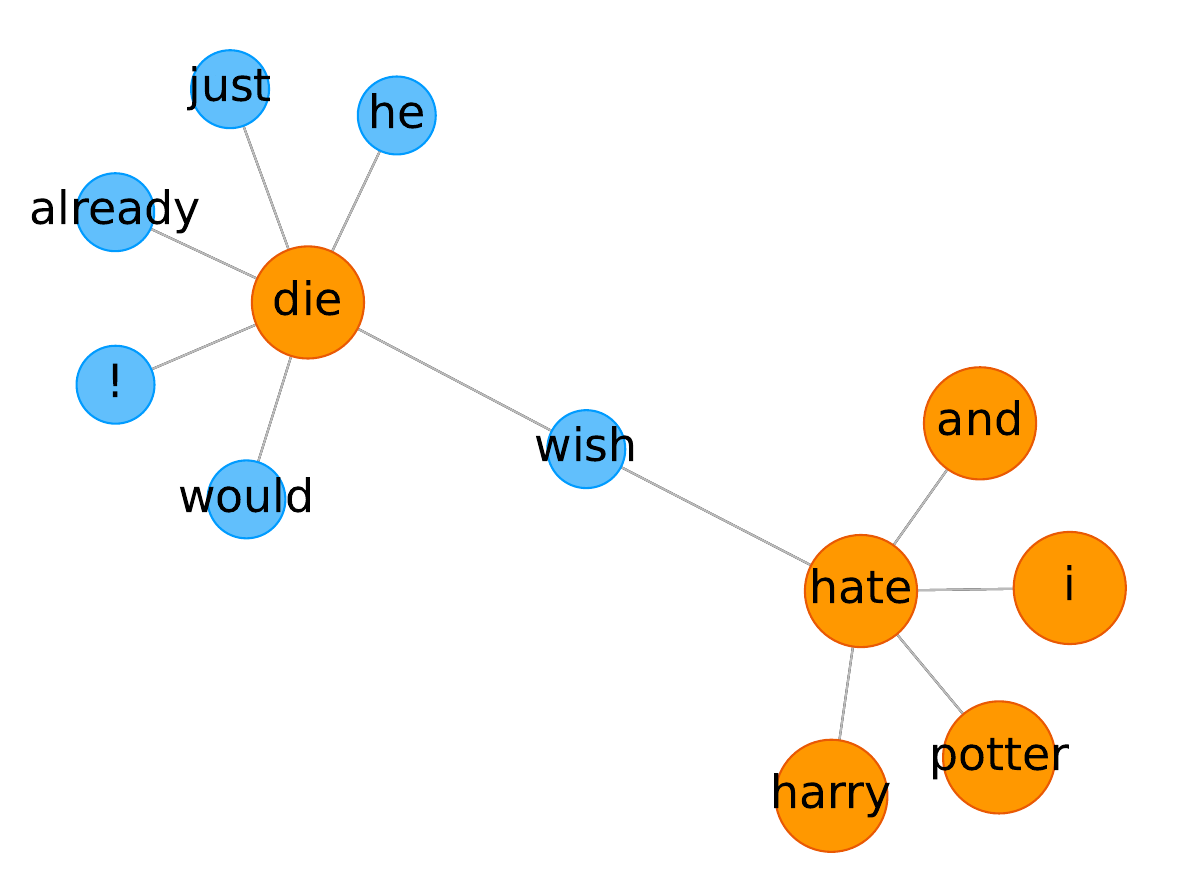} \newline  
			\makecell{\small I \underline{hate harry potter and} wish \\ he would just \underline{die} already \underline{!}}
		} 
		
		& \multicolumn{1}{m{3.7cm}}{
			\includegraphics[clip,trim=1cm 0.5cm 1cm 0.6cm,width=3.5cm]{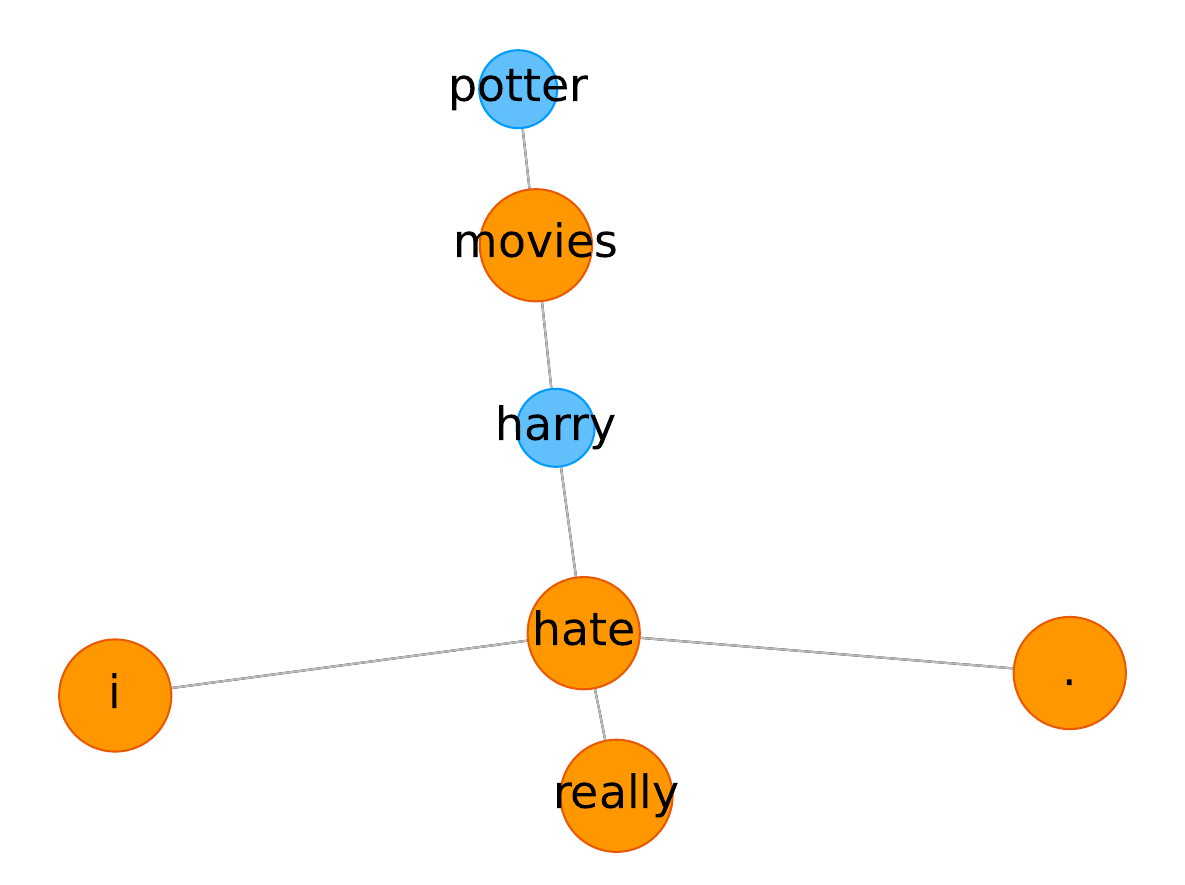} \newline
			\makecell{
				$\mathbf{a_e = 0.230}$ \\ $\mathbf{a_c = 0.468}$ 
			}
		} 
		
		\\ 
		
		\multicolumn{1}{m{4.2cm}}{\includegraphics[clip,trim=1cm 0.7cm 1cm 0.7cm,width=3.5cm]{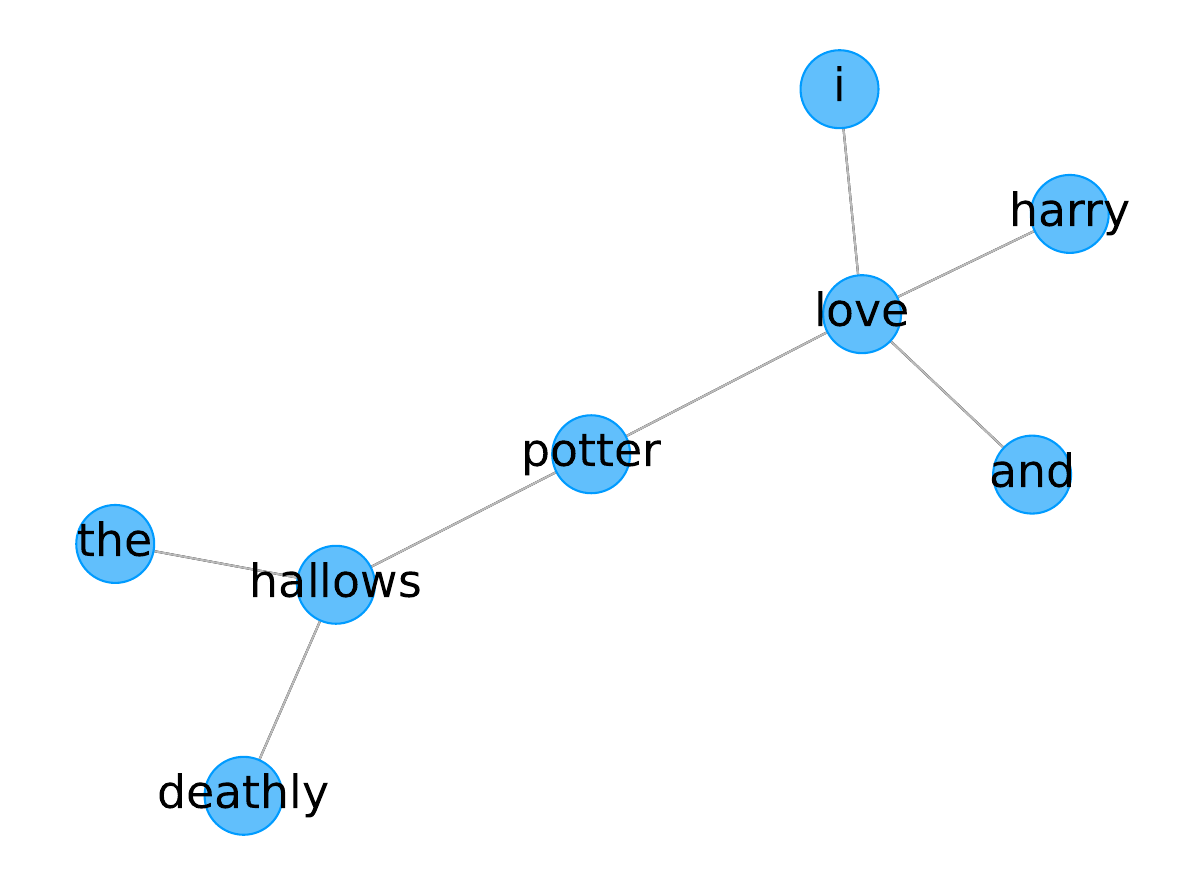} 
			\newline \newline 
			\makecell{\small I love harry potter \\ and the deathly hallows}}
		& 
		\multicolumn{1}{m{4.2cm}}{
			\includegraphics[clip,trim=1cm 0.5cm 1cm 0.6cm,width=3.5cm]{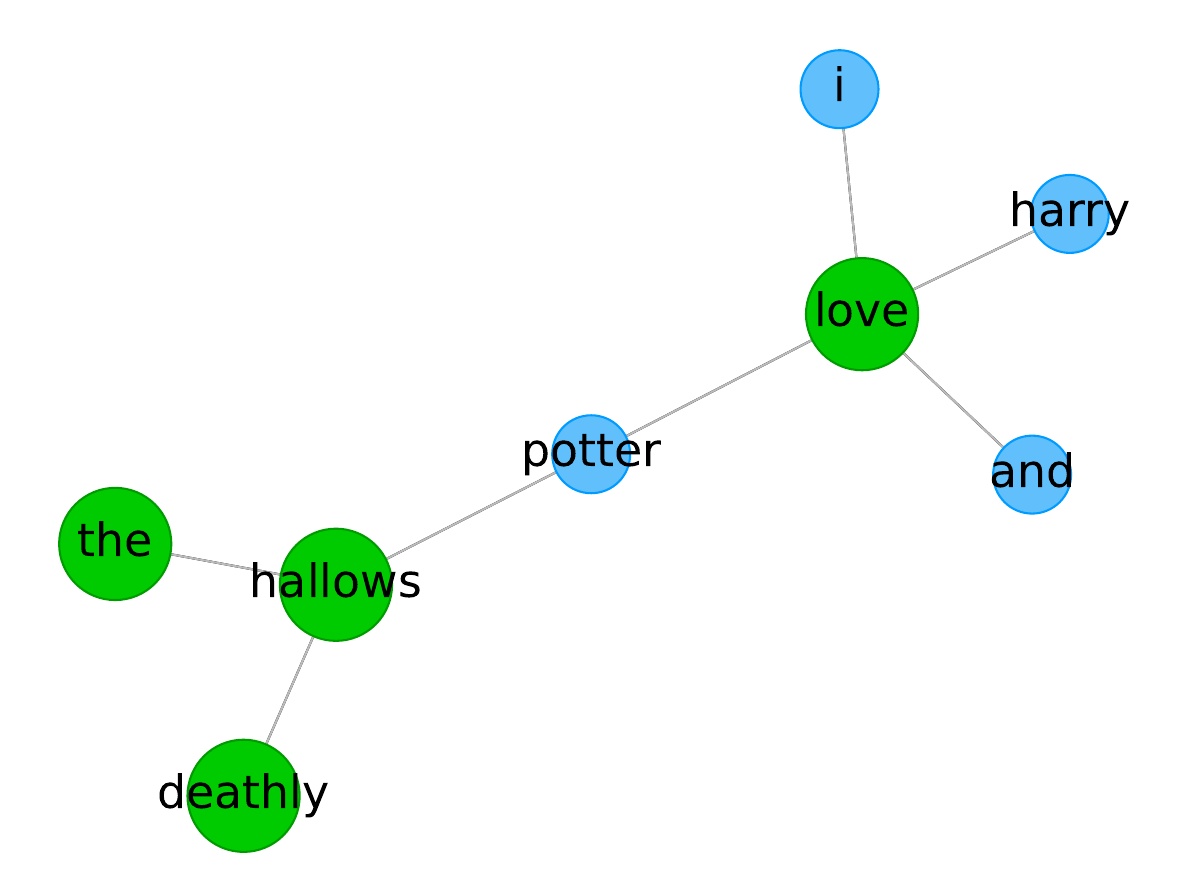}
			\newline 
			\makecell{\small I \underline{love} harry potter \\ and \underline{the deathly hallows}}
		} 
		& 
		\multicolumn{1}{m{3.7cm}}{
			\includegraphics[clip,trim=1cm 0.5cm 1cm 0.6cm,width=3.5cm]{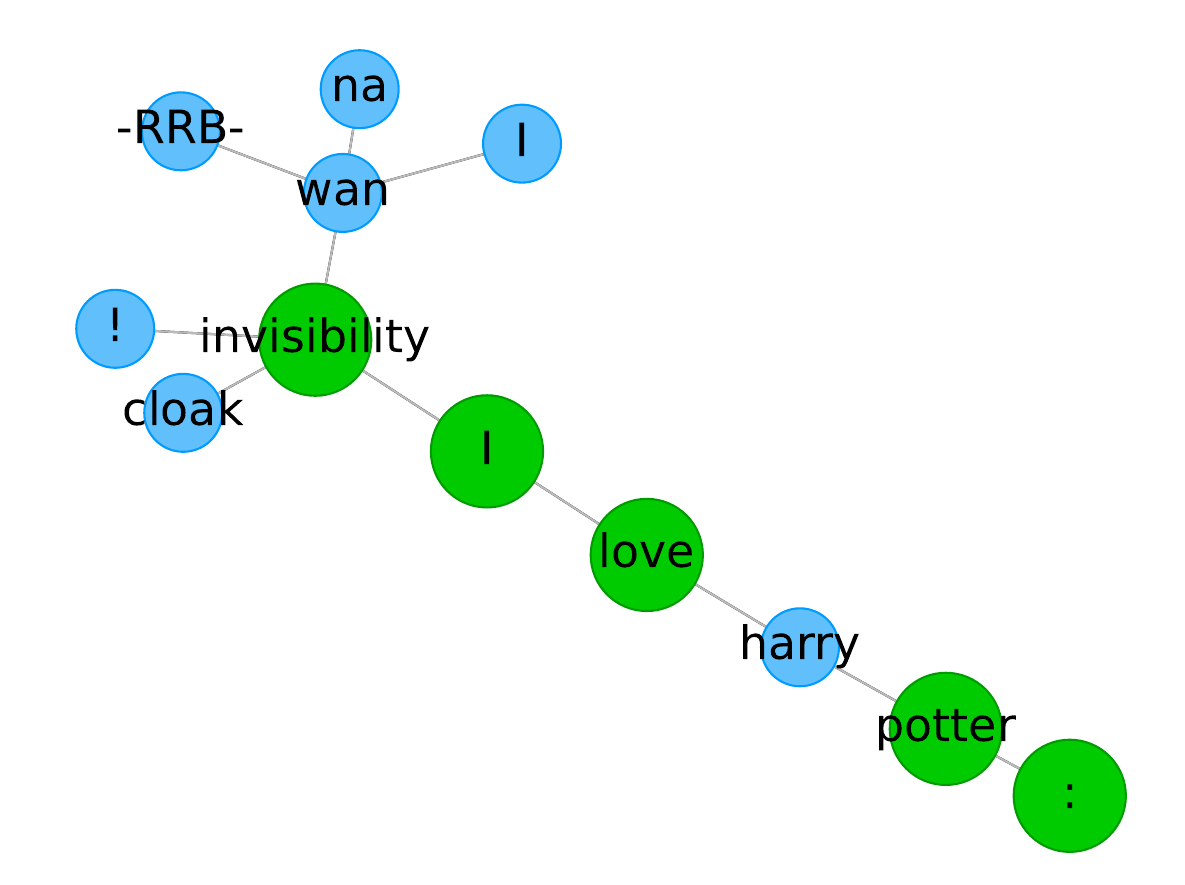}
			\newline
			\makecell{
				$\mathbf{a_e = 0.231}$ \\ 
				$\mathbf{a_c = 0.471}$  
			}
		}
		
	\end{tabular}
%	\caption{Visualizing Concept-based Prediction using KNN\_Class with $K = 3$. Here, $a_e$ denotes a reference's attributional score, while $a_c$ represents the total scores of all references in a class. The \textcolor{DarkGreen}{green} and \textcolor{DarkOrange}{orange} emphasize nodes selected for concepts corresponding to \textcolor{DarkGreen}{positive} and \textcolor{DarkOrange}{negative} classes, respectively. The underlined text highlights corresponding parts in sentences of colored nodes. Due to space constraints, only the highest-score references are displayed. }
	\caption{KNN\_Class Predictions with $K = 3$. In this visualization, $a_e$ signifies the attributional score of a reference, and $a_c$ indicates the cumulative scores for all references within a class. Nodes belonging to concepts linked to \textcolor{DarkGreen}{positive} and \textcolor{DarkOrange}{negative} classes are highlighted in \textcolor{DarkGreen}{green} and \textcolor{DarkOrange}{orange}, respectively. Corresponding portions of colored nodes are underlined in sentences. Only references with the highest scores are displayed due to space constraints.}
	\label{fig:cm_interpretation}
\end{figure}

%Interpretation analyses were conducted on the Twitter dataset, where graphs were randomly selected. Extracted concepts were visualized with highlighted nodes in different colors to signify class associations, as shown in \cref{fig:cm_interpretation}. Additionally, KNN\_Class and K\_centroids strategies were employed to identify and present relevant reference concepts closely aligned with the input concepts, showcasing their influence on the model's predictions. Furthermore, the contribution of these reference concepts was quantified through attributional measurements. Visualizing concepts, references, and attributions provided interpretable insights into the GNN model's reasoning process and effectively demonstrated interpretative capabilities.

Interpretation analyses were performed using Twitter graphs, which were selected at random. The concepts extracted from these graphs were depicted using nodes marked in various colors to indicate different class associations, as illustrated in \cref{fig:cm_interpretation}. Furthermore, strategies such as KNN\_Class and K\_centroids were utilized to retrieve and display reference concepts that closely corresponded with the input concepts, thereby highlighting their impact on the model’s predictive outcomes. The influence of these reference concepts was also quantified using attributional metrics. The depiction of these concepts, along with their references and attributions, provided clear insights into the reasoning process of the GNN model and effectively illustrated its explainability strengths.

\begin{figure}
	\centering
	\subfloat[Concept-based Correspondence]{
		\includegraphics[width=0.47\linewidth]{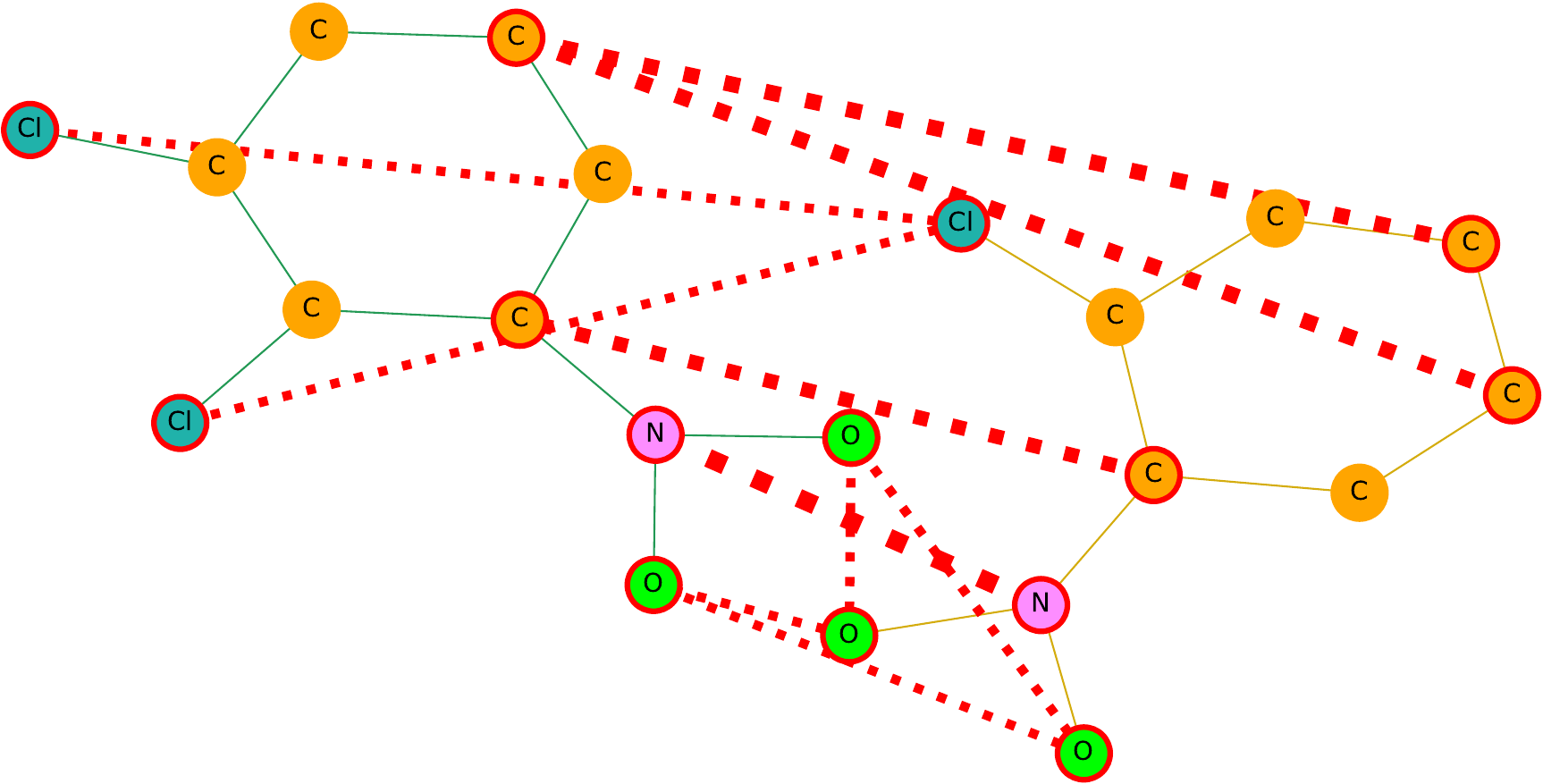}
		\label{fig:viz_con_uni_a}
	}
	\subfloat[Uniform-based Correspondence]{
		\includegraphics[width=0.47\linewidth]{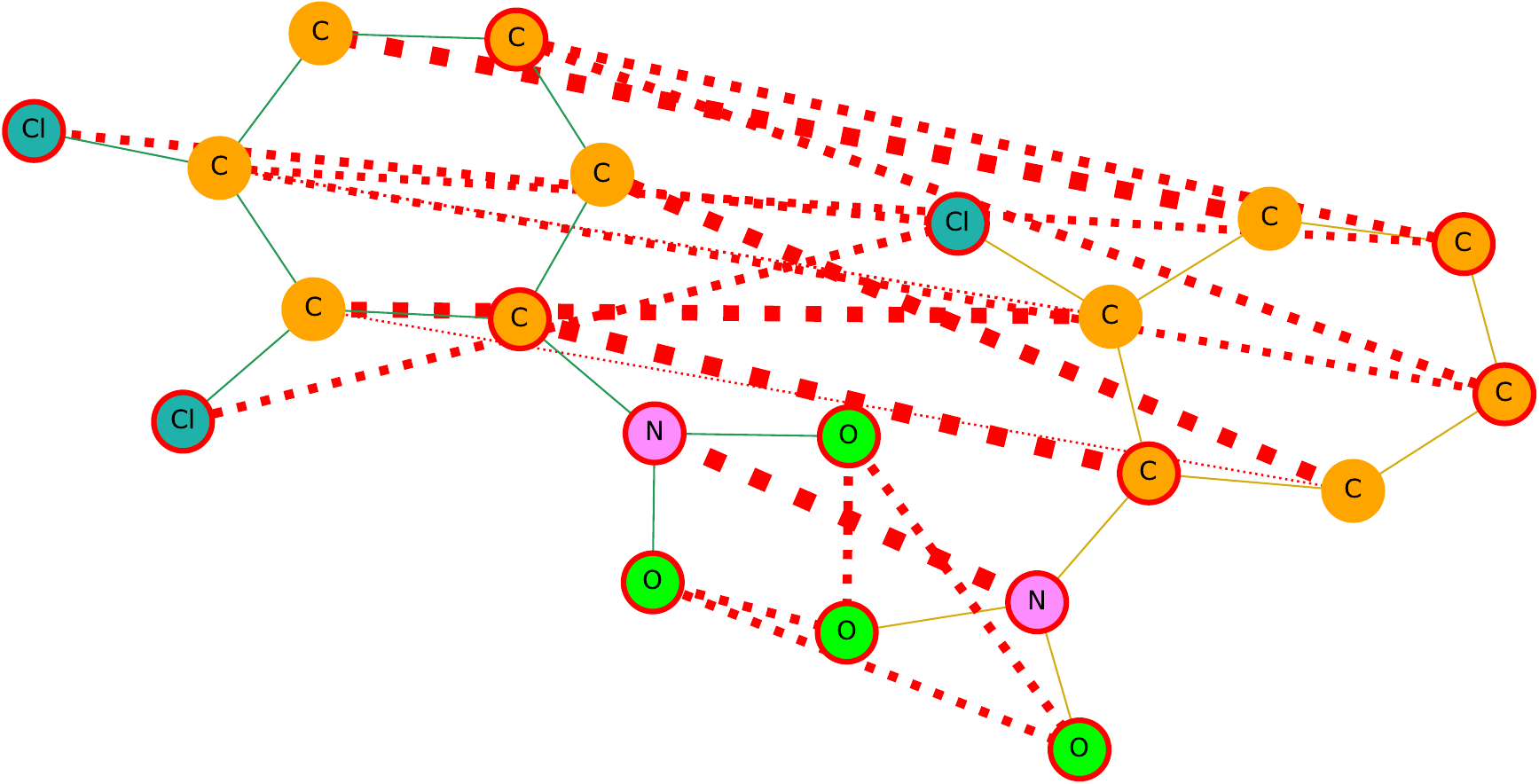}
		\label{fig:viz_con_uni_b}
	}
%	\caption{Visualization of Transport Matrices on Mutag Graphs. Nodes in frequent concepts are highlighted with red borders. Only transport edges (red ones) with $t \geq 0.1$ after min-max normalization are visualized. Edge widths correspond to the magnitude.}
	\caption{Transport Flow Visualizations Using Mutag Graphs. Visualizations utilize red borders to highlight vertices belonging to concepts and only display transport (red) edges with $t \geq 0.1$ following min-max normalization. Edge widths are proportional to the magnitude of similarity between two interconnected nodes.}
	\label{fig:viz_con_uni}
\end{figure}

%\noindent\textbf{Concept-based vs. Uniform-based:} The concept-based method concentrated on essential parts, providing clear visualization, while the uniform-based approach considered all nodes, resulting in a less distinct representation, as shown in \cref{fig:viz_con_uni}.

\noindent\textbf{Concept-based vs. Uniform-based:} The concept-based approach focused on only important elements, yielding a clear visualization, whereas the uniform-based method included all nodes, leading to a more complex representation, as illustrated in \cref{fig:viz_con_uni}.

\subsection{Node-based vs. Edge-based Concept Discovery}

\pgfplotsset{width=9cm,height=4.4cm}
\begin{figure}[ht]
	\centering
	\begin{tikzpicture}
		\begin{axis}[
			ybar,
			bar width=8pt,
			legend cell align=left,
			area legend,
			enlarge x limits={abs=1cm},
			ylabel={Accuracy Score},
			y label style={at={(axis description cs:-0.06,0.5)},anchor=south},
			symbolic x coords={Node-based Mutag, Edge-based Mutag, Node-based Proteins, Edge-based Proteins},
			x tick label style= {text width=1.4cm,align=center},
			legend style={draw=none, font=\scriptsize, at={(0.85,0.39)}, anchor=south,legend columns=1,/tikz/every even column/.append style={column sep=0.5cm}},
			xtick=data,
			]
			\addplot[fill=Blue!80] coordinates{(Node-based Mutag, 0.878) (Edge-based Mutag, 0.830) (Node-based Proteins, 0.756) (Edge-based Proteins, 0.718) };
			\addplot[fill=red!30] coordinates{(Node-based Mutag, 0.772) (Edge-based Mutag, 0.734) (Node-based Proteins, 0.718) 
				(Edge-based Proteins, 0.673)};
			\addplot[fill=yellow!30] coordinates{(Node-based Mutag, 0.756) (Edge-based Mutag, 0.730) (Node-based Proteins, 0.726) (Edge-based Proteins, 0.704)};
			\addplot[fill=green!30] coordinates{(Node-based Mutag, 0.787) (Edge-based Mutag, 0.745) (Node-based Proteins, 0.731) (Edge-based Proteins, 0.726) };
			\legend{GIN,GCN,GraphSage,GAT}
		\end{axis}
	\end{tikzpicture}
%	\caption{Accuracy Comparison between Node-based and Edge-based Concept Discovery. All models are trained with the $I(\hat{Y}, \mathcal{G})$ constraint.}
	\caption{An Assessment of Node-based and Edge-based Concept Discovery Approaches. All settings utilized $I(\hat{Y}, \mathcal{G})$ constrain in training.}
	\label{fig:cm_edge_node}
\end{figure}
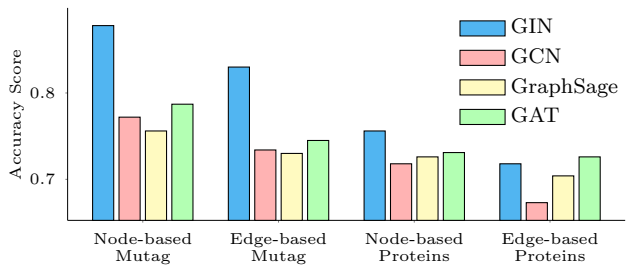

%This experiment evaluated the prediction accuracy gap between two concept discovery approaches. Here, Tanh was utilized as the non-linear function for edge assignment. As demonstrated in \cref{fig:cm_edge_node}, experimental results revealed a clear trend: the node-based approach consistently outperformed the edge-based approach regarding prediction accuracy across all models and datasets. This finding suggested that assigning selection probabilities to nodes yields superior results compared to assigning probabilities to edges. Therefore, based on experimental results, the node-based concept discovery approach was preferable for concept extraction and subsequent prediction tasks. However, it is possible to improve the performance of the edge-based approach in the future.

This study assessed the discrepancy in predictive performance between two methods of concept discovery. In this analysis, the Tanh function was employed for the assignment of edges. Illustrated in \cref{fig:cm_edge_node}, the findings indicated a distinct pattern: the approach focusing on node assignments markedly surpassed the edge-focused strategy in predictive performance across all examined settings. These results imply that allocating selection probabilities to nodes is more effective than to edges in graph classification tasks. Consequently, the node-centric emerged as the preferred choice for concept extraction and related prediction activities. Nonetheless, there remains potential to enhance the effectiveness of the edge-based method in future investigations.

\subsection{Effects of MI Constraint between Input Graph and Outcome on Prediction Accuracy}
\pgfplotsset{width=9cm,height=4.4cm}
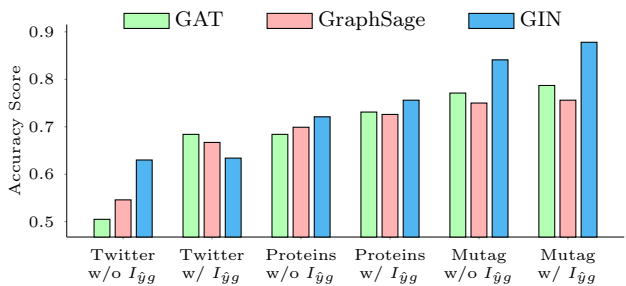
\begin{figure}[ht]
	\centering
	\begin{tikzpicture}
		\begin{axis}[
			ybar,
			bar width=6pt,
			legend cell align=left,
			area legend,
			enlarge x limits={abs=0.75cm},
			ylabel={Accuracy Score},
			y label style={at={(axis description cs:-0.06,0.5)},anchor=south},
			symbolic x coords={Twitter w/o $I_{\hat{y}g}$, Twitter w/ $I_{\hat{y}g}$, Proteins w/o $I_{\hat{y}g}$, Proteins w/ $I_{\hat{y}g}$, Mutag w/o $I_{\hat{y}g}$, Mutag w/ $I_{\hat{y}g}$},
			x tick label style= {text width=1.2cm,align=center},
			legend style={draw=none,font=\scriptsize, at={(0.5,0.92)}, anchor=south,legend columns=-1,/tikz/every even column/.append style={column sep=0.5cm}},
			xtick=data,
			]
			\addplot[fill=green!30] coordinates{(Twitter w/o $I_{\hat{y}g}$, 0.505) (Twitter w/ $I_{\hat{y}g}$, 0.684) (Proteins w/o $I_{\hat{y}g}$, 0.684) (Proteins w/ $I_{\hat{y}g}$, 0.731) (Mutag w/o $I_{\hat{y}g}$, 0.771) (Mutag w/ $I_{\hat{y}g}$, 0.787)};
			\addplot[fill=red!30] coordinates{(Twitter w/o $I_{\hat{y}g}$, 0.546) (Twitter w/ $I_{\hat{y}g}$, 0.667) (Proteins w/o $I_{\hat{y}g}$, 0.699) (Proteins w/ $I_{\hat{y}g}$, 0.726) (Mutag w/o $I_{\hat{y}g}$, 0.750) (Mutag w/ $I_{\hat{y}g}$, 0.756)};
			\addplot[fill=Blue!80] coordinates{(Twitter w/o $I_{\hat{y}g}$, 0.630) (Twitter w/ $I_{\hat{y}g}$, 0.634) (Proteins w/o $I_{\hat{y}g}$, 0.721) (Proteins w/ $I_{\hat{y}g}$, 0.756) (Mutag w/o $I_{\hat{y}g}$, 0.841) (Mutag w/ $I_{\hat{y}g}$, 0.878)};
			
			\legend{GAT, GraphSage, GIN}
		\end{axis}
	\end{tikzpicture}
%	\caption{Effects of $I(\hat{Y}, \mathcal{G})$ on Prediction Accuracy. All models are trained with the node-based concept discovery approach.}
	\caption{An Assessment of $I(\hat{Y}, \mathcal{G})$'s Impacts on Model Performance. Node-based concepts are used in all settings.}
	\label{fig:cm_igg}
\end{figure}

%This study evaluated the effectiveness of the MI constraint between the input graph and the prediction outcome on the prediction accuracy of \textit{CONG}. The node-based concept discovery method was utilized in this experiment. A hypothesis was that this constraint stabilizes model training by providing a shortcut feedback loop to the message-passing processes. Specifically, including this constraint establishes a valuable feedback mechanism, enabling the GNN encoder to understand better the relationship between the graph structure and the desired outcome. Furthermore, this constraint facilitates the graph-embedding learning process by not paying too much attention to only selected nodes. As \cref{fig:cm_igg} demonstrates, incorporating $I(\hat{Y}, \mathcal{G})$ constraint increases the prediction accuracy of graph classification in three datasets: Mutag, Proteins, and Twitter. 

This research assessed the impact of $I(\hat{Y}, \mathcal{G})$ constraint on the predictive performance. The node-based discovery approach was employed in this analysis. A hypothesis was that this constraint aids in stabilizing training processes by creating shortcut paths for feedback information to flow back to message-passing mechanisms. Specifically, the inclusion of this constraint sets up an effective feedback system, which helps the GNN encoder to more clearly comprehend the correlation between graph data and labels. Additionally, this constraint assists in the learning process of graph embedding by ensuring that not too much emphasis is placed on selected nodes only. As shown in \cref{fig:cm_igg}, integrating $I(\hat{Y}, \mathcal{G})$ in training enhances the predictive performance in graph classification across study datasets.

\subsection{Comparison of Different Reference Strategies}
\pgfplotsset{width=9cm,height=4.4cm}
\begin{figure}[ht]
	\centering
	\begin{tikzpicture}
		\begin{axis}[
			ybar,
			bar width=10pt,
			legend cell align=left,
			area legend,
			enlarge x limits={abs=1.5cm},
			ylabel={Accuracy Score},
			y label style={at={(axis description cs:-0.08,0.5)},anchor=south},
			symbolic x coords={KNN, K\_Centroids, KNN\_Class},
			x tick label style= {text width=1.2cm,align=center},
			legend style={draw=none,font=\scriptsize, at={(0.5,0.92)}, anchor=south,legend columns=-1,/tikz/every even column/.append style={column sep=0.5cm}},
			xtick=data,
			]
			\addplot[fill=green!30] coordinates{(KNN, 0.687) (K\_Centroids, 0.684) (KNN\_Class, 0.682)};
			\addplot[fill=red!30] coordinates{(KNN, 0.672) (K\_Centroids, 0.663) (KNN\_Class, 0.660)};
			\addplot[fill=Blue!80] coordinates{(KNN, 0.633) (K\_Centroids, 0.627) (KNN\_Class, 0.627)};
			
			\legend{GAT, GraphSage, GIN}
		\end{axis}
	\end{tikzpicture}
	% updated
	\caption{An Assessment of Three Reference Strategies on Prediction Accuracy}
	\label{fig:cm_ref_strat}
\end{figure}
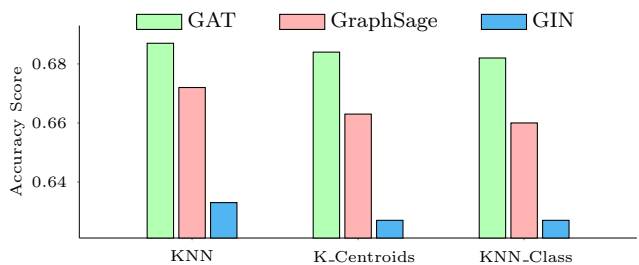

%This experiment investigated three reference strategies. As shown in \cref{fig:cm_ref_strat}, KNN showed the highest predictive accuracy, followed closely by K\_Centroids and KNN\_Class. Each strategy offered distinct benefits for users' understanding of predictions. KNN enabled local similarity examination by selecting the nearest reference concepts. KNN\_Class enhanced understanding by showcasing reference group attributional scores and comparative concepts. Lastly, K\_Centroids highlighted the contributions of concept groups to decisions while being computationally cost-effective, as it didn't require finding references every time. 

This study aimed to assess three reference strategies, as illustrated in \cref{fig:cm_ref_strat}. KNN demonstrated the greatest predictive precision, with K\_Centroids and KNN\_Class closely trailing. Each strategy provided unique advantages for users' comprehension of predictions. KNN allowed for the analysis of local similarities by identifying the closest reference concepts. KNN\_Class improved comprehension by displaying class-related contribution scores and comparative visualizations. K\_Centroids, meanwhile, emphasized the role of concept groups in decision-making and offered a computationally efficient option by eliminating the need to identify references repeatedly.

\subsection{Evaluating User Perception of Explanations}
This study evaluated the user perception of different types of explanations for model predictions. This study sought to answer the following question: How effective were explanations in enhancing user understanding of model predictions? The study was organized as a small competition wherein participants, whoever won, received a 20\$ voucher. The competition had 21 contestants with background knowledge in ML models. Each one predicted model outcomes of four sets of ten graphs given one of the following explanations.

\begin{enumerate}[label=(\arabic*)]
	\item Extracted concept visualization only
	\item Visualizing the extracted concept with \textbf{KNN-based} references and attributional scores
	\item Visualizing the extracted concept with \textbf{KNN\_Class-based} references with attributional scores 
	\item Visualizing the extracted concept with \textbf{K\_Centroids-based} references with attributional scores
\end{enumerate}
The first explanation is similar to the third column in \cref{fig:cm_interpretation} without highlighted colors. The last three types are equivalent to all columns of \cref{fig:cm_interpretation}. After contestants finished the test, they were asked to grade the usefulness of explanations and the confidence of their predictions on a 10-point scale. 

\pgfplotsset{width=9cm,height=4.4cm}
\begin{figure}[ht]
	\centering
	\begin{tikzpicture}
		\begin{axis}[
			axis y line*=left,
			ybar,
			bar width=0.3cm,
			area legend,
			enlarge x limits={abs=1cm},
			ylabel={Accuracy Score},
			y label style={at={(axis description cs:-0.08,0.5)},anchor=south},
			symbolic x coords={(1), (2), (3), (4)},
			x tick label style= {text width=1.2cm,align=center},
			legend style={draw=none, fill=none, font=\scriptsize, at={(0.16,0.95)}, anchor=south, legend columns=-1,/tikz/every even column/.append style={column sep=0.5cm}},
			xtick=data,
			]
			
			\addplot[fill=green!30,  bar shift=-0.4cm, error bars/.cd, y dir=both, y explicit] coordinates{
				((1), 0.643) +- (0,  0.0449)
				((2), 1.) +- (0, 0)
				((3), 1.) +- (0, 0)
				((4), 1.) +- (0, 0)
			};
			\addlegendentry{User Pred}
			
			% \legend{User Prediction, Usefulness, Confidence}
		\end{axis}
		
		\begin{axis}[
			ybar,
			axis y line*=right,
			bar width=0.3cm,
			area legend,
			enlarge x limits={abs=1cm},
			ylabel={10-point Score},
			y label style={at={(axis description cs:1.11,0.5)},anchor=south},
			legend style={draw=none,font=\scriptsize, at={(0.68,0.95)}, anchor=south,legend columns=-1,/tikz/every even column/.append style={column sep=0.5cm}},
			xticklabels={}
			]
			\addplot[fill=red!30, bar shift=0cm, error bars/.cd, y dir=both, y explicit] coordinates{
				((1), 5.095238095238095) +- (0, 0.346837564713363)
				((2), 8.333333333333334) +- (0, 0.20162304395046676)
				((3), 8.333333333333334) +- (0, 0.20162304395046676)
				((4), 7.619047619047619) +- (0, 0.20770349236529645)
			};
			
			\addlegendentry{Usefulness}
			\addplot[fill=Blue!80, bar shift=0.4cm, error bars/.cd, y dir=both, y explicit] coordinates{
				((1), 4.428571428571429) +- (0, 0.21166010488516723)
				((2), 8.523809523809524) +- (0, 0.21361093050129626)
				((3), 8.619047619047619) +- (0, 0.2077034923652964)
				((4), 7.476190476190476) +- (0, 0.21361093050129626)
			};
			\addlegendentry{Confidence}
			
		\end{axis}
		
	\end{tikzpicture}
%	\caption{User Perception of Various Explanation Types. User prediction is measured using an accuracy score, while usefulness and confidence are measured using a 10-point scale.}
	\caption{An Assessment of Users' Comprehension on Explanation Modalities with Visible Labels. Their comprehension is measured through the ability to predict the model outcomes given explanations. A 10-point rating system is used to determine the usefulness and confidence scores.}
	\label{fig:cm_user_study1}
\end{figure}
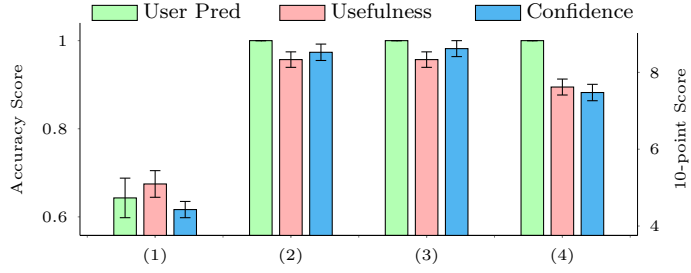

Several noteworthy observations emerged from the results presented in \cref{fig:cm_user_study1}. Firstly, it was evident that solely presenting extracted concept visualizations had a limited impact on users' comprehension and confidence in the model's predictions. Various reference strategies led to a significant improvement in users' understanding, resulting in a high level of consensus with the model's predictions. However, K\_Centroids-based explanations occasionally caused user confusion due to the equivalent of class attributional scores. These results effectively demonstrated the effectiveness and usefulness of incorporating multiple features within a single explanation, as it greatly enhanced users' understanding of the model's predictions.

\pgfplotsset{width=9cm,height=4.4cm}
\begin{figure}[ht]
	\centering
	\begin{tikzpicture}
		\begin{axis}[
			axis y line*=left,
			ybar,
			bar width=0.4cm,
			area legend,
			enlarge x limits={abs=1cm},
			ylabel={Accuracy (\%)},
			y label style={at={(axis description cs:-0.08,0.5)},anchor=south},
			symbolic x coords={(1), (2), (3), (4)},
			x tick label style= {text width=1.2cm,align=center},
			legend style={draw=none, fill=none, font=\scriptsize, at={(0.16,0.95)}, anchor=south, legend columns=-1,/tikz/every even column/.append style={column sep=0.5cm}},
			xtick=data,
			]
			
			\addplot[fill=green!30,  bar shift=-0.5cm, error bars/.cd, y dir=both, y explicit] coordinates{
				((1), 67.999) +- (0,  4.92) 
				((2), 68.5) +- (0,  3.98)
				((3), 82.0) +- (0, 4.92)
			};
			\addlegendentry{User Pred}
		\end{axis}
		
		\begin{axis}[
			ybar,
			axis y line*=right,
			bar width=0.4cm,
			area legend,
			enlarge x limits={abs=1cm},
			ylabel={10-point Score},
			y label style={at={(axis description cs:1.11,0.5)},anchor=south},
			legend style={draw=none,font=\scriptsize, at={(0.68,0.95)}, anchor=south,legend columns=-1,/tikz/every even column/.append style={column sep=0.5cm}},
			xticklabels={}
			]
			\addplot[fill=red!30, bar shift=0cm, error bars/.cd, y dir=both, y explicit] coordinates{
				((1), 5.995) +- (0, 0.35)
				((2), 6.250) +- (0, 0.46)
				((3), 8.450) +- (0, 0.22)
			};
			
			\addlegendentry{Usefulness}
			\addplot[fill=Blue!80, bar shift=0.5cm, error bars/.cd, y dir=both, y explicit] coordinates{
				((1), 5.829) +- (0, 0.21)
				((2), 5.980) +- (0, 0.32)
				((3), 8.5) +- (0, 0.22)
			};
			\addlegendentry{Confidence}
			
		\end{axis}
		
	\end{tikzpicture}
%	\caption{User Performance with Explanation Modalities while labels are hidden. Participants assessed prediction confidence and explanation usefulness on a 10-point scale.}
	\caption{An Assessment on Users' Comprehension on Explanation Modalities with Invisible Labels.}
	\label{fig:cm_user_study2}
\end{figure}
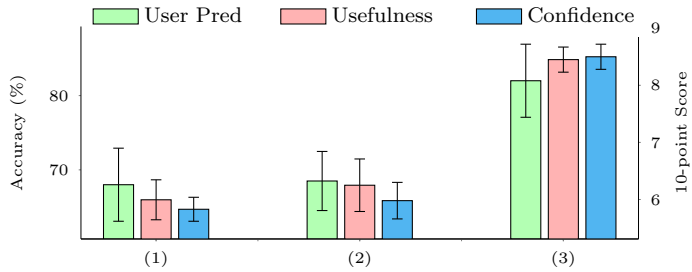

%A second user study, with 20 participants, was conducted to evaluate the hypothesized direct correlation between the amount of context provided to users and their prediction accuracy \cite{lai2019human}. This study followed a similar procedure to the first, except aggregated attributions $a_c$ were hidden. Participants guessed model predictions based on one of three explanation types: (1) PGExplainer subgraph visualization; (2) Concept-focused subgraph visualization; (3) (2) + presenting KNN-based references and their attributional scores $a_e$. As presented in \cref{fig:cm_user_study2}, subgraph visualization alone had minimal impact on users' prediction confidence, resulting in low scores across metrics for the first two explanation types. Combining essential subgraphs with relevant references significantly improved user comprehension and confidence. Therefore, this combination led to a notable increase in prediction accuracy and user ratings. Importantly, hiding aggregated attributions significantly decreased prediction accuracy compared to results in \cref{fig:cm_user_study1}. 

A second user study, with 20 participants, was conducted to evaluate the hypothesized direct correlation between the amount of context provided to users and their prediction accuracy \cite{lai2019human}. This study followed a similar procedure to the first, except aggregated attributions $a_c$ were hidden. Participants guessed model predictions given one of the following modalities: (1) PGExplainer subgraph visualization; (2) Visualizing extracted concepts; (3) Concept-focused visualization coupled with KNN-based references and corresponding attribution scores $a_e$. As depicted in \cref{fig:cm_user_study2}, solely visualizing subgraphs was minimally effective in enhancing users' confidence in their predictions, yielding low predictive performance for the first two explanation modalities. However, showing key subgraphs with pertinent references markedly boosted user understanding and confidence. Consequently, this integrated approach resulted in a significant improvement in both prediction accuracy and user assessments. Notably, the omission of aggregated attributions led to a marked reduction in prediction accuracy when compared to the results shown in \cref{fig:cm_user_study1}.

These results demonstrate a direct correlation between the amount of contextual information provided and user perception. Visualizing essential subgraphs alone was insufficient to improve user understanding; insights into how these subgraphs relate to key concepts proved crucial.  Presenting similar concepts likely enabled comparative analysis, helping users grasp a graph's classification by relating it to familiar examples. Additionally, attributional scores, especially aggregated values, offered quantitative measures that further clarified and enhanced the explanations' interpretability. These findings highlight a promising avenue for research into how GNN explanations influence human decision-making.

\subsection{Shortcomings of User Study}
% no need param
Organizing user studies through competitions has been a practical approach in our research to gather valuable insights into how users predict model outcomes based on explanations and reference visualizations. However, we acknowledge certain limitations associated with this method. First, the competitive nature of the study may introduce biases, as participants might be driven by the desire to outperform their peers rather than providing authentic predictions. This behavior could potentially impact the accuracy of user responses. Second, the participants in such competitions may not represent a diverse cross-section of the intended user base, potentially limiting the generalizability of our findings. Moreover, the competition setup might not fully capture real-world scenarios where users engage with AI systems without a competitive backdrop. Despite these limitations, organizing competitions remains a valuable approach to our framework evaluation, and we continue to work toward refining our methodology to address these challenges.

\section{Conclusion} \label{conclusion}
%This chapter addressed the challenge of designing an interpretable GNN architecture by proposing a novel concept-matching model. Existing methods had limitations that affected the quality and user-centric nature of GNN explanations. The proposed model overcame these challenges by leveraging the graph information bottleneck theory with modified constraints to extract concepts from input graphs. These concepts were efficiently managed in a concept corpus, enabling quick inference lookups and meaningful explanation generation. This chapter proposed various explanation modalities based on the concept corpus and discovery module to fulfill diverse user preferences. Extensive experiments and a user study were conducted to assess the proposed approach's performance. The obtained results provided compelling evidence for the efficacy of the proposed model in enhancing GNN interpretability and prediction accuracy, thus paving the way for further advancements in XAI methods for GNN architectures. 

This research tackled the challenge of developing an interpretable GNN framework by introducing an innovative concept-matching model. Prior methodologies faced limitations that impacted the quality and user-centric aspects of GNN explanations. The proposed framework addressed these issues by applying the graph information bottleneck theory with adjusted constraints to derive concepts from input graphs. These concepts were systematically organized in a concept repository, facilitating rapid inference lookups and the generation of meaningful explanations. This research also introduced various explanation modalities grounded in the concept repository and discovery module to cater to varied user needs. Thorough experiments and a user study were conducted to evaluate the effectiveness of the proposed approach. The results from these assessments provided strong support for the model’s capability to enhance GNN interpretability and predictive precision, thereby setting the stage for future enhancements in XAI methodologies for GNN models.

This chapter's insights and findings pave the way for several promising extensions. Integrating human constraints into concept discovery would ensure greater alignment with domain knowledge. Structuring the corpus hierarchically would streamline concept exploration, allowing users to navigate different levels of abstraction more efficiently. An interactive, user-friendly interface would further enhance user-centric explanations. These advancements would collectively improve the system's overall interpretability and usability. The following chapter will build upon these considerations, extending the capabilities of the proposed framework to incorporate these enhancements.

	\chapter{Trustworthy Graph Classification via Active Human Verification} \label{chap:human}

\section{Introduction}
%Conventional graph classification models primarily learn the mapping $\mathcal{G} \rightarrow \mathcal{Y}$, where $\mathcal{G}$ represents an input graph, and $\mathcal{Y}$ is the ground truth. While significant efforts \cite{yuan2022explainability, ragno2022prototype, zhang2022protgnn} have been made in improving the interpretability of these models, they often prioritize algorithmic evaluations and model performance metrics, sometimes neglecting the critical role of human collaboration. Valuable insights offered by domain experts can be overlooked, leading to discrepancies between human and model decision-making. Therefore, integrating domain knowledge into these models is not only beneficial but also essential, as it has the potential to enhance both model performance and interpretability, ultimately bridging the gap between human understanding and machine learning outcomes.

Graph classification methods like GNNs focus on establishing a relationship $\mathcal{G} \rightarrow \mathcal{Y}$, wherein $\mathcal{G}$ and $\mathcal{Y}$ denote an input graph and the outcome, respectively. Although substantial progress \cite{yuan2022explainability, ragno2022prototype, zhang2022protgnn} has been achieved in enhancing the interpretability of these methods, the emphasis frequently remains on algorithmic assessments like model accuracy, often overlooking the vital importance of human collaboration. Knowledge and feedback provided by experts in a particular field might be neglected, resulting in a misalignment between human and AI decision-making processes. Thus, the incorporation of domain expertise into graph classification models is not merely advantageous but imperative. It can significantly improve both the interpretability and the performance of the models, effectively narrowing the gap between human comprehension and the decisions of AI models.

%The collaborative approach between humans and AI \cite{mosqueira2023human, ramos2020interactive} has garnered significant attention within the research community. This approach holds promise in enhancing model performance and reliability by harnessing the complementary strengths of both entities. However, integrating human knowledge and feedback into ML models poses significant challenges. While RLHF \cite{ouyang2022training} has shown promise in learning from human feedback, it faces difficulties with inconsistent reward signals due to the inherent subjectivity of human preferences. Contrastive learning, on the other hand, simplifies the feedback process through pairwise comparisons, making it a more favorable approach in certain scenarios, especially when feedback is limited. Notably, Liu et al. \cite{liu2022learning} introduced a case-based reasoning framework that effectively combines the advantages of deep representation learning and case-based decision support, incorporating human feedback through contrastive learning. Similarly, Hejna et al. \cite{hejna2023contrastive} proposed contrastive preference learning as an alternative approach for learning from human feedback. These contributions collectively underscore the potential of contrastive learning as an effective method for encoding human feedback into models. 

The interaction between AI and humans \cite{mosqueira2023human, ramos2020interactive} has captured considerable interest within research communities. This collaboration is believed to improve the predictive performance and reliability of models by leveraging the unique strengths of both sides. Nonetheless, encoding knowledge and feedback into AI models presents substantial obstacles due to the discrete and non-differentiable nature of input information. Lately, reinforcement learning has demonstrated potential in adapting to human feedback, resulting in a famous method named RLHF \cite{ouyang2022training}. However, this approach struggles with the variability of reward signals, which arise from the subjective nature of human preferences. In contrast, contrastive learning offers a more streamlined method for integrating feedback via pairwise comparisons, presenting a preferred option in scenarios where the number of feedback samples is small. Additionally, Liu et al. \cite{liu2022learning} developed a framework that empowers case-based decision support with deep representation learning, utilizing contrastive learning to integrate human feedback. Sharing the same approach, Hejna et al. \cite{hejna2023contrastive} introduced a novel method to incorporate feedback into learning processes through contrastive preference learning. These advancements highlight the efficacy of contrastive learning in encoding knowledge and feedback into AI models.

%The motivation for this research originates from the inherent transparency of case-based reasoning processes \cite{slade1991case}, coupled with the enormous challenges domain experts face when reviewing extensive training samples and offering guidance for the learning process. Practically, they can focus on a select set of prominent examples that effectively represent entire groups of training samples. This intuitive strategy aligns with the idea that training samples should be situated close to at least one of these representative samples in representation learning. Moreover, experts can offer valuable insights, specifying which samples should be close to one another and far from certain others, thus enriching the learning process. 

This work's motivation stems from the transparent and interpretable nature of case-based reasoning \cite{slade1991case} and the significant challenges that domain experts encounter when analyzing large training datasets and examining the model learning process. This work holds the premise that experts can manually pre-define representative samples based on insights from previous experiments or their knowledge. In training, representation models must learn to push data points to at least one of these key samples similar to the update process of the K-mean clustering algorithm. On a fine-grained level, experts can further adjust the position of a data point in the latent space by defining its closest and distant friends. Representation models are trained by measuring the relevance between the control sample and its references.

%This chapter presents a novel method named \textbf{HVG}, facilitating efficient and reliable \textbf{G}raph classification by active \textbf{H}uman \textbf{V}erification. The proposed method centers around a pivotal concept: developing a human-aligned representation learning component capable of generating graph representations crucial for interpretable predictions. The proposed method integrates GNN architectures as functions responsible for encoding graph data. This approach harnesses human knowledge and feedback by imposing dual layers of constraints: class-level knowledge and instance-level feedback. Furthermore, this work introduces an iterative human-AI interaction approach for representation learning, a method that significantly elevates predictive performance and model stability. To enhance interpretability, the method incorporates two interpretable predictors based on the widely recognized k-nearest-neighbor algorithm. Additionally, it introduces multiple prediction explanation formats based on the resources provided by these interpretable components. Extensive experiments and user studies validate the correctness and efficiency of the proposed method.

This work introduces a breakthrough method called \textbf{HVG} designed to enhance the accuracy and transparency of \textbf{G}raph classification through \textbf{H}uman \textbf{V}erification. The cornerstone of this method is a representation learning process that aligns with human understanding, crucial for generating graph representations that aid in making transparent and interpretable predictions. Specifically, in the learning process, a GNN encoder is trained to transform a graph into vector embeddings in a latent space. The interactive collaboration approach is adaptable to diverse GNN architectures. Additionally, it leverages knowledge as a class-level constraint and feedback as an instance-level constraint to achieve the human-alignment graph representations. Moreover, this research incorporates an iterative process of human-AI interaction in the learning process, which substantially improves both the predictive accuracy and stability of classification models. To boost transparency and interpretability, the proposed technique employs two predictors based on the established KNN algorithm and introduces various formats for explanations of predictions, drawing on the capabilities of designed interpretable features. Comprehensive experiments and analyses confirm the method's effectiveness and efficiency.

%The research presented in this chapter, including the proposed method and experimental results, was published in \cite{bui2024human}. The paper's remainder is as follows. \cref{sec:related_work} presents related work. \cref{sec:methodology} describes the methodology. Experiments are reported in \cref{sec:exp}. Possible fairness issues are discussed in \cref{sec:hm_discussion}. The paper is concluded in \cref{sec:conclusion}.

The work described in this chapter, encompassing the proposed methodology and its experimental validations, has been documented in \cite{bui2024human}. The structure of the remaining content is organized as follows. \cref{sec:related_work} reviews the literature related to this work. The methodology employed is detailed in \cref{sec:methodology}. The experimental findings are presented in \cref{sec:exp}. Discussion of potential fairness concerns is demonstrated in \cref{sec:hm_discussion}. Finally, the chapter concludes with \cref{sec:conclusion}.

\section{Related Work} \label{sec:related_work}
\subsection{Human-in-the-loop AI}
%Human-in-the-loop ML \cite{mosqueira2023human} is an approach where human expertise collaborates with machine learning algorithms, enabling both iterative feedback loops and improved model performance. Ramos et al., 2020 \cite{ramos2020interactive} introduced a comprehensive framework for facilitating human-AI interactions. Lately, Liu et al., 2022 \cite{liu2022learning} discovered that algorithmic representations might be incompatible with human intuitions requiring human constraints in the training process. Taesiri et al., 2022 \cite{taesiri2022visual} proposed a framework for humans and AI to collaborate toward a mutual decision-making process. These approaches leverage the strengths of both humans and machines to build more effective and reliable systems.

In the AI area, the HITL concept \cite{mosqueira2023human} embodies a collaborative strategy where human knowledge is integrated with algorithmic processes, enhancing model accuracy through repetitive feedback mechanisms. The framework delineated by Ramos et al.  \cite{ramos2020interactive} aims to optimize interactions between humans and artificial intelligence. Recent investigations by Liu et al. \cite{liu2022learning} highlighted a discord between algorithmic outputs and human intuition, advocating for the incorporation of human-guided constraints during model training. Similarly, Taesiri et al. \cite{taesiri2022visual} advanced a cooperative framework wherein humans and AI engage jointly in decision-making processes. These methodologies underscore the combined capabilities of human and machine contributions in developing more robust and efficient systems.

\subsection{Deep-learning-enhanced Case-based Reasoning}

%In traditional machine learning, the case-based reasoning paradigm is notable \cite{slade1991case}, serving as a foundational component in decision-support systems. This approach hinges on utilizing past experiences to address new problems. The inherent ability of deep learning models to recognize patterns and transform data into latent representations significantly enhances the process of retrieving past instances. Recent contributions from Li et al. \cite{li2018deep}, Chen et al. \cite{chen2019looks}, and Davoudi et al. \cite{davoudi2021toward} adopt a prototype-centric method in which prototypes are discovered during the training phase. This study resembles \cite{davoudi2021toward}, particularly in differentiating between deep representation learning and the phase of prototype determination.

Among conventional ML algorithms, case-based reasoning \cite{slade1991case} is a significant approach, acting as a cornerstone in decision-support systems. In this approach, new challenges are solved by referring to past experiences. The capacity of DL models to transform data into hidden representations and identify patterns markedly improves the retrieval of previous cases. Innovative works \cite{li2018deep, chen2019looks, davoudi2021toward} followed prototype-centric approaches, wherein prototypes are identified during the training period. This research is similar to the study by Davoudi et al. \cite{davoudi2021toward}, especially in separating the deep representation learning and the retrieval phases. 

\subsection{Interpretable Graph Neural Networks}
%Interpretable GNNs \cite{li2021braingnn, dai2021towards, feng2022kergnns, ragno2022prototype, zhang2022protgnn} aim for improved interpretability via mechanisms like node pooling, similarity modules, subgraph aggregation, and prototypes. The method by Dai et al. \cite{dai2021towards} encountered training difficulties and did not thoroughly address explanation generation methods. Methods by Ragno et al. \cite{ragno2022prototype} and Zhang et al. \cite{zhang2022protgnn} utilize prototype-based prediction techniques but differ in the prototype projection phase. It is pivotal to note that contemporary approaches prioritize prediction accuracy over users' perception of explanations.

Interpretable GNNs \cite{li2021braingnn, dai2021towards, feng2022kergnns, ragno2022prototype, zhang2022protgnn} strive to enhance model transparency and interpretability through various methods such as node pooling, similarity assessment, subgraph extraction, and prototype mapping. The approach presented by Dai et al. \cite{dai2021towards} faced challenges during the training phase and was insufficient in developing robust explanation techniques. Techniques by Zhang et al. \cite{zhang2022protgnn} and Ragno et al. \cite{ragno2022prototype} employ strategies that make predictions based on prototypes, yet they vary in the projection processes. Notably, existing methodologies often concentrate on predictive performance, while overlooking how explanations are perceived by users.

\section{Methodology} \label{sec:methodology}

\begin{figure}[ht]
	\centering
	\includegraphics[width=\linewidth]{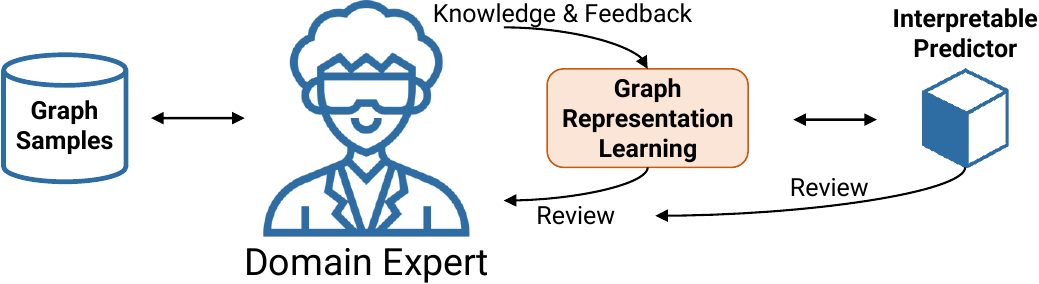}
%	\caption{HVG Overview. It centers around a pivotal concept: learning human-aligned graph representations through the empowerment of human active verification.}
	\caption{HVG Framework. It revolves around a central principle: developing graph representations that align with human understanding by incorporating human verification in the training process.}
	\label{fig:hm_framework}
\end{figure}

\subsection{Methodology}
%The graph classification problem is formalized in the context of representation learning and case-based reasoning. The goal is to find a mapping $P: \mathcal{G} \rightarrow \mathcal{Y}$. This work assumes there exists a representation model $f$, which takes a graph $\mathcal{G} \in \mathcal{D} = \{(\mathcal{G}_1, y_1),...,(\mathcal{G}_N, y_N)\}$ as an input and outputs a $d$-dimensional representation $h_\mathcal{G} \in \mathbb{R}^d$. Given a GNN model $g_{\phi,\theta}: \mathcal{G} \rightarrow \mathcal{Y}$, the representation model is the last layer before the classifier, referred to as $f = e(g)$, where $e$ is a selection function. For each instance $\mathcal{G}$, a reference policy $\pi$ selects $K$-labeled samples from the training set $\mathcal{D}_{\text{train}}$ and presents them to humans. This work's primary focus is the effectiveness of $f$ for case-based graph classification. 

In this study, the problem of graph classification is defined based on case-based reasoning and representation learning, with the objective of establishing a mapping function $P: \mathcal{G} \rightarrow \mathcal{Y}$. The hypothesis is that a representational model $f$ exists, capable of processing a graph $\mathcal{G} \in \mathcal{D} = \{(\mathcal{G}_1, y_1),...,(\mathcal{G}_N, y_N)\}$ and generating a $d$-dimensional vector $h_\mathcal{G} \in \mathbb{R}^d$. Given a GNN $g_{\phi,\theta}: \mathcal{G} \rightarrow \mathcal{Y}$, $f$ functions as the layer preceding the classification stage, denoted as $f = e(g)$, where $e$ serves as a function that selects a specific layer. For every graph $\mathcal{G}$, a policy $\pi$ identifies $K$ references from $\mathcal{D}_{\text{train}}$, the training set of $\mathcal{D}$. A prediction is then derived by weighted voting on labels of these references, wherein weights are measured by the proximity of the input graph to references. The primary focus of this work is on the effectiveness of $f$ in facilitating human-alignment graph classification.

%As presented in \cref{fig:hm_framework}, the essential component is graph representation learning, which turns experts' knowledge and feedback into constraints to achieve human-compatible graph representations. Interpretable predictors utilize these representations to make predictions. Domain experts inspect learned representations and predictions to ensure that AI models and human intuitions are correspondence. \cref{alg:overview_algorithm} presents an overview of the execution pipeline.

As shown in \cref{fig:hm_framework}, the fundamental process of the framework is learning graph representations, which integrates the expertise and feedback into constraints to train graph representations compatible with human comprehension. These representation vectors are then employed for interpretable predictions. Additionally, domain experts can review predictions alongside learned representations to verify the alignment of human and model decisions. \cref{alg:overview_algorithm} outlines the process flow of the execution pipeline.

\begin{algorithm}[ht]
  \caption{General Procedure}
  \begingroup
      \raggedright
      \textbf{Input}: GNN $g$ with $\phi,\theta$, reference policy $\pi$, dataset $\mathcal{D}$, and \#epochs $T$ \\
      \textbf{Output}: Representation model $f$, interpretable predictions \\
  \endgroup
  \begin{algorithmic}[1]
    \FOR{$i = 1$ to $T$}
    \STATE{Execute $g$ on $\mathcal{D}$} 
    \STATE{Update $\phi,\theta$ via \cref{eq:hm_objective}} \COMMENT{\cref{sec:rel}}
    \STATE{Suggest new centroids by interval} \COMMENT{\cref{sec:rel}}
    \STATE{Encode knowledge and feedback dynamically} \COMMENT{\cref{sec:rel}}
    \STATE{Break if meeting early stopping criteria}
    \ENDFOR
    \STATE{Execute $h_\mathcal{G} = f(\mathcal{G})$} \COMMENT {Obtain representations}
    \STATE{Retrieve $\mathcal{G}$'s closest references via $\pi$} \COMMENT {\cref{sec:predictor}}
    \STATE{Execute $P$} \COMMENT {\cref{sec:predictor}}
    \STATE{Construct explanations} \COMMENT {\cref{sec:expl}}
  \end{algorithmic}
\label{alg:overview_algorithm}
\end{algorithm}

\subsection{Human-alignment Representation Learning} \label{sec:rel}
%The graph representation learning component incorporates a GNN encoder to transform graph data into a latent space. The proposed method is open to various GNN architectures, abstracted as $\mathbf{H}^l = \text{GNN}(\mathcal{G}, \mathbf{A}, \mathbf{H}^{l-1})$, where $l$ is the layer index, $\mathbf{A}$ is the adjacency matrix, and $\mathbf{H}$ is a representation matrix. Sum pooling is performed over $\mathbf{H}$ to compute the graph representation vector $h_\mathcal{G}$. 

%The goal is to learn representations that are efficient in classification tasks and compatible with human intuitions. First, we adopt the cross entropy loss function to encourage samples to be well-separated in the latent space. This loss function can be replaced in real-world problems with other objective types.
Human-alignment graph representations not only excel in classifications but also align with human comprehension. Initially, the cross-entropy loss function is applied to promote the distinct separation of samples within this latent space. In practical applications, this loss function may be substituted with alternative objectives tailored to specific real-world scenarios.

To achieve the objectives above, this component employs a GNN encoder to map graphs into a latent space. This approach is adaptable to a variety of GNNs, described by the equation $\mathbf{H}^l = \text{GNN}(\mathcal{G}, \mathbf{A}, \mathbf{H}^{l-1})$, where $\mathbf{H}$ is a matrix of node representations, $l$ denotes the layer index, and $\mathbf{A}$ represents the adjacency matrix. Sum pooling across $\mathbf{H}$ is utilized to derive the graph representation vector $h_\mathcal{G}$.

\begin{equation}
    \mathcal{L}_{pred} = -\frac{1}{N} \sum_{i=1}^{N} y_i \cdot \textrm{log}(p_{\theta}(\hat{y}|h_\mathcal{G})), 
\end{equation}
wherein $p_{\theta}$ serves as a variational approximation function that estimates predictive probabilities based on a graph representation. Practically, $\theta$ denotes the weights associated with the predictive layer in a GNN model.

%where $p_{\theta}$ is a variational approximation function estimating class probabilities given a graph representation. Practically, $\theta$ represents the weights of a prediction layer in a DL model. 

\noindent\textbf{Class-level Knowledge:} 
This work operates on the premise that domain experts can identify representative instances that exhibit distinct characteristics, representing a variety of sample groups within a given problem. These key instances are defined as a prototype set $\mathcal{P} = \{p_1, p_2, ..., p_M\}$. It is reasoned that a graph $\mathcal{G} \in \mathcal{D}$ should be proximate to at least one prototype in the latent space. Additionally, the sample ought to be distanced from prototypes that have a different label. Consequently, the secondary objective constraint is constructed using the triplet loss principle, outlined as follows:

%A hypothesis is that domain experts can point out prominent cases with specific features representing diverse families of samples given a problem. These representative samples are referred to as a prototype set $\mathcal{P} = \{p_1, p_2, ..., p_M\}$.
%Intuitively, a training sample must be located near at least one of the representative points in the latent space. Moreover, a sample should also be distant from points that do not belong to the same class. Therefore, the second objective function is designed based on the triplet loss as follows:

\begin{equation}
    \mathcal{L}_{ck} = \frac{1}{N} \sum_{i=1}^{N} \min_{j:p_j \in \mathcal{P}_{y_i}} || f(\mathcal{G}_i) - f(p_j) ||_2^2 - \frac{1}{N} \sum_{i=1}^N \min_{j:p_j \not\in \mathcal{P}_{y_i}} || f(\mathcal{G}_i) - f(p_j) ||_2^2,
\end{equation}
where $\mathcal{P}_{y_i}$ represents a subset of $\mathcal{P}$ with respect to the class $y_i$.

%To enhance discriminative power, domain experts can examine representations in visualization interfaces and provide additional feedback via a contrastive style. Specifically, experts can generate triplets, each consisting of an input graph, a positive graph, and a negative one, denoted as $(\mathcal{G}, \mathcal{G}^{+}, \mathcal{G}^{-})$. Two graphs are considered a positive/negative pair if they are similar/dissimilar to each other on a specific criterion. The instance-level feedback is beneficial for error analysis scenarios or scenarios when experts want to ensure that the reference policy doesn't select human-incompatible examples. The instance-level constraint is presented as follows:

\noindent\textbf{Instance-level Feedback:} Domain experts are able to scrutinize representations through visualization tools and provide further adjustments using a contrastive approach to increase the discriminative capability of the model. In particular, ones may create triplets comprising an input graph, a positive reference, and a negative one, represented as $(\mathcal{G}, \mathcal{G}^{+}, \mathcal{G}^{-})$. A graph is regarded as positive or negative based on its similarity or dissimilarity to the input with respect to a specific metric. Such instance-level feedback proves invaluable in error analysis or in situations where one aims to prevent the selection of examples by a policy $\pi$ that are incompatible with human comprehension. The instance-level feedback is represented in a constraint formula as follows:

\begin{equation}
    \mathcal{L}_{ik} = \sum_{(\mathcal{G}, \mathcal{G}^{+}, \mathcal{G}^{-}) \in \mathcal{T}} \text{max}(0, || f(\mathcal{G}) - f(\mathcal{G}^{+})||_2^2 - || f(\mathcal{G}) - f(\mathcal{G}^{-})||_2^2 + \epsilon)
    \label{eq:hm_instance_level}
\end{equation}

\begin{equation}
    \min_{\phi,\theta}\quad  \mathcal{L}_{pred} + \alpha \mathcal{L}_{ck} + \beta \mathcal{L}_{ik}
\label{eq:hm_objective}
\end{equation}

\noindent\textbf{Iterative Interaction:} As discussed in \cite{ramos2020interactive}, an iterative paradigm facilitates the alignment between human and AI models. Particularly, humans have the flexibility to halt the training at any point to assess if their contributed knowledge and feedback are proving beneficial. Moreover, AI models are equipped to suggest alternative centroids for human consideration, which could assist in better adjustments in optimization processes. These suggestions can either be accepted or rejected by human operators. For each class, $K_c$ centroids are established as $\mu = \{\mu_i,...,\mu_{K_c}\}$, outlined by the equation provided below:

%An iterative representation learning process \cite{ramos2020interactive} facilitates human-AI alignment. Specifically, experts can stop the training process anytime and review whether their injected knowledge/feedback is beneficial. Additionally, AI models can propose new centroid candidates to humans to improve weight adjustments in optimization processes. Humans can either accept or reject the suggestions. For each class, $K_c$ centroids $\mu = \{\mu_i,...,\mu_{K_c}\}$ are defined using the following equation:

\begin{equation}
    \argmin_{\mu} \sum_{i=1}^{K_c} \sum_{j=1}^{N_i} || h_{\mathcal{G}_{ij}} - \mu_i||,
    \label{eq:hm_centroid_candidate}
\end{equation}
where $N_i$ is the number of graphs in a cluster $i$ of the class $c$.

\subsection{Interpretable Predictor} \label{sec:predictor}

%Case-based prediction \cite{slade1991case} aligns naturally with human cognition, drawing inspiration from a human innate ability to tackle new challenges by referencing similar past experiences. This approach aims to define a predictor $P: h_\mathcal{G} \rightarrow \mathcal{Y} $ that outputs a class for an input $\mathcal{G}$, given a representation $h_\mathcal{G}$ and a reference policy $\pi$. This work focuses on two distinct reference policies rooted in the nearest-neighbor algorithm.

As discussed by  \cite{slade1991case}, case-based reasoning closely mirrors the way humans process information, leveraging our intrinsic capacity to address novel problems by recalling analogous prior situations. This method endeavors to establish a mapping $P: h_\mathcal{G} \rightarrow \mathcal{Y} $, which assigns an outcome to an input $\mathcal{G}$, based on a representation vector $h_\mathcal{G}$ and a policy $\pi$. This work specifically concentrates on two different policies derived from the nearest-neighbor approach.

\begin{equation}
\begin{aligned}
    \pi_a &= \text{KNN}(\mathcal{G}, f, \mathcal{D}_{\textrm{train}}) \\
    \pi_c &= \{ \text{KNN\_CLASS}(\mathcal{G}, f, \mathcal{D}^c_{\textrm{train}}) \}_{c=1}^C
\end{aligned}
\label{eq:hm_ref}
\end{equation}

%As shown in \cref{eq:hm_ref}, different subscripts are employed to denote two reference policies. $\pi_a$ represents a conventional k-nearest-neighbors algorithm, while $\pi_c$ adopts a strategy that selects an equal number of references for each class based on the corresponding sub-training set $D^c_{\text{train}}$. Strategy selection depends on the characteristics of representation spaces. Practically, $\pi_a$ is appropriate for scenarios marked by well-separated representations, low noise, and homogeneous neighbors. In contrast, $\pi_c$ excels in situations with complex decision boundaries, where overlapping representations are prevalent.

As illustrated in \cref{eq:hm_ref}, two distinct reference policies are indicated by varying subscripts. $\pi_a$ signifies the traditional KNN algorithm, whereas $\pi_c$ implements a method that ensures an equal representation of references from each class, utilizing the specific subset $D^c_{\text{train}}$. The choice of strategy depends on the properties of representations. Typically, $\pi_a$ is suitable for environments characterized by distinct, low-noise, and uniform representation spaces. Conversely, $\pi_c$ is more effective in contexts with intricate decision boundaries and overlapping representations.

\begin{equation}
    P(\hat{Y} | \mathcal{G}, \pi) = \sum_{R_i \in \pi} a(\mathcal{G}, R_i) y_i \quad \textrm{s.t} \quad
    a(\mathcal{G},  R_i) = \text{softmax} (\text{sim}(\mathcal{G}, R_i)),
    \label{eq:hm_predictor}
\end{equation}
where $y_i$ is the ground-truth label represented in a one-hot format, and sim is a similarity function. Practically, $\text{sim}(\mathcal{G}, R) = \text{exp}\bigl( {-\frac{||h_\mathcal{G} - h_{R}||^2} {2\sigma^2}} \bigr)$, where $\sigma = 2$.

\subsection{Explanation Construction} \label{sec:expl}
%Explanations are necessary for improving humans' understanding of model predictions \cite{doshi2017towards}. This module organizes information provided by the interpretable prediction function into user-friendly explanations. Explanations are presented to users in the following types:

Explanations are essential for improving human comprehension of model predictions, as posited by Doshi et al. \cite{doshi2017towards}. In this work, explanations are generated based on information from the interpretable predictor. Additionally, generated explanations are user-friendly and comprehensive,  and are formatted for users in various types:

\begin{itemize}
%    \item \textbf{Comparative Analysis:} Visualizing references enables users to comprehend the model's decisions more easily. Additionally, it facilitates the examination of model errors by comparing incorrect predictions with correct ones in similar cases, aiding model refinement based on instance-level constraints.
    \item \textbf{Comparative Analysis:} Visualization of references enhances understanding of the model's rationale. This method also supports the scrutiny of model prediction errors by contrasting incorrect and correct predictions in analogous scenarios, thus contributing to model improvement through instance-specific adjustments.
 
%    \item \textbf{Reference attribution:} This feature gives users quantitative insights into the decision-making process, revealing the most influential references in shaping the current decision. This feature fosters transparency and interpretability in decision-making.
	\item \textbf{Reference Attributions:} This functionality provides quantitative insights into the decision-making mechanism by identifying the most significant references that influence the current prediction. It promotes clarity and interpretability in the decision process.

%    \item \textbf{Subgraph visualization:} This function highlights essential components within execution graphs, thus enhancing user understanding. These essential components typically represent common patterns within a family of graphs and are extracted using PGExplainer \cite{luo2020parameterized}.
	\item \textbf{Visualization of Essential Patterns:} This feature accentuates critical elements within execution graphs, improving user comprehension. These patterns generally signify recurring patterns across a series of graphs and are identified through the use of techniques like those presented in the previous chapter.
	
\end{itemize}
 
\section{Experiments} \label{sec:exp}

\subsection{Datasets and Benchmark Models}

\begin{table}[ht]
	\centering
	\caption{Statistical Information on Datasets}
	\begin{tabular}{c|c|c|c|c|c}
		\toprule
		Dataset Name & Graphs & Avg Nodes & Avg Edges & Features & Classes  \\
		\midrule
		Mutag & 188 & 17.93 & 19.79 & 7 & 2\\
		Proteins & 1113 & 39.06 & 72.82 & 29 & 2\\
		IMDB-Binary & 1000 & 19.77 & 96.53 & 271 & 2 \\
		DD & 1178 & 284.32 & 715.66 & 89 & 2 \\
		Twitter & 6940 & 21.10 & 20.10 & 768 & 3 \\
		\bottomrule
	\end{tabular}
	\label{tab:hm_data_stat}
\end{table}

% para
This work utilized five graph classification datasets: Mutag, IMDB-Binary (IMDB), DD, Proteins \cite{rossi2015network}, and Graph-Twitter (Twitter) \cite{yuan2022explainability} for experiments. It selected four fundamental GNNs as baselines: GCN \cite{kipf2016semi}, GraphSage (Sage) \cite{hamilton2017inductive}, GIN \cite{xu2018powerful}, and GAT \cite{velickovic2017graph}. Each architecture included two message-passing layers, a subsequent hidden layer, and a final prediction layer. Based on these baselines, a group of models was trained with the objective function \cref{eq:hm_objective}, referring to these configurations as HVG. Subsequently, a predictor $P$ was deployed on this HVG group, resulting in interpretable models collectively termed \textbf{HVG}$^\mathbf{+}$.

\subsection{Configuration Details}

%An 8:1:1 data-splitting strategy and 10-fold cross-validation were employed similar to experimental configurations in previous chapters. Note that in IMDB and DD datasets, node features were represented as one-hot vectors based on node degrees, whereas the features for the Proteins dataset were standardized. 

Similar to methodologies outlined in previous chapters, an 8:1:1 data-separation method along with 10-fold cross-validation was implemented. It's important to note that in graphs of the DD and IMDB datasets, node features were represented through one-hot vectors related to node degrees, while those in graphs of the Proteins dataset were subjected to standard normalization.

%All models underwent training for 100 epochs, with an initial learning rate of 0.01 reduced by a factor of 0.5 after 50 epochs. Additionally, the early stopping technique was adopted for model training. Hidden numbers were set to 32, except for the Twitter dataset, which was 16. GAT employed eight attention heads and utilized ReLU activation. Meanwhile, for GraphSage, Mean aggregators were utilized, except for the Twitter dataset, where GCN aggregators were applied. The parameters $\alpha$ and $\beta$ were selected between $10^{-2}$ and $10^{-5}$. 

Each model was trained for 100 epochs, starting with a learning rate of 0.01, which was halved after the 50th epoch. Early stopping was also utilized during model training. The number of hidden units in transformation layers was set at 32 for all datasets except for Twitter, where it was reduced to 16. The GAT model incorporated 8 attention heads and used ReLU as the activation function. In GraphSage models, Mean was used as the aggregation function, except in the Twitter dataset where GCN was utilized instead. The hyper parameters $\alpha$ and $\beta$ ranged between $10^{-2}$ and $10^{-5}$ depending on particular situations.

%Similar to the \cref{chap:concept}, reference selection methods and \cref{eq:hm_centroid_candidate} were implemented based on Faiss 1.7.4. Specifically, the number of references $K$ was set to 10 and 3 in $\pi_a$ and $\pi_c$, respectively. 

Consistent with the strategies discussed in \cref{chap:concept}, reference policies and \cref{eq:hm_centroid_candidate} were developed using Faiss \cite{johnson2019billion}. This work configured the number of references $K$  to 10 for $\pi_a$ and 3 for $\pi_c$.

\subsection{Predictive Performance Comparison} 

\begin{sidewaystable}
	\begin{minipage}[c][\textheight]{\linewidth}
		\caption{Comparative Analysis of Predictive Performance Across Datasets}
		\centering
		\begin{tabular}{c|c|c|c|c|c}
			\toprule
			\textbf{Method}                                & \textbf{Mutag} & \textbf{Proteins} & \textbf{IMDB} & \textbf{DD} & \textbf{Twitter} \\
			\midrule
			BASE-GCN & 0.718 $\pm$ 0.094 & 0.714 $\pm$ 0.051 & 0.710 $\pm$ 0.049 & 0.715 $\pm$ 0.040 & 0.642 $\pm$ 0.017 \\
			BASE-SAGE & 0.730 $\pm$ 0.096 & 0.694 $\pm$ 0.049 & 0.715 $\pm$ 0.051 & 0.743 $\pm$ 0.038 & 0.636 $\pm$ 0.021 \\        
			BASE-GIN & 0.862 $\pm$ 0.096 & 0.750 $\pm$ 0.052 & 0.726 $\pm$ 0.029 & 0.699 $\pm$ 0.035 & 0.651 $\pm$ 0.013 \\
			BASE-GAT & 0.750 $\pm$ 0.112 & 0.672 $\pm$ 0.120 & 0.726 $\pm$ 0.034 & 0.699 $\pm$ 0.035 & 0.652 $\pm$ 0.016 \\
			\hline
			
			HVG-GCN & 0.740 $\pm$ 0.085 & 0.722 $\pm$ 0.038 & 0.755 $\pm$ 0.039 & 0.729 $\pm$ 0.037 & 0.643 $\pm$ 0.018 \\
			HVG-SAGE & 0.751 $\pm$ 0.087 & 0.730 $\pm$ 0.046 & 0.749 $\pm$ 0.038 & 0.740 $\pm$ 0.034 & 0.653 $\pm$ 0.014 \\
			HVG-GIN & \underline{0.873 $\pm$ 0.071} & \underline{0.772 $\pm$ 0.036} & \underline{0.773 $\pm$ 0.043} & \textbf{0.772} $\pm$ \textbf{0.030} & 0.651 $\pm$ 0.018 \\
			HVG-GAT & 0.782 $\pm$ 0.068 & 0.753 $\pm$ 0.040 & 0.762 $\pm$ 0.027 & 0.706 $\pm$ 0.032 & \underline{0.658 $\pm$ 0.012} \\
			\hline
			\textbf{HVG}$^{+}$\textbf{-GCN} & 0.766 $\pm$ 0.100 & 0.746 $\pm$ 0.041 & 0.748 $\pm$ 0.035 & 0.728 $\pm$ 0.037 & 0.646 $\pm$ 0.018 \\
			\textbf{HVG}$^{+}$\textbf{-SAGE} & 0.767 $\pm$ 0.090 & 0.748 $\pm$ 0.033 & 0.744 $\pm$ 0.043 & 0.724 $\pm$ 0.034 & 0.642 $\pm$ 0.014 \\
			\textbf{HVG}$^{+}$\textbf{-GIN} & \textbf{0.882} $\pm$ \textbf{0.063} & \textbf{0.781} $\pm$ \textbf{0.033} & \textbf{0.777} $\pm$ \textbf{0.041} & \underline{0.761 $\pm$ 0.022} & 0.651 $\pm$ 0.021 \\
			\textbf{HVG}$^{+}$\textbf{-GAT} & 0.777 $\pm$ 0.073 & 0.753 $\pm$ 0.040 & 0.762 $\pm$ 0.026 & 0.708 $\pm$ 0.031 & \textbf{0.660} $\pm$ \textbf{0.013} \\
			\bottomrule
		\end{tabular}
		\label{tab:hm_acc_comp}
	\end{minipage}
\end{sidewaystable}	

%Analysis of \cref{tab:hm_acc_comp} reveals several noteworthy findings. The proposed HVG method significantly enhances the performance of GNN backbones, achieving up to 8\% higher accuracies than baselines. Additionally, the introduction of class-level knowledge constraints results in reduced accuracy variances across all datasets and models. Furthermore, the KNN-based interpretable predictor, which leverages GNN representations, contributes to an overall improvement in predictive performance. The observations indicate that KNN is particularly effective for the Mutag and Proteins datasets, while KNN\_Class exhibits higher performance with these others. This disparity in performance can be attributed to the nature of the node features and graph complexity. Unlike Mutag and Proteins, IMDB and DD solely rely on vertex degrees. Twitter graphs are characterized by their noise and complexity. Consequently, graphs are not well-separated in latent space, thereby hindering the efficiency of KNN.

\cref{tab:hm_acc_comp} presents several important observations. The proposed human-AI interaction technique boosts the efficacy of GNN architectures substantially, recording accuracies up to 8\% higher than those of baseline models. Furthermore, integrating general knowledge constraints tends to decrease the variability of accuracy across various configurations. Additionally, the KNN-based predictor, empowered by GNN representations, achieves significant predictive performance in all scenarios. Notably, KNN proves especially effective for the Mutag and Proteins datasets, whereas KNN\_Class shows superior performance with other datasets. This variation in effectiveness is linked to differences in network complexity and specific characteristics like node features. For example, IMDB and DD depend exclusively on vertex degrees, while Twitter graphs are distinguished by their noise and difficulty, which complicates the separation of graphs in latent space and reduces the effectiveness of KNN strategies.

\subsection{Benefits of Human-AI Interactions}

\pgfplotsset{width=8.5cm,height=4.3cm}
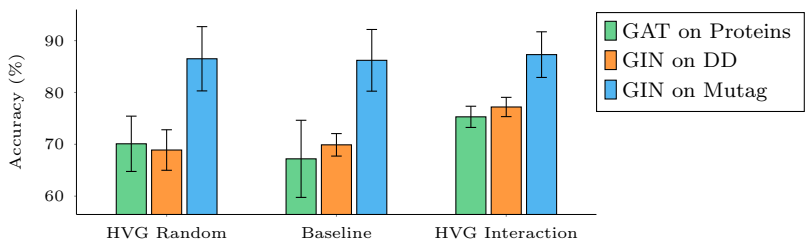
\begin{figure}[ht]
    \centering
    \begin{tikzpicture}
    \begin{axis}[
        ybar,
        bar width=0.4cm,
        legend cell align=left,
        enlarge x limits={abs=1.2cm},
        ylabel={Accuracy (\%)},
        y label style={at={(axis description cs:-0.08,0.5)},anchor=south},
        symbolic x coords={HVG Random,Baseline,HVG Interaction},
        xtick=data,
        x tick label style= {text width=3cm,align=center},
        legend style={font=\scriptsize, at={(1.2,0.5)}, anchor=south,legend columns=1,/tikz/every even column/.append style={column sep=0.5cm}},
        legend image code/.code={\draw [#1] (0cm,-0.1cm) rectangle (0.15cm,0.15cm);},
        ]
    \addplot[fill=Green!80,  error bars/.cd, y dir=both, y explicit] coordinates{
        (HVG Random, 70.1) +- (0, 5.3333)
        (Baseline, 67.2) +- (0, 7.438)
        (HVG Interaction, 75.3) +- (0, 2.045)
    };

    \addplot[fill=Orange!80,  error bars/.cd, y dir=both, y explicit] coordinates{
        (HVG Random, 68.9) +- (0,  3.907)
        (Baseline, 69.9) +- (0,  2.169)
        (HVG Interaction, 77.2) +- (0,  1.859)
    };

    \addplot[fill=Blue!80,  error bars/.cd, y dir=both, y explicit] coordinates{
        (HVG Random, 86.5) +- (0, 6.198)
        (Baseline, 86.2) +- (0, 5.95)
        (HVG Interaction, 87.3) +- (0, 4.401)
    };
    \legend{GAT on Proteins, GIN on DD, GIN on Mutag}
    \end{axis}
    
    \end{tikzpicture}
%    \caption{Comparing the accuracy of three training configurations on two GNN backbones and three datasets}
	\caption{Evaluating the Accuracy of Three Configurations on Human-AI Interaction}
    \label{fig:hm_human_ai}
\end{figure}

%This experiment aimed to validate the hypothesis that the proposed interaction strategy could enhance HVG predictive performance and the stability of training processes. \cref{fig:hm_human_ai} demonstrates the experiments' results of two different backbones and three datasets. For HVG Interaction, centroid candidates were identified via \cref{eq:hm_centroid_candidate}, and user selections were simulated by controlling a rejection threshold. Notably, the HVG Interaction strategy significantly improved the predictive capabilities of GNN backbones compared with Baseline and HVG Random approaches. Since the author assumed that datasets possessed IID properties and employed random centroid selection, the model accuracies of the HVG Random strategy showed slightly more significant variability than others, and its accuracies were only comparable to those of baseline models. These findings underscore the importance of human-AI interactions as a critical and beneficial factor for achieving superior model performance and alignment between human and AI systems.

This study was conducted to test the hypothesis that the interactive strategy could improve both the predictive accuracy of GNN architectures and the stability of training procedures. \cref{fig:hm_human_ai} presents the outcomes of experiments involving three datasets and two distinct model architectures. In the interaction scenario, centroid candidates were defined using \cref{eq:hm_centroid_candidate}, and user choices were simulated through the adjustment of a rejection threshold. The interactive strategy notably enhanced the predictive performance of models over both the random and baseline scenarios. Given the assumption that datasets exhibited IID characteristics and the use of arbitrary centroid selection, the random strategy demonstrated slightly higher variability in predictive accuracy compared to other methods, with its performance only mirroring that of the baseline models. These results highlight the crucial and positive role of human-AI collaboration in elevating model effectiveness and fostering alignment between human operators and AI systems.

\subsection{Assessing the Efficacy of Instance-Level Feedback}

%Based on \cref{sec:expl}, the author investigated the benefits of incorporating instance-level user feedback. An HVG-GIN model, fine-tuned with \cref{eq:hm_instance_level}, generated triplets from Mutag's training set. After being explained the task quickly, 19 volunteers predicted the labels of ten graphs based on supported information like subgraph visualizations and reference graphs. Ultimately, user prediction accuracy in the non-fine-tuned setting was compared with that in the fine-tuned configuration.
Referencing \cref{sec:expl}, the author explored the advantages of integrating instance-level user feedback. Utilizing an HVG-GIN model refined with \cref{eq:hm_instance_level}, triplet data points were created from the training set of Mutag. Nineteen volunteers were explained briefly on the task and then predicted outcomes of ten graphs, supported by tools such as pattern visualizations and references. Finally, the prediction accuracy of users and models using the non-fine-tuned and fine-tuned versions were compared against each other.

\pgfplotsset{width=8.5cm,height=4.2cm}
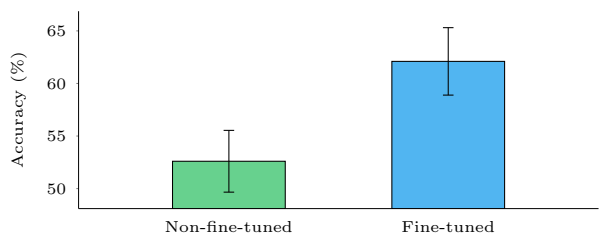
\begin{figure}[ht]
    \centering
    \begin{tikzpicture}
    \begin{axis}[
        ybar,
        bar width=1.5cm,
        bar shift=0pt,
        enlarge x limits={abs=2cm},
        ylabel={Accuracy (\%)},
        y label style={at={(axis description cs:-0.08,0.5)},anchor=south},
        % symbolic x coords={Non-fine-tuned, Fine-tuned},
        xtick={1,2},
        xticklabels={Non-fine-tuned,Fine-tuned},
        x tick style={draw=none},
        x tick label style= {text width=3cm,align=center},
        legend style={draw=none, fill=none, font=\scriptsize, at={(0.17,0.85)}, anchor=south,
        legend columns=-1,/tikz/every even column/.append style={column sep=0.5cm}}
        ]

    \addplot[fill=Green!80,  error bars/.cd, y dir=both, y explicit] coordinates{
        (1, 52.6) +- (0,  2.93793813)
    };
    \addplot[fill=Blue!80,  error bars/.cd, y dir=both, y explicit] coordinates{
        (2, 62.1) +- (0, 3.207309288)
    };

    \end{axis}
    
    \end{tikzpicture}
%    \caption{User predictive performance with references retrieved from non-fine-tuned and fine-tuned models using the instance-level feedback constraint}
	\caption{A Comparison of Predictive Performance With References Retrieved From Two Model Versions}
    \label{fig:hm_instance_level}
\end{figure}

%As depicted in \cref{fig:hm_instance_level}, the fine-tuned model outperformed its non-fine-tuned counterpart, improving user performance. However, participants faced significant challenges due to limited domain knowledge, resulting in relatively low accuracy in both scenarios. Notably, the instance-level feedback adjustments brought target graph representations closer to actual neighbors in the latent space, making references from the fine-tuned model more helpful. This study highlighted the potential of instance-level feedback in specific scenarios, such as human-model alignment enhancement.

As illustrated in \cref{fig:hm_instance_level}, the fine-tuned version demonstrated superior performance compared to its non-fine-tuned counterpart, leading to enhanced user outcomes. Nonetheless, participants encountered considerable difficulties due to inadequate domain knowledge, which led to relatively low accuracy in both conditions. Importantly, modifications based on instance-level feedback made the target graph representations more closely approximate actual neighbors in the latent space, thereby retrieving references based on the fine-tuned version more beneficial. This study underscored the effectiveness of instance-level feedback in certain situations, particularly in improving alignment between humans and models.

\section{Discussions of Fairness and Ethical Issues} \label{sec:hm_discussion}
%The proposed method holds potential in various applications, but it is crucial to acknowledge the fairness and ethical concerns. Firstly, the prototype selection process, driven by domain experts, can introduce biases, potentially leading to systematic errors. There is also a risk of malicious actors manipulating the system to guide users toward incorrect or harmful decisions by introducing specific prototypes or triplet samples. Second, feedback loops can exacerbate biases, mainly if the system continually receives feedback from a particular viewpoint. Third, reference policies are unintelligible to some extent since graph representations can be different from the original data. Moreover, if certain groups are underrepresented in the training data or among selected prototypes, the system may perform poorly for those groups, resulting in potentially discriminatory outcomes. Lastly, while experts are responsible for addressing ethical concerns, they are also susceptible to their own biases and errors, necessitating ongoing vigilance and mitigation measures.

The interactive approach between humans and AI proposed in this work offers promise for diverse applications, yet it is essential to consider the issues of fairness and ethics it raises. Initially, the process of selecting prototypes, orchestrated by domain experts, may inadvertently introduce biases, potentially causing systematic errors. Additionally, malevolent entities can exploit the system, steering users towards detrimental or erroneous choices by inserting specific prototypes and references. Furthermore, feedback loops could amplify biases, especially when the system persistently receives input from a singular perspective. Compounding this issue is the obscurity of reference policies since representations possibly diverge from original graph structures. Moreover, if certain groups are underrepresented in either training data or chosen representative samples, the system's performance could degrade for these groups, leading to possibly discriminatory outcomes. Finally, although experts bear the responsibility for addressing these ethical issues, they are also prone to their own biases and mistakes, requiring continuous attentiveness and preventative strategies.

\section{Conclusion} \label{sec:conclusion}
%In conclusion, this research has yielded a promising method for bridging the gap between powerful yet opaque graph representation learning models and human decision-making. An iterative incorporation of human expertise improved the interpretability and reliability of graph classification models. Experiments and user studies have shown the effectiveness of this approach, demonstrating the potential for broader adoption in real-world applications where transparency and collaboration between humans and AI systems are crucial.

In summary, this study introduces an effective approach for integrating the accurate yet less transparent representation of learning models with human decision-making processes. The iterative engagement of human insights has enhanced the transparency and interpretability of graph classification models. Both experiments and user studies validate the efficacy of this method, underscoring its potential for widespread implementation in scenarios where transparent and collaborative interactions between humans and AI are essential.

%Future research endeavors should explore more advanced techniques for combining human feedback with representation learning, possibly utilizing reinforcement learning strategies. Additionally, extending this method to handle more extensive and complex graph datasets will address scalability challenges. Lastly, opportunities exist for more effectively integrating domain-specific knowledge and enhancing the adaptability of this method across various application domains. 

Future research should investigate advanced methods for integrating human inputs with representation learning, potentially through the application of reinforcement learning techniques. Furthermore, expanding the proposed method to accommodate larger and more intricate graph datasets will tackle scalability issues. Finally, there are promising prospects for more seamlessly incorporating domain-specific knowledge and increasing the versatility of this methodology across different fields.
	\chapter{Future Work} \label{chap:llm}

\section{Novel Combinations of Methods}

\begin{figure}[ht]
	\centering
	\includegraphics[width=0.9\linewidth]{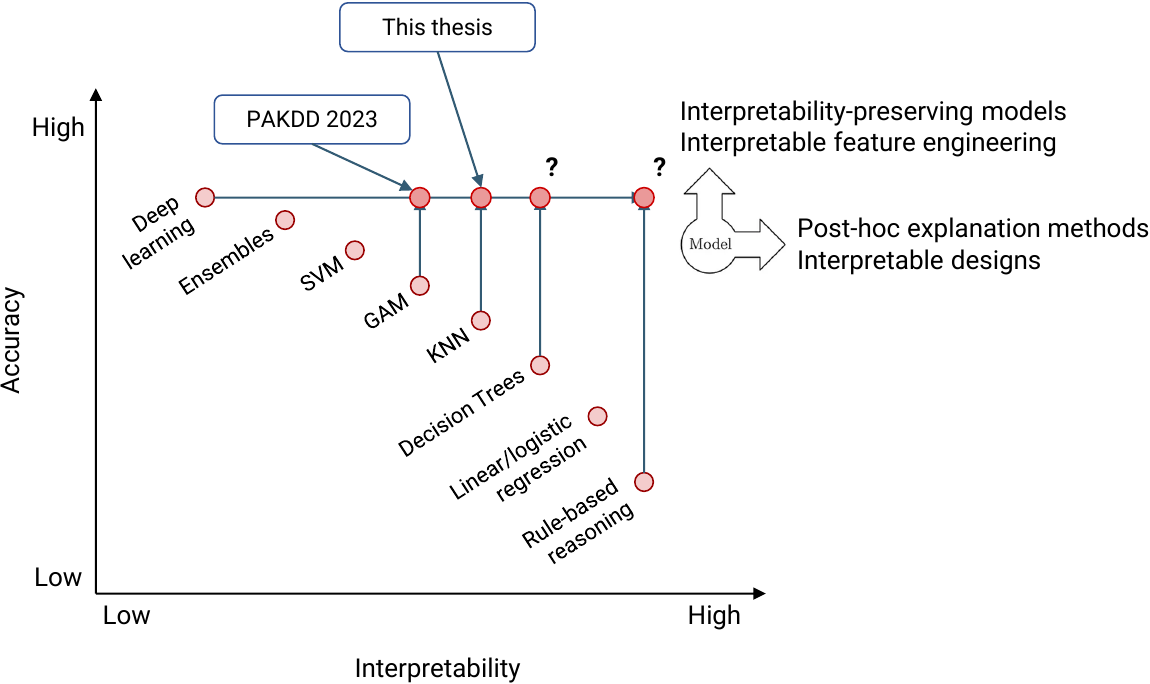}
	\caption{Novel Combinations of Methods for Enhancing GNN Interpretability. This figure refers to \cite{nesvijevskaia2021accuracy}.}
	\label{fig:lm_novel_combine}
\end{figure}

This thesis has been limited to the combination of deep graph representation learning and KNN. As presented in \cref{fig:lm_novel_combine}, the findings herein suggest several potential extensions. Future research can explore integrating representation learning with rule-based methods or decision trees to create more interpretable and accurate models. However, the non-differentiable nature and potential scalability issues of rule-based and tree-based methods pose formidable challenges. Further research into addressing these challenges is warranted and may yield significant advances in the field.

\section{Applications of Interpretable GNN Frameworks}
The proposed interpretable GNN frameworks in this thesis can give rise to several real-world applications in various domains, including but not limited to recommendation, finance, and e-commerce, where interpretability and fairness are significant concerns. This section discusses a few of these applications.
\subsection{Dynamic Interpretable Graph-based Recommendation Systems}
The proposed frameworks have the potential for extension to dynamic graph settings, particularly within recommendation systems \cite{wu2020graph, gao2023survey}. As user satisfaction is paramount in these systems, interpretable GNNs are crucial for enhancing user understanding of recommendations. By integrating interpretable frameworks, systems can elucidate their reasoning behind suggestions over time, increasing credibility and fostering user adoption. Moreover, understanding the rationale behind recommendations empowers businesses to refine strategies based on model insights. Additionally, interpretable GNN frameworks facilitate bias detection and model debugging. Analyzing explanations can expose biases embedded within training data or reveal the causes of inappropriate recommendations.

\subsection{Hybrid Human-GNN Decision Support Systems for E-commerce}
The proposed frameworks in this thesis have the potential to revolutionize decision support systems in e-commerce \cite{gao2023survey}. By enhancing model interpretability, businesses can understand the complex relationships driving product recommendations, user behavior, and market trends. Explainable GNNs could provide clear justifications for why certain products are suggested to customers, facilitating user trust and leading to increased conversions. Moreover, the HITL approach allows experts to fine-tune GNN models with their domain knowledge, ensuring recommendations and predictions align with business strategies and goals. The integration of LLMs could provide even more nuanced explanations tailored to specific customer inquiries, enhancing the overall shopping experience. Ultimately, these advancements promise more transparent, trustworthy, and effective decision-making in the dynamic world of e-commerce.

\subsection{Fairness-aware Financial Systems}
The proposed approaches, with their emphasis on interpretability and HITL processes, have significant potential for enhancing fairness within financial systems \cite{wang2021review}. In areas like credit scoring and loan approvals, understanding the rationale behind a GNN model's decisions is paramount to avoid discriminatory biases. The thesis's focus on explainability, particularly through structural analysis and human feedback, can help identify and mitigate potential biases within GNN models. Moreover, the integration of LLMs could streamline the process of clearly explaining model decisions to end-users, fostering trust and transparency within financial decision-making systems.

\subsection{Anomaly Detection in Fraudulent Activities}
The proposed frameworks offer versatile and promising solutions for anomaly detection within the realm of fraudulent activities \cite{dou2020enhancing, lu2022bright, motie2023financial}. GNNs are particularly well-suited to analyze the complex relationships and interconnectedness often present in financial transaction data. By enhancing the interpretability of GNN models, investigators can gain valuable insights into the factors driving a model's fraud classification.  XAI and HITL methodologies allow experts to provide domain knowledge, refining detection capabilities and reducing false positives and false negatives. Incorporating LLMs could generate user-friendly summaries of suspicious activity patterns in plain language, aiding in swift investigation and remediation. 
This comprehensive approach can improve the efficiency and accuracy of fraud detection processes significantly.

\section{Complex Reasoning with GNN-Empowered LLMs}
\subsection{What are LLMs?}
LLMs \cite{zhao2023survey} have emerged as a transformative force in the field of artificial intelligence, symbolizing a paradigm shift in the way machines understand and generate human language. These models are usually trained via unsupervised paradigms on an enormous amount of data followed by fine-tuning processes, enabling them to capture the nuances and complexities of natural language. This proficiency has far-reaching implications, as LLMs are not only redefining human-computer interactions but also offering unprecedented opportunities and challenges across diverse fields like linguistics, ethics, and information technology. The evolution of LLMs, marked by their growing sophistication and applicability, raises compelling questions about their future role in society, the ethical considerations they entail, and the balance between their benefits and potential risks. 

\subsection{Integration of LLMs and GNNs}
Given the remarkable capabilities of LLMs, there has been increasing interest in applying them to graph-related problems. The integration of LLMs with GNNs \cite{li2023survey} exhibits two distinct trends, influenced by the emergence of generative pre-trained models like ChatGPT \cite{ChatGPT}. Pre-ChatGPT, LLM architectures (such as Transformers \cite{vaswani2017attention} and BERT \cite{devlin2018bert}) were primarily employed to develop expressive graph encoders, enabling their use in multi-modal applications. ChatGPT's generalized abilities, demonstrated through its success in diverse AI tasks via chat interactions, have stimulated novel graph learning frameworks. However, as LLMs are fundamentally trained on sequential text data, directly applying them to complex graph structures is challenging. Two main strategies address this issue: Graph2Text and GNN-enhancement. The Graph2Text approach converts graphs into textual representations (e.g., graph description language, adjacency lists, edge lists, or domain-specific formats like SMILES \cite{weininger1988smiles}). While simple and interpretable, this method may encounter token limitations and suboptimal performance when handling complex graph structures. GNN enhancement holds promise by enabling LLMs to comprehend graph structures through the expressive power of GNNs.

\subsection{Strategies for Applying LLMs to Graph Data}
This section elaborates on strategies for utilizing LLMs on top of graphs. These strategies can be categorized into four groups: Hard prompt tuning, soft prompt tuning, instruction fine-tuning, and LLM as a controller. Each strategy is appropriate for different scenarios and has specific drawbacks and advantages.

\begin{figure}[ht]
	\centering
	\subfloat[Hard Prompting]{
		\includegraphics[width=0.8\linewidth]{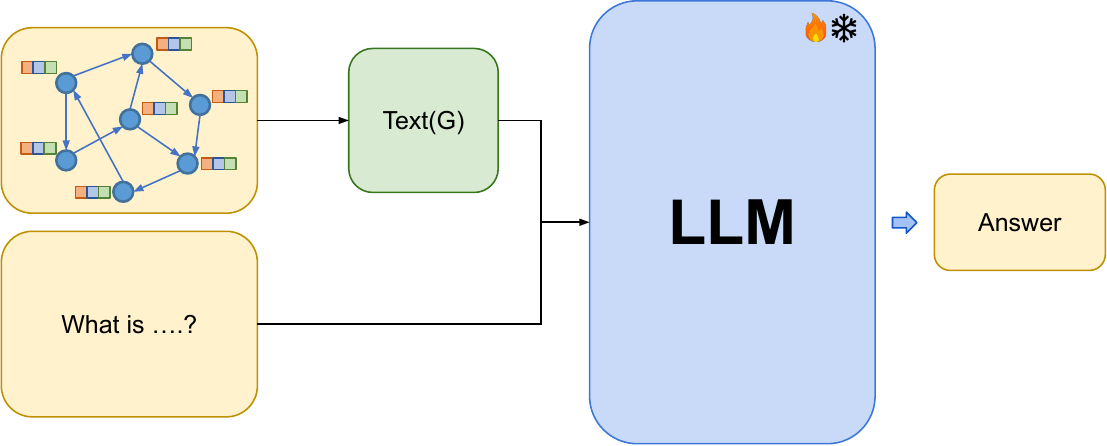}
		\label{fig:lm_hard_prompting}
	}
	\hfil
	\subfloat[Soft Prompting]{
		\includegraphics[width=0.8\linewidth]{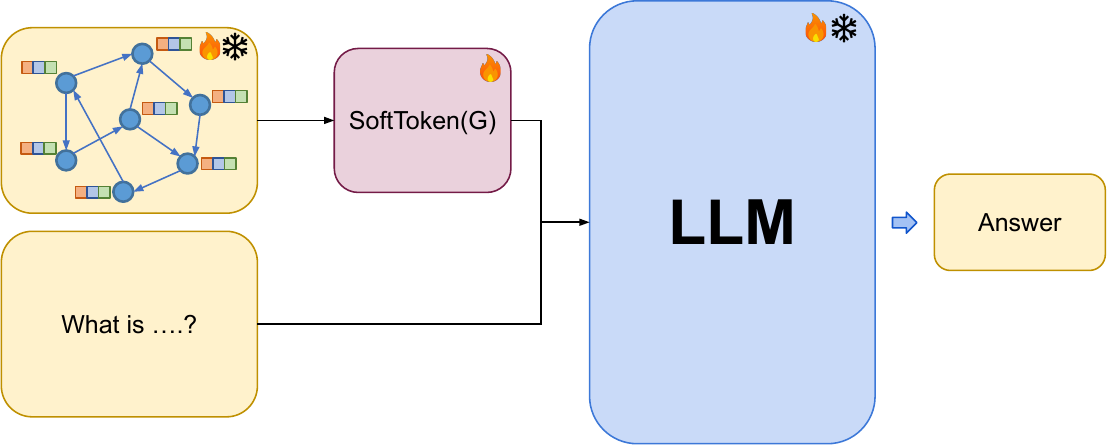}
		\label{fig:lm_soft_prompting}
	}
	\caption{Soft Prompting vs. Hard Prompting Techniques. The fire and snow symbols represent trainable and frozen components during the fine-tuning processes. This figure is referred to \cite{perozzi2024let}.}
\end{figure}

\noindent\textbf{Hard Prompt Tuning.}
Recent achievements have demonstrated significant interest in leveraging ``hard prompts'' to enhance the performance of LLMs \cite{jang2025prompt}, particularly for graph-based tasks. Hard prompts consist of explicit textual instructions that guide the reasoning process of LLMs. Studies like \cite{fatemi2023talk, guo2023gpt4graph} indicate their effectiveness, particularly in fundamental graph tasks such as node/edge/triangle counting, cycle detection, and other basic reasoning tasks. These prompts incorporate structural information from the graph, offering crucial context and constraints for the LLMs. Current research focuses on designing hard prompt formats and effective graph2text conversion methods. This approach still has much room for improvement, especially in complex reasoning problems. Methods like \cite{wei2022chain, zhao2024retrieval, gao2023retrieval} can be applied to enhance the prompt quality.

\noindent\textbf{Soft Prompt Tuning.} 
Soft prompt tuning has emerged as a powerful technique to enhance the performance of LLMs on graph-related tasks. Soft prompts are optimized based on LLMs' outputs for specific tasks instead of handcrafting textual input like hard prompt tuning. This flexibility allows for greater adaptability and the potential to capture subtle nuances within the graph data. Literature in this domain focuses on methods for encoding graph structures into soft prompts via GNNs. Studies like \cite{perozzi2024let, chen2024llaga} explore how soft prompt tuning enables LLMs to effectively perform fundamental graph-based tasks similar to hard prompt tuning. One promising direction is developing specialized soft prompt tuning architectures tailored for graph data, potentially leading to even more significant performance gains. 

\noindent\textbf{Instruction Fine-tuning.} 
Instruction fine-tuning \cite{zhang2025survey, le2024auto} is also an effective approach to improving the performance of LLMs by aligning LLM reasoning with provided context from carefully designed instructions. In graph research, Graph2Text instruction, where structural information from graphs is translated into natural language instructions that the LLM can readily understand, is a common approach. Additionally, the integration of soft prompt tuning within the instruction fine-tuning paradigm adds further flexibility. Soft prompts act as learnable parameters that can be fine-tuned in conjunction with the instructions themselves.  Studies like \cite{wang2024instructgraph, wang2024can, tang2023graphgpt, ye2024language} investigate the optimal design of Graph2Text instruction, effective strategies for combining them with soft prompt tuning, and how these techniques can improve LLM performance on diverse graph-related tasks.

\noindent\textbf{LLM as a Controller.} LLMs can benefit from leveraging GNNs and other graph analytics algorithms to extend their reasoning capabilities for intricate graph-based tasks. By utilizing these techniques, LLMs can be empowered to perform intermediate reasoning steps. LLM-GNN integration holds immense potential as an LLM can act as a powerful controller due to their ability to process and generate step-by-step actions given context information. For instance, Zhang et al. \cite{zhang2023graph} proposed a novel approach to enhancing LLM reasoning capability on graph data by learning to use external toolkits.

\subsection{A Real-world Scenario of LLM-GNN Integration}
Traditional graph databases and query languages, such as Neo4J \cite{neo4j}, excel at handling structured graph data. However, they encounter limitations when faced with complex natural language queries that require a nuanced understanding of graph structures and diverse node attributes.  Recent advancements in LLMs \cite{zhao2023survey} have demonstrated their exceptional capabilities in understanding and responding to natural language. Yet, LLMs typically lack an inherent understanding of graph structures \cite{fatemi2023talk}, a crucial element for addressing complex graph-related queries.

A potential approach to this problem is leveraging GNNs to empower LLMs with graph-structural awareness \cite{li2023survey, perozzi2024let, chen2024llaga}. GNNs are designed to learn representations that capture the inherent relationships and rich attributes within graph data. By integrating the understanding of graph structure encoded by GNNs with the natural language processing prowess of LLMs, the proposed method aims to enable flexible and accurate querying of complex graphs. The GNN-LLM integration has potential applications in diverse domains. For example, in scientific literature analysis, the proposed approach could facilitate identifying influential papers, potential collaborators, and emerging research trends within intricate citation networks and collaboration graphs. Similarly, when applied to large-scale knowledge graphs, the proposed method could support sophisticated entity search, fact verification, and complex question answering.
	\chapter{Conclusion} \label{chap:conclusion}
This thesis has presented a novel XAI framework specifically designed to enhance the interpretability of GNNs through structural and conceptual analyses and extensions. The proposed framework addressed the shortcomings of existing XAI methods with data having graph structures. It also successfully addresses the limitations of both post-hoc GNN explanation methods and intrinsically interpretable GNN models. It offers adaptability and computational efficiency and moves beyond basic feature analysis to provide insights into how graph structure influences GNN predictions.  Additionally, the framework provides accurate predictions alongside compact, user-centric explanations by leveraging the interpretability of KNN enhanced by a concept discovery module. The incorporation of domain knowledge further aligns GNN representations with human understanding, fostering trust and responsible application in high-stakes domains. Comprehensive experiments demonstrate the framework's effectiveness and efficiency. Future work will focus on enhancing the framework's interpretability through innovative combinations of methods, exploring real-world applications, and investigating potential integration with cutting-edge technologies like large language models. These extensions will ultimately promote the responsible and beneficial use of GNNs across a wide range of fields.
	
	%%%%%%%%%%%%%%%%%%%%%%%%%%%%%%%%%%%%%%%%%%%%%%%%%%%%%%%%%%%%%%%%%%%%
	%% END OF MAIN MATTER
	%%%%%%%%%%%%%%%%%%%%%%%%%%%%%%%%%%%%%%%%%%%%%%%%%%%%%%%%%%%%%%%%%%%%
	
	%%%%%%%%%%%%%%%%%%%%%%%%%%%%%%%%%%%%%%%%%%%%%%%%%%%%%%%%%%%%%%%%%%%%
	%% BACK MATTER
	%%%%%%%%%%%%%%%%%%%%%%%%%%%%%%%%%%%%%%%%%%%%%%%%%%%%%%%%%%%%%%%%%%%%
	
	%% BIBLIOGRAPHY
	\clearpage
	\begingroup
	\setstretch{1.1} % remove to make all references double spacing 
	%\nohyphenation % turn on to prevent hyphenation in a bibliography
	% (remember to remove \setlength\bibitemsep in the preamble if you choose to use double spacing)
	\phantomsection\clearpage\addcontentsline{toc}{chapter}{Bibliography}	
	\printbibliography

@article{bui2023generating,
  title={Generating Real-time Explanations for GNNs via Multiple Specialty Learners and Online Knowledge Distillation},
  author={Bui, Tien-Cuong and Le, Van-Duc and Li, Wen-Syan},
  journal={IEEE Access},
  year={2023},
  publisher={IEEE}
}

@article{zhang2020deep,
  title={Deep learning on graphs: A survey},
  author={Zhang, Ziwei and Cui, Peng and Zhu, Wenwu},
  journal={IEEE Transactions on Knowledge and Data Engineering},
  year={2020},
  publisher={IEEE}
}

@misc{dgldata,
  publtype={informal},
  author={Minjie Wang and Lingfan Yu and Da Zheng and Quan Gan and Yu Gai and Zihao Ye and Mufei Li and Jinjing Zhou and Qi Huang and Chao Ma and Ziyue Huang and Qipeng Guo and Hao Zhang and Haibin Lin and Junbo Zhao and Jinyang Li and Alexander J. Smola and Zheng Zhang},
  title={Deep Graph Library: Towards Efficient and Scalable Deep Learning on Graphs.},
  year={2019},
  cdate={1546300800000},
  journal={CoRR},
  volume={abs/1909.01315}
}

@inproceedings{zhang2022protgnn,
  title={Protgnn: Towards self-explaining graph neural networks},
  author={Zhang, Zaixi and Liu, Qi and Wang, Hao and Lu, Chengqiang and Lee, Cheekong},
  booktitle={Proceedings of the AAAI Conference on Artificial Intelligence},
  volume={36},
  pages={9127--9135},
  year={2022}
}

@article{wang2020unifying,
  title={Unifying graph convolutional neural networks and label propagation},
  author={Wang, Hongwei and Leskovec, Jure},
  journal={arXiv preprint arXiv:2002.06755},
  year={2020}
}

@inproceedings{dai2021towards,
  title={Towards self-explainable graph neural network},
  author={Dai, Enyan and Wang, Suhang},
  booktitle={Proceedings of the 30th ACM International Conference on Information \& Knowledge Management},
  pages={302--311},
  year={2021}
}

@article{gou2021knowledge,
  title={Knowledge distillation: A survey},
  author={Gou, Jianping and Yu, Baosheng and Maybank, Stephen J and Tao, Dacheng},
  journal={International Journal of Computer Vision},
  volume={129},
  number={6},
  pages={1789--1819},
  year={2021},
  publisher={Springer}
}

@misc{deng2021graph,
  title={Graph-free knowledge distillation for graph neural networks},
  author={Deng, Xiang and Zhang, Zhongfei},
  journal={arXiv preprint arXiv:2105.07519},
  year={2021}
}

@misc{joshi2021representation,
  title={On Representation Knowledge Distillation for Graph Neural Networks},
  author={Joshi, Chaitanya K and Liu, Fayao and Xun, Xu and Lin, Jie and Foo, Chuan-Sheng},
  journal={arXiv preprint arXiv:2111.04964},
  year={2021}
}

@article{li2022egnn,
  title={EGNN: Constructing explainable graph neural networks via knowledge distillation},
  author={Li, Yuan and Liu, Li and Wang, Guoyin and Du, Yong and Chen, Penggang},
  journal={Knowledge-Based Systems},
  pages={108345},
  year={2022},
  volume={241},
  issn = {0950-7051},
  publisher={Elsevier}
}

@article{tsang2020does,
  title={How does this interaction affect me? interpretable attribution for feature interactions},
  author={Tsang, Michael and Rambhatla, Sirisha and Liu, Yan},
  journal={Advances in neural information processing systems},
  volume={33},
  pages={6147--6159},
  year={2020}
}

@misc{ribeiro2016should,
      title={"Why Should I Trust You?": Explaining the Predictions of Any Classifier}, 
      author={Marco Tulio Ribeiro and Sameer Singh and Carlos Guestrin},
      year={2016},
      eprint={1602.04938},
      archivePrefix={arXiv},
      primaryClass={cs.LG}
}

@inproceedings{lundberg2017unified,
 author = {Lundberg, Scott M and Lee, Su-In},
 booktitle = {Advances in Neural Information Processing Systems},
 journal={Advances in neural information processing systems},
 publisher = {Curran Associates, Inc.},
 title = {A Unified Approach to Interpreting Model Predictions},
 volume = {30},
 year = {2017}
}

@article{nori2019interpretml,
  title={InterpretML: A Unified Framework for Machine Learning Interpretability},
  author={Nori, Harsha and Jenkins, Samuel and Koch, Paul and Caruana, Rich},
  journal={arXiv preprint arXiv:1909.09223},
  year={2019}
}

@inproceedings{ying2019gnnexplainer,
 author = {Ying, Zhitao and Bourgeois, Dylan and You, Jiaxuan and Zitnik, Marinka and Leskovec, Jure},
 booktitle = {Advances in Neural Information Processing Systems},
 publisher = {Curran Associates, Inc.},
 title = {GNNExplainer: Generating Explanations for Graph Neural Networks},
 volume = {32},
 year = {2019}
}

@article{luo2020parameterized,
  title={Parameterized explainer for graph neural network},
  author={Luo, Dongsheng and Cheng, Wei and Xu, Dongkuan and Yu, Wenchao and Zong, Bo and Chen, Haifeng and Zhang, Xiang},
  journal={Advances in neural information processing systems},
  volume={33},
  pages={19620--19631},
  year={2020}
}

@misc{schlichtkrull2020interpreting,
  title={Interpreting Graph Neural Networks for NLP With Differentiable Edge Masking},
  author={Sejr Schlichtkrull, Michael and De Cao, Nicola and Titov, Ivan},
  journal={arXiv e-prints},
  year={2020}
}

@inproceedings{yuan2021explainability,
  title={On explainability of graph neural networks via subgraph explorations},
  author={Yuan, Hao and Yu, Haiyang and Wang, Jie and Li, Kang and Ji, Shuiwang},
  booktitle={International Conference on Machine Learning},
  pages={12241--12252},
  year={2021},
  organization={PMLR}
}

@inproceedings{alharbi2021learning,
  title={Learning Interpretation with Explainable Knowledge Distillation},
  author={Alharbi, Raed and Vu, Minh N and Thai, My T},
  booktitle={2021 IEEE International Conference on Big Data (Big Data)},
  pages={705--714},
  year={2021},
  organization={IEEE}
}

@misc{hinton2015distilling,
  title={Distilling the knowledge in a neural network},
  author={Hinton, Geoffrey and Vinyals, Oriol and Dean, Jeff and others},
  journal={arXiv preprint arXiv:1503.02531},
  volume={2},
  number={7},
  year={2015}
}

@inproceedings{dou2020enhancing,
  title={Enhancing graph neural network-based fraud detectors against camouflaged fraudsters},
  author={Dou, Yingtong and Liu, Zhiwei and Sun, Li and Deng, Yutong and Peng, Hao and Yu, Philip S},
  booktitle={Proceedings of the 29th ACM International Conference on Information \& Knowledge Management},
  pages={315--324},
  year={2020}
}

@article{baldassarre2019explainability,
  title={Explainability techniques for graph convolutional networks},
  author={Baldassarre, Federico and Azizpour, Hossein},
  journal={arXiv preprint arXiv:1905.13686},
  year={2019}
}

@article{schnake2020higher,
  title={Higher-order explanations of graph neural networks via relevant walks},
  author={Schnake, Thomas and Eberle, Oliver and Lederer, Jonas and Nakajima, Shinichi and Sch{\"u}tt, Kristof T and M{\"u}ller, Klaus-Robert and Montavon, Gr{\'e}goire},
  journal={arXiv preprint arXiv:2006.03589},
  year={2020}
}

@article{velickovic2017graph,
  title={Graph attention networks},
  author={Velickovic, Petar and Cucurull, Guillem and Casanova, Arantxa and Romero, Adriana and Lio, Pietro and Bengio, Yoshua},
  journal={stat},
  volume={1050},
  pages={20},
  year={2017}
}

@article{kipf2016semi,
  title={Semi-supervised classification with graph convolutional networks},
  author={Kipf, Thomas N and Welling, Max},
  journal={arXiv preprint arXiv:1609.02907},
  year={2016}
}

@inproceedings{zhang2020gcn,
  title={Gcn-based user representation learning for unifying robust recommendation and fraudster detection},
  author={Zhang, Shijie and Yin, Hongzhi and Chen, Tong and Hung, Quoc Viet Nguyen and Huang, Zi and Cui, Lizhen},
  booktitle={Proceedings of the 43rd international ACM SIGIR conference on research and development in information retrieval},
  pages={689--698},
  year={2020},
  publisher={Association for Computing Machinery},
  address = {New York, NY, USA}
}

@inproceedings{rayana2015collective,
author = {Rayana, Shebuti and Akoglu, Leman},
title = {Collective Opinion Spam Detection: Bridging Review Networks and Metadata},
year = {2015},
isbn = {9781450336642},
publisher = {Association for Computing Machinery},
address = {New York, NY, USA},
doi = {10.1145/2783258.2783370},
booktitle = {Proceedings of the 21th ACM SIGKDD International Conference on Knowledge Discovery and Data Mining},
pages = {985–994},
numpages = {10},
location = {Sydney, NSW, Australia},
series = {KDD '15}
}

@article{lpa_gcn,
  title={Combining graph convolutional neural networks and label propagation},
  author={Wang, Hongwei and Leskovec, Jure},
  journal={ACM Transactions on Information Systems (TOIS)},
  volume={40},
  number={4},
  pages={1--27},
  year={2021},
  publisher={ACM New York, NY}
}

@inproceedings{shrikumar2017learning,
  title={Learning important features through propagating activation differences},
  author={Shrikumar, Avanti and Greenside, Peyton and Kundaje, Anshul},
  booktitle={International conference on machine learning},
  pages={3145--3153},
  year={2017},
  organization={PMLR}
}

@inproceedings{tong2006fast,
  title={Fast random walk with restart and its applications},
  author={Tong, Hanghang and Faloutsos, Christos and Pan, Jia-Yu},
  booktitle={Sixth international conference on data mining (ICDM'06)},
  pages={613--622},
  year={2006},
  organization={IEEE}
}

@article{sun2022human,
  title={Human action recognition from various data modalities: A review},
  author={Sun, Zehua and Ke, Qiuhong and Rahmani, Hossein and Bennamoun, Mohammed and Wang, Gang and Liu, Jun},
  journal={IEEE transactions on pattern analysis and machine intelligence},
  year={2022},
  publisher={IEEE}
}

@article{wu2020graph,
  title={Graph neural networks in recommender systems: a survey},
  author={Wu, Shiwen and Sun, Fei and Zhang, Wentao and Xie, Xu and Cui, Bin},
  journal={ACM Computing Surveys (CSUR)},
  year={2020},
  publisher={ACM New York, NY}
}

@inproceedings{fan2019graph,
  title={Graph neural networks for social recommendation},
  author={Fan, Wenqi and Ma, Yao and Li, Qing and He, Yuan and Zhao, Eric and Tang, Jiliang and Yin, Dawei},
  booktitle={The world wide web conference},
  pages={417--426},
  year={2019}
}

@article{zhang2021graph,
  title={Graph neural networks and their current applications in bioinformatics},
  author={Zhang, Xiao-Meng and Liang, Li and Liu, Lin and Tang, Ming-Jing},
  journal={Frontiers in genetics},
  volume={12},
  year={2021},
  publisher={Frontiers Media SA}
}

@article{zhou2020graph,
  title={Graph neural networks: A review of methods and applications},
  author={Zhou, Jie and Cui, Ganqu and Hu, Shengding and Zhang, Zhengyan and Yang, Cheng and Liu, Zhiyuan and Wang, Lifeng and Li, Changcheng and Sun, Maosong},
  journal={AI Open},
  volume={1},
  pages={57--81},
  year={2020},
  publisher={Elsevier}
}

@inproceedings{yang2021extract,
  title={Extract the knowledge of graph neural networks and go beyond it: An effective knowledge distillation framework},
  author={Yang, Cheng and Liu, Jiawei and Shi, Chuan},
  booktitle={Proceedings of the Web Conference 2021},
  pages={1227--1237},
  year={2021}
}

@inproceedings{wang2020personalized,
  title={Personalized pagerank to a target node, revisited},
  author={Wang, Hanzhi and Wei, Zhewei and Gan, Junhao and Wang, Sibo and Huang, Zengfeng},
  booktitle={Proceedings of the 26th ACM SIGKDD International Conference on Knowledge Discovery \& Data Mining},
  pages={657--667},
  year={2020}
}

@inproceedings{park2017comparative,
  title={A comparative study of matrix factorization and random walk with restart in recommender systems},
  author={Park, Haekyu and Jung, Jinhong and Kang, U},
  booktitle={2017 IEEE International Conference on Big Data (Big Data)},
  pages={756--765},
  year={2017},
  organization={IEEE}
}

@article{chiang2013exploring,
  title={Exploring heterogeneous information networks and random walk with restart for academic search},
  author={Chiang, Meng-Fen and Liou, Jiun-Jiue and Wang, Jen-Liang and Peng, Wen-Chih and Shan, Man-Kwan},
  journal={Knowledge and information systems},
  volume={36},
  number={1},
  pages={59--82},
  year={2013},
  publisher={Springer}
}

@article{brin1998anatomy,
  title={The anatomy of a large-scale hypertextual web search engine},
  author={Brin, Sergey and Page, Lawrence},
  journal={Computer networks and ISDN systems},
  volume={30},
  number={1-7},
  pages={107--117},
  year={1998},
  publisher={Elsevier}
}

@article{doshi2017towards,
  title={Towards a rigorous science of interpretable machine learning},
  author={Doshi-Velez, Finale and Kim, Been},
  journal={arXiv preprint arXiv:1702.08608},
  year={2017}
}

@inproceedings{ioffe2015batch,
  title={Batch normalization: Accelerating deep network training by reducing internal covariate shift},
  author={Ioffe, Sergey and Szegedy, Christian},
  booktitle={International conference on machine learning},
  pages={448--456},
  year={2015},
  organization={PMLR}
}

@misc{shapDeep19:online,
author = {},
title = {SHAP DeepExplainer Documentation},
howpublished = {\url{https://shap-lrjball.readthedocs.io/en/latest/generated/shap.DeepExplainer.html}},
month = {},
year = {},
note = {(Accessed on 10/08/2022)}
}

@misc{shap_doc,
author = {},
title = {SHAP Documentation},
howpublished = {\url{https://shap-lrjball.readthedocs.io/}},
month = {},
year = {},
note = {(Accessed on 10/08/2022)}
}

@misc{Cytoscape,
author = {},
title = {Cytoscape: An Open Source Software Platform For Visualizing Complex Networks},
howpublished = {\url{https://js.cytoscape.org/}},
month = {},
year = {},
note = {(Accessed on 10/14/2022)}
}

@article{johnson2019billion,
  title={Billion-scale similarity search with {GPUs}},
  author={Johnson, Jeff and Douze, Matthijs and J{\'e}gou, Herv{\'e}},
  journal={IEEE Transactions on Big Data},
  volume={7},
  number={3},
  pages={535--547},
  year={2019},
  publisher={IEEE}
}

@article{jeyakumar2020can,
  title={How can i explain this to you? an empirical study of deep neural network explanation methods},
  author={Jeyakumar, Jeya Vikranth and Noor, Joseph and Cheng, Yu-Hsi and Garcia, Luis and Srivastava, Mani},
  journal={Advances in Neural Information Processing Systems},
  volume={33},
  pages={4211--4222},
  year={2020}
}

@inproceedings{cho2019efficacy,
  title={On the efficacy of knowledge distillation},
  author={Cho, Jang Hyun and Hariharan, Bharath},
  booktitle={Proceedings of the IEEE/CVF international conference on computer vision},
  pages={4794--4802},
  year={2019}
}

@article{shen2021label,
  title={Is label smoothing truly incompatible with knowledge distillation: An empirical study},
  author={Shen, Zhiqiang and Liu, Zechun and Xu, Dejia and Chen, Zitian and Cheng, Kwang-Ting and Savvides, Marios},
  journal={arXiv preprint arXiv:2104.00676},
  year={2021}
}

@article{zhou2021rethinking,
  title={Rethinking soft labels for knowledge distillation: A bias-variance tradeoff perspective},
  author={Zhou, Helong and Song, Liangchen and Chen, Jiajie and Zhou, Ye and Wang, Guoli and Yuan, Junsong and Zhang, Qian},
  journal={arXiv preprint arXiv:2102.00650},
  year={2021}
}

@article{balestriero2022batch,
  title={Batch Normalization Explained},
  author={Balestriero, Randall and Baraniuk, Richard G},
  journal={arXiv preprint arXiv:2209.14778},
  year={2022}
}

@inproceedings{palowitch2022graphworld,
  title={Graphworld: Fake graphs bring real insights for gnns},
  author={Palowitch, John and Tsitsulin, Anton and Mayer, Brandon and Perozzi, Bryan},
  booktitle={Proceedings of the 28th ACM SIGKDD Conference on Knowledge Discovery and Data Mining},
  pages={3691--3701},
  year={2022}
}

@article{bonnici2013subgraph,
  title={A subgraph isomorphism algorithm and its application to biochemical data},
  author={Bonnici, Vincenzo and Giugno, Rosalba and Pulvirenti, Alfredo and Shasha, Dennis and Ferro, Alfredo},
  journal={BMC bioinformatics},
  volume={14},
  number={7},
  pages={1--13},
  year={2013},
  publisher={BioMed Central}
}

@inproceedings{jun2017sequential,
  title={Sequential graph matching with sequential Monte Carlo},
  author={Jun, Seong-Hwan and Wong, Samuel WK and Zidek, James and Bouchard-C{\^o}t{\'e}, Alexandre},
  booktitle={Artificial Intelligence and Statistics},
  pages={1075--1084},
  year={2017},
  organization={PMLR}
}

@article{lou2020neural,
  title={Neural subgraph matching},
  author={Lou, Zhaoyu and You, Jiaxuan and Wen, Chengtao and Canedo, Arquimedes and Leskovec, Jure and others},
  journal={arXiv preprint arXiv:2007.03092},
  year={2020}
}

@inproceedings{roy2022interpretable,
  title={Interpretable Neural Subgraph Matching for Graph Retrieval},
  author={Roy, Indradyumna and Velugoti, Venkata Sai Baba Reddy and Chakrabarti, Soumen and De, Abir},
  booktitle={Proceedings of the AAAI Conference on Artificial Intelligence},
  volume={36},
  number={7},
  pages={8115--8123},
  year={2022}
}

@inproceedings{li2019graph,
  title={Graph matching networks for learning the similarity of graph structured objects},
  author={Li, Yujia and Gu, Chenjie and Dullien, Thomas and Vinyals, Oriol and Kohli, Pushmeet},
  booktitle={International conference on machine learning},
  year={2019},
  organization={PMLR}
}

@article{ahmed2017graphlet,
  title={Graphlet decomposition: Framework, algorithms, and applications},
  author={Ahmed, Nesreen K and Neville, Jennifer and Rossi, Ryan A and Duffield, Nick G and Willke, Theodore L},
  journal={Knowledge and Information Systems},
  volume={50},
  pages={689--722},
  year={2017},
  publisher={Springer}
}

@article{degen2008art,
  title={On the Art of Compiling and Using'Drug-Like'Chemical Fragment Spaces},
  author={Degen, J{\"o}rg and Wegscheid-Gerlach, Christof and Zaliani, Andrea and Rarey, Matthias},
  journal={ChemMedChem: Chemistry Enabling Drug Discovery},
  volume={3},
  number={10},
  pages={1503--1507},
  year={2008},
  publisher={Wiley Online Library}
}

@inproceedings{huang2021towards,
  title={Towards efficient motif-based graph partitioning: An adaptive sampling approach},
  author={Huang, Shixun and Li, Yuchen and Bao, Zhifeng and Li, Zhao},
  booktitle={2021 IEEE 37th International Conference on Data Engineering (ICDE)},
  pages={528--539},
  year={2021},
  organization={IEEE}
}

@book{Theodoridis_Koutroumbas_2009,
title={Pattern recognition, 4th Edition},
publisher={Academic Press}, author={Theodoridis, Sergios and Koutroumbas, Konstantinos},
year={2009}
}

@article{yu2020graph,
  title={Graph information bottleneck for subgraph recognition},
  author={Yu, Junchi and Xu, Tingyang and Rong, Yu and Bian, Yatao and Huang, Junzhou and He, Ran},
  journal={arXiv preprint arXiv:2010.05563},
  year={2020}
}

@article{fang2022densest,
  title={Densest subgraph discovery on large graphs: Applications, challenges, and techniques},
  author={Fang, Yixiang and Luo, Wensheng and Ma, Chenhao},
  journal={Proceedings of the VLDB Endowment},
  volume={15},
  number={12},
  pages={3766--3769},
  year={2022},
  publisher={VLDB Endowment}
}

@inproceedings{yan2002gspan,
  title={gspan: Graph-based substructure pattern mining},
  author={Yan, Xifeng and Han, Jiawei},
  booktitle={2002 IEEE International Conference on Data Mining, 2002. Proceedings.},
  pages={721--724},
  year={2002},
  organization={IEEE}
}

@article{vinyals2016matching,
  title={Matching networks for one shot learning},
  author={Vinyals, Oriol and Blundell, Charles and Lillicrap, Timothy and Wierstra, Daan and others},
  journal={Advances in neural information processing systems},
  volume={29},
  year={2016}
}

@article{ghorbani2019towards,
  title={Towards automatic concept-based explanations},
  author={Ghorbani, Amirata and Wexler, James and Zou, James Y and Kim, Been},
  journal={Advances in Neural Information Processing Systems},
  volume={32},
  year={2019}
}

@article{rudin2019stop,
  title={Stop explaining black box machine learning models for high stakes decisions and use interpretable models instead},
  author={Rudin, Cynthia},
  journal={Nature machine intelligence},
  volume={1},
  number={5},
  pages={206--215},
  year={2019},
  publisher={Nature Publishing Group UK London}
}

@article{hamilton2017inductive,
  title={Inductive representation learning on large graphs},
  author={Hamilton, Will and Ying, Zhitao and Leskovec, Jure},
  journal={Advances in neural information processing systems},
  volume={30},
  year={2017}
}

@article{xu2018powerful,
  title={How powerful are graph neural networks?},
  author={Xu, Keyulu and Hu, Weihua and Leskovec, Jure and Jegelka, Stefanie},
  journal={arXiv preprint arXiv:1810.00826},
  year={2018}
}

@article{rupp2012fast,
  title={Fast and accurate modeling of molecular atomization energies with machine learning},
  author={Rupp, Matthias and Tkatchenko, Alexandre and M{\"u}ller, Klaus-Robert and Von Lilienfeld, O Anatole},
  journal={Physical review letters},
  volume={108},
  number={5},
  pages={058301},
  year={2012},
  publisher={APS}
}

@inproceedings{rossi2015network,
  title={The network data repository with interactive graph analytics and visualization},
  author={Rossi, Ryan and Ahmed, Nesreen},
  booktitle={Proceedings of the AAAI conference on artificial intelligence},
  volume={29},
  number={1},
  year={2015}
}

@article{borgwardt2005protein,
  title={Protein function prediction via graph kernels},
  author={Borgwardt, Karsten M and Ong, Cheng Soon and Sch{\"o}nauer, Stefan and Vishwanathan, SVN and Smola, Alex J and Kriegel, Hans-Peter},
  journal={Bioinformatics},
  volume={21},
  number={suppl\_1},
  pages={i47--i56},
  year={2005},
  publisher={Oxford University Press}
}

@article{morris2020tudataset,
  title={Tudataset: A collection of benchmark datasets for learning with graphs},
  author={Morris, Christopher and Kriege, Nils M and Bause, Franka and Kersting, Kristian and Mutzel, Petra and Neumann, Marion},
  journal={arXiv preprint arXiv:2007.08663},
  year={2020}
}

@inproceedings{cai2019effects,
  title={The effects of example-based explanations in a machine learning interface},
  author={Cai, Carrie J and Jongejan, Jonas and Holbrook, Jess},
  booktitle={Proceedings of the 24th international conference on intelligent user interfaces},
  year={2019}
}

@inproceedings{li2018deep,
  title={Deep learning for case-based reasoning through prototypes: A neural network that explains its predictions},
  author={Li, Oscar and Liu, Hao and Chen, Chaofan and Rudin, Cynthia},
  booktitle={Proceedings of the AAAI Conference on Artificial Intelligence},
  volume={32},
  year={2018}
}

@article{chen2019looks,
  title={This looks like that: deep learning for interpretable image recognition},
  author={Chen, Chaofan and Li, Oscar and Tao, Daniel and Barnett, Alina and Rudin, Cynthia and Su, Jonathan K},
  journal={Advances in neural information processing systems},
  volume={32},
  year={2019}
}

@inproceedings{davoudi2021toward,
  title={Toward Faithful Case-based Reasoning through Learning Prototypes in a Nearest Neighbor-friendly Space.},
  author={Davoudi, Seyed Omid and Komeili, Majid},
  booktitle={International Conference on Learning Representations},
  year={2021}
}

@article{cuturi2013sinkhorn,
  title={Sinkhorn distances: Lightspeed computation of optimal transport},
  author={Cuturi, Marco},
  journal={Advances in neural information processing systems},
  volume={26},
  year={2013}
}

@article{li2021braingnn,
  title={Braingnn: Interpretable brain graph neural network for fmri analysis},
  author={Li, Xiaoxiao and Zhou, Yuan and Dvornek, Nicha and Zhang, Muhan and Gao, Siyuan and Zhuang, Juntang and Scheinost, Dustin and Staib, Lawrence H and Ventola, Pamela and Duncan, James S},
  journal={Medical Image Analysis},
  volume={74},
  pages={102233},
  year={2021},
  publisher={Elsevier}
}

@inproceedings{feng2022kergnns,
  title={Kergnns: Interpretable graph neural networks with graph kernels},
  author={Feng, Aosong and You, Chenyu and Wang, Shiqiang and Tassiulas, Leandros},
  booktitle={Proceedings of the AAAI Conference on Artificial Intelligence},
  volume={36},
  pages={6614--6622},
  year={2022}
}

@article{ragno2022prototype,
  title={Prototype-based Interpretable Graph Neural Networks},
  author={Ragno, Alessio and La Rosa, Biagio and Capobianco, Roberto},
  journal={IEEE Transactions on Artificial Intelligence},
  year={2022},
  publisher={IEEE}
}

@inproceedings{nikolentzos2017matching,
  title={Matching node embeddings for graph similarity},
  author={Nikolentzos, Giannis and Meladianos, Polykarpos and Vazirgiannis, Michalis},
  booktitle={Proceedings of the AAAI Conference on Artificial Intelligence},
  volume={31},
  number={1},
  year={2017}
}

@article{togninalli2019wasserstein,
  title={Wasserstein weisfeiler-lehman graph kernels},
  author={Togninalli, Matteo and Ghisu, Elisabetta and Llinares-L{\'o}pez, Felipe and Rieck, Bastian and Borgwardt, Karsten},
  journal={Advances in neural information processing systems},
  volume={32},
  year={2019}
}

@article{vincent2021semi,
  title={Semi-relaxed Gromov-Wasserstein divergence with applications on graphs},
  author={Vincent-Cuaz, C{\'e}dric and Flamary, R{\'e}mi and Corneli, Marco and Vayer, Titouan and Courty, Nicolas},
  journal={arXiv preprint arXiv:2110.02753},
  year={2021}
}

@article{vincent2022template,
  title={Template based graph neural network with optimal transport distances},
  author={Vincent-Cuaz, C{\'e}dric and Flamary, R{\'e}mi and Corneli, Marco and Vayer, Titouan and Courty, Nicolas},
  journal={Advances in Neural Information Processing Systems},
  volume={35},
  pages={11800--11814},
  year={2022}
}

@article{rubner2000earth,
  title={The earth mover's distance as a metric for image retrieval},
  author={Rubner, Yossi and Tomasi, Carlo and Guibas, Leonidas J},
  journal={International journal of computer vision},
  volume={40},
  pages={99--121},
  year={2000},
  publisher={Springer}
}

@article{ramos2020interactive,
  title={Interactive machine teaching: a human-centered approach to building machine-learned models},
  author={Ramos, Gonzalo and Meek, Christopher and Simard, Patrice and Suh, Jina and Ghorashi, Soroush},
  journal={Human--Computer Interaction},
  volume={35},
  number={5-6},
  pages={413--451},
  year={2020},
  publisher={Taylor \& Francis}
}

@article{mosqueira2023human,
  title={Human-in-the-loop machine learning: A state of the art},
  author={Mosqueira-Rey, Eduardo and Hern{\'a}ndez-Pereira, Elena and Alonso-R{\'\i}os, David and Bobes-Bascar{\'a}n, Jos{\'e} and Fern{\'a}ndez-Leal, {\'A}ngel},
  journal={Artificial Intelligence Review},
  volume={56},
  number={4},
  pages={3005--3054},
  year={2023},
  publisher={Springer}
}

@inproceedings{liu2022learning,
  title={Learning Human-Compatible Representations for Case-Based Decision Support},
  author={Liu, Han and Tian, Yizhou and Chen, Chacha and Feng, Shi and Chen, Yuxin and Tan, Chenhao},
  booktitle={The Eleventh International Conference on Learning Representations},
  year={2022}
}

@article{taesiri2022visual,
  title={Visual correspondence-based explanations improve AI robustness and human-AI team accuracy},
  author={Taesiri, Mohammad Reza and Nguyen, Giang and Nguyen, Anh},
  journal={Advances in Neural Information Processing Systems},
  volume={35},
  pages={34287--34301},
  year={2022}
}

@article{slade1991case,
  title={Case-based reasoning: A research paradigm},
  author={Slade, Stephen},
  journal={AI magazine},
  volume={12},
  number={1},
  pages={42--42},
  year={1991}
}

@article{xia2021graph,
  title={Graph learning: A survey},
  author={Xia, Feng and Sun, Ke and Yu, Shuo and Aziz, Abdul and Wan, Liangtian and Pan, Shirui and Liu, Huan},
  journal={IEEE Transactions on Artificial Intelligence},
  volume={2},
  number={2},
  pages={109--127},
  year={2021},
  publisher={IEEE}
}

@article{yuan2022explainability,
	title={Explainability in graph neural networks: A taxonomic survey},
	author={Yuan, Hao and Yu, Haiyang and Gui, Shurui and Ji, Shuiwang},
	journal={IEEE transactions on pattern analysis and machine intelligence},
	volume={45},
	number={5},
	pages={5782--5799},
	year={2022},
	publisher={IEEE}
}

@inproceedings{yuan2020xgnn,
  title={Xgnn: Towards model-level explanations of graph neural networks},
  author={Yuan, Hao and Tang, Jiliang and Hu, Xia and Ji, Shuiwang},
  booktitle={Proceedings of the 26th ACM SIGKDD International Conference on Knowledge Discovery \& Data Mining},
  pages={430--438},
  year={2020}
}

@article{vaswani2017attention,
  title={Attention is all you need},
  author={Vaswani, Ashish and Shazeer, Noam and Parmar, Niki and Uszkoreit, Jakob and Jones, Llion and Gomez, Aidan N and Kaiser, {\L}ukasz and Polosukhin, Illia},
  journal={Advances in neural information processing systems},
  volume={30},
  year={2017}
}

@misc{neo4j,
  title = {{Neo4j: the graph database}},
  howpublished = {\url{http://neo4j.com}},
  note = {Accessed: 2019-11-21}
}

@inproceedings{sankar2021graph,
  title={Graph neural networks for friend ranking in large-scale social platforms},
  author={Sankar, Aravind and Liu, Yozen and Yu, Jun and Shah, Neil},
  booktitle={Proceedings of the Web Conference 2021},
  pages={2535--2546},
  year={2021}
}

@article{devlin2018bert,
  title={Bert: Pre-training of deep bidirectional transformers for language understanding},
  author={Devlin, Jacob and Chang, Ming-Wei and Lee, Kenton and Toutanova, Kristina},
  journal={arXiv preprint arXiv:1810.04805},
  year={2018}
}

@inproceedings{lai2019human,
	title={On human predictions with explanations and predictions of machine learning models: A case study on deception detection},
	author={Lai, Vivian and Tan, Chenhao},
	booktitle={Proceedings of the conference on fairness, accountability, and transparency},
	pages={29--38},
	year={2019}
}

@article{fatemi2023talk,
	title={Talk like a graph: Encoding graphs for large language models},
	author={Fatemi, Bahare and Halcrow, Jonathan and Perozzi, Bryan},
	journal={arXiv preprint arXiv:2310.04560},
	year={2023}
}

@article{guo2023gpt4graph,
	title={Gpt4graph: Can large language models understand graph structured data? an empirical evaluation and benchmarking},
	author={Guo, Jiayan and Du, Lun and Liu, Hengyu},
	journal={arXiv preprint arXiv:2305.15066},
	year={2023}
}

@article{wei2022chain,
	title={Chain-of-thought prompting elicits reasoning in large language models},
	author={Wei, Jason and Wang, Xuezhi and Schuurmans, Dale and Bosma, Maarten and Xia, Fei and Chi, Ed and Le, Quoc V and Zhou, Denny and others},
	journal={Advances in neural information processing systems},
	volume={35},
	pages={24824--24837},
	year={2022}
}

@article{zhao2024retrieval,
	title={Retrieval-Augmented Generation for AI-Generated Content: A Survey},
	author={Zhao, Penghao and Zhang, Hailin and Yu, Qinhan and Wang, Zhengren and Geng, Yunteng and Fu, Fangcheng and Yang, Ling and Zhang, Wentao and Cui, Bin},
	journal={arXiv preprint arXiv:2402.19473},
	year={2024}
}

@article{gao2023retrieval,
	title={Retrieval-augmented generation for large language models: A survey},
	author={Gao, Yunfan and Xiong, Yun and Gao, Xinyu and Jia, Kangxiang and Pan, Jinliu and Bi, Yuxi and Dai, Yi and Sun, Jiawei and Wang, Haofen},
	journal={arXiv preprint arXiv:2312.10997},
	year={2023}
}

@article{perozzi2024let,
	title={Let Your Graph Do the Talking: Encoding Structured Data for LLMs},
	author={Perozzi, Bryan and Fatemi, Bahare and Zelle, Dustin and Tsitsulin, Anton and Kazemi, Mehran and Al-Rfou, Rami and Halcrow, Jonathan},
	journal={arXiv preprint arXiv:2402.05862},
	year={2024}
}

@article{chen2024llaga,
	title={LLaGA: Large Language and Graph Assistant},
	author={Chen, Runjin and Zhao, Tong and Jaiswal, Ajay and Shah, Neil and Wang, Zhangyang},
	journal={arXiv preprint arXiv:2402.08170},
	year={2024}
}

@article{wang2024instructgraph,
	title={InstructGraph: Boosting Large Language Models via Graph-centric Instruction Tuning and Preference Alignment},
	author={Wang, Jianing and Wu, Junda and Hou, Yupeng and Liu, Yao and Gao, Ming and McAuley, Julian},
	journal={arXiv preprint arXiv:2402.08785},
	year={2024}
}

@article{wang2024can,
	title={Can language models solve graph problems in natural language?},
	author={Wang, Heng and Feng, Shangbin and He, Tianxing and Tan, Zhaoxuan and Han, Xiaochuang and Tsvetkov, Yulia},
	journal={Advances in Neural Information Processing Systems},
	volume={36},
	year={2024}
}

@article{tang2023graphgpt,
	title={Graphgpt: Graph instruction tuning for large language models},
	author={Tang, Jiabin and Yang, Yuhao and Wei, Wei and Shi, Lei and Su, Lixin and Cheng, Suqi and Yin, Dawei and Huang, Chao},
	journal={arXiv preprint arXiv:2310.13023},
	year={2023}
}

@article{dwivedi2023explainable,
	title={Explainable AI (XAI): Core ideas, techniques, and solutions},
	author={Dwivedi, Rudresh and Dave, Devam and Naik, Het and Singhal, Smiti and Omer, Rana and Patel, Pankesh and Qian, Bin and Wen, Zhenyu and Shah, Tejal and Morgan, Graham and others},
	journal={ACM Computing Surveys},
	volume={55},
	number={9},
	pages={1--33},
	year={2023},
	publisher={ACM New York, NY}
}

@article{zhao2023survey,
	title={A survey of large language models},
	author={Zhao, Wayne Xin and Zhou, Kun and Li, Junyi and Tang, Tianyi and Wang, Xiaolei and Hou, Yupeng and Min, Yingqian and Zhang, Beichen and Zhang, Junjie and Dong, Zican and others},
	journal={arXiv preprint arXiv:2303.18223},
	year={2023}
}

@article{hassija2024interpreting,
	title={Interpreting black-box models: a review on explainable artificial intelligence},
	author={Hassija, Vikas and Chamola, Vinay and Mahapatra, Atmesh and Singal, Abhinandan and Goel, Divyansh and Huang, Kaizhu and Scardapane, Simone and Spinelli, Indro and Mahmud, Mufti and Hussain, Amir},
	journal={Cognitive Computation},
	volume={16},
	number={1},
	pages={45--74},
	year={2024},
	publisher={Springer}
}

@article{wu2022survey,
	title={A survey of human-in-the-loop for machine learning},
	author={Wu, Xingjiao and Xiao, Luwei and Sun, Yixuan and Zhang, Junhang and Ma, Tianlong and He, Liang},
	journal={Future Generation Computer Systems},
	volume={135},
	pages={364--381},
	year={2022},
	publisher={Elsevier}
}

@article{taylor2023human,
	title={Human-centric AI: philosophical and community-centric considerations},
	author={Taylor, Randon R and O’Dell, Bessie and Murphy, John W},
	journal={AI \& SOCIETY},
	pages={1--8},
	year={2023},
	publisher={Springer}
}

@article{li2023survey,
	title={A survey of graph meets large language model: Progress and future directions},
	author={Li, Yuhan and Li, Zhixun and Wang, Peisong and Li, Jia and Sun, Xiangguo and Cheng, Hong and Yu, Jeffrey Xu},
	journal={arXiv preprint arXiv:2311.12399},
	year={2023}
}

@inproceedings{ye2024language,
	title={Language is all a graph needs},
	author={Ye, Ruosong and Zhang, Caiqi and Wang, Runhui and Xu, Shuyuan and Zhang, Yongfeng},
	booktitle={Findings of the Association for Computational Linguistics: EACL 2024},
	pages={1955--1973},
	year={2024}
}

@article{zhang2023graph,
	title={Graph-toolformer: To empower llms with graph reasoning ability via prompt augmented by chatgpt},
	author={Zhang, Jiawei},
	journal={arXiv preprint arXiv:2304.11116},
	year={2023}
}

@article{gao2023survey,
	title={A survey of graph neural networks for recommender systems: Challenges, methods, and directions},
	author={Gao, Chen and Zheng, Yu and Li, Nian and Li, Yinfeng and Qin, Yingrong and Piao, Jinghua and Quan, Yuhan and Chang, Jianxin and Jin, Depeng and He, Xiangnan and others},
	journal={ACM Transactions on Recommender Systems},
	volume={1},
	number={1},
	pages={1--51},
	year={2023},
	publisher={ACM New York, NY, USA}
}

@article{wang2021review,
	title={A review on graph neural network methods in financial applications},
	author={Wang, Jianian and Zhang, Sheng and Xiao, Yanghua and Song, Rui},
	journal={arXiv preprint arXiv:2111.15367},
	year={2021}
}

@inproceedings{lu2022bright,
	title={Bright-graph neural networks in real-time fraud detection},
	author={Lu, Mingxuan and Han, Zhichao and Rao, Susie Xi and Zhang, Zitao and Zhao, Yang and Shan, Yinan and Raghunathan, Ramesh and Zhang, Ce and Jiang, Jiawei},
	booktitle={Proceedings of the 31st ACM International Conference on Information \& Knowledge Management},
	pages={3342--3351},
	year={2022}
}

@article{motie2023financial,
	title={Financial fraud detection using graph neural networks: A systematic review},
	author={Motie, Soroor and Raahemi, Bijan},
	journal={Expert Systems With Applications},
	pages={122156},
	year={2023},
	publisher={Elsevier}
}

@misc{ChatGPT,
	author = "OpenAI",
	title = "ChatGPT",
	url = "chat.openai.com",
	date = "2024-05-11"  
}

@article{weininger1988smiles,
	title={SMILES, a chemical language and information system. 1. Introduction to methodology and encoding rules},
	author={Weininger, David},
	journal={Journal of chemical information and computer sciences},
	volume={28},
	number={1},
	pages={31--36},
	year={1988},
	publisher={ACS Publications}
}

@article{wu2023graph,
	title={Graph neural networks for natural language processing: A survey},
	author={Wu, Lingfei and Chen, Yu and Shen, Kai and Guo, Xiaojie and Gao, Hanning and Li, Shucheng and Pei, Jian and Long, Bo and others},
	journal={Foundations and Trends{\textregistered} in Machine Learning},
	volume={16},
	number={2},
	pages={119--328},
	year={2023},
	publisher={Now Publishers, Inc.}
}

@article{jiao2022graph,
	title={Graph representation learning meets computer vision: A survey},
	author={Jiao, Licheng and Chen, Jie and Liu, Fang and Yang, Shuyuan and You, Chao and Liu, Xu and Li, Lingling and Hou, Biao},
	journal={IEEE Transactions on Artificial Intelligence},
	volume={4},
	number={1},
	pages={2--22},
	year={2022},
	publisher={IEEE}
}

@article{jin2023spatio,
	title={Spatio-temporal graph neural networks for predictive learning in urban computing: A survey},
	author={Jin, Guangyin and Liang, Yuxuan and Fang, Yuchen and Shao, Zezhi and Huang, Jincai and Zhang, Junbo and Zheng, Yu},
	journal={IEEE Transactions on Knowledge and Data Engineering},
	year={2023},
	publisher={IEEE}
}

@article{xue2022quantifying,
	title={Quantifying the spatial homogeneity of urban road networks via graph neural networks},
	author={Xue, Jiawei and Jiang, Nan and Liang, Senwei and Pang, Qiyuan and Yabe, Takahiro and Ukkusuri, Satish V and Ma, Jianzhu},
	journal={Nature Machine Intelligence},
	volume={4},
	number={3},
	pages={246--257},
	year={2022},
	publisher={Nature Publishing Group UK London}
}

@article{li2022graph,
	title={Graph neural networks in urban intelligence},
	author={Li, Yanhua and Zhou, Xun and Pan, Menghai},
	journal={Graph Neural Networks: Foundations, Frontiers, and Applications},
	pages={579--593},
	year={2022},
	publisher={Springer}
}

@article{wang2021towards,
	title={Towards multi-grained explainability for graph neural networks},
	author={Wang, Xiang and Wu, Yingxin and Zhang, An and He, Xiangnan and Chua, Tat-Seng},
	journal={Advances in Neural Information Processing Systems},
	volume={34},
	pages={18446--18458},
	year={2021}
}

@article{huang2022graphlime,
	title={Graphlime: Local interpretable model explanations for graph neural networks},
	author={Huang, Qiang and Yamada, Makoto and Tian, Yuan and Singh, Dinesh and Chang, Yi},
	journal={IEEE Transactions on Knowledge and Data Engineering},
	year={2022},
	publisher={IEEE}
}

@inproceedings{zhang2021relex,
	title={Relex: A model-agnostic relational model explainer},
	author={Zhang, Yue and Defazio, David and Ramesh, Arti},
	booktitle={Proceedings of the 2021 AAAI/ACM Conference on AI, Ethics, and Society},
	pages={1042--1049},
	year={2021}
}

@article{vu2020pgm,
	title={Pgm-explainer: Probabilistic graphical model explanations for graph neural networks},
	author={Vu, Minh and Thai, My T},
	journal={Advances in neural information processing systems},
	volume={33},
	pages={12225--12235},
	year={2020}
}

@inproceedings{pope2019explainability,
	title={Explainability methods for graph convolutional neural networks},
	author={Pope, Phillip E and Kolouri, Soheil and Rostami, Mohammad and Martin, Charles E and Hoffmann, Heiko},
	booktitle={Proceedings of the IEEE/CVF conference on computer vision and pattern recognition},
	pages={10772--10781},
	year={2019}
}

@inproceedings{gao2021gnes,
	title={Gnes: Learning to explain graph neural networks},
	author={Gao, Yuyang and Sun, Tong and Bhatt, Rishab and Yu, Dazhou and Hong, Sungsoo and Zhao, Liang},
	booktitle={2021 IEEE International Conference on Data Mining (ICDM)},
	pages={131--140},
	year={2021},
	organization={IEEE}
}

@inproceedings{tang2023explainable,
	title={Explainable Spatio-Temporal Graph Neural Networks},
	author={Tang, Jiabin and Xia, Lianghao and Huang, Chao},
	booktitle={Proceedings of the 32nd ACM International Conference on Information and Knowledge Management},
	pages={2432--2441},
	year={2023}
}

@inproceedings{nian2024globally,
	title={Globally Interpretable Graph Learning via Distribution Matching},
	author={Nian, Yi and Chang, Yurui and Jin, Wei and Lin, Lu},
	booktitle={Proceedings of the ACM on Web Conference 2024},
	pages={992--1002},
	year={2024}
}

@article{wang2022gnninterpreter,
	title={Gnninterpreter: A probabilistic generative model-level explanation for graph neural networks},
	author={Wang, Xiaoqi and Shen, Han-Wei},
	journal={arXiv preprint arXiv:2209.07924},
	year={2022}
}

@article{shin2024page,
	title={PAGE: prototype-based model-level explanations for graph neural networks},
	author={Shin, Yong-Min and Kim, Sun-Woo and Shin, Won-Yong},
	journal={IEEE Transactions on Pattern Analysis and Machine Intelligence},
	year={2024},
	publisher={IEEE}
}

@article{spinelli2022meta,
	title={A meta-learning approach for training explainable graph neural networks},
	author={Spinelli, Indro and Scardapane, Simone and Uncini, Aurelio},
	journal={IEEE Transactions on Neural Networks and Learning Systems},
	year={2022},
	publisher={IEEE}
}

@article{seo2024interpretable,
	title={Interpretable Prototype-based Graph Information Bottleneck},
	author={Seo, Sangwoo and Kim, Sungwon and Park, Chanyoung},
	journal={Advances in Neural Information Processing Systems},
	volume={36},
	year={2024}
}

@article{hejna2023contrastive,
	title={Contrastive preference learning: Learning from human feedback without rl},
	author={Hejna, Joey and Rafailov, Rafael and Sikchi, Harshit and Finn, Chelsea and Niekum, Scott and Knox, W Bradley and Sadigh, Dorsa},
	journal={arXiv preprint arXiv:2310.13639},
	year={2023}
}

@article{ouyang2022training,
	title={Training language models to follow instructions with human feedback},
	author={Ouyang, Long and Wu, Jeffrey and Jiang, Xu and Almeida, Diogo and Wainwright, Carroll and Mishkin, Pamela and Zhang, Chong and Agarwal, Sandhini and Slama, Katarina and Ray, Alex and others},
	journal={Advances in neural information processing systems},
	volume={35},
	pages={27730--27744},
	year={2022}
}

@article{nesvijevskaia2021accuracy,
	title={The accuracy versus interpretability trade-off in fraud detection model},
	author={Nesvijevskaia, Anna and Ouillade, Sophie and Guilmin, Pauline and Zucker, Jean-Daniel},
	journal={Data \& Policy},
	volume={3},
	pages={e12},
	year={2021},
	publisher={Cambridge University Press}
}

@inproceedings{jang2023toward,
	title={Toward Interpretable Machine Learning: Constructing Polynomial Models Based on Feature Interaction Trees},
	author={Jang, Jisoo and Kim, Mina and Bui, Tien-Cuong and Li, Wen-Syan},
	booktitle={Pacific-Asia Conference on Knowledge Discovery and Data Mining},
	pages={159--170},
	year={2023},
	organization={Springer}
}

@article{swiechowski2023monte,
	title={Monte Carlo tree search: A review of recent modifications and applications},
	author={{\'S}wiechowski, Maciej and Godlewski, Konrad and Sawicki, Bartosz and Ma{\'n}dziuk, Jacek},
	journal={Artificial Intelligence Review},
	volume={56},
	number={3},
	pages={2497--2562},
	year={2023},
	publisher={Springer}
}

@inproceedings{bui2023toward,
	title={Toward Interpretable Graph Neural Networks via Concept Matching Model},
	author={Bui, Tien-Cuong and Li, Wen-Syan},
	booktitle={2023 IEEE International Conference on Data Mining (ICDM)},
	pages={950--955},
	year={2023},
	organization={IEEE}
}

@inproceedings{bui2024human,
	title={Human-Driven Active Verification for Efficient and Trustworthy Graph Classification},
	author={Bui, Tien-Cuong and Li, Wen-Syan},
	booktitle={Pacific-Asia Conference on Knowledge Discovery and Data Mining},
	pages={105--116},
	year={2024},
	organization={Springer}
}

@inproceedings{bui2024toward,
	title={Toward Interpretable Graph Classification via Concept-Focused Structural Correspondence},
	author={Bui, Tien-Cuong and Li, Wen-Syan},
	booktitle={Pacific-Asia Conference on Knowledge Discovery and Data Mining},
	pages={20--31},
	year={2024},
	organization={Springer}
}

@article{dig,
	author  = {Meng Liu and Youzhi Luo and Limei Wang and Yaochen Xie and Hao Yuan and Shurui Gui and Haiyang Yu and Zhao Xu and Jingtun Zhang and Yi Liu and Keqiang Yan and Haoran Liu and Cong Fu and Bora M Oztekin and Xuan Zhang and Shuiwang Ji},
	title   = {{DIG}: A Turnkey Library for Diving into Graph Deep Learning Research},
	journal = {Journal of Machine Learning Research},
	year    = {2021},
	volume  = {22},
	number  = {240},
	pages   = {1-9},
	url     = {http://jmlr.org/papers/v22/21-0343.html}
}

@article{jang2025prompt,
	title={Prompt Tuning for Natural Language to SQL with Embedding Fine-Tuning and RAG},
	author={Jang, Jisoo and Bui, Tien-Cuong and Choi, Yunjun and Li, Wen-Syan},
	journal={arXiv preprint arXiv:2511.08245},
	year={2025}
}

@article{le2024auto,
	title={Auto-Generating Earnings Report Analysis via a Financial-Augmented LLM},
	author={Le, Van-Duc},
	journal={arXiv preprint arXiv:2412.08179},
	year={2024}
}

@article{zhang2025survey,
	title={A survey on data selection for llm instruction tuning},
	author={Zhang, Bolin and Wang, Jiahao and Du, Qianlong and Zhang, Jiajun and Tu, Zhiying and Chu, Dianhui},
	journal={Journal of Artificial Intelligence Research},
	volume={83},
	year={2025}
}

@article{le2023spatiotemporal,
	title={Spatiotemporal graph convolutional recurrent neural network model for citywide air pollution forecasting},
	author={Le, Van-Duc},
	journal={arXiv preprint arXiv:2304.12630},
	year={2023}
}

@inproceedings{duong2022towards,
	title={Towards an error-free deep occupancy detector for smart camera parking system},
	author={Duong, Tung-Lam and Le, Van-Duc and Bui, Tien-Cuong and To, Hai-Thien},
	booktitle={European Conference on Computer Vision},
	pages={163--178},
	year={2022},
	organization={Springer}
}

@inproceedings{to2021real,
	title={Real-time social distancing alert system using pose estimation on smart edge devices},
	author={To, Hai-Thien and Bui, Khac-Hoai Nam and Le, Van-Duc and Bui, Tien-Cuong and Li, Wen-Syan and Cha, Sang Kyun},
	booktitle={Asian conference on intelligent information and database systems},
	pages={291--300},
	year={2021},
	organization={Springer}
}
	\endgroup
	
	% Turn on for acknowledgment (English) (ssss)
	\cleardoublepage
	\acknowledgment
	
	Foremost, my deepest gratitude goes to my wife and family. My wife endured countless difficult times by my side. Her kindness, patience, and understanding through challenging times have been immeasurable. My family's unwavering support has been the foundation for my entire academic journey.

I am thankful to Professor Sin-Doo Lee for recommending me to my first advisor, Professor Sang Kyun Cha. This introduction set me on the course of my academic journey.
 
I am profoundly grateful to my advisors, Professor Sang Kyun Cha and Professor Wen-Syan Li. Professor Cha, my first advisor, took a chance on me as his student, imparting not only engineering knowledge but also an entrepreneurial spirit. His guidance on academia, markets, and the Korean-Vietnamese relationship has been invaluable. Professor Li's extraordinary support, especially after Professor Cha's retirement, enabled me to persevere in my doctoral studies. His exceptional mentorship, insightful discussions, and genuine friendship have shaped my academic and professional development.

I want to express my sincere gratitude to Professor Kyomin Jung and Professor Insoon Yang for their invaluable contributions in co-authoring my first paper. Their guidance was essential in shaping my understanding of academic writing. My sincere thanks extend to all esteemed professors who served on my committee. 
 
My lab mates, who supported me throughout seven challenging years, deserve my appreciation. In particular, I want to thank Duc and Tommy for countless conversations that helped me navigate both life's difficulties and personal struggles.

	{\raggedleft \hspace*{\fill}Bui Tien Cuong\\\hspace*{\fill}Seoul, June 2024}
	
	% Turn on to include index (ssss)
	\cleardoublepage
	\begingroup
	\setstretch{1.2}
	\phantomsection\clearpage\addcontentsline{toc}{chapter}{Index}
	\printindex
	\endgroup
	
	% Turn on for Korean abstract (ssss)
	\cleardoublepage
	\addtolength{\topskip}{0pt plus-10pt}
	\addtolength{\parskip}{0pt plus-10pt}
	\keywordalt{그래프 신경망, 설명 가능한 AI, 인간 중심 기계 학습, 사례 기반 추론, 지식 증류, 대형 언어 모델}
		
	\begin{abstractalt}
		\hspace{\parindent} 
		그래프 신경망 (GNNs)은 그래프 구조의 데이터를 모델링하고 분석하는 강력한 도구로 자리 잡았습니다. 이러한 모델의 광범위한 적용은 그 가치를 부각시킵니다. 그러나 이러한 방법의 복잡성은 종종 결정 과정을 이해하는 데 장애가 됩니다. 현재 설명 가능한 인공지능 (XAI) 방법은 그래프 내의 복잡한 관계와 상호작용을 풀어내는 데 어려움을 겪고 있습니다. 여러 방법들이 사후 접근 (post-hoc approach) 또는 자체 해석 가능한 설계를 통해 이 격차를 메우려 시도했습니다. 대부분은 예측 결과와 관련된 핵심 패턴을 파악하기 위해 그래프 구조 분석에 초점을 맞춥니다. 사후 설명 방법은 적응 가능하지만 추가적인 계산 자원을 요구하며, 모델 내부 작동에 대한 접근이 제한되어 있어 신뢰성이 떨어질 수 있습니다. 반면, 해석 가능한 모델은 즉각적인 설명을 제공할 수 있지만 다양한 시나리오에 일반화하는 것은 주요한 우려사항입니다.
		
		이러한 단점을 해결하기 위해, 이 논문은 그래프 기반 기계 학습을 위한 새로운 XAI 프레임워크를 개발하고자 합니다. 제안된 프레임워크는 개별 특성 분석을 넘어서 그래프 구조가 예측에 미치는 영향을 포착하는 적응 가능하고 계산적으로 효율적인 설명을 제공하고자 합니다. 이는 기존 GNN 아키텍처의 해석성을 강화하기 위해 특정 유형의 상호작용 (예: 특성 또는 메시지 전달 과정)을 포착하는 여러 전문 학습자를 훈련함으로써 일반적인 접근 방식을 제시합니다. 이후에는 훈련된 전문 학습자를 기반으로 다양한 설명 모달리티를 제공하는 여러 설명자를 구축합니다. 예시 기반 설명의 효과성과 KNN 알고리즘의 자연스러운 해석 가능성은 새로운 해석 가능한 GNN을 창조하도록 동기를 부여합니다. 이 프레임워크는 훈련 그래프에서 자주 발생하는 ``개념'' (하위 구조)을 추출하여 예측을 추론하고 설명을 생성하는 기반이 됩니다. 목표는 사용자 중심의 간결한 통찰을 제공하는 다면적 설명 시스템입니다. 또한, 프레임워크는 두 그래프 간의 구조 유사성을 지구 이동 거리 (Earth Mover Distance) 최적 운송을 통해 근사하는 방법을 제안하여 예측 성능과 사용자의 참조 선택 이해를 향상시킵니다. 다양한 설명 모달리티는 사용자에게 모델의 내부 논리에 대한 의미 있는 통찰을 제공하여 모델 디버깅, 편향 제거 및 개선에 활용할 수 있습니다. 이러한 직관에 기반하여, 프레임워크는 도메인 지식을 통합하여 GNN을 더욱 인간이 이해할 수 있는 표현으로 안내하고 이 기술의 신뢰성과 윤리적 사용을 촉진하고자 합니다. 구체적으로는 도메인 전문가가 표현 학습과 참조 선택 과정을 적극적으로 검증하고 제어할 수 있도록 다중 수준의 지식 가이드 제약을 제공합니다. 이 논문은 제안된 프레임워크의 효율성과 효과성을 강조하는 광범위한 실험 결과와 발견을 제시합니다. 마지막으로, 미래 작업, 실용적인 응용 프로그램 및 대규모 언어 모델 같은 최신 고급 분야로의 잠재적 확장에 대한 가능한 방향에 대한 철저한 논의로 결론짓습니다.
		
	\end{abstractalt}
	
	% %% Turn on for Korean acknowledgment (ssss)
	% \cleardoublepage
	% \acknowledgement
	% \hspace{\parindent}한국어로 감사합니다!
	
	% \bibliographystyle{IEEEtran}
	% \bibliography{references}

\end{document}